\documentclass{article}
\usepackage[dvipsnames]{xcolor}
\usepackage[numbers]{natbib}
\usepackage[preprint]{neurips_2024}

\usepackage{hyperref}
\usepackage{url}
\usepackage{graphicx}
\usepackage{amsmath, nccmath}
\usepackage{bbm}
\usepackage{multirow}
\usepackage{soul}
\usepackage{makecell}
\usepackage{empheq}
\usepackage{listings}
\usepackage{epigraph}
\usepackage{pdfpages}
\usepackage{wrapfig}
\hypersetup{hidelinks}
\usepackage{booktabs}
\usepackage{dashrule}
\usepackage{tabularx,ragged2e}

\usepackage{caption}
\usepackage{subcaption}
\usepackage{diagbox}

\usepackage{longtable}

\newcommand\scalemath[2]{\scalebox{#1}{\mbox{\ensuremath{\displaystyle #2}}}}

\usepackage[skins,theorems]{tcolorbox}
\tcbset{highlight math style={enhanced,
  colframe=red,colback=white,arc=0pt,boxrule=1pt}}

\newcommand{\arcsinh}{\text{arcsinh}}
\newcommand{\arccosh}{\text{arccosh}}
\newcommand{\arctanh}{\text{arctanh}}

\newcommand{\mb}{\mathbf}
\newcommand{\bs}{\boldsymbol}
\newcommand{\mc}{\mathcal}

\newcommand{\ie}{\textit{i.e.,}}
\newcommand{\eg}{\textit{e.g.,}}
\newcommand{\etal}{\textit{et al.}}

\newtheorem{definition}{\textsc{Definition}}
\newtheorem{theorem}{\textsc{Theorem}}
\newtheorem{example}{\textsc{Example}}

\newcommand{\our}{\textsc{RPN}}
\newcommand{\toolkit}{\href{https://www.tinybig.org}{\textcolor{Plum}{\textbf{\textsc{tinyBIG}}}}}

\newcommand{\insertquote}[2]{%
    \begin{quote}
        \textit{#1}
        \par
        \hfill --- #2
    \end{quote}%
}

\usepackage{array}
\usepackage{makecell}

\usepackage[edges]{forest}
\tikzset{%
    parent/.style =          {align=center,yshift=-0cm,text width=1.2cm,rounded corners=3pt},
    child/.style =           {align=center,text width=1.3cm,rounded corners=3pt},
    grandchild/.style =      {align=left,text width=2.7cm,rounded corners=3pt},
    greatgrandchild/.style = {align=left,text width=6.3cm,rounded corners=3pt},
    referenceblock/.style =  {align=left,text width=2.5cm,rounded corners=3pt}
}

\DeclareMathAlphabet{\mathcal}{OMS}{cmsy}{m}{n}
\SetMathAlphabet{\mathcal}{bold}{OMS}{cmsy}{b}{n}

\title{{\our}: Reconciled Polynomial Network\\ [1ex]
\Large{
Towards Unifying PGMs, Kernel SVMs, MLP and KAN
}}
\author{%
  Jiawei Zhang\\
  IFM Lab\thanks{\textcopyright\ 2024 IFM Lab. All rights reserved. The {\our} project and {\toolkit} toolkit is developed and maintained by IFM Lab.}\\
  Department of Computer Science\\
  University of California, Davis\\
  \texttt{jiawei@ifmlab.org} \\
  {Project Website: \color{magenta}{https://www.tinybig.org}}\\
  {Github: \color{magenta}{https://github.com/jwzhanggy/tinyBIG}}\\
  (Initial Version: \today)
 }
  
\begin{document}
\maketitle

%------------------------------------
\begin{figure*}[h]
    \begin{minipage}{\textwidth}
    \centering
    	\vspace{-25pt}
	\href{https://www.tinybig.org}{
    	\includegraphics[width=0.5\linewidth]{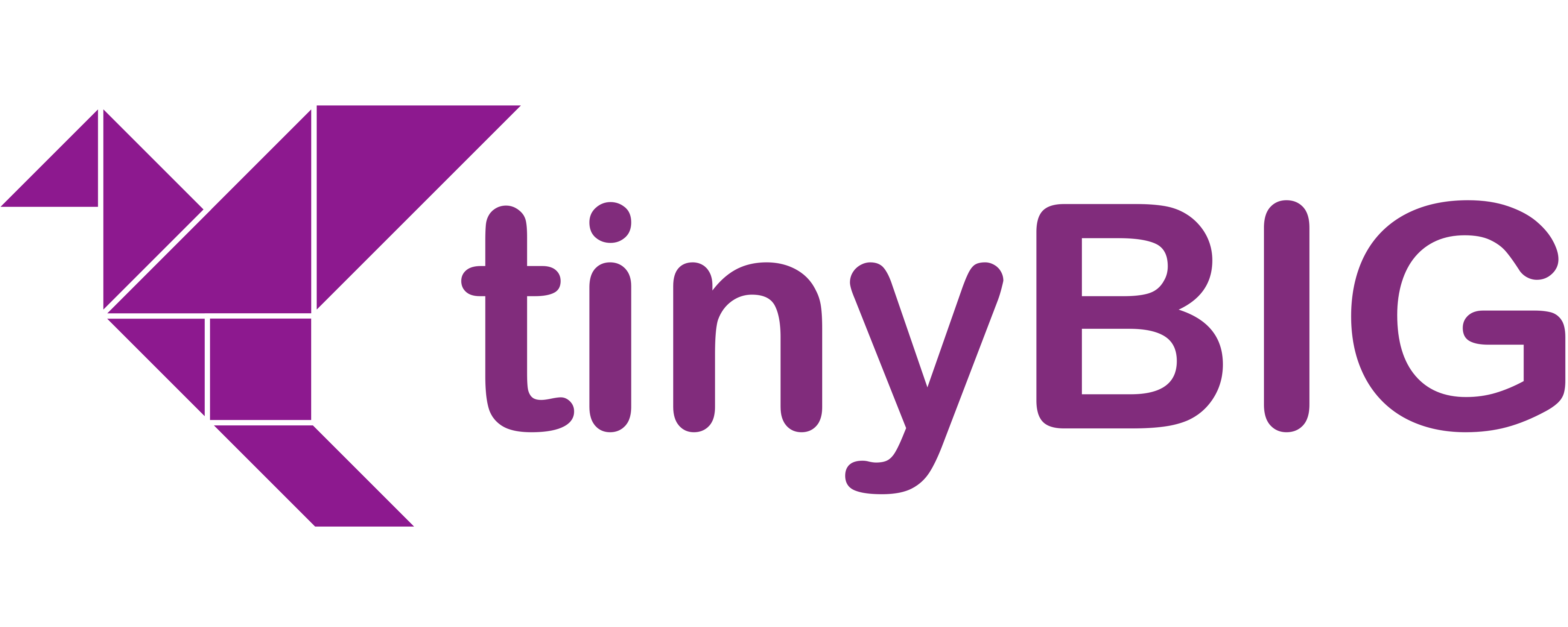}
	}
    \end{minipage}%
\end{figure*}
%------------------------------------

\insertquote{``With four parameters I can fit an elephant, and with five I can make him wiggle his trunk."\\}{Von Neumann}
%\vspace{10pt}

\begin{abstract}

In this paper, we will introduce a novel deep model named Reconciled Polynomial Network ({\our}) for deep function learning. {\our} has a very general architecture and can be used to build models with various complexities, capacities, and levels of completeness, which all contribute to the correctness of these models. As indicated in the subtitle, {\our} can also serve as the backbone to unify different base models into one canonical representation. This includes non-deep models, like probabilistic graphical models (PGMs) - such as Bayesian network and Markov network - and kernel support vector machines (kernel SVMs), as well as deep models like the classic multi-layer perceptron (MLP) and the recent Kolmogorov-Arnold network (KAN).

Technically, inspired by the Taylor's Theorem, {\our} proposes to disentangle the underlying function to be inferred into the inner product of a {data expansion function} and a {parameter reconciliation function}. Together with the {remainder function}, {\our} accurately approximates the underlying functions that governs data distributions. The {data expansion functions} in {\our} project data vectors from the input space to a high-dimensional intermediate space, specified by the expansion functions in definition. Meanwhile, {\our} also introduces the {parameter reconciliation functions} to fabricate a small number of parameters into a higher-order parameter matrix to address the ``curse of dimensionality'' problem caused by the data expansions. In the intermediate space, the expanded vectors are polynomially integrated and further projected into the low-dimensional output space via the inner product with the {reconciled parameters} generated by these {parameter reconciliation functions}. Moreover, the {remainder functions} provide {\our} with additional complementary information to reduce potential approximation errors.

We conducted extensive empirical experiments on numerous benchmark datasets across multiple modalities, including continuous function datasets, discrete vision and language datasets, and classic tabular datasets, to investigate the effectiveness of {\our}. The experimental results demonstrate that, {\our} outperforms MLP and KAN with mean squared errors at least $\times 10^{-1}$ lower (and even $\times 10^{-2}$ lower in some cases) for continuous function fitting. On both vision and language benchmark datasets, using much less learnable parameters, {\our} consistently achieves higher accuracy scores than Naive Bayes, kernel SVMs, MLP, and KAN for discrete image and text data classifications. In addition, equipped with the probabilistic data expansion functions, {\our} learns better probabilistic dependency relationships among variables and outperforms other probabilistic models, including Naive Bayes, Bayesian networks, and Markov networks, for learning on classic tabular benchmark datasets. 

Reconciled Polynomial Network ({\our}) proposed in this paper provides the opportunity to represent and interpret current machine and deep learning models as sequences of vector space expansions and parameter reconciliations. These functions can all deliver concrete physical meanings about both the input data and model parameters. Furthermore, the application of simple inner-product and summation operations to these functions significantly enhances the interpretability of {\our}. This paper presents not only empirical experimental investigations but also in-depth discussions on {\our}, addressing its interpretations, merits, limitations, and potential future developments.

What's more, to facilitate the implementation of {\our}-based models, we have developed and released a toolkit named {\toolkit}. This toolkit encompasses all functions, modules, and models introduced in this paper, accompanied by comprehensive documentation and tutorials. Detailed information about {\toolkit} is available at the project's GitHub repository and the dedicated project webpage, with their respective URLs provided above.

\end{abstract}


%\noindent \textbf{KEY WORDS}: Function Learning; Data expansion; Parameter Reconciliation; Remainder; Deep Learning
%\keywords{TBD}
%%%%% NIPS %%%%%

{
\newpage
\vspace{3em}
\hrule
\vspace{1em}
\setcounter{tocdepth}{2}
\tableofcontents
\vspace{3em}
\hrule
\vspace{1em}
%--------------------------------------------------------------------------
}

{
\newpage
%--------------------------------------------------------------------------

%--------------------------------------------------------------------------
\section{Introduction}\label{sec:introduction}

Over the past 70 years, the field of artificial intelligence has experienced dramatic changes in both the problems studied and the models used. With the emergence of new learning tasks, various machine learning models, each designed based on different prior assumptions, have been proposed to address these problems. As shown in Figure~\ref{fig:timeline}, we illustrate the timeline about three types of machine learning models that have dominated the field of artificial intelligence in the past 50 years, including {probabilistic graphical models} \cite{kindermann1980markov, Pearl_CSS85, 10.5555/1795555}, {support vector machines} \cite{10.1023/A:1022627411411, 10.5555/2998981.2999021, 10.1145/130385.130401} and {deep neural networks} \cite{Rumelhart1986LearningRB, GoodBengCour16}. Along with important technological breakthroughs, these models each had their moments of prominence and have been extensively explored and utilized in various research and application tasks related to data and learning nowadays. Besides these three categories of machine learning models, there are many other models ({\eg} the tree based models and clustering models) that do not fit into these categories, but we will not discuss them in this paper and will leave them for future investigation instead.

%------------------------------------
\definecolor{darkgreen}{rgb}{0.0, 0.5, 0.0}
\begin{figure*}[h!]
\vspace{1em}
    \begin{minipage}{\textwidth}
    \centering
    	\includegraphics[width=1.0\linewidth]{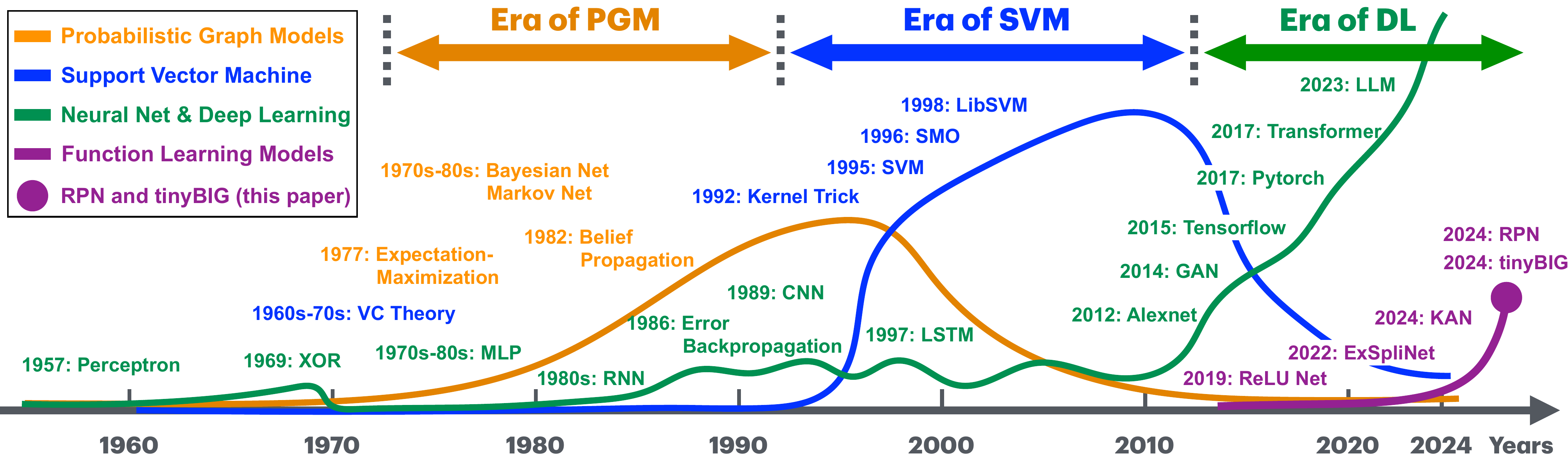}
    	\caption{The timeline illustrates the development of various dominant machine learning base models over the past 70 years, with different colors representing different models. \textcolor{orange}{\textbf{Orange Color}}: probabilistic graphical models (1980s to mid-2000s); \textcolor{blue}{\textbf{Blue Color}}: support vector machine (mid 1990s to early 2010s); \textcolor{darkgreen}{\textbf{Green Color}}: deep learning models (mid-2010s to present); and \textcolor{Plum}{\textbf{Purple Color}}: deep function learning (2020s to present).}
    	\label{fig:timeline}
    \end{minipage}%
\end{figure*}
%------------------------------------

In this paper, we will introduce a novel deep model, namely \textbf{Reconciled Polynomial Network} (\textbf{\our}), that can potentially unify these different aforementioned base models into one shared representation. In terms of model architecture, {\our} consists of three component functions: \textbf{data expansion function}, \textbf{parameter reconciliation function} and \textbf{remainder function}. Inspired by the Taylor's theorem, {\our} disentangles the input data from model parameters, and approximates the target functions to be inferred as the inner product of the {data expansion function} with the {parameter reconciliation function}, subsequently summed with the {remainder function}.

Based on architecture of {\our}, inferring the diverse underlying mapping that governs data distributions (from inputs to outputs) is actually equivalent to inferring these three compositional functions. This inference process of the diverse data distribution mappings based on {\our} is named as the \textbf{function learning task} in this paper. Specifically, the ``function'' term mentioned in the task name refers to not only the \textbf{mathematical function} components composing the {\our} model but also the \textbf{cognitive function} of {\our} as an intelligent system to relate input signals with desired output response. Function learning has been long-time treated as equivalent to the continuous function fitting and approximation for regression tasks only. Actually, in psychology and cognitive science, researchers have also used the function learning concept for modeling the mental induction process of stimulus-response relations of human and other intelligent subjects \cite{carroll1963functional, koh1991function}, involving the acquisition of knowledge, manipulation of information and reasoning. In this paper, we argue that {function learning} is the most fundamental task in intelligent model learning, encompassing \textbf{continuous function approximation}, \textbf{discrete vision and language data recognition and prediction}, and \textbf{cognitive and logic dependency relation induction}. The following Section~\ref{sec:function_learning} will provide an in-depth discussion of {function learning} and offer a comparative analysis of function learning with the currently prevailing paradigm of {representation learning}.

Determined by the definitions of the data expansion functions, {\our} will project data vectors from the input space to an intermediate (higher-dimensional) space represented with new basis vectors. To address the ``curse of dimensionality'' issue stemming from the data expansions, the {parameter reconciliation function} in {\our} fabricates a reduced set of parameters into a higher-order parameter matrix. These expanded data vectors are then polynomially integrated via the inner product with these generated reconciled parameters, which further projects these expanded data vectors back to the desired lower-dimensional output space. Moreover, the {remainder function} provides {\our} with additional complementary information to further reduce potential approximation errors. All these component functions within {\our} embody concrete physical meanings. These functions, coupled with the straightforward application of simple inner product and summation operators, provide {\our} with greater interpretability compared to other existing base models. 

{\our} possesses a highly versatile architecture capable of constructing models with diverse complexities, capacities, and levels of completeness. In this paper, to provide {\our} with greater modeling capabilities in design, we enable {\our} to incorporate both a wide architecture featuring multi-heads and multi-channels (within each layer), as well as a deep architecture comprising multi-layers. Additionally, we further offer {\our} with a more adaptable and lightweight mechanism for constructing models with comparable capabilities through the nested and extended data expansion functions. These powerful yet flexible design mechanisms provide {\our} with greater modeling capability, enabling it to serve as the backbone for unifying various base models mentioned above into a single representation. This includes non-deep models, like probabilistic graphical models (PGMs) - such as Bayesian network \cite{Pearl_CSS85} and Markov network \cite{kindermann1980markov} - and kernel support vector machines (kernel SVMs) \cite{10.1145/130385.130401}, as well as deep models like the classic multi-layer perceptron (MLP) \cite{Rumelhart1986LearningRB} and the recent Kolmogorov-Arnold network (KAN) \cite{Liu2024KANKN}.

To investigate the effectiveness of {\our} for deep function learning tasks, this paper will present extensive empirical experiments conducted on numerous benchmark datasets. Given {\our}'s status as a general base model for function learning, we evaluate its performance with datasets in various modalities, including numerical function datasets (for continuous function fitting and approximation), image and text datasets (for discrete vision and language data classification), and classic tabular datasets (for variable dependency relationship inference and induction). The experimental results demonstrate that, {\our} outperforms MLP and KAN with mean squared errors at least $\times 10^{-1}$ lower (and even $\times 10^{-2}$ lower in some cases) on continuous function fitting tasks. On both vision and language benchmark datasets, using much less learnable parameters, {\our} consistently achieves higher accuracy scores than Naive Bayes, kernel SVM, MLP, and KAN for these discrete data classifications. Moreover, equipped with probabilistic data expansion functions, {\our} also learns better probabilistic dependency relationships among variables and outperforms probabilistic models, including Naive Bayes, Bayesian networks, and Markov networks, for learning on the tabular benchmark datasets. 

We summarize the contributions of this paper as follows:
\begin{itemize}

\item \textbf{{\our} for Deep Function Learning}: In this paper, we propose the task of ``deep function learning'' and introduce a novel deep function learning base model, {\ie} the Reconciled Polynomial Network ({\our}). {\our} has a versatile model architecture and attains superior modeling capabilities for diverse deep function learning tasks on various multi-modality datasets. Moreover, by disentangling input data from model parameters with the expansion, reconciliation and remainder functions, {\our} achieves greater interpretability than existing deep and non-deep base models. 

\item \textbf{Component Functions}: In this paper, we introduce a tripartite set of compositional functions - data expansion, parameter reconciliation, and remainder functions - that serve as the building blocks for the {\our} model. By strategically combining these component functions, we can construct a multi-head, multi-channel, and multi-layer architecture, enabling {\our} to address a wide spectrum of learning challenges across diverse function learning tasks.

\item \textbf{Base Model Unification}: This paper demonstrates that {\our} provides a unifying framework for several influential base models, including Bayesian networks, Markov networks, kernel SVMs, MLP, and KAN. We show that, through specific selections of component functions, each of these models can be unified into {\our}'s canonical representation, characterized by the inner product of a data expansion function with a parameter reconciliation function, summed with a remainder function.

\item \textbf{Experimental Investigations}: This paper presents a series of extensive empirical experiments conducted across numerous benchmark datasets for various deep function learning tasks, including numerical function fitting tasks, discrete image and language data classification tasks, and tabular data based dependency relation inference and induction tasks. The results demonstrate {\our}'s consistently superior performance compared to other existing base models, providing strong empirical validations of our proposed model.

\item \textbf{The {\toolkit} Toolkit}: To facilitate the adoption, implementation and experimentation of {\our}, we have released {\toolkit}, a comprehensive toolkit for {\our} model construction. {\toolkit} offers a rich library of pre-implemented functions, including $25$ categories of data expansion functions, $10$ parameter reconciliation functions, and $5$ remainder functions, along with the complete model framework and optimized model training pipelines. This integrated toolkit enables researchers to rapidly design, customize, and deploy {\our} models across a wide spectrum of deep function learning tasks.

\end{itemize}

This paper provides a comprehensive investigation of the proposed {Reconciled Polynomial Network} model. The remaining parts of this paper will be organized as follows. In Section~\ref{sec:function_learning}, we will first introduce the novel {function learning} concept and compare {\our} with several existing base models. In Section~\ref{sec:formulation}, we will cover notations, task formulations, and essential background knowledge on Taylor's theorem. In Section~\ref{sec:method}, we will provide detailed descriptions of {\our} model's architecture and design mechanisms. Our library of expansion, reconciliation, and remainder functions will be presented in Section~\ref{sec:functions}. In Section~\ref{sec:backbone_unification}, we demonstrate how {\our} unifies and represents existing base models. The experimental evaluation of {\our}'s performance on numerous benchmark datasets will be provided in Section~\ref{sec:experiments}. After that, we will discuss {\our}'s interpretations from both machine learning and biological neuroscience perspectives in Section~\ref{sec:interpretation}. In Section~\ref{sec:merits_limitations}, we will critically discuss the merits, limitations and potential future works of {\our}. Finally, we will introduce the related works in Section~\ref{sec:related_work} and conclude this paper in Section~\ref{sec:conclusion}.

%--------------------------------------------------------------------------

%--------------------------------------------------------------------------
%--------------------------------------------------------------------------
\section{Deep Function Learning}\label{sec:function_learning}

In this section, we will first introduce the concept of \textbf{deep function learning} task. After that, we will provide the detailed clarifications about how deep function learning differs from existing deep representation learning tasks. Based on this concept, we will further compare {\our}, the deep function learning model proposed in this paper, with other existing non-deep and deep base models to illustrate their key differences.

\subsection{What is Deep Function Learning?}

As its name suggests, \textbf{deep function learning}, as the most fundamental task in machine learning, aims to build general deep models composed of a sequence of component functions that infer the relationships between inputs and outputs. These component functions define the mathematical projections across different data and parameter spaces. In deep function learning, without any prior assumptions about the data modalities, the corresponding input and output data can also appear in different forms, including but not limited to continuous numerical values (such as continuous functions), discrete categorical features (such as images and language data), probabilistic variables (defining the dependency relationships between inputs and outputs), and others.

\begin{definition}
(\textbf{Deep Function Learning}): Formally, given the input and output spaces $\mathbbm{R}^m$ and $\mathbbm{R}^n$, the underling mapping that governs the data projection between these two spaces can be denoted as:
\begin{equation}
f: \mathbbm{R}^m \to \mathbbm{R}^n.
\end{equation}

Deep function learning aims to build a model $g$ as a composition of deep mathematical function sequences $g_1, g_2, \cdots, g_K$ to project data cross different vector spaces, which can be represented as
\begin{equation}
g: \mathbbm{R}^m \to \mathbbm{R}^n, \text{ and } g = g_1 \circ g_2 \circ \cdots \circ g_k,
\end{equation}
where the $\circ$ notation denotes the component function integration and composition operators. The component functions $g_i$ can be defined on either input data or the model parameters. 

For input $\mb{x} \in \mathbbm{R}^m$, if the output generated by the model can approximate the desired output, {\ie} 
\begin{equation}
g(\mb{x} | \mb{w}, \bs{\theta}) \approx f(\mb{x}),
\end{equation} 
then model is $g$ can serve as an approximated mapping of $f$. Notations $\mb{w} \in \mathbbm{R}^{l}$ and $\bs{\theta} \in \mathbbm{R}^{l'}$ denote the learnable parameters and hyper-parameters of the function learning model, respectively.
\end{definition}

Below, we will further clarify the distinctions between deep function learning and current deep model-based data representation learning tasks. After that, we will compare our {\our} model, which is grounded in deep function learning, against other existing base models.

%--------------------------------------------------------------------------

\subsection{Deep Function Learning vs Deep Representation Learning}

%As a special type of machine learning, both deep function learning and deep representation learning are focused on inferring the underlying distributions of data. In contrast to representation learning,  deep function learning narrows down the model architecture to a sequence of concrete mathematical functions defined on both data and parameter spaces. Deep function learning disentangles data from parameters and aims to infer and learn these compositional functions, each bearing a concrete physical interpretation for mathematical projections between various data and parameter domains. In contrast, existing deep representation learning models inextricably mix data and parameters together, rendering model interpretability virtually impossible. Moreover, since there are no prior assumptions about data modalities or categories of input-output relations, deep function learning models must be general and adaptable to various application scenarios. This requirement greatly enforces the generalizability of deep function learning models, an aspect overlooked by existing representation learning models, and is critical for achieving future artificial general intelligence, which also renders deep function learning tasks and models significantly different from other learning tasks and methods.

As mentioned previously, the {function learning} tasks and models examined in this paper encompass not only continuous function approximation, but also discrete data classification and the induction of dependency relations. Besides the literal differences indicated by their names - representation learning is data oriented but function learning is model oriented - deep function learning significantly differs from the current deep representation learning in several critical perspectives discussed below.

\begin{itemize}
\item \textbf{Generalizability}: Representation learning, to some extent, has contributed to the current fragmentation within the AI community, as data - the carrier of information - is collected, represented, and stored in disparate modalities. Existing deep models, specifically designed for certain modalities, tend to overfit to these modality-specific data representations in addition to learning the underlying information. Applying a model proposed for one modality to another typically necessitates significant architectural redesigns. Recently, there have been efforts to explore the cross-modal applicability of certain models, {\eg} CNNs for language and Transformers for vision, but replicating such cross-modality migration explorations across all current and future deep models is extremely expensive and unsustainable. Furthermore, to achieve the future artificial general intelligence (AGI), the available data in a single modality is no longer sufficient for training larger models. Deep function learning, without any prior assumptions on data modalities, will pave the way for improving the model generalizability. These learned functions should demonstrate their generalizability and applicability to multi-modal data from the outset, during their design and investigation phases. 

\item \textbf{Interpretability}: Representation learning primarily aims to learn and extract latent patterns or salient features from data, aligning with the technological advancements in data science and big data analytics over the past two decades. However, the learned data representations often lack interpretable physical meanings, rendering most current AI models to be black boxes. In contrast, to realize the goal of explainable AI (xAI), greater emphasis must be placed on developing new model architectures with concrete physical meanings and mathematical interpretations in the future. The {\our} based deep function learning, on the other hand, aims to learn compositional functions with inherent physical interpretability for building general-purpose models across various tasks, thereby bridging the interpretability gap of current and future deep models.

\item \textbf{Reusability}: Representation learning converts input data into embedding vectors that can be stored in vector databases and reused in future applications ({\eg} in the retrieval-augmented generation (RAG) models). However, in practical real-world scenarios, the direct usability of such pre-computed embedding representations in vector databases can be quite limited, both in terms of use cases and transactional queries or operations. Moreover, as new data arrives, new architectures are designed, and new model checkpoints are updated in the dynamically evolving online and offline worlds, we may need to re-learn all these embedding representation vectors via fine-tuning or retraining from scratch to maintain consistency, greatly impacting the reusability of representation learning results. In contrast, function learning focuses on learning compositional functions for underlying mapping inference, whose disentangled design is inherently well-suited for reusability and continual learning in future AI systems.
\end{itemize}

As a special type of machine learning, both deep function learning and deep representation learning are focused on inferring the underlying distributions of data. In contrast to representation learning, deep function learning narrows down the model architecture to a sequence of concrete mathematical functions defined on both data and parameter spaces. The {\our} based deep function learning model also disentangles data from parameters and aims to infer and learn these compositional functions, each bearing a concrete physical interpretation for mathematical projections between various data and parameter domains. In contrast, existing deep representation learning models inextricably mix data and parameters together, rendering model interpretability virtually impossible. Below, we will further illustrate the differences of {\our} with several existing base models. 

%--------------------------------------------------------------------------

\subsection{{\our} vs Other Base Models}

%------------------------------------
\begin{figure*}[t]
    \begin{minipage}{\textwidth}
    \centering
    	\includegraphics[width=1.0\linewidth]{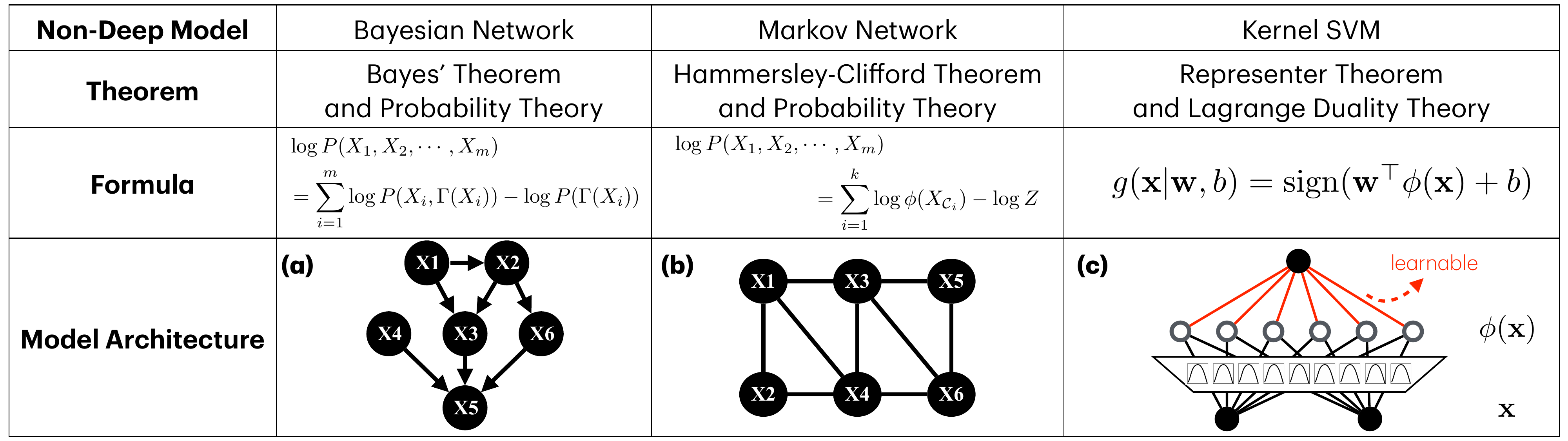}\\
	\vspace{-3pt}
	\hdashrule[0.5ex][x]{\linewidth}{1pt}{1.5mm}\\
	%\rule{\textwidth}{0.05cm}\\
    	\includegraphics[width=1.0\linewidth]{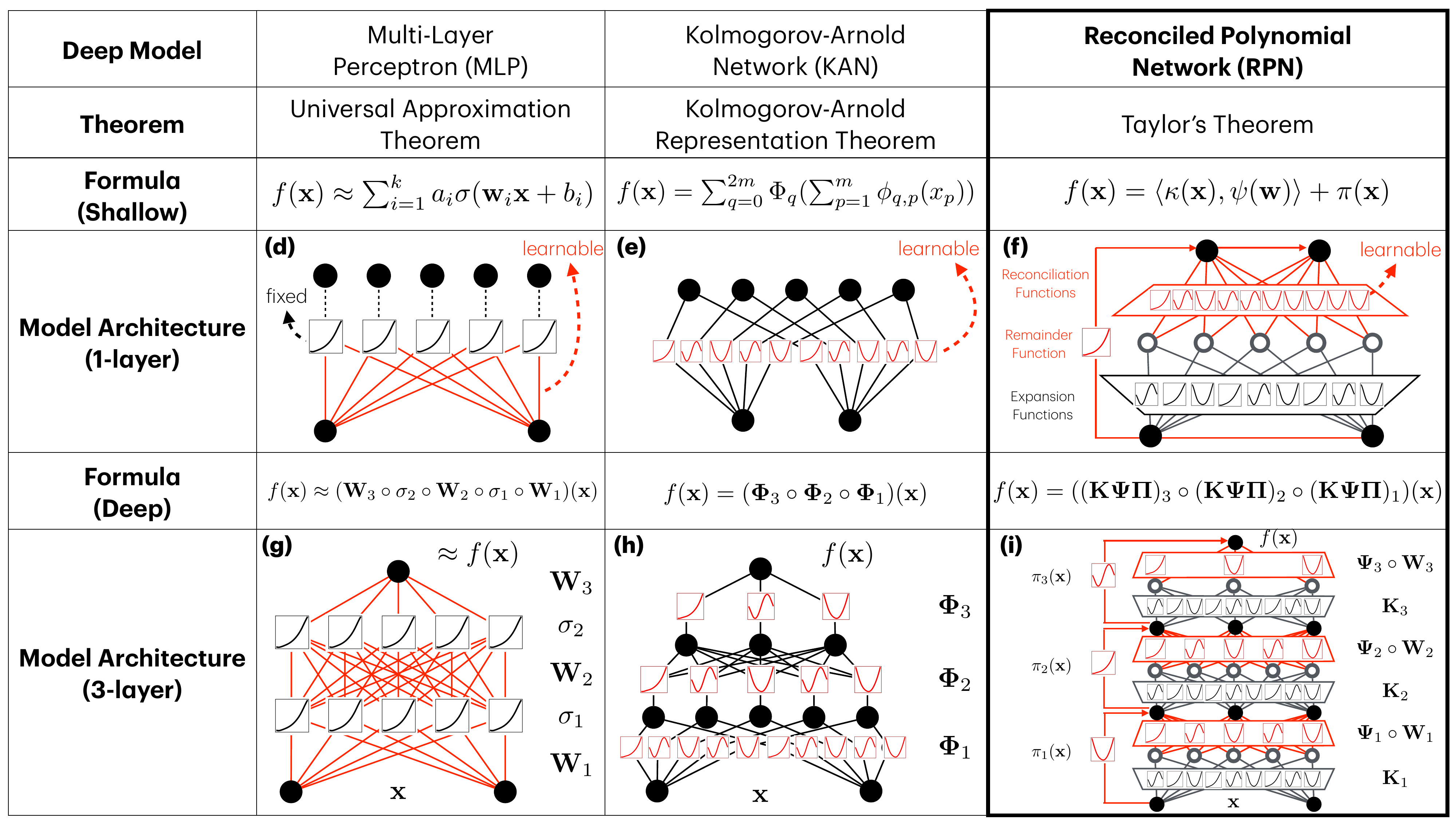}
    	\caption{A comparison of {\our} with Bayesian Network, Markov Network, Kernel SVM, MLP and KAN in terms of mathematical theorem foundation, formula and model architecture. In the plots, we represent the learnable parameters and functions in the red color, while the unlearnable/fixed ones are represented in the dark/gray colors instead. The inputs and outputs are represented with the solid circles, while the expansions are represented as the hollow circles instead.}
    	\label{fig:comparison}
    \end{minipage}%
\end{figure*}
%------------------------------------

Figure~\ref{fig:comparison} compares the {\our} model proposed for deep function learning with several base models in terms of mathematical theorem foundations, formula representations, and model architectures. The top three Plots (a)-(c) describe the non-deep base models: Bayesian Networks, Markov Networks, and Kernel SVMs; while the Plots (d)-(i) at the bottom illustrate the architectures of deep base models: MLPs, KANs, and {\our}. For MLPs and KANs, Plots (d)-(e) and (g)-(h) illustrate their two-layer and three-layer architectures, respectively. Similarly, for {\our}, we present its one-layer and three-layer architectures in Plots (f) and (i).

Based on the plots shown in Figure~\ref{fig:comparison}, we can observe significant differences of {\our} compared against these base models, which are summarized as follows:

\begin{itemize}

\item \textbf{{\our} vs Non-Deep Base Models}: Examining the model plots, we observe that all these base model architectures can be represented as graph structures composed of variables and their relationships. The model architecture of the Markov network is undirected, while that of the Bayesian network is directed. Similarly, for MLP, KAN, and {\our}, although we haven't shown the variable connection directions, their model architecture are also directed, flowing from bottom to top. The model architectures of both Markov network and Bayesian network consist of variable nodes that correspond only to input features and output labels. In contrast, for kernel SVM, MLP, KAN, and {\our}, their model architectures involve not only nodes representing inputs and outputs, but also those representing expansions and hidden layers. Both {\our} and kernel SVM involve a data expansion function to project input data into a high-dimensional space. However, their approaches diverge thereafter. Kernel SVM directly defines parameters within this high-dimensional space to integrate expansion vectors into outputs. In contrast, {\our} fabricates these high-dimensional parameters via a reconciliation function from a reduced set of parameters instead. 

\item \textbf{{\our} vs Deep Base Models}: The difference between {\our} and MLP is easy to observe. MLP involves neither data expansion nor parameter reconciliation. Instead, they apply activation functions to neurons after input integration, which is parameterized by neuron connection weights. Unlike MLPs with predefined, static activation functions, the recent KAN model proposes to learn the activation functions attached to neuron-neuron connections, with their outputs being directly summed together. {\our}, in contrast, integrates the strengths of both kernel SVM and KAN: it employs the expansion function from kernel SVM and adopts learnable functions attached to neuron-to-neuron connections, similar to KANs. Meanwhile, different from kernel SVM, MLP and KAN, {\our} introduces a novel approach to fabriate a large number of parameters from a small set. This technique helps address both the ``curse of dimension'' and the model generalization problems. We have briefly mentioned the ``curse of dimension'' problem before already, and will discuss about the model generalization issue later in Section~\ref{sec:interpretation} from the VC-theory perspective.

\end{itemize}

Here, we briefly compare these base models with {\our}. In the following Section~\ref{sec:backbone_unification}, after we introduce the {\our} model architecture and the component functions, we will further discuss how to unify these base models into {\our}'s canonical representations. More comprehensive information about these base models and other related work will also be provided in Section~\ref{sec:related_work}.

%--------------------------------------------------------------------------
%----------------------------------------------------------------------------------------------------
%----------------------------------------------------------------------------------------------------

\section{Notations and Background Knowledge on Taylor's Theorem}\label{sec:formulation}

This section first introduces the notation system used throughout this paper. Based on the notations, we then briefly present Taylor's theorem as the preliminary knowledge of the {\our} model, which will be introduced in the following Section~\ref{sec:method}.

%----------------------------------------------------------------------------------------------------
\subsection{Notation System}

In the sequel of this paper, we will use the lower case letters (e.g., $x$) to represent scalars, lower case bold letters (e.g., $\mathbf{x}$) to denote column vectors, bold-face upper case letters (e.g., $\mathbf{X}$) to denote matrices and high-order tensors, and upper case calligraphic letters (e.g., $\mathcal{X}$) to denote sets. Given a matrix $\mathbf{X}$, we denote $\mathbf{X}(i,:)$ and $\mathbf{X}(:,j)$ as its $i_{th}$ row and $j_{th}$ column, respectively. The ($i_{th}$, $j_{th}$) entry of matrix $\mathbf{X}$ can be denoted as $\mathbf{X}(i,j)$. We use $\mathbf{X}^\top$ and $\mathbf{x}^\top$ to represent the transpose of matrix $\mathbf{X}$ and vector $\mathbf{x}$. For vector $\mathbf{x}$, we represent its $L_p$-norm as $\left\| \mathbf{x} \right\|_p = \left( \sum_i |\mathbf{x}(i)|^p \right)^{\frac{1}{p}}$. The Frobenius-norm of matrix $\mathbf{X}$ is represented as $\left\| \mathbf{X} \right\|_F = \left( \sum_{i,j} |\mathbf{X}(i,j)|^2 \right)^{\frac{1}{2}}$. The elementwise product of vectors $\mathbf{x}$ and $\mathbf{y}$ of the same dimension is represented as $\mathbf{x} \odot \mathbf{y}$, their inner product is represented as $\left\langle \mb{x}, \mb{y} \right\rangle$, and their Kronecker product is $\mb{x} \otimes \mb{y}$. The elementwise product and Kronecker product operators can also be applied to matrices $\mb{X}$ and $\mb{Y}$ as $\mathbf{X} \odot \mathbf{Y}$ and $\mb{X} \otimes \mb{Y}$, respectively.

\subsection{Taylor's Theorem for Univariate Function Approximation}

Taylor's theorem approximates a $d$-times differentiable function around a given point using polynomials up to degree $d$, commonly referred to as the $d_{\text{th}}$-order Taylor polynomial. In this section, we first introduce Taylor's theorem for univariate functions, which also generalizes to multivariate and vector valued functions. We will briefly describe its extension to multivariate functions in the subsequent Subsection~\ref{subsec:multivariate_taylor}, and then introduce the {\our} model designed based on Taylor's theorem with vector valued functions in Section~\ref{sec:method}. 

\begin{theorem}\label{theo:taylor_theorem}
(Taylor's Theorem): Let $d \ge 1$ be an integer and let function $f: \mathbbm{R} \to \mathbbm{R}$ be $d$ times differentiable at the point $a \in \mathbbm{R}$. As illustrated in Figure~\ref{fig:taylor_example}, then there exists a function $h_d: \mathbbm{R} \to \mathbbm{R}$ such that
\begin{equation}\label{equ:taylor_theorem}
\begin{aligned}
f(x) &= \frac{f(a)}{0!} (x-a)^0 + \frac{f'(a)}{1!} (x-a)^1 + \frac{f''(a)}{2!} (x-a)^2 + \cdots + \frac{f^{(d)}(a)}{d!} (x-a)^d + R_d(x),\\
&= \sum_{i=0}^d \frac{f^{(i)}(a)}{i!} (x-a)^i + R_d(x).
\end{aligned}
\end{equation}
In the equation, $R_d(x)$ is also normally called the ``remainder'' term and can be represented as
\begin{equation}\label{equ:peano_remainder}
R_d(x) = h_d(x) (x-a)^d \text{ , where }\lim_{x \to a} h_d (x) = 0.
\end{equation}
\end{theorem}

%------------------------------------
\begin{figure*}[t]
\vspace{1em}
    \begin{minipage}{\textwidth}
    \centering
    	\includegraphics[width=0.6\linewidth]{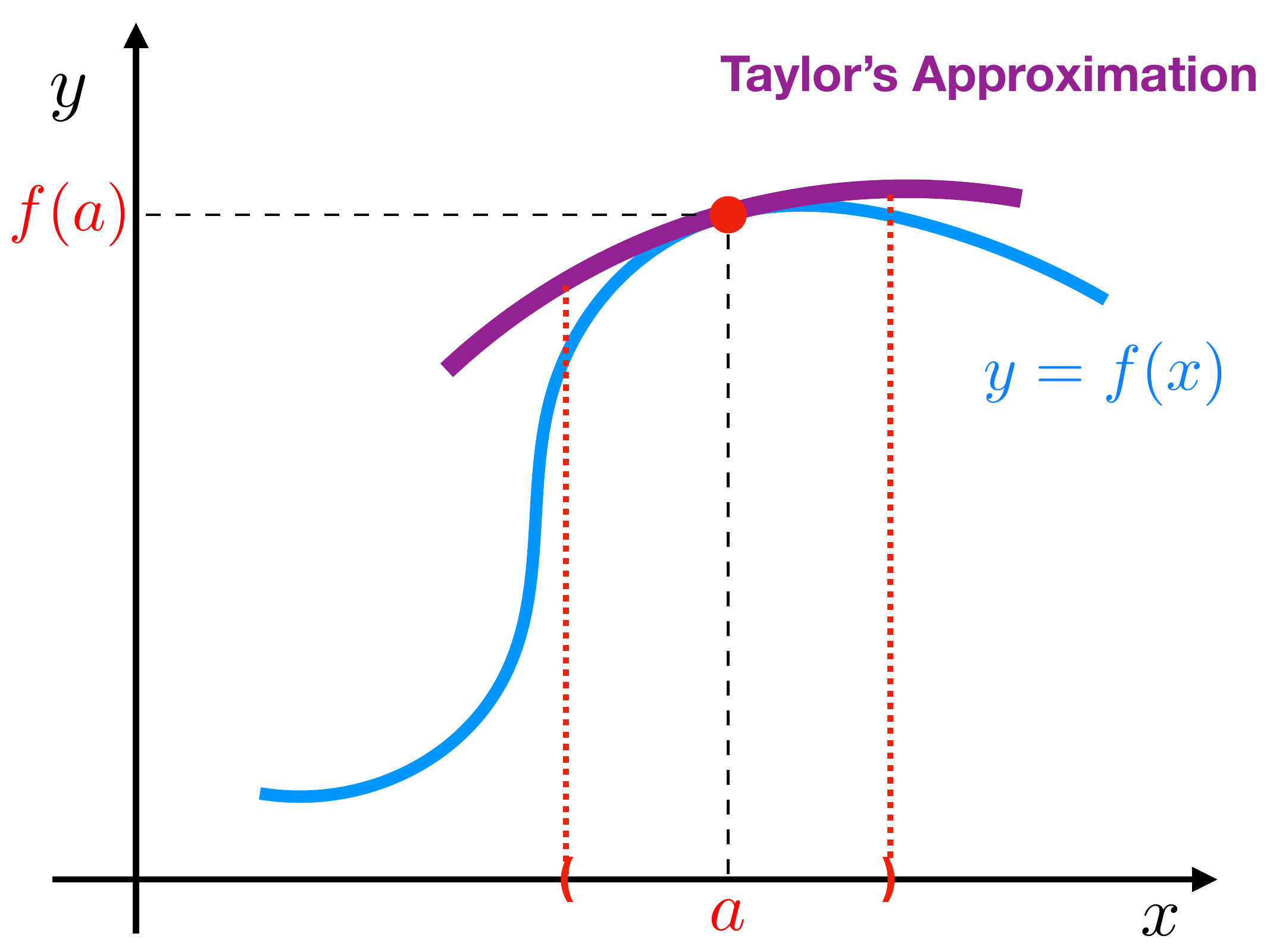}
    	\caption{An illustration of Taylor's approximation of continuous functions.}
    	\label{fig:taylor_example}
    \end{minipage}%
\end{figure*}
%------------------------------------

According to the above description of the Taylor's Theorem, the function output $f(x)$ can be represent as a summation of polynomials of high degrees of the $(x-a)$. What's more, in this paper, we propose to further disentangle the variable $x$ from the given constant point $a$. Terms like $(x-a)^k$ can be decomposed into summations of polynomials in $x$ alone, with $a$ serving as the coefficients:
\begin{equation}
\begin{aligned}
(x-a)^d &= {d \choose 0} (-a)^{d-0} x^0 + {d \choose 1} (-a)^{d-1} x^1 + \cdots +  {d \choose d} (-a)^{d-d} x^{d}.
%&= \sum_{i=0}^k {k \choose i} (-a)^{k-i} x^i.
\end{aligned}
\end{equation}

Based on the decomposition, we can rewrite the above Equation~(\ref{equ:taylor_theorem}) as follows:

\begin{equation}\label{equ:new_taylor_theorem}
%\tcbhighmath[fuzzy halo=1mm with blue!50!white,arc=2pt, boxrule=0pt,frame hidden]
{
\begin{aligned}
f(x | a) &= \left\langle {\mb{x}}, \mb{c} \right\rangle + R_d(x),
%&= \mb{x} \mb{c}^\top + R_k(x),
\end{aligned}
}
\end{equation}

where $\left\langle \cdot, \cdot \right\rangle$ denotes the inner product operator. The expanded data vector $\mb{x} = \kappa(x) = [x^0, x^1, x^2, \cdots, x^d] \in \mathbbm{R}^{d+1}$ contains the high-order polynomials of $x$, where the created coefficient vector $\mb{c} = \psi(a) = [c_0, c_1, c_2, \cdots, c_d] \in \mathbbm{R}^{d+1}$ has the same dimension as $\mb{x}$. Each coefficient term, such as $c_i$ (where $i \in \{0, 1, \cdots, d\})$, is fabricated with $a$ as follows:
\begin{equation}
c_i = \sum_{j=i}^d \frac{f^{(j)}(a)}{j!} {j \choose i} (-a)^{j-i}.
\end{equation}

The remainder term measure the error in approximating $f$ with Taylor's polynomials. The representation illustrated in Equation~(\ref{equ:peano_remainder}) above is known as the ``Peano Remainder''. In addition, mathematicians have introduced many different forms of remainder representations, some of which are listed below:\\

%--------------------------------
\noindent\begin{minipage}{.42\linewidth}
\begin{fleqn}
%\tcbhighmath[fuzzy halo=1mm with blue!50!white,arc=2pt, boxrule=0pt,frame hidden]
{
\begin{equation}
\begin{aligned}
&\underline{\textbf{(a) Peano Remainder:}}\\[3pt]
&R_d(x) = h_d(x) (x-a)^d,\\[5pt]
&\text{where } \lim_{x \to a} h_d (x) = 0.
\end{aligned}
\end{equation}
}
\end{fleqn}
\end{minipage}
\hfill
\noindent\begin{minipage}{.48\linewidth}
\begin{fleqn}
%\tcbhighmath[fuzzy halo=1mm with blue!50!white,arc=2pt, boxrule=0pt,frame hidden]
{
\begin{equation}
\begin{aligned}
&\underline{\textbf{(b) Lagrange Remainder:}}\\
&R_d(x) = \frac{f^{(d+1)}(\xi)}{(d+1)!}{(x-a)^{d+1}},\\
&\text{for some } \xi \text{ between $a$ and $x$}.
\end{aligned}
\end{equation}
}
\end{fleqn}
\end{minipage}

\noindent\begin{minipage}{.42\linewidth}
\begin{fleqn}
%\tcbhighmath[fuzzy halo=1mm with blue!50!white,arc=2pt, boxrule=0pt,frame hidden]
{
\begin{equation}
\begin{aligned}
&\underline{\textbf{(c) Cauchy Remainder:}}\\
&R_d(x) = \frac{f^{(d+1)}(\xi)}{d!}{(x-\xi)^{d}(x-a)},\\
&\text{for some } \xi \text{ between $a$ and $x$}.
\end{aligned}
\end{equation}
}
\end{fleqn}
\end{minipage}
\hfill
\noindent\begin{minipage}{.48\linewidth}
\begin{fleqn}
%\tcbhighmath[fuzzy halo=1mm with blue!50!white,arc=2pt, boxrule=0pt,frame hidden]
{
\begin{equation}
\begin{aligned}
&\underline{\textbf{(d) Schl\"omilch Remainder:}}\\
&R_d(x) = \frac{f^{(d+1)}(\xi)}{d!}{(x-\xi)^{d+1-p}\frac{(x-a)^p}{p}},\\
&\text{for some } p>0 \text{ and } \xi \text{ between $a$ and $x$}.
\end{aligned}
\end{equation}
}
\end{fleqn}
\end{minipage}
%--------------------------------

%----------------------------------------------------------------------------------------------------

\subsection{Taylor's Theorem for Multivariate Function Approximation}\label{subsec:multivariate_taylor}

Representing multivariate continuous functions with Taylor's polynomials is more intricate. In this part, we use a multivariate function $f: \mathbbm{R}^m \to \mathbbm{R}$ as an example to illustrate how to disentangle the input variables and function parameters via Taylor's formula. Similar as the above single-variable function, assuming function $f$ is $d_{th}$-time continuously differentiable at point $\mb{a} \in \mathbbm{R}^m$, then for the inputs near the point can be approximated as

\begin{equation}\label{equ:multivariate_taylor_1}
f(\mb{x}) = \sum_{|\alpha| \le d} \frac{D^{|\alpha|} f(\mb{a})}{\alpha!} (\mb{x} - \mb{a})^{\alpha} + R_d(\mb{x}).
\end{equation}

The notation $D^{|\alpha|} f$ denotes the $|\alpha|_{th}$ partial derivatives of function $f$ and $R_d(\mb{x})$ denotes the remainder term:
\begin{equation}\label{equ:multivariate_taylor_2}
\begin{cases}
D^{|\alpha|} f &= \frac{\partial^{|\alpha|} f}{\partial x_{1}^{\alpha_1} \cdots \partial x_{m}^{\alpha_m}}\\
R_d(\mb{x}) &= \sum_{|\alpha| = d} h_{\alpha}(\mb{x}) (\mb{x} - \mb{a})^{\alpha}
\end{cases}, \text{ where }
\begin{cases}
&|\alpha| \ \ = \alpha_1 + \alpha_2 + \cdots + \alpha_m;\\
&\alpha! \ \ = \alpha_1 !  \alpha_2!  \cdots  \alpha_m !;\\
&(\mb{x} - \mb{a})^{\alpha} \ \ = (x_1-a_1)^{\alpha_1} \cdots (x_m-a_m)^{\alpha_m};\\
&\lim_{\mb{x} \to \mb{a}}  h_{\alpha}(\mb{x}) \ \  = 0.
\end{cases}
\end{equation}

Similar to the single-variable case, the variables $\mb{x} = [x_1, x_2, \cdots, x_m]$ involved in the multivariate polynomials can also be decoupled from the data point $\mb{a} = [a_1, a_2, \cdots, a_m]$, leading to the following representation:

\begin{equation}\label{equ:multivariate_taylor_expansion}
f(\mb{x} | \mb{a}) = \left\langle \bar{\mb{x}}, {\mb{c}} \right\rangle + R_d(\mb{x}),
\end{equation}

where $\bar{\mb{x}}$ denotes the data expansion vector, and $\mb{c}$ represents the created coefficient vector. Their detailed representations are provided as follows.

\begin{itemize}
\item \textit{Data Expansion}: Function $\kappa: \mathbbm{R}^m \to \mathbbm{R}^{D}$ will expand the input vector $\mb{x} = [x_1, x_2, \cdots, x_m]^\top$ to $\bar{\mb{x}} \in \mathbbm{R}^{D}$ as follows:
\begin{equation}
\bar{\mb{x}} = \kappa(\mb{x}) = [\underbrace{1_{_{_{_{}}}}}_{\text{1 term of order 0}}, \underbrace{x_1, x_2, \cdots, {x_m} _{_{_{}}}}_{\text{$m$ terms of order 1}}, \underbrace{x_1^2, x_1x_2, \cdots, {x_m^2}_{}}_{\text{$m^2$ terms of order 2}}, \underbrace{\cdots_{_{}}}_{\cdots}, \underbrace{\cdots, x_m^d}_{\text{$m^d$ terms of order $d$}}]^\top,
\end{equation}
where the expansion output vector has a dimension of $D = \sum_{i=0}^d m^i$.

\item \textit{Parameter Fabrication}: Function $\psi: \mathbbm{R}^m \to \mathbbm{R}^D$ will fabricate the constant $\mb{a}$ to a coefficient vector of dimension $D$ as follows:
\begin{equation}
{\mb{c}} = \psi(\mb{a}) = [\underbrace{c_0}_{\text{coeff. of constant}}, \underbrace{c_1, c_2, \cdots, c_m}_{\text{coeff. of terms with order 1}}, \underbrace{c_{1,1}, c_{1,2}, \cdots, c_{m,m}}_{\text{coeff. of terms with order 2} }, \underbrace{\cdots}_{\cdots}, \underbrace{\cdots, c_{m,m,\cdots, m}}_{\text{coeff. of terms with order d} }]^\top.
\end{equation}
The coefficient vector $\mb{c}$ has the same dimension as $\bar{\mb{x}}$, and the coefficient $c_{i_1,i_2, \cdots, i_k}$ corresponds to the polynomial term $x_{i_1} x_{i_2} \cdots x_{i_k}$. As to the specific representation of $c_{i_1,i_2, \cdots, i_k}$, it can be obtained by decomposing the above Equations~(\ref{equ:multivariate_taylor_1})-(\ref{equ:multivariate_taylor_2}).

\item \textit{Lagrange Remainder}: The remainder $R_d(\mb{x})$ will include all the terms with order higher than $d$, which can reduce the approximation errors.
\end{itemize}

%----------------------------------------------------------------------------------------------------

\subsection{Taylor's Theorem based Machine Learning Models}\label{subsec:taylor_theorem_machine_learning}

In real-world problems, the underlying functional mappings are often more intricate, such as $f: \mathbbm{R}^m \to \mathbbm{R}^n$ with multiple input variables and multiple outputs. Representing these functions with Taylor's polynomials requires more cumbersome derivations, and the coefficient fabrication outputs should be a two-dimensional matrix, such as $\psi(\mb{a}) \in \mathbbm{R}^{n \times D}$. To avoid getting bogged down in unnecessary mathematical details, we will not repeat those derivations here. 

In recent years, there has been a growing interest in designing machine learning and deep learning models based on Taylor's theorem. For binary data inputs, Zhang {\etal} \cite{Zhang2018ReconciledPM} introduce the reconciled polynomial machine to unify shallow and deep learning models, which is also the prior work that this paper is based on. Balduzzi {\etal} \cite{Balduzzi2016NeuralTA} investigate the convergence and exploration in rectifier networks with neural Taylor approximations, while Chrysos {\etal} \cite{Chrysos2020DeepPN} propose a new class of function approximation method based on polynomial expansions. Zhao {\etal} \cite{zhao2023taylornet} propose a generic neural architecture TaylorNet based on tensor decomposition to initialize the models, and Nivron {\etal} \cite{Nivron2023TaylorformerPP} introduce to incorporate use Taylor's expansion as a wrapper of transformer for the probabilistic predictions for time series and other random processes. Beyond time series and continuous function approximation, Taylor's expansion has found applications in reinforcement learning and computer vision. \cite{Tang2020TaylorEP} investigates the application of Taylor's expansions in reinforcement learning and introduces the Taylor expansion policy optimization to generalize prior work; and \cite{Pfrommer2022TaSILTS} introduces a simple augmentation to standard behavior cloning losses in the context of continuous control for Taylor series imitation learning. In image processing, \cite{Zhou2021UnfoldingTA} proposes to use Taylor’s formula to construct a novel framework for image restoration, and \cite{10070581} proposes the Taylor neural net for image super-resolution.

Different these prior work, since the underlying function $f$ is unknown, we cannot directly employ the above derivations and Taylor's expansions to define approximated polynomial representations of $f$ for practical applications. Drawing inspiration from the approximation architecture delineated in Equation~\ref{equ:new_taylor_theorem} and Equation~\ref{equ:multivariate_taylor_expansion}, we propose a novel approach that defines distinct component functions to substitute the data vector, coefficient vector, and remainder terms. Additionally, as the input variable $\mb{x}$ varies across instances, instead of manually selecting one single fixed constant $\mb{a}$, we propose to define it as multi-channel parameters and learn them instead. These innovations form the foundation of our proposed Reconciled Polynomial Network ({our}) model to be introduced in the following section.

%--------------------------------------------------------------------------
%----------------------------------------------------------------------------------------------------
%----------------------------------------------------------------------------------------------------

\section{{\our}: Reconciled Polynomial Network for Deep Function Learning}\label{sec:method}

Based on the preliminary background introduced above and inspired by the work of \cite{Zhang2018ReconciledPM}, we will introduce the Reconciled Polynomial Network ({\our}) model for function learning in this section.

%---------------------------------------------------

\subsection{{\our}: Reconciled Polynomial Network}

Formally, given the underlying data distribution mapping $f: \mathbbm{R}^m \to \mathbbm{R}^n$, we represent the {\our} model proposed to approximate function $f$ as follows:

\begin{equation}\label{equ:rpn_layer}
%\tcbhighmath[fuzzy halo=1mm with blue!50!white,arc=2pt, boxrule=0pt,frame hidden]
{
g(\mb{x} | \mb{w}) = \left\langle \kappa(\mb{x}), \psi(\mb{w}) \right\rangle + \pi(\mb{x}),
}
\end{equation}
where
\begin{itemize}
\item $\kappa: \mathbbm{R}^m \to \mathbbm{R}^{D}$ is named as the \textbf{data expansion function} and $D$ is the target expansion space dimension.
\item $\psi: \mathbbm{R}^l \to \mathbbm{R}^{n \times D}$ is named as the \textbf{parameter reconciliation function}, which is defined only on the parameters without any input data.
\item $\pi: \mathbbm{R}^m \to \mathbbm{R}^n$ is named as the \textbf{remainder function}. 
\end{itemize}

The architecture of {\our} is also illustrated in Figure~\ref{fig:architecture}. The {\our} model disentangles input data from model parameters through the expansion functions $\kappa$ and reconciliation function $\psi$. More detailed information about all these components and modules mentioned in Figure~\ref{fig:architecture} will be introduced in the following parts of this section.

\subsection{{\our} Component Functions}

The \textbf{data expansion function} $\kappa$ projects input data into a new space with different basis vectors, where the target vector space dimension $D$ is determined when defining $\kappa$. In practice, the function $\kappa$ can either expand or compress the input to a higher- or lower-dimensional space. The corresponding function, $\kappa$, can also be referred to as the {data expansion function} (if $D > m$) and {data compression function} (if $D < m$), respectively. Collectively, these can be unified under the term ``\textbf{data transformation functions}''. In this paper, we focus on expanding the inputs to a higher-dimensional space, and will use the function names ``data transformation'' and ``data expansion'' interchangeably in the following sections. 

Meanwhile, the \textbf{parameter reconciliation function} $\psi$ adjusts the available parameter vector of length $l$ by fabricating a new parameter matrix of size $n \times D$ to accommodate the expansion space dimension $D$ defined by function $\kappa$. In most of the cases studied in this paper, the parameter vector length $l$ is much smaller than the output matrix size $n \times D$, {\ie} $l \ll n \times D$. Meanwhile, in practice, we can also define function $\psi$ to fabricate a longer parameter vector into a smaller parameter matrix, {\ie} $l > n \times D$. To unify these different cases, the data reconciliation function can also be referred to as the ``\textbf{parameter fabrication function}'', and these function names will be used interchangeably in this paper.

Without specific descriptions, the \textbf{remainder function} $\pi$ defined here is based solely on the input data $\mb{x}$. However, in practice, we also allow $\pi$ to include learnable parameters for output dimension adjustment. In such cases, it should be rewritten as $\pi(\mb{x} | \mb{w}')$, where $\mb{w}'$ is one extra fraction of the model's learnable parameters. Together with the parameter vector $\mb{w}$ ({\ie} the input to the parameter reconciliation function $\psi$), they will define the complete set of learnable parameters for the model.

%---------------------------------------------------

%------------------------------------
\begin{figure*}[t]
    \begin{minipage}{\textwidth}
    \centering
    	\includegraphics[width=1.0\linewidth]{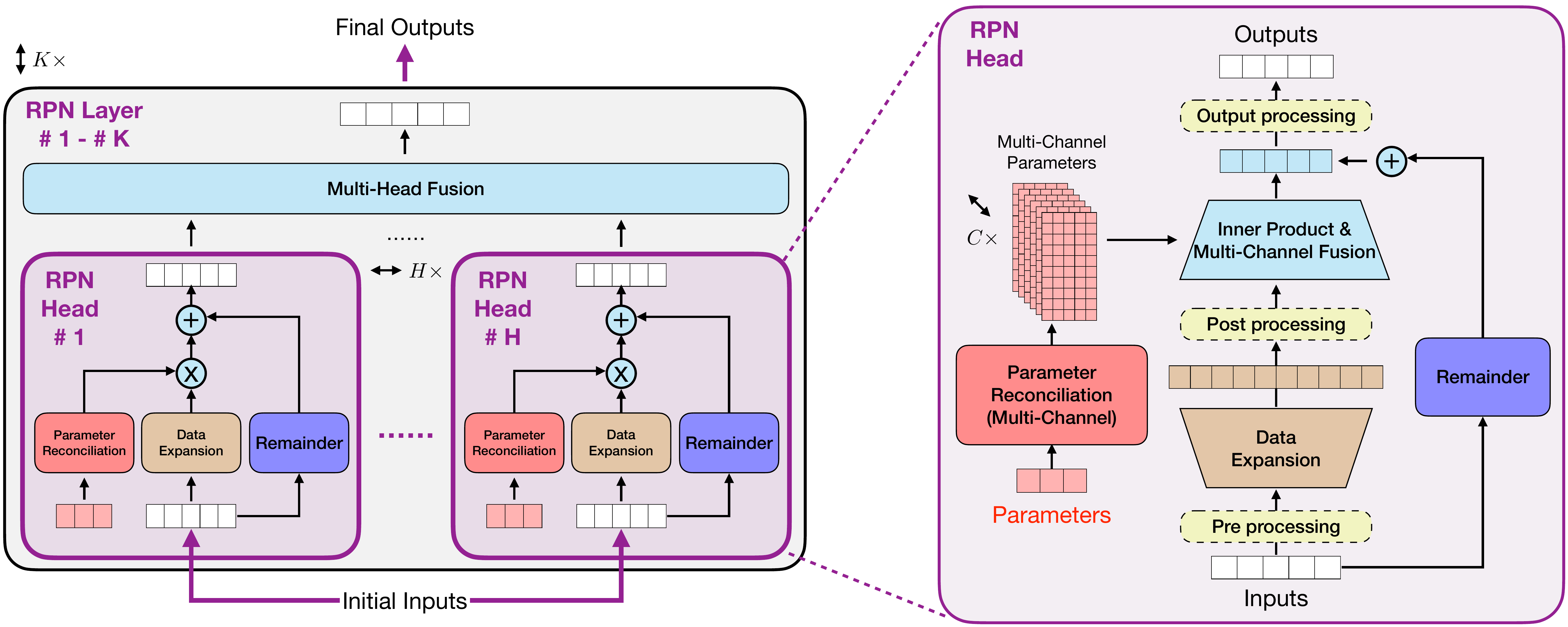}
    	\caption{An illustration of the {\our} framework. The left plot illustrates the multi-layer ($K$-layer) architecture of {\our}. Each layer involves multi-head for function learning, whose outputs will be fused together. The right plot illustrates the detailed architecture of the {\our} head, involving data expansion, multi-channel parameter reconciliation, remainder, and their internal operations. The components with yellow color in dashed lines denote the optional data processing functions ({\eg} activation functions and norm functions) for the inputs, expansions and outputs.}
    	\label{fig:architecture}
    \end{minipage}%
\end{figure*}
%------------------------------------

\subsection{Wide {\our}: Multi-Head and Multi-Channel Model Architecture}

Similar to the Transformer with multi-head attention \cite{Vaswani2017AttentionIA}, as shown in Figure~\ref{fig:architecture}, the {\our} model employs a multi-head architecture, where each head can disentangle the input data and model parameters using different expansion, reconciliation and remainder functions, respectively:

\begin{equation}
g(\mb{x} | \mb{w}, H) = \sum_{h=0}^{H-1} \left\langle \kappa^{(h)}(\mb{x}), \psi^{(h)}(\mb{w}^{(h)}) \right\rangle + \pi^{(h)}(\mb{x}),
\end{equation}

where the superscript ``$h$'' indicates the head index and $H$ denotes the total head number. By default, we use summation to combine the results from all these heads.

Moreover, in the {\our} model shown in Figure~\ref{fig:architecture}, similar to convolutional neural networks (CNNs) employing multiple filters, we allow each head to have multiple channels of parameters applied to the same data expansion. For example, for the $h_{th}$ head, we define its multi-channel parameters as $\mb{w}^{(h),0}, \mb{w}^{(h),1}, \cdots, \mb{w}^{(h), C-1}$, where $C$ denotes the number of channels. These parameters will be reconciled using the same parameter reconciliation function, as shown below:

\begin{equation}
g(\mb{x} | \mb{w}, H, C) = \sum_{h=0}^{H-1} \sum_{c=0}^{C-1} \left\langle \kappa^{(h)}(\mb{x}), \psi^{(h)}(\mb{w}^{(h), c}) \right\rangle + \pi^{(h)}(\mb{x}),
\end{equation}

The multi-head, multi-channel design of the {\our} model allows it to project the same input data into multiple different high-dimensional spaces simultaneously. Each head and channel combination may potentially learn unique features from the data. The unique parameters at different heads can have different initialized lengths, and each of them will be processed in unique ways to accommodate the expanded data. This multi-channel approach provides our model with more flexibility in model design. In the following parts of this paper, to simplify the notations, we will illustrate the model's functional components using a single-head, single-channel architecture by default. However, readers should note that these components to be introduced below can be extended to their multi-head, multi-channel designs in practical implementations.

%---------------------------------------------------

\subsection{Deep {\our}: Multi-Layer Model Architecture}

The wide model architecture introduced above provides {\our} with greater capabilities for approximating functions with diverse expansions concurrently. However, such shallow architectures can be insufficient for modeling complex functions. In this paper, as illustrated in Figure~\ref{fig:architecture}, we propose to stack {\our} layers on top of each other to build a deeper architecture, where the Equation~(\ref{equ:rpn_layer}) actually defines one single layer of the model. Formally, we can represent the deep {\our} with multi-layers as follows:

\begin{equation}\label{equ:deep_rpn}
\begin{cases}
\text{Input: } & \mb{h}_0  = \mb{x},\\
\text{Layer 1: } & \mb{h}_1 = \left\langle \kappa_1(\mb{h}_0), \psi_1(\mb{w}_1) \right\rangle + \pi_1(\mb{h}_0),\\
\text{Layer 2: } & \mb{h}_2 = \left\langle \kappa_2(\mb{h}_1), \psi_2(\mb{w}_2) \right\rangle + \pi_2(\mb{h}_1),\\
\cdots & \cdots \ \cdots\\
\text{Layer K: } & \mb{h}_K = \left\langle \kappa_K(\mb{h}_{K-1}), \psi_K(\mb{w}_K) \right\rangle + \pi_K(\mb{h}_{K-1}),\\
\text{Output: } & \hat{\mb{y}}  = \mb{h}_K.
\end{cases}
\end{equation}

The subscripts used above denote the layer index. The dimensions of the outputs at each layer can be represented as a list $[d_0, d_1, \cdots, d_{K-1}, d_K]$, where $d_0 = m$ and $d_K = n$ denote the input and the desired output dimensions, respectively. Therefore, if the component functions at each layer of our model have been predetermined, we can just use the dimension list $[d_0, d_1, \cdots, d_{K-1}, d_K]$ to represent the architecture of the {\our} model. 

%---------------------------------------------------

\subsection{Versatile {\our}: Nested and Extended Expansion Functions}

%------------------------------------
\begin{figure*}[t]
    \begin{minipage}{\textwidth}
    \centering
    	\includegraphics[width=1.0\linewidth]{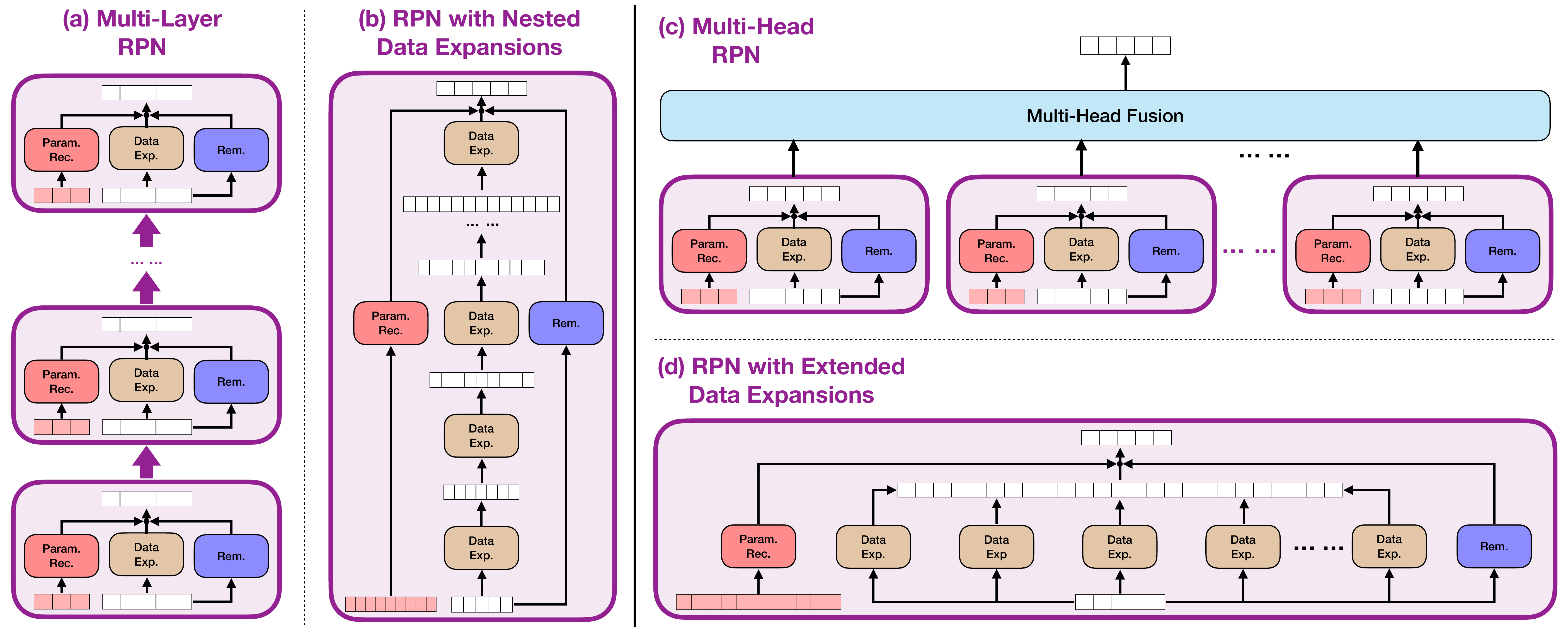}
    	\caption{An illustration of the {\our} layer with nested and extended data expansions. Plot (a): multi-layer {\our}; Plot (b): single-layer {\our} with nested data expansions; Plot (c): multi-head {\our}; Plot (d): single-head {\our} with extended data expansions.}
    	\label{fig:nested_extended_expansions}
    \end{minipage}%
\end{figure*}
%------------------------------------

The data expansion function introduced earlier projects the input data to a higher-dimensional space. There exist different ways to define the data expansion function, and a list of such basic expansion functions will be introduced in the following Section~\ref{subsec:data_expansion_function}. The multi-head, multi-channel and multi-layer architecture also provides {\our} with more capacity to build wider and deeper architectures for projecting input data to the desired target space. In addition to these designs, as illustrated in Figure~\ref{fig:nested_extended_expansions}, {\our} also provides a more flexible and lightweight mechanism to build models with similar capacities via the nested and extended data expansion functions.

\noindent \textbf{Nested expansions}: Formally, given a list of $n$ data expansion functions $\kappa_1: \mathbbm{R}^{d_0} \to \mathbbm{R}^{d_1}$,  $\kappa_2: \mathbbm{R}^{d_1} \to \mathbbm{R}^{d_2}$, $\cdots$, $\kappa_n: \mathbbm{R}^{d_{n-1}} \to \mathbbm{R}^{d_{n}}$, as shown in Plots (a)-(b) of Figure~\ref{fig:nested_extended_expansions}, the nested calls of these functions will project a data vector from the input space $\mathbbm{R}^{d_0}$ to the desired output space $\mathbbm{R}^{d_{n}}$, defining the nested data expansion function $\kappa: \mathbbm{R}^{m} \to \mathbbm{R}^{D}$ as follows:

\begin{equation}
\kappa(\mb{x}) = \kappa_{n} \left( \kappa_{n-1} \left( \cdots \kappa_2 \left( \kappa_{1} \left( \mb{x} \right) \right) \right) \right) \in \mathbbm{R}^{D}.
\end{equation}

where the function input and output dimensions should be $d_0 = m$ and $d_n = D$.

\noindent \textbf{Extended expansions}: In addition to nesting these $n$ expansion functions, as shown in Plots (c)-(d) of Figure~\ref{fig:nested_extended_expansions}, they can also be concatenated and applied concurrently, with their extended outputs allowing the model to leverage multiple expansion functions simultaneously. Formally, we can represent the extended data expansion function $\kappa: \mathbbm{R}^{m} \to \mathbbm{R}^D$ defined based on $\kappa_1: \mathbbm{R}^{m} \to \mathbbm{R}^{d_1}$,  $\kappa_2: \mathbbm{R}^{m} \to \mathbbm{R}^{d_2}$, $\cdots$, $\kappa_n: \mathbbm{R}^{m} \to \mathbbm{R}^{d_{n}}$ as follows:

\begin{equation}
\kappa(\mb{x}) = \left[ \kappa_1\left( \mb{x} \right), \kappa_2\left( \mb{x} \right), \cdots, \kappa_n\left( \mb{x} \right) \right] \in \mathbbm{R}^{D},
\end{equation}

where the extended expansion's output dimension is equal to the sum of the output dimensions from all the individual expansion functions, {\ie} $D = \sum_{i=1}^n d_i$.

As illustrated in Figure~\ref{fig:nested_extended_expansions}, the nested expansion functions can define complex expansions akin to the multi-layer architecture of {\our} mentioned above. Meanwhile, the extended expansion functions can define expansions similar to the multi-head architecture of {\our}. Both nested and extended expansions allow for faster data expansions, circumventing cumbersome parameter inference and remainder function calculation, and can reduce the additional learning costs associated with training deep and wide architectures of our model. This flexibility afforded by nested and extended expansions provides us with greater versatility in designing the {\our} model.

%---------------------------------------------------

\subsection{Learning Correctness of {\our}: Complexity, Capacity and Completeness}

The \textbf{learning correctness} of {\our} is fundamentally determined by the compositions of its component functions, each contributing from different perspectives:
\begin{itemize}
\item \textbf{Model Complexity}: The data expansion function $\kappa$ expands the input data by projecting its representations using basis vectors in the new space. In other words, function $\kappa$ determines the upper bound of the {\our} model's complexity.

\item \textbf{Model Capacity}: The reconciliation function $\psi$ processes the parameters to match the dimensions of the expanded data vectors. The reconciliation function and parameters jointly determine the learning capacity and associated training costs of the {\our} model.

\item \textbf{Model Completeness}: The remainder function $\pi$ completes the approximation as a residual term, governing the learning completeness of the {\our} model.
\end{itemize}

In the following Section~\ref{sec:functions}, we will introduce several different representations for the data expansion function $\kappa$, parameter reconciliation function $\psi$, and remainder function $\pi$.  By strategically combining these component functions, we can construct a multi-head, multi-channel, and multi-layer architecture, enabling {\our} to address a wide spectrum of learning challenges across diverse learning tasks.

%---------------------------------------------------

%---------------------------------------------------

\subsection{Learning Cost of {\our}: Space, Time and Parameter Number}

To analyze the learning costs of {\our}, we can take a batch input $\mb{X} \in \mathbbm{R}^{B \times m}$ of batch size $B$ as an example, which will be fed to the {\our} model with $K$ layers, each with $H$ heads and each head has $C$ channels. Each head will project the data instance from a vector of length $m$ to an expanded vector of length $D$ and then further projected to the desired output of length $n$. Each channel reconciles parameters from length $l$ to the sizes determined by both the expansion space and output space dimensions, {\ie} $n \times D$.

Based on the above hyper-parameters, assuming the input and output dimensions at each layer are comparable to $m$ and $n$, then the space, time costs and the number of involved parameters in learning the {\our} model are calculated as follows:
\begin{itemize}
\item \textbf{Space Cost}: The total space cost for data (including the inputs, expansions and outputs) and parameter (including raw parameters, fabricated parameters generated by the reconciliation function and optional remainder function parameters) can be represented as $\mathcal{O}( K H (\underbrace{B (\underbrace{m}_{\text{input}} + \underbrace{D}_{\text{expansion}} + \underbrace{n}_{\text{output}} ) }_{\text{space cost for data}} + \underbrace{C (\underbrace{l}_{\text{raw param.}} + \underbrace{nD}_{\text{reconciled param.}}) + \underbrace{m n}_{\text{(optional) remainder param.}}}_{\text{space cost for parameters}}))$.

%\item \textbf{Space Cost}: The total space cost for data (including input data vector, expanded data vector and output data vector) can be represented as $\mathcal{O}( B K H (\underbrace{m}_{\text{input dim.}} + \underbrace{D}_{\text{expansion dim.}} + \underbrace{n}_{\text{output dim,}} ) )$. Meanwhile, the space cost for parameters (including raw parameters and fabricated parameters generated by the reconciliation function) can be represented as $\mathcal{O}(K H C (\underbrace{l}_{\text{raw param. dim.}} + \underbrace{nD}_{\text{reconciled param. dim.}}) )$. For the remainder, its optional parameter space cost will be $\mathcal{O}(K H m n)$. 

%space cost for data $\mathcal{O}\left( B K H m \right)$, space cost for expanded data $\mathcal{O}\left( B K H D \right)$, space cost for output data $\mathcal{O}\left( B K H n \right)$; space cost for parameters $\mathcal{O}\left(K H C l \right)$, space cost for reconciled parameters $ \mathcal{O}\left(K H C n D \right)$.

\item \textbf{Time Cost}: Depending on the expansion and reconciliation functions used for building {\our}, the total time cost of {\our} can be represented as $\mathcal{O}( K H (\underbrace{t_{exp}(m, D)}_{\text{time cost for data exp.}} + \underbrace{C t_{rec}(l, D)}_{\text{time cost for param. rec.}} + \underbrace{C m n D}_{\text{time cost for inner product}} + \underbrace{m n}_{\text{(optional) time cost for remainder}}))$, where notations $t_{exp}(m, D)$ and $t_{rec}(l, D)$ denote the expected time costs for data expansion and parameter reconciliation functions, respectively.

%time cost for data expansion $\mathcal{O}\left( K H o(m, D) ) \right)$; time cost for parameter reconciliation $\mathcal{O}\left( K H C o(l, D)\right)$; time cost for inner product $\mathcal{O}\left( K H C m n D) \right)$; time cost for remainder $\mathcal{O}\left( K H m n \right)$

\item \textbf{Learnable parameters}: The total number of parameters in {\our} will be $\mathcal{O}(K H C l + K H m n)$, where $\mathcal{O}( K H m n)$ denotes the optional parameter number used for defining the remainder function.

\end{itemize}

%---------------------------------------------------

%--------------------------------------------------------------------------

%=================================================
%=================================================

%---------------------------------------------------
%\newpage
%------------------------------------------------------------------------------------
%\vspace{2em}
%\hrule
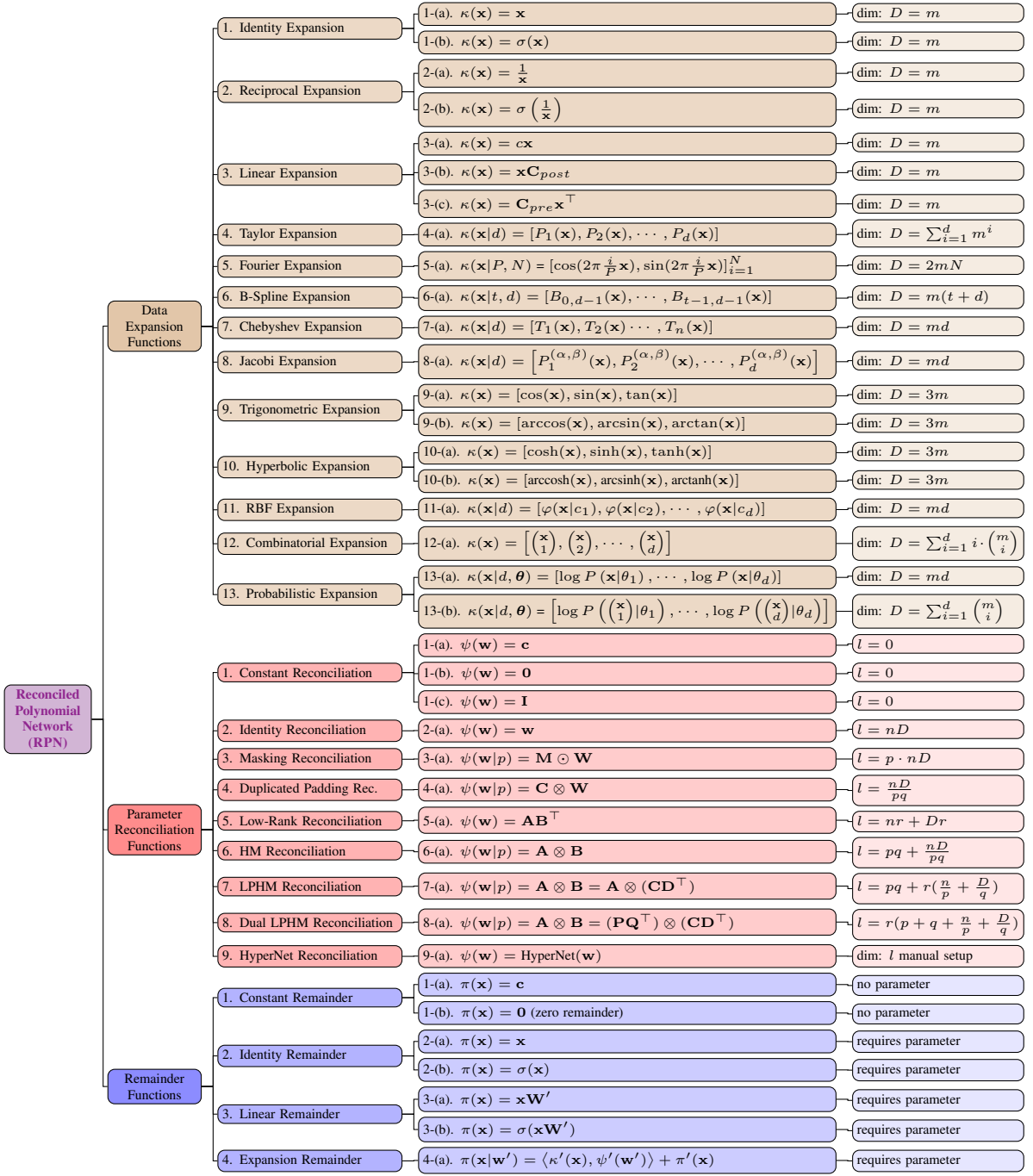
\begin{figure*}[h]
    \begin{center}
    \tiny
            \begin{forest}
                for tree={
                    forked edges,
                    grow'=0,
                    s sep=2.1pt,
                    draw,
                    rounded corners,
                    node options={align=center,},
                    text width=2.7cm,
                },
                %[\rotatebox{90}{\textcolor{Plum}{\textbf{Reconciled Polynomial Network ({\our})}}}, fill=Plum!30, parent
                [{\textcolor{Plum}{\textbf{Reconciled\\ Polynomial\\ Network\\ ({\our})}}}, fill=Plum!30, parent
                %------------------------------------------------------------
                    [Data\\ Expansion\\ Functions, for tree={fill=brown!45, child}
                    	%-------------------------------------------------------------
                    	[1. Identity Expansion, fill=brown!30, grandchild
                    		[{1-(a). $\kappa(\mb{x}) = \mb{x}$}, fill=brown!30, greatgrandchild
                                		[dim: {$D=m$}, fill=brown!15, referenceblock]
                            	]
                            	[{1-(b). $\kappa(\mb{x}) = \sigma(\mb{x})$}, fill=brown!30, greatgrandchild
                                		[{dim: $D=m$}, fill=brown!15, referenceblock]
                            	]
                        ]
                        [2. Reciprocal Expansion, fill=brown!30, grandchild
                            	[{2-(a). $\kappa(\mb{x}) = \frac{{1}}{\mb{x}}$}, fill=brown!30, greatgrandchild
                                		[dim: {$D=m$}, fill=brown!15, referenceblock]
                            	]
                            	[{2-(b). $\kappa(\mb{x}) = \sigma\left(\frac{{1}}{\mb{x}} \right)$}, fill=brown!30, greatgrandchild
                                		[{dim: $D=m$}, fill=brown!15, referenceblock]
                            	]
                        ]
                        [3. Linear Expansion, fill=brown!30, grandchild
                            	[{3-(a). $\kappa(\mb{x}) = c \mb{x}$}, fill=brown!30, greatgrandchild
                                		[dim: {$D=m$}, fill=brown!15, referenceblock]
                            	]
                            	[{3-(b). $\kappa(\mb{x}) = \mb{x} \mb{C}_{post}$}, fill=brown!30, greatgrandchild
                                		[{dim: $D=m$}, fill=brown!15, referenceblock]
                            	]
				[{3-(c). $\kappa(\mb{x}) = \mb{C}_{pre} \mb{x}^\top$}, fill=brown!30, greatgrandchild
                                		[{dim: $D=m$}, fill=brown!15, referenceblock]
                            	]
                        ]
                        [4. Taylor Expansion, fill=brown!30, grandchild
                            	[{4-(a). $\kappa (\mb{x}|d) = [P_1(\mb{x}), P_2(\mb{x}), \cdots, P_d(\mb{x})]$}, fill=brown!30, greatgrandchild
                                		[dim: {$D = \sum_{i=1}^d m^i$}, fill=brown!15, referenceblock]
                            	]
                        ]
                        [5. Fourier Expansion, fill=brown!30, grandchild
                            	[{5-(a). $\kappa (\mb{x} | P, N)$ = $[ \cos (2\pi \frac{i}{P} \mb{x} ), \sin(2\pi \frac{i}{P} \mb{x} )]_{i=1}^N$}, fill=brown!30, greatgrandchild
                                		[dim: {$D = 2 m N$}, fill=brown!15, referenceblock]
                            	]
                        ]
                        [6. B-Spline Expansion, fill=brown!30, grandchild
                            	[{6-(a). $\kappa (\mb{x}|t, d)=[ B_{0,d-1}(\mb{x}), \cdots, B_{t-1,d-1}(\mb{x}) ]$}, fill=brown!30, greatgrandchild
                                		[dim: {$D = m (t + d)$}, fill=brown!15, referenceblock]
                            	]
                        ]
                        [7. Chebyshev Expansion, fill=brown!30, grandchild
                            	[{7-(a). $\kappa(\mb{x} | d) = [ T_1(\mb{x}), T_2(\mb{x}) \cdots, T_n(\mb{x}) ]$}, fill=brown!30, greatgrandchild
                                		[dim: {$D = md$}, fill=brown!15, referenceblock]
                            	]
                        ]
                        [8. Jacobi Expansion, fill=brown!30, grandchild
                            	[{8-(a). $\kappa(\mb{x} | d) = \left[ P_1^{(\alpha, \beta)}(\mb{x}), P_2^{(\alpha, \beta)}(\mb{x}), \cdots, P_d^{(\alpha, \beta)}(\mb{x})\right]$}, fill=brown!30, greatgrandchild
                                		[dim: {$D = md$}, fill=brown!15, referenceblock]
                            	]
                        ]
                        [9. Trigonometric Expansion, fill=brown!30, grandchild
                            	[{9-(a). $\kappa(\mb{x})=[\cos(\mb{x}),\sin(\mb{x}),\tan(\mb{x})]$}, fill=brown!30, greatgrandchild
                                		[dim: {$D = 3m$}, fill=brown!15, referenceblock]
                            	]
				[{9-(b). $\kappa(\mb{x})=[\arccos(\mb{x}),\arcsin(\mb{x}),\arctan(\mb{x})]$}, fill=brown!30, greatgrandchild
                                		[dim: {$D = 3m$}, fill=brown!15, referenceblock]
                            	]
                        ]
                        [10. Hyperbolic Expansion, fill=brown!30, grandchild
                            	[{10-(a). $\kappa(\mb{x})=[\cosh(\mb{x}),\sinh(\mb{x}),\tanh(\mb{x})]$}, fill=brown!30, greatgrandchild
                                		[dim: {$D = 3m$}, fill=brown!15, referenceblock]
                            	]
				[{10-(b). $\kappa(\mb{x})=[\arccosh(\mb{x}),\arcsinh(\mb{x}),\arctanh(\mb{x})]$}, fill=brown!30, greatgrandchild
                                		[dim: {$D = 3m$}, fill=brown!15, referenceblock]
                            	]
                        ]
                        [11. RBF Expansion, fill=brown!30, grandchild
                            	[{11-(a). $\kappa(\mb{x} | d) = \left[ \varphi (\mb{x} | c_1), \varphi (\mb{x} | c_2), \cdots, \varphi (\mb{x} | c_d) \right]$}, fill=brown!30, greatgrandchild
                                		[dim: {$D = md$}, fill=brown!15, referenceblock]
                            	]
                        ]
                        [12. Combinatorial Expansion, fill=brown!30, grandchild
                            	[{12-(a). $\kappa(\mb{x}) = \left[ {\mb{x} \choose 1}, {\mb{x} \choose 2}, \cdots, {\mb{x} \choose d} \right]$}, fill=brown!30, greatgrandchild
                                		[dim: {$D = \sum_{i=1}^d i \cdot {m \choose i}$}, fill=brown!15, referenceblock]
                            	]
                        ]
                        [13. Probabilistic Expansion, fill=brown!30, grandchild
                            	[{13-(a). $\kappa(\mb{x} |d, \bs{\theta}) = \left[ \log P\left({\mb{x}} | \theta_1\right), \cdots, \log P\left({\mb{x} } | \theta_d\right)  \right]$}, fill=brown!30, greatgrandchild
                                		[dim: {$D = md$}, fill=brown!15, referenceblock]
                            	]
				[{13-(b). $\kappa(\mb{x} | d, \bs{\theta})$ = $\left[ \log P\left({\mb{x} \choose 1} | \theta_1\right), \cdots, \log P\left({\mb{x} \choose d} | \theta_d\right)  \right]$}, fill=brown!30, greatgrandchild
                                		[dim: {$D = \sum_{i=1}^d {m \choose i}$}, fill=brown!15, referenceblock]
                            	]
                        ]
                    ]
                    %------------------------------------------------------------
                    [Parameter Reconciliation\\ Functions, for tree={fill=red!45,child}
                    	[1. Constant Reconciliation, fill=red!30, grandchild
				[{1-(a). $\psi(\mb{w}) = \mb{c}$}, fill=red!20, greatgrandchild
                                		[{$l = 0$}, fill=red!10, referenceblock]
                            	]
				[{1-(b). $\psi(\mb{w}) = \mb{0}$}, fill=red!20, greatgrandchild
                                		[{$l = 0$}, fill=red!10, referenceblock]
                            	]
				[{1-(c). $\psi(\mb{w}) = \mb{I}$}, fill=red!20, greatgrandchild
                                		[{$l = 0$}, fill=red!10, referenceblock]
                            	]
			]
			[2. Identity Reconciliation, fill=red!30, grandchild
				[{2-(a). $\psi(\mb{w}) = \mb{w}$}, fill=red!20, greatgrandchild
                                		[{$l = nD$}, fill=red!10, referenceblock]
                            	]
			]
			[3. Masking Reconciliation, fill=red!30, grandchild
				[{3-(a). $\psi({\mb{w}} | p) = \mb{M} \odot \mb{W}$}, fill=red!20, greatgrandchild
                                		[{$l = p \cdot n D$}, fill=red!10, referenceblock]
                            	]
			]
			[4. Duplicated Padding Rec., fill=red!30, grandchild
				[{4-(a). $\psi(\mb{w}|p) = \mb{C} \otimes \mb{W}$}, fill=red!20, greatgrandchild
                                		[{$l= \frac{n D}{pq}$}, fill=red!10, referenceblock]
                            	]
			]
			[5. Low-Rank Reconciliation, fill=red!30, grandchild
				[{5-(a). $\psi(\mb{w}) = \mb{A} \mb{B}^\top$}, fill=red!20, greatgrandchild
                                		[{$l = n r + D r$}, fill=red!10, referenceblock]
                            	]
			]
			[6. HM Reconciliation, fill=red!30, grandchild
				[{6-(a). $\psi(\mb{w}|p) = \mb{A} \otimes \mb{B}$}, fill=red!20, greatgrandchild
                                		[{$l = pq + \frac{nD}{pq}$}, fill=red!10, referenceblock]
                            	]
			]
			[7. LPHM Reconciliation, fill=red!30, grandchild
				[{7-(a). $\psi(\mb{w}|p) = \mb{A} \otimes \mb{B} = \mb{A} \otimes ( \mb{C} \mb{D}^\top)$}, fill=red!20, greatgrandchild
                                		[{$l = pq +r(\frac{n}{p}+\frac{D}{q})$}, fill=red!10, referenceblock]
                            	]
			]
			[8. Dual LPHM Reconciliation, fill=red!30, grandchild
				[{8-(a). $\psi(\mb{w}|p) = \mb{A} \otimes \mb{B} = (\mb{P} \mb{Q}^\top) \otimes ( \mb{C} \mb{D}^\top)$}, fill=red!20, greatgrandchild
                                		[{$l = r(p+ q + \frac{n}{p}+\frac{D}{q})$}, fill=red!10, referenceblock]
                            	]
			]
			[9. HyperNet Reconciliation, fill=red!30, grandchild
				[{9-(a). $\psi(\mb{w}) = \text{HyperNet}(\mb{w})$}, fill=red!20, greatgrandchild
                                		[dim: {$l$ manual setup}, fill=red!10, referenceblock]
                            	]
			]
                    ]
                    %------------------------------------------------------------
                    [Remainder Functions, for tree={fill=blue!45, child}
                        [1. Constant Remainder, fill=blue!30, grandchild
				[{1-(a). $\pi(\mb{x}) = \mb{c}$}, fill=blue!20, greatgrandchild
                                		[no parameter, fill=blue!10, referenceblock]
                            	]
				[{1-(b). $\pi(\mb{x}) = \mb{0}$ (zero remainder)}, fill=blue!20, greatgrandchild
                                		[no parameter, fill=blue!10, referenceblock]
                            	]
			]
			[2. Identity Remainder, fill=blue!30, grandchild
				[{2-(a). $\pi(\mb{x}) = \mb{x}$}, fill=blue!20, greatgrandchild
                                		[requires parameter, fill=blue!10, referenceblock]
                            	]
				[{2-(b). $\pi(\mb{x}) = \sigma(\mb{x})$}, fill=blue!20, greatgrandchild
                                		[requires parameter, fill=blue!10, referenceblock]
                            	]
			]
			[3. Linear Remainder, fill=blue!30, grandchild
				[{3-(a). $\pi(\mb{x}) = \mb{x} \mb{W}'$}, fill=blue!20, greatgrandchild
                                		[requires parameter, fill=blue!10, referenceblock]
                            	]
				[{3-(b). $\pi(\mb{x}) = \sigma(\mb{x} \mb{W}')$}, fill=blue!20, greatgrandchild
                                		[requires parameter, fill=blue!10, referenceblock]
                            	]
			]
			[4. Expansion Remainder, fill=blue!30, grandchild
				[{4-(a). $\pi(\mb{x} | \mb{w}') = \left\langle \kappa'(\mb{x}), \psi'(\mb{w}') \right\rangle + \pi'(\mb{x})$}, fill=blue!20, greatgrandchild
                                		[requires parameter, fill=blue!10, referenceblock]
                            	]
			]
                    ]
                ]
            \end{forest}
    \end{center}
    \caption{An overview of data expansion, parameter reconciliation, and remainder functions implemented in the {\toolkit} toolkit for constructing the {\our} model architecture. }\label{fig:function_instances}
\end{figure*}
%\vspace{2em}
%\hrule
%------------------------------------------------------------------------------------

%\newpage
%---------------------------------------------------

\section{List of Expansion, Reconciliation and Remainder Functions for {\our} Model}\label{sec:functions}

This section introduces the expansion, reconciliation, and remainder functions that can be used to design the {\our} model, all of which have been implemented in the {\toolkit} toolkit and are readily available. Readers seeking a concise overview can refer to Figure~\ref{fig:function_instances}, which summarizes the lists of expansion, reconciliation and remainder functions to be introduced in this section.

%=================================================

\subsection{Data Expansion Functions}\label{subsec:data_expansion_function}

The \textbf{data expansion function} determines the complexity of {\our}. We will introduce several different data expansion functions below. In real-world practice, these individual data expansions introduced below can also be nested and extended to define more complex expansions, which provides more flexibility in the design of our {\our} model.

%---------------------------------------------------

\subsubsection{Identity and Reciprocal Data Expansion}

The simplest data expansion methods are the {identity data expansion} and {reciprocal data expansion}, which project the input data vector $\mb{x} \in \mathbbm{R}^m$ onto itself and its reciprocal, potentially with minor transformations via some activation functions, as denoted below:
\begin{equation}\label{equ:identity_data_expansion_1}
\kappa(\mb{x}) = \mb{x} \in \mathbbm{R}^D \text{, and } \kappa(\mb{x}) = \frac{{1}}{\mb{x}} \in \mathbbm{R}^D,
\end{equation}
or
\begin{equation}\label{equ:identity_data_expansion_2}
\kappa(\mb{x}) = \sigma(\mb{x}) \in \mathbbm{R}^D \text{, and } \kappa(\mb{x}) = \sigma\left( \frac{{1}}{\mb{x}} \right) \in \mathbbm{R}^D.
\end{equation}
In the above equations, $\sigma$ denotes an optional activation function (e.g., sigmoid, ReLU, SiLU) or normalization function (e.g., layer-norm, batch-norm, instance-norm). For both the identity and reciprocal expansion functions, their output dimension is equal to the input dimension, {\ie} $D=m$.

For all the other expansion functions introduced hereafter, as mentioned in the previous Figure~\ref{fig:architecture}, we can also apply the (optional) activation and normalization functions both before and after the expansion by default.

%---------------------------------------------------

\subsubsection{Linear Data Expansion}

In certain cases, we may need to adjust the value scales of $\mb{x}$ linearly without altering the basis vectors or the dimensions of the space. This can be accomplished through the linear data expansion function. Formally, the linear data expansion function projects the input data vector $\mb{x} \in \mathbbm{R}^m$ onto itself via linear projection, as follows:
\begin{equation}
\kappa(\mb{x}) = c \mb{x} \in \mathbbm{R}^D,
\end{equation}
or
\begin{equation}
\kappa(\mb{x}) = \mb{x} \mb{C}_{post} \in \mathbbm{R}^D,
\end{equation}
or
\begin{equation}
\kappa(\mb{x}) = \mb{C}_{pre} \mb{x}^\top \in \mathbbm{R}^{D},
\end{equation}
where the activation function or norm function $\sigma$ is optional, and $c \in \mathbbm{R}$, $\mb{C}_{post}, \mb{C}_{pre} \in \mathbbm{R}^{m \times m}$ denote the provided constant scalar and linear transformation matrices, respectively. Linear data expansion will not change the data vector dimensions, and the output data vector dimension $D=m$. 

%---------------------------------------------------

\subsubsection{Taylor's Polynomials based Data Expansions}\label{subsubsec:taylor_polynomials}

Given a vector $\mb{x} = [x_1, x_2, \cdots, x_m] \in \mathbbm{R}^m$ of dimension $m$, the multivariate composition of order $d$ defined based on $\mb{x}$ can be represented as a list of potential polynomials composed by the product of the vector elements $x_1$, $x_2$, $\cdots$, $x_m$, where the sum of the degrees equals $d$, {\ie}

\begin{equation}
P_d(\mb{x}) = [x_1^{d_1} x_2^{d_2} \cdots x_m^{d_m}]_{d_1, d_2,\cdots, d_m \in \{0, 1, \cdots m\} \land \sum_{i=1}^m d_i = d}.
\end{equation}

Some examples of the multivariate polynomials are provided as follows:

\begin{equation}
\begin{aligned}
P_0(\mb{x}) &= [1] \in \mathbbm{R}^{1},\\
P_1(\mb{x}) &= [x_1, x_2, \cdots, x_m] \in \mathbbm{R}^{m},\\
P_2(\mb{x}) &= [x_1^2,  x_1 x_2, x_1 x_3, \cdots, x_1 x_m, x_2 x_1, x_2^2, x_2 x_3, \cdots, x_{m} x_m] \in \mathbbm{R}^{m^2}.
\end{aligned}
\end{equation}

We observe that the above representation of $P_2(\mb{x})$ may contain duplicated elements, e.g., $x_1 x_2$ and $x_2 x_1$. However, this representation simplifies the implementation, and high-order polynomials can be recursively calculated using the Kronecker product operator based on the lower-order ones.

\begin{definition}
%------------------------------------
\begin{figure*}[h]
    \begin{minipage}{\textwidth}
    \centering
    	\includegraphics[width=0.5\linewidth]{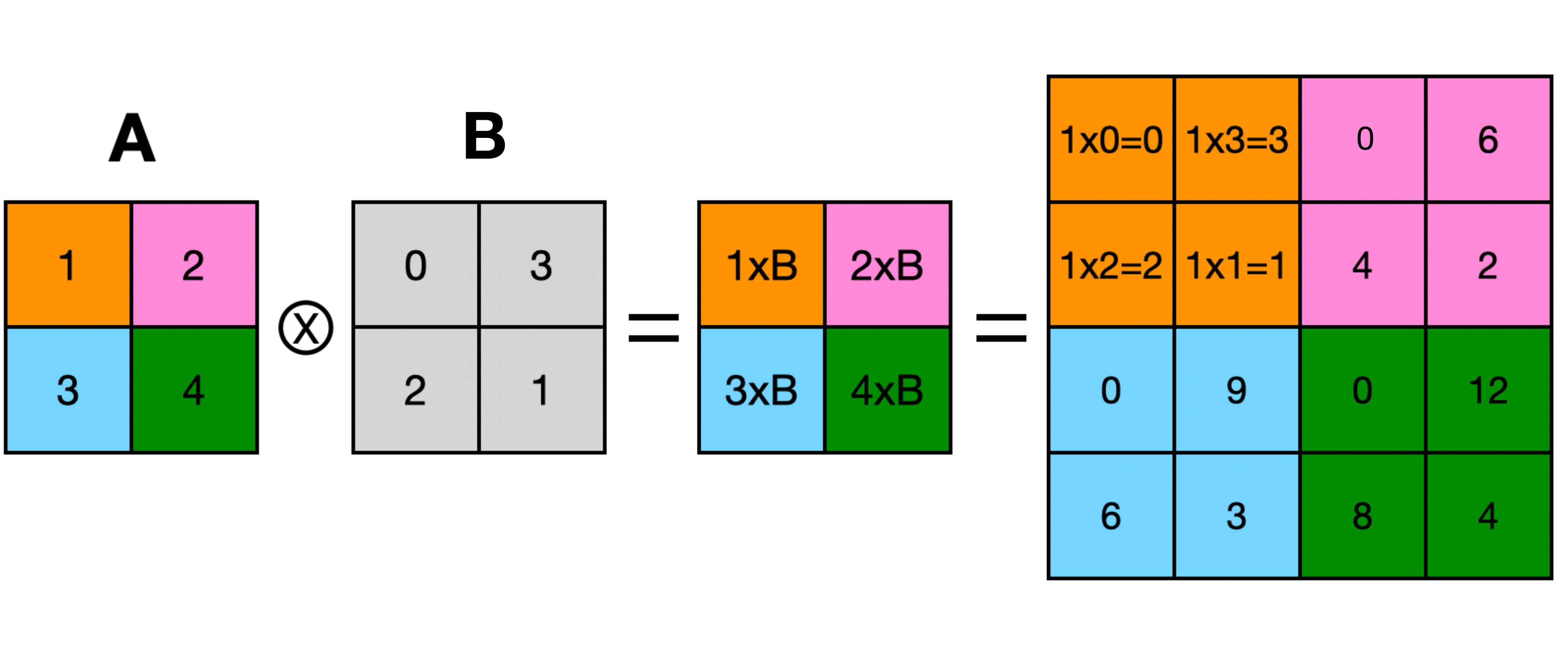}
    	\caption{An illustration of Kronecker product on matrices $\mb{A}$ and $\mb{B}$.}
    	\label{fig:kronecker_product}
    \end{minipage}%
\end{figure*}
%------------------------------------
(Kronecker product): Formally, as illustrated in Figure~\ref{fig:kronecker_product}, given two matrices $\mb{A} \in \mathbbm{R}^{p \times q}$ and $\mb{B} \in \mathbbm{R}^{s \times t}$, the Kronecker product of $\mb{A}$ and $\mb{B}$ is defined as

\begin{equation}
\mb{A} \otimes \mb{B} = 
\begin{bmatrix}
a_{1,1} \mb{B} & a_{1,2} \mb{B}      & \cdots & a_{1,q} \mb{B}      \\
a_{2,1} \mb{B} & a_{2,2} \mb{B}      & \cdots & a_{2,q} \mb{B}      \\
\vdots & \vdots & \ddots & \vdots \\
a_{p,1} \mb{B} & a_{p,2} \mb{B}      & \cdots & a_{p,q} \mb{B}      \\
\end{bmatrix} \in \mathbbm{R}^{ps \times qt},
\end{equation}
where the output will be a larger matrix with $p \times s$ rows and $q \times t$ columns.
\end{definition}

The Kronecker product can also be applied to vectors, with the base polynomial vectors $P_0(\mb{x})$ and $P_1(\mb{x})$, the other Taylor's polynomials with higher orders can all be recursively defined as follows:

\begin{equation}
P_d(\mb{x}) = P_1(\mb{x}) \otimes P_{d-1}(\mb{x}) \text{, for } \forall d \ge 2.
\end{equation}

With the notation $P_d(\mb{x})$, we can define the \textit{Taylor's polynomials} based data expansion function as the list of polynomial terms with orders no greater than $d$ (where $d$ is a hyper-parameter of the function) as follows:

\begin{equation}
\kappa (\mb{x} | d) = [P_1(\mb{x}), \cdots, P_d(\mb{x}) ] \in \mathbbm{R}^D.
\end{equation}

Since {\our} has a deep architecture, the hyper-parameter $d$ is normally set to a small value ({\eg} $d=2$) to avoid excessively large expansions at each layer. Expanding the input data to a Taylor's polynomial of order $4$ can be achieved either by stacking two layers of {\our} layer or by nesting two Taylor's expansion functions with $d=2$. Additionally, the base term $P_0(\mb{x})$ containing the constant value `$1$' can be subsumed by the bias term in the inner product implementation, and thus need not be explicitly included in the expansion. The output dimension will then be $D = \sum_{i=1}^d m^i$. 

Taylor's polynomials are known to approximate functions very well, and an illustrative example of their approximation correctness on function $y = \exp(x)$ is provided below.

%In real implementation, to avoid expanding the data instances to extremely high-dimensions, we also allow the filtering conditions to exclude terms from the expansion results for taylor's polynomial expansion function. 

%--------------------------------
\noindent
\begin{minipage}[h]{0.4\textwidth}
    \textbf{Example:} In the right plot, we illustrate the approximation of function $f(x) = \exp(x)$ with Taylor's polynomials of different orders, where the notation $T_d(x)$ denotes the Taylor's polynomials with degrees up to $d$. According to the plot, increasing the degree allows the Taylor's polynomials to approximate the function $f(x)$ more accurately. Among all these illustrated Taylor's polynomials in the plot, $T_4(x) = 1 + x + \frac{x^2}{2} + \frac{x^3}{6} + \frac{x^4}{24}$ outperforms the others.
\end{minipage}%
\hfill
\begin{minipage}[h]{0.6\textwidth}
    \includegraphics[width=\textwidth]{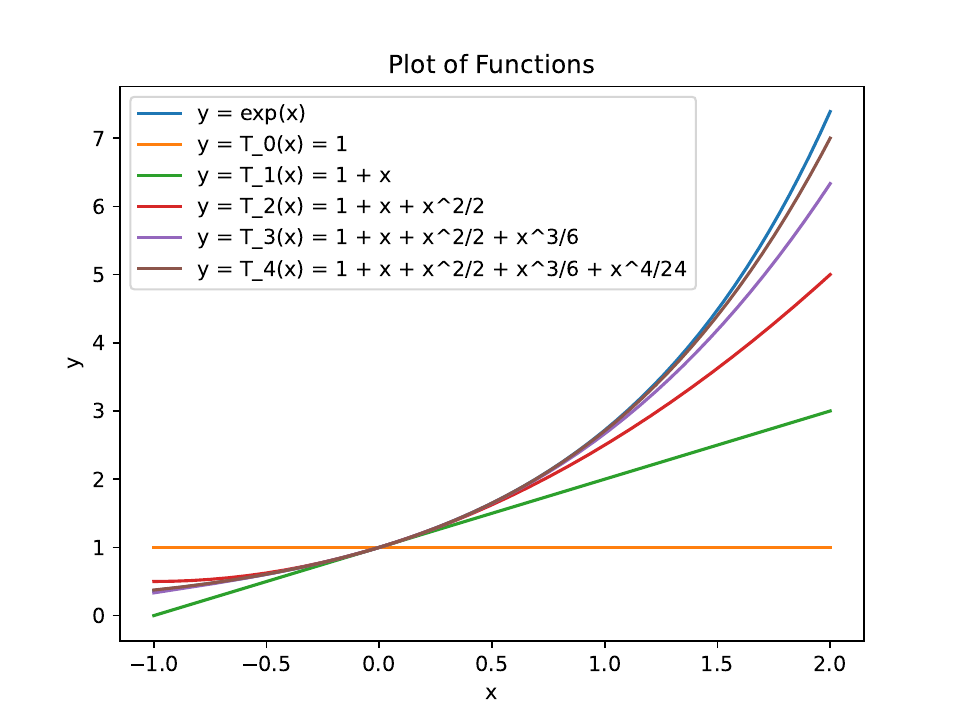}
\end{minipage}
%--------------------------------

%---------------------------------------------------

\subsubsection{Fourier Series based Data Expansions}

A Fourier series is an expansion of a periodic function into the sum of Fourier series. In mathematics, the Dirichlet-Jordan test gives sufficient conditions for a real-valued, periodic function to be equal to the sum of its Fourier series at a point of continuity. Fourier series can be represented in several forms, and in this paper, we will utilize the sine-cosine representation.

Based on the hyper-parameters $P$ and $N$, we can represent the Fourier series based data expansion function $\kappa$ for the input vector $\mb{x}$ as follows:
{
\scriptsize
\begin{equation}
\kappa (\mb{x} | P, N) = \left[ \cos (2\pi \frac{1}{P} \mb{x} ), \sin(2\pi \frac{1}{P} \mb{x} ), \cos(2\pi \frac{2}{P} \mb{x} ), \sin(2\pi \frac{2}{P} \mb{x} ), \cdots, \cos(2\pi \frac{N}{P} \mb{x} ), \sin(2\pi \frac{N}{P} \mb{x} ) \right] \in \mathbbm{R}^D,
\end{equation}
}
where the output dimension $D = 2 m N$.

%---------------------------------------------------

\subsubsection{B-Splines based Data Expansion}

Formally, a B-spline of degree $d+1$ is defined as a collection of piecewise polynomial functions $\left\{B_{i,d}(x) \right\}_{i\in \{0, 2, \cdots, t-1\}}$ of degree $d$ over the variable $x$, which takes values from a pre-defined value range $[x_0, x_t)$. The value range $[x_0, x_t]$ is divided into smaller pieces by points $x_0, x_1, x_2, \cdots, x_t$ sorted in a non-decreasing order, and these points are also known as the \textbf{knots}. These knots partition the value range $[x_0, x_t]$ into $t$ disjoint intervals: $[x_0, x_1), [x_1, x_2), \cdots, [x_{t-1}, x_t)$.

As to the specific representations of B-splines, they can be defined recursively based on the lower-degree terms according to the following equations:

\noindent \textbf{Base B-splines with degree $d = 0$}:
\begin{equation}
\left\{ B_{0,0}(x), B_{1,0}(x), \cdots, B_{t-1,0}(x) \right\},
\end{equation}
where
\begin{equation}
B_{i,0}(x) = \begin{cases}
1, &\text{if } x_i \le x < x_{i+1};\\
0, &\text{otherwise}.
\end{cases}
\end{equation}

\noindent \textbf{Higher-degree B-splines with $d>0$}:
\begin{equation}
\left\{ B_{0,d}(x), B_{1,d}(x), \cdots, B_{t-1,d}(x) \right\},
\end{equation}
where
\begin{equation}
B_{i,d}(x) = \frac{x - x_i}{x_{i+d} - x_i} B_{i, d-1}(x) + \frac{x_{i+d+1} - x}{x_{i+d+1} - x_{i+1}} B_{i+1, d-1}(x).
\end{equation}
According to the representations, term $B_{i,d}(x)$ recursively defined above will have non-zero outputs if and only if the inputs lie within the value range $x_i \le x < x_{i+d+1}$. 

B-splines have been extensively used in curve-fitting and numerical differentiation of experimental data, including their recent application in the design of KAN \cite{Liu2024KANKN}. In this paper, we define the B-spline-based data expansion function with degree $d$, which can be represented as follows:
\begin{equation}
\kappa (\mb{x} | d) = \left[ B_{0,d}(\mb{x}), B_{1,d}(\mb{x}), \cdots, B_{t-1,d}(\mb{x}) \right] \in \mathbbm{R}^D,
\end{equation}
where the output dimension can be calculated as $D = m (t + d)$.

%---------------------------------------------------

\subsubsection{Chebyshev Polynomials based Data Expansion}

In addition to B-splines, we observe several other similar basis functions recursively defined based on those of lower degrees, including Chebyshev polynomials and Jacobi polynomials. 

Chebyshev polynomials have been demonstrated to be important in approximation theory for the solution of linear systems. They can be represented as two sequences of polynomials related to the cosine and sine functions (also known as the first-kind and second-kind), with recursive calculation equations that are quite similar to each other, differing only in scalar coefficients. In this paper, we will use the {Chebyshev polynomials} of the first kind ({\ie} defined based on the cosine function) and it can be represented with the following recursive equations. 

\noindent \textbf{Base cases $d=0$ and $d=1$}:
\begin{equation}
T_0(x) = 1 \text{, and } T_1(x) = x
\end{equation}

\noindent \textbf{High-order cases with degree $d \ge 2$}:
\begin{equation}
T_d(x) = 2x \cdot T_{d-1}(x) - T_{d-2}(x).
\end{equation}

Based on the above representations, in this paper, we define the {Chebyshev polynomials} based data expansion function with degree hyper-parameter $d \ge 1$ as follows:
\begin{equation}
\kappa(\mb{x} | d) = \left[ T_1(\mb{x}), T_2(\mb{x}) \cdots, T_d(\mb{x}) \right] \in \mathbbm{R}^D.
\end{equation}
Similar to the aforementioned Taylor's polynomial-based data expansions, since the output of $T_0(x)$ is a constant, it will not be included in the expansion function definition by default, and the output dimension will be $D = m d$.

%---------------------------------------------------

\subsubsection{Jacobi Polynomials based Data Expansion}
%https://en.wikipedia.org/wiki/Jacobi_polynomials

Different from the Chebyshev polynomial, the Jacobi polynomials have a more complicated recursive representation. Formally, the Jacobi polynomials parameterized by $\alpha$ and $\beta$ of degree $d \ge 2$ on variable $x$ can be represented as $P_d^{(\alpha, \beta)}(x)$, which can be recursively defined based on the lower-order cases:
\begin{equation}
\begin{aligned}
P_d^{(\alpha, \beta)}(x) =& \frac{(2d + \alpha + \beta -1) \left[ (2d + \alpha + \beta)(2d + \alpha + \beta -2) x + (\alpha^2 - \beta^2) \right]}{2d(d + \alpha + \beta)(2d + \alpha + \beta - 2) } P_{d-1}^{(\alpha, \beta)}(x)\\
& - \frac{2(d+\alpha-1)(d+\beta-1)(2d+\alpha+\beta)}{2d(d + \alpha + \beta)(2d + \alpha + \beta - 2) }P_{d-2}^{(\alpha, \beta)}(x).
\end{aligned}
\end{equation}

As to the base case, we list some of them as follows:
\begin{equation}
\begin{aligned}
P_0^{(\alpha, \beta)}(x) &= 1,\\
P_1^{(\alpha, \beta)}(x) &= (\alpha + 1) + (\alpha + \beta + 2) \frac{(x-1)}{2},\\
P_2^{(\alpha, \beta)}(x) &= \frac{(\alpha+1)(\alpha+2)}{2} + (\alpha+2)(\alpha+\beta+3) \frac{x-1}{2} + \frac{(\alpha + \beta + 3)(\alpha + \beta + 4)}{2} \left( \frac{x-1}{2} \right)^2.
\end{aligned}
\end{equation}

In this paper, we define the Jacobi polynomial based data expansion function with degree $d$ as follows:
\begin{equation}
\kappa(\mb{x} | d) = \left[ P_1^{(\alpha, \beta)}(\mb{x}), P_2^{(\alpha, \beta)}(\mb{x}),  \cdots, P_d^{(\alpha, \beta)}(\mb{x})\right] \in \mathbbm{R}^D,
\end{equation}
where the output dimension $D = m d$.

The Jacobi polynomials belong to the family of classic orthogonal polynomials, where two different polynomials in the sequence are orthogonal to each other under some inner product. Meanwhile, the Gegenbauer polynomials form the most important class of Jacobi polynomials, and Chebyshev polynomial is a special case of the Gegenbauer polynomials. In addition to these, we will also gradually incorporate other classic orthogonal polynomials into our {\toolkit} toolkit.

%---------------------------------------------------

\subsubsection{Hyperbolic Function and Trigonometric Function based Data Expansions}

The Fourier series introduced above is actually an example of a trigonometric series. In addition to Fourier series, we also include several other types of trigonometric functions based data expansion approach in this paper, such as \textit{hyperbolic functions}, shown as follows:

\begin{equation}
\kappa (\mb{x}) = \left[ \cosh(\mb{x}), \sinh(\mb{x}), \tanh(\mb{x}) \right] \in \mathbbm{R}^D \text{, where } D = 3 m.
\end{equation}

In addition to the hyperbolic functions, we can also define the data expansion function using inverse hyperbolic functions, trigonometric functions, and inverse trigonometric functions, as follows:
\begin{align}
\kappa (\mb{x}) &= \left[ \arccosh(\mb{x}), \sinh(\mb{x}), \tanh(\mb{x}) \right] \in \mathbbm{R}^D;\\
\kappa (\mb{x}) &= \left[ \cos(\mb{x}), \sin(\mb{x}), \tan(\mb{x}) \right] \in \mathbbm{R}^D;\\
\kappa (\mb{x}) &= \left[ \arccos(\mb{x}), \arcsin(\mb{x}), \arctan(\mb{x}) \right] \in \mathbbm{R}^D,
\end{align}
where the output dimensions of these expansion functions are all $D = 3 m$.

Unlike hyperbolic functions, trigonometric functions are periodic, and different input values of $x$ may be projected to identical outputs, rendering them indistinguishable. This can potentially lead to degraded performance. Nonetheless, the above trigonometric function-based data expansion function can be employed as intermediate layers or complementary heads within a layer to compose more complex functions and construct more powerful models.

%---------------------------------------------------

\subsubsection{Radial Basis Functions based Data Expansion}\label{subsubsec:rbf}
%https://en.wikipedia.org/wiki/Radial_basis_function

In mathematics, a radial basis function (RBF) is a real-valued function $\varphi$ defined based on the distance between the input ${x}$ and some fixed point, {\eg} ${c}$, which can be represented as follows:
\begin{equation}
\varphi({x} | c) = \varphi (x - c).
\end{equation}

There are different ways to define the function $\varphi$ in practice, two of which studied in this paper are shown as follows:

\noindent\begin{minipage}{.42\linewidth}
\begin{fleqn}
%\tcbhighmath[fuzzy halo=1mm with blue!50!white,arc=2pt, boxrule=0pt,frame hidden]
{
\begin{equation}
\begin{aligned}
&\underline{\textbf{(a) Gaussian RBF:}}\\[6pt]
&\varphi (x | c) = e^{-(\epsilon (x - c) )^2},\\[6pt]
&\text{ where } \epsilon \text{ is a hyperparameter}.
\end{aligned}
\end{equation}
}
\end{fleqn}
\end{minipage}
\hfill
\noindent\begin{minipage}{.48\linewidth}
\begin{fleqn}
{
\begin{equation}
\begin{aligned}
&\underline{\textbf{(b) Inverse Quadratic RBF:}}\\[0pt]
&\varphi (x | c) = \frac{1}{1 + (\epsilon (x - c))^2},\\[0pt]
&\text{ where } \epsilon \text{ is a hyperparameter}.
\end{aligned}
\end{equation}
}
\end{fleqn}
\end{minipage}

Given a set of $d$ different fixed points, {\eg} $\mb{c} = [c_1, c_2, \cdots, c_d]$, a sequence of $d$ different such RBF can be defined, which compare the input against these $d$ different fixed points shown as follows:
\begin{equation}
{\varphi} (x | \mb{c}) = \left[ \varphi (x | c_1), \varphi (x | c_2), \cdots, \varphi (x | c_d) \right] \in \mathbbm{R}^d.
\end{equation}

With the above notations, we can represent the Gaussian RBF and Inverse Quadratic RBF based data expansion functions both with the following equation:
\begin{equation}
\kappa(\mb{x}) = {\varphi} (\mb{x} | \mb{c}) = \left[ \varphi (\mb{x} | c_1), \varphi (\mb{x} | c_2), \cdots, \varphi (\mb{x} | c_d) \right] \in \mathbbm{R}^D,
\end{equation}
where the output dimension will be $D = m  d$.

%\subsubsection{Other Polynomials}

%https://github.com/lif314/X-KANeRF

%---------------------------------------------------

\subsubsection{{Combinatorial Data Expansion}}

{Combinatorial data expansion function} expands input data by enumerating the potential combinations of elements from the input vector, with the number of elements to be combined ranging from $1$, $2$, $\cdots$, $d$, where $d$ is a hyper-parameter. Formally, given a data instance featured by a variable set $\mc{X} = \{X_1, X_2, \cdots, X_m\}$ (here, we use the upper-case $X_i$ to denote the variable of the $i_{th}$ feature), we can represent the possible combinations of $d$ terms selected from $\mc{X}$ with notation:
\begin{equation}
{\mc{X} \choose d} = \left\{ \mc{C} | \mc{C} \subset \mc{X} \land |\mc{C}| = d \right\},\\
\end{equation}
where $\mc{C}$ denotes a subset of $\mc{X}$ containing no duplicated elements and the size of the output set ${\mc{X} \choose d}$ will be equal to ${m \choose d}$. Some simple examples with $d=1$, $d=2$ and $d=3$ are illustrated as follows:
\begin{equation}
\begin{aligned}
d = 1:\ \  &{\mc{X} \choose 1} = \left\{\{X_i\} | X_i \in \mc{X} \right\},\\
d = 2:\ \  &{\mc{X} \choose 2} = \left\{\{X_i, X_j\} | X_i, X_j \in \mc{X} \land X_i \neq X_j \right\},\\
d = 3:\ \  &{\mc{X} \choose 3} = \left\{\{X_i, X_j, X_k\} | X_i, X_j, X_k \in \mc{X} \land X_i \neq X_j \land X_i \neq X_k \land X_j \neq X_k \right\}.\\
\end{aligned}
\end{equation}

By applying the above notations to concrete data instances, given a data instance with values $\mb{x} = [x_1, x_2, \cdots, x_m]$ (the lower-case $x_i$ denotes the feature value), we can also represent the combinations of $d$ selected features from $\mb{x}$ as ${\mb{x} \choose d}$, which can be used to define the combinatorial data expansion function as follows:
\begin{equation}\label{equ:combinatorial}
\kappa(\mb{x}) = \left[ {\mb{x} \choose 1}, {\mb{x} \choose 2}, \cdots, {\mb{x} \choose d} \right].
\end{equation}

Similar as the above Taylor's expansions, the output dimension of the combinatorial expansion will increase exponentially. Given an input data vector $\mb{x} \in \mathbbm{R}^m$, with hyper-parameter $d$, its expansion output of combinations with up to $d$ elements will be $\kappa(\mb{x}) \in \mathbbm{R}^D$, where $D = \sum_{i=1}^d i \cdot {m \choose i}$.

%---------------------------------------------------

\subsubsection{Probabilistic Data Expansion}\label{subsubsec:probabilistic_data_expansion}

An important category of data expansion functions that surprisingly performed very well, even outperforming many of the extension approaches mentioned above, are probability density function based data expansions. Formally, in probability theory and statistics, a probability density function (PDF) is a function that describes the relative likelihood for a random variable to take on a given value within its sample space. Formally, given a probabilistic distribution parameterized by $\theta$, we can represent its probability density function as
\begin{equation}
P(x | \theta) \in [0, 1] \text{, where } \int_x P(x | \theta) \mathrm{d}x = 1.
\end{equation}

Lots of probabilistic distributions have been proposed by mathematicians and statisticians, such as \textit{Gaussian distribution} $\mc{N}(\mu, \sigma)$, \textit{Exponential distribution} $\mc{E}(\lambda)$, \textit{Laplace distribution} $\mc{L}(\mu, b)$, \textit{Cauchy distribution} $\mc{C}(x_0, \gamma)$, \textit{Chi-squared distribution} $\mc{X}^2(k)$ and \textit{Gamma distribution} $\Gamma(k, \theta)$, etc. The PDFs of these distributions are also provided as follows:

%---------------------------------------------------
\begin{minipage}{.48\linewidth}
\begin{fleqn}
{
\begin{equation}
\begin{aligned}
&\underline{\textbf{(a) Gaussian Distribution:}}\\[5pt]
&P(x| \mu, \sigma) = \frac{1}{\sigma \sqrt{2 \pi}}\exp^{-\frac{1}{2} (\frac{x-\mu}{\sigma})^2},\\[5pt]
&\text{ where } \mu, \sigma \text{ are the mean and std parameters}.
\end{aligned}
\end{equation}
}
\end{fleqn}
\end{minipage}
\hfill
\noindent\begin{minipage}{.45\linewidth}
\begin{fleqn}
{
\begin{equation}
\begin{aligned}
&\underline{\textbf{(b) Exponential Distribution:}}\\[1pt]
&P(x | \lambda) = \begin{cases}
\lambda \exp^{- \lambda x} & \text{ for } x \ge 0,\\
0 & \text{ otherwise},
\end{cases},\\[1pt]
&\text{ where } \lambda > 0 \text{ is the rate parameter}.
\end{aligned}
\end{equation}
}
\end{fleqn}
\end{minipage}
%---------------------------------------------------

%---------------------------------------------------
\begin{minipage}{.48\linewidth}
\begin{fleqn}
{
\begin{equation}
\begin{aligned}
&\underline{\textbf{(c) Laplace Distribution:}}\\[10pt]
&P(x| \mu, b) = \frac{1}{2b} \exp^{\left(- \frac{|x-\mu|}{b} \right)},\\[10pt]
&\text{ where } \mu, b>0 \text{ are the location, scale parameters}.
\end{aligned}
\end{equation}
}
\end{fleqn}
\end{minipage}
\hfill
\noindent\begin{minipage}{.48\linewidth}
\begin{fleqn}
{
\begin{equation}
\begin{aligned}
&\underline{\textbf{(d) Cauchy Distribution:}}\\[3pt]
&P(x | x_0, \gamma) = \frac{1}{\pi \gamma \left[1 +\left( \frac{x-x_0}{\gamma} \right)^2 \right]},\\[3pt]
&\text{ where } x_0, \gamma \text{ are the location and scale parameters}.
\end{aligned}
\end{equation}
}
\end{fleqn}
\end{minipage}
%---------------------------------------------------

%---------------------------------------------------
\begin{minipage}{.48\linewidth}
\begin{fleqn}
{
\begin{equation}
\begin{aligned}
&\underline{\textbf{(e) Chi-Squared Distribution:}}\\[5pt]
&P(x| k) = \frac{1}{2^{\frac{k}{2}} \Gamma(\frac{k}{2})} x^{(\frac{k}{2}-1)} \exp^{-\frac{x}{2}},\\[5pt]
&\text{ where } k \in \mathbbm{N}^+ \text{ is the dof parameter}.
\end{aligned}
\end{equation}
}
\end{fleqn}
\end{minipage}
\hfill
\noindent\begin{minipage}{.48\linewidth}
\begin{fleqn}
{
\begin{equation}
\begin{aligned}
&\underline{\textbf{(f) Gamma Distribution:}}\\[8pt]
&P(x | k, \theta) = \frac{1}{\Gamma(k) \theta^k} x^{k-1} \exp^{- \frac{x}{\theta}},\\[8pt]
&\text{ where } k, \theta>0 \text{ are the shape and scale parameters}.
\end{aligned}
\end{equation}
}
\end{fleqn}
\end{minipage}
%---------------------------------------------------

When feeding inputs to a probability density function, its output is typically a very small number, and the curve of many distribution PDFs can be quite flat ({\ie} with a very small slope). In this paper, we propose using the log-likelihood to expand the input data instead, which makes it possible to unify the representations of probabilistic graphical models with {\our}, more information of which will be introduced in Section~\ref{subsec:pm}. 

In this paper, we introduce two different expansions based on the probabilistic distributions, {\ie} \textbf{naive probabilistic expansion} and \textbf{combinatorial probabilistic expansion} introduced as follows.

\subsubsubsection{\textbf{Naive Probabilistic Data Expansion}}

The naive probabilistic expansion assumes the input features are independent and directly applies the distribution PDFs to the input data vector to calculate the corresponding log-likelihood scores as the outputs. To expand the inputs, a set of $d$ identical (or different) distribution PDFs with different hyper-parameters can be used to concurrently calculate the log-likelihood scores. The concatenated outputs from these PDFs are then returned as the expansions.

Formally, given the input $\mb{x} \in \mathbbm{R}^m$, the naive probabilistic function assuming all the elements in $\mb{x}$ to be independent will compute the log-likelihood to sample each of the features in the instance according to certain distributions:
\begin{equation}
\kappa(\mb{x} | \bs{\theta}) = \left[ \log P\left({\mb{x}} | \theta_1\right), \log P\left({\mb{x} } | \theta_2\right), \cdots, \log P\left({\mb{x} } | \theta_d\right)  \right] \in \mathbbm{R}^D,
\end{equation}
where $\bs{\theta} = [\theta_1, \theta_2, \cdots, \theta_d]$ denotes $d$ different hyper-parameters used for the PDFs. For the input data vector of length $m$, it is easy to obtain its expansion output dimension via the above function will be $D = m d$.

In addition to using a single distribution PDF (with different hyper-parameters), we also introduce a hybrid naive probabilistic expansion approach that simultaneously employs PDFs of different distributions. For example, $P_1(\cdot | \theta_1)$ denotes the normal distribution with mean/std denoted by $\theta_1$, while $P_2(\cdot | \theta_2)$ denotes the Cauchy distribution with location/scale denoted by $\theta_2$, and so on. The above expansion function can also be rewritten as follows:
\begin{equation}
\kappa(\mb{x} | \bs{\theta}) = \left[ \log P_1\left({\mb{x}} | \theta_1\right), \log P_2\left({\mb{x} } | \theta_2\right), \cdots, \log P_d\left({\mb{x} } | \theta_d\right)  \right] \in \mathbbm{R}^D,
\end{equation}
where $d$ different distribution PDFs $P_1, P_2, \cdots, P_d$ are concatenated to define the expansion function here. The output dimension will remain the same as the above, {\ie} $D = m d$.

For the PDF of the Gamma, Chi-square and Exponential distributions, they may require non-negative inputs (some may also require the input to be non-zero). Therefore, prior to feeding the input vector $\mb{x}$ to their PDF for expansion, we need to pre-transform $\mb{x}$ into positive vectors (with either activation functions or normalization functions).

%The above combinatorial expansion function projects the vector to a high-dimensional vector with duplicated elements actually didn't project the instance to a new space with new basis vectors. Besides the above random combinatorial expansions, we can also define the above function with minor changes to only do the combinations of nearby elements, which will be used when representing the CNN model with {\our} (we will talk about that later).

\subsubsubsection{\textbf{Combinatorial Probabilistic Data Expansion}}

Based on the above combinatorial expansions and multivariate distributions, we can introduce the combinatorial probabilistic expansion function. Distinct from naive probabilistic data expansion functions, combinatorial probabilistic data expansion function considers the relationships among variables in the multivariate distribution PDFs, which can model complex data distributions better.

Formally, given the input data vector $\mb{x} \in \mathbbm{R}^m$, the combinatorial probabilistic expansion function defined based on the multivariate distribution PDF can be represented as follows:
\begin{equation}
\kappa(\mb{x} | \bs{\theta}) = \left[ \log P\left({\mb{x} \choose 1} | \theta_1\right), \log P\left({\mb{x} \choose 2} | \theta_2\right), \cdots, \log P\left({\mb{x} \choose d} | \theta_d\right)  \right] \in \mathbbm{R}^D,
\end{equation}
where the output vector containing the log-likelihood has a dimension of $D = \sum_{i=1}^d {m \choose i}$. In real applications, the hyper-parameter $d$ are usually set with a small number ({\eg} $d=2$) to avoid extremely high-dimensional expansions. Different from the probability density functions used for naive probabilistic expansion, the above function $P(\cdot | \theta_d)$ is a multivariate probability density function with $d$ variables. Notations $\theta_d$ denotes the hyper-parameters ({\eg} encompassing both the mean vector and covariance matrix if $P(\cdot | \theta_d)$ denotes the PDF of the multivariate normal distribution) of the distribution PDF for combinatorial terms with $d$ elements.

This combinatorial probabilistic functions enable the unification of current deep learning models with classic probabilistic models ({\eg} Bayesian networks and Markov networks) into a single framework for data processing and learning. This is because the summation/subtraction of the output terms directly calculates the conditional and joint probabilities of the feature variables. Of course, the parameters of the distributions used above are pre-defined and frozen constants for the current expansion functions. In the future, we will investigate to learn the optimal distribution hyper-parameters instead for better data expansions.

%=================================================

\subsection{Parameter Reconciliation Functions}

To approximate the underlying mapping $f: \mathbbm{R}^m \to \mathbbm{R}^n$, the {data expansion functions} $\kappa: \mathbbm{R}^m \to \mathbbm{R}^{D}$ introduced above projects data instances from input dimension $m$ to an intermediate space of dimension $D$, where $D > m$. When learning on such expanded data, directly applying existing models and learning approaches with similar parameter scales may suffer from the ``curse of dimensionality'' and overfitting issues, leading to practical failures. Instead of directly defining a parameter of a scale of $D$, we propose defining functions $\psi: \mathbbm{R}^l \to \mathbbm{R}^{n \times D}$ to fabricate a parameter vector $\mb{w} \in \mathbbm{R}^l$ of length $l$ to the target dimensions using advanced techniques, where $l \ll n \times D$. 

This process is referred to as \textbf{parameter reconciliation} in this paper. The inner product of the expanded data vectors and the reconciled parameter matrices will project the data vectors from input dimension $m$ to an intermediate dimension $D$ and then back to the desired output dimension $n$. The parameter reconciliation function determines both the learning capacity and costs of the {\our} model, and we will introduce several practical ways to fabricate the parameters in defining the parameter reconciliation functions in this section. In addition to the summary provided in Figure~\ref{fig:function_instances}, we also illustrate the parameter reconciliation functions to be introduced here in Figure~\ref{fig:reconciliations} as well.

%------------------------------------
\begin{figure*}[t]
    \begin{minipage}{\textwidth}
    \centering
    	\includegraphics[width=\linewidth]{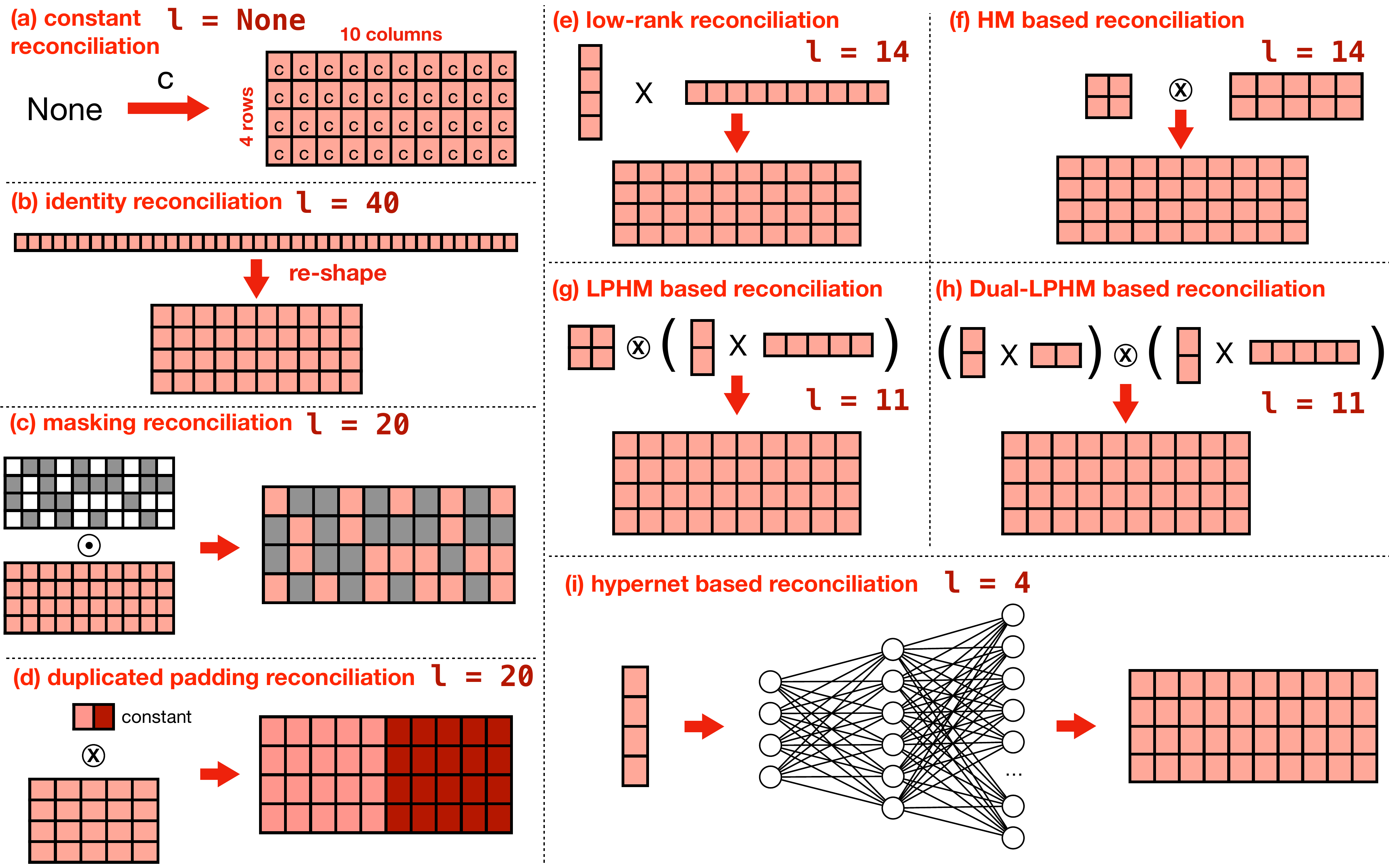}
    	\caption{An illustration of different parameter reconciliation functions introduced in this paper. For each of the reconciliation function, we also indicate the number of required parameter length $l$ to generate the desired parameter matrix of size $4 \times 10$ in the plots.}
    	\label{fig:reconciliations}
    \end{minipage}%
\end{figure*}
%------------------------------------

%---------------------------------------------------

\subsubsection{Constant Parameter Reconciliation}

The simplest parameter reconciliation function will be the {constant parameter reconciliation}, which projects any input parameters to constants ({\eg} zeros or ones) as follows:
\begin{equation}
\psi(\mb{w} | c) = c \cdot \mb{1}^{n \times D} = \mb{C} \in \mathbbm{R}^{n \times D},
\end{equation}
where the output matrix $\mb{C}$ of size $n \times D$ is filled with the provided constant $c$.

For constant parameter reconciliation, the input parameter $\mb{w}$ is not required, which together with its dimension hyper-parameter $l$ can both be set to \textit{none} in implementation. If the output constant $\mb{C} = \mb{0}$ or $\mb{C} = \mb{1}$, we can also name the functions as \textbf{zero reconciliation} and \textbf{one reconciliation}, respectively. Constant parameter reconciliation functions can accommodate outputs according to requirements. For example, we can set the output to be an identity matrix $\mb{I}$ with dimensions $D \times D$, which can be used if and only if the layer input and output dimensions are identical, {\ie} $m=n$. We can name such a function as the \textbf{constant eye reconciliation} to differentiate it from the identity reconciliation defined below.

Constant reconciliation contributes almost nothing to model learning since it involves no parameters, but it provides our approach with substantial flexibility in representing and designing many models, such as the probabilistic models introduced later in the following Section~\ref{sec:backbone_unification}.

%In implementation, we don't need to strictly follow the above equation to define the real constant outputs, like the \textit{constant square identity reconciliation}, and we will use nested expansions to avoid unnecessary space and time costs.

%---------------------------------------------------

\subsubsection{Identity Parameter Reconciliation}

Another simple parameter reconciliation function is the {identity parameter reconciliation}, which defines the identity reconciliation function $\psi: \mathbbm{R}^{l} \to \mathbbm{R}^{n \times D}$ as follows:
\begin{equation}\label{equ:identity_reconciliation}
\psi(\mb{w}) = \text{reshape}(\mb{w}) = \mb{W},
\end{equation}
where the function will resize the parameters from vector $\mb{w}$ of length $l = n \times D$ to the matrix $\mb{W}$ of size $n \times D$.

Identity parameter reconciliation is straightforward and may work well for some expansion functions whose output dimension $D$ is not very large. However, when used with expansion functions that produce a large output dimension (such as the high-order Taylor's polynomial expansions), the identity parameter reconciliation function may fail due to the ``curse of dimensionality'' issues. In such cases, the learning cost would also be extremely high.

%---------------------------------------------------

\subsubsection{Masking based Parameter Reconciliation}

% Fig 3 of https://arxiv.org/pdf/2403.14608

To mitigate the identified limitation of identity parameter reconciliation function, one prospective approach entails the initial definition of a parameter matrix denoted as $\mb{W} \in \mathbbm{R}^{n \times D}$, with a subsequent strategic masking of a substantial proportion of its elements. This strategy is implemented to curtail the count of learnable parameters in $\mb{W}$ to a reduced number of $l$:
\begin{equation}
\psi({\mb{w}}) = (\mb{M} \odot \text{reshape}(\mb{w})) = (\mb{M} \odot \mb{W}) \in \mathbbm{R}^{n \times D},
\end{equation}
where the term $\mb{M} \in \{0, 1\}^{n \times D}$ denotes the binary masking matrix only with $l$ non-zero entries and $\odot$ denotes the element-wise product operator. The notation $\mb{w}$ is the vector representation of the parameter matrix $\mb{W}$.

This reconciliation operator is equivalent to: (a) first defining a parameter vector $\mb{w}$ of length $l$, and (b) then scatting these $l$ parameters to a larger matrix of size $n \times D$. Moreover, only these scattered parameters are learnable while all the remaining ones are constant zeros with no gradients. However, current programming toolkits such as PyTorch lack the capability to selectively assign gradients to specific entries within a tensor, rendering the above masking based reconciliation a more pragmatic choice for implementation. 

What's more, to facilitate practical adoption, instead of pre-define the parameter dimension $l$, we advocate for the definition of the masking ratio $p \in [0, 1]$ as a hyper-parameter of the masking based reconciliation function instead. This parameter, in conjunction with the output dimensions $n \times D$, computes the requisite parameter vector dimension, given by $l = p \times n \times D$.

%---------------------------------------------------

\subsubsection{Duplicated Padding based Parameter Reconciliation}\label{subsubsec:duplicated_padding_reconciliation}

In addition to masking, an alternative straightforward approach for fabricating the parameter vector $\mb{w}$ from length $l$ to size $n \times D$ involves recursively duplicating $\mathbf{w}$ and sequentially padding them to form the larger parameter matrix. Such a reconciliation function can be efficiently implemented using the matrix Kronecker product operator introduced before in Section~\ref{subsubsec:taylor_polynomials}.

Specifically, for the parameter vector $\mb{w} \in \mathbbm{R}^{l}$ of length $l$, it can be reshaped into a matrix $\mb{W}$ comprising $s$ rows and $t$ columns, where $l = s \times t$. Through the multiplication of $\mb{W}$ with a constant matrix $\mb{C} \in \mathbbm{R}^{p \times q}$ populated with the constant value of ones, we can define the {duplicated padding based parameter reconciliation} function as follows:
\begin{equation}
\psi(\mb{w}) = \mb{C} \otimes \mb{W} = \begin{bmatrix}
C_{1,1} \mb{W} & C_{1,2} \mb{W}      & \cdots & C_{1,q} \mb{W}      \\
C_{2,1} \mb{W} & C_{2,2} \mb{W}      & \cdots & C_{2,q} \mb{W}      \\
\vdots & \vdots & \ddots & \vdots \\
C_{p,1} \mb{W} & C_{p,2} \mb{W}      & \cdots & C_{p,q} \mb{W}      \\
\end{bmatrix} \in \mathbbm{R}^{ps \times qt},
\end{equation}
where $\mb{W} = \text{reshape}(\mb{w})$ and $\otimes$ denotes the Kronecker product operator.

The resulting matrix will encompass $p \times q$ duplicates of the reshaped parameter matrix $\mathbf{W}$. By adjusting the dimensions of the constant matrix $\mathbf{C}$ - specifically, $p$ and $q$ - we can ensure that $p \times s = n$, $q \times t = D$, and $l = s \times t$, aligning with the desired target parameter dimensions. Regarding the constant matrix $\mathbf{C}$, beyond being filled with all ones, its elements can also adopt a binary form, comprising zeros and ones. In this scenario, the output will exhibit sparsity, featuring only a few replicated copies of $\mathbf{W}$. This flexibility in matrix construction augments the versatility of {\our} model design and learning process. Notably, the parameter length $l$ is not predetermined but computed during the function definition as $l= st = \frac{n \times D}{pq}$, where $p$ and $q$ are the hyper-parameters of this reconciliation function to be set manually.

%Moreover, to simplify the implementation process, we can define the constant matrix $\mathbf{C}$ as a square matrix of size $p \times p$, where $p$ represents the size hyper-parameter. Then, the desired dimensions of $\mathbf{W}$, denoted as $s$ and $t$, can be calculated to spread across all entries of the $n \times D$ matrix. Notably, the parameter length $l$ is not predetermined but computed during the function definition as $l= \frac{n \times D}{p^2}$.
 
%---------------------------------------------------

\subsubsection{Low-Rank Parameter Reconciliation (LoRR)}

The practice of low-rank adaption, as investigated in \cite{Hu2021LoRALA}, is commonly employed in contemporary parameter-efficient fine-tuning (PEFT) methodologies for language models. Although our paper does not center on PEFT, and our {\our} model doesn't incorporate adapters, the principles derived from existing low-rank adaption techniques can be leveraged to define the parameter reconciliation function, effectively addressing our current challenge. Consequently, we define this reconciliation method as LoRR (Low-Rank Reconciliation) in this paper.

Formally, given the parameter vector $\mb{w} \in \mathbbm{R}^{l}$ and a rank hyper-parameter $r$, we partition $\mathbf{w}$ into two sub-vectors and subsequently reshape them into two matrices $\mb{A} \in \mathbbm{R}^{n \times r}$ and $\mb{B} \in \mathbbm{R}^{D \times r}$, each possessing a rank of $r$. These two sub-matrices $\mb{A}$ and $\mb{B}$ help define the low-rank reconciliation function as follows:
\begin{equation}
\psi(\mb{w}) = \mb{A} \mb{B}^\top \in \mathbbm{R}^{n \times D}.
\end{equation}

In implementation, similar to the aforementioned duplicated padding reconciliation, we will solely define $r$ as the hyper-parameter, which in turn determines the desired parameter length $l$ in accordance with the stated constraints. This necessitates imposing certain limitations on these dimension and rank parameters, specifically, $l = (n + D) \times r$.

%---------------------------------------------------

\subsubsection{Hypercomplex Multiplication (HM) based Parameter Reconciliation}

The Kronecker product operator described above can also be directly applied to parameter for defining new reconciliation functions, {\ie} the hypercomplex parameter reconciliation function:
\begin{equation}
\psi(\mb{w}) = \mb{A} \otimes \mb{B} \in \mathbbm{R}^{n \times D}.
\end{equation}
Similar to the aforementioned low-rank reconciliation, both matrices $\mathbf{A}$ and $\mathbf{B}$ are derived from the parameter vector $\mathbf{w}$ through partitioning and subsequent reshaping. However, instead of computing regular matrix multiplication as in low-rank reconciliation, hypercomplex multiplication-based reconciliation suggests computing the Kronecker product of these two parameter matrices instead.

In implementation, to reduce the number of hyper-parameters and accommodate the parameter dimensions, we can maintain the size of matrix $\mathbf{A}$ as fixed by two hyper-parameters $p$ and $q$, {\ie} $\mb{A} \in \mathbbm{R}^{p \times q}$. Subsequently, the desired size of matrix $\mb{B}$ can be directly calculated as $s \times t$, where $s =\frac{n}{p}$ and $t = \frac{D}{q}$. The hyper-parameters $p$ and $q$ need to be divisors of $n$ and $D$, respectively. Since both $\mathbf{A}$ and $\mathbf{B}$ originate from $\mathbf{w}$, the desired parameter length defining $\mathbf{w}$ can be obtained as $l = p \times q + \frac{n}{p} \times \frac{D}{q}$.

%---------------------------------------------------

\subsubsection{Low-Rank Parameterized Hypercomplex Multiplication (LPHM) based Parameter Reconciliation}

% https://arxiv.org/pdf/2106.04647

In the aforementioned hypercomplex multiplication-based parameter reconciliation, ensuring that ``parameters $p$ and $q$ divide both $n$ and $D$'' results in a limited number of choices for $p$ and $q$, typically leading to a small value ({\eg} $p=4$ and $q=8$). Consequently, the size of matrix $\mb{B}$ and the parameter length $l$ can exceed expectations. To further diminish the parameter count, inspired by \cite{Davison2021CompacterEL}, we can additionally transform matrix $B$ into its low-rank representations during the reconciliation definition. Specifically,
\begin{equation}
\psi(\mb{w}) = \mb{A} \otimes \mb{B} = \mb{A} \otimes ( \mb{S} \mb{T}^\top) \in \mathbbm{R}^{n \times D},
\end{equation}
where $\mb{S} \in \mathbbm{R}^{\frac{n}{p} \times r}$ and $\mb{T} \in \mathbbm{R}^{\frac{D}{q} \times r}$ represent the low-rank matrices for composing $\mb{B}$. This parameter fabrication approach is coined as the {Low-Rank Parameterized Hypercomplex Multiplication} (LPHM) based parameter reconciliation, enabling further reduction of the required parameter vector length to $l = p \times q + r( \frac{n}{p} + \frac{D}{p} )$.

%---------------------------------------------------

\subsubsection{Dual Low-Rank Parameterized Hypercomplex Multiplication (Dual LPHM) based Parameter Reconciliation}

% https://arxiv.org/pdf/2106.04647

To provide {\our} with more flexibility in defining the parameter reconciliation function, based on the above LPHM function, we further introduce the dual LPHM parameter reconciliation function allowing low-rank representations of both sub-matrices $\mb{A}$ and $\mb{B}$, {\ie}
\begin{equation}
\psi(\mb{w}) = \mb{A} \otimes \mb{B} = ( \mb{P} \mb{Q}^\top) \otimes ( \mb{S} \mb{T}^\top) \in \mathbbm{R}^{n \times D},
\end{equation}
where $\mb{P} \in \mathbbm{R}^{p \times r}$ and $\mb{Q} \in \mathbbm{R}^{q \times r}$ represent the low-rank matrices for composing $\mb{A}$, and $\mb{S} \in \mathbbm{R}^{\frac{n}{p} \times r}$ and $\mb{T} \in \mathbbm{R}^{\frac{D}{q} \times r}$ represent the low-rank matrices for composing $\mb{B}$. This parameter fabrication approach is named as the Dual {Low-Rank Parameterized Hypercomplex Multiplication} (Dual LPHM) based parameter reconciliation, which reduces the required parameter vector length to $l = r( p + q +\frac{n}{p} + \frac{D}{q} )$.

%---------------------------------------------------

\subsubsection{HyperNets based Parameter Reconciliation}

% https://arxiv.org/pdf/1609.09106
% https://arxiv.org/pdf/2106.04489
% https://arxiv.org/pdf/1906.00695

Besides these above matrix fabrication based reconciliations defined above, another viable approach for parameter reconciliation is through the utilization of hypernets \cite{Ha2016HyperNetworks, Mahabadi2021ParameterefficientMF}. These works employ a hypernet model, such as a Multi-Layer Perceptron (MLP), to project the input parameter vector $\mathbf{w} \in \mathbbm{R}^l$ from length $l$ to a significantly higher-dimensional output:
\begin{equation}
\psi(\mb{w}) = \text{HyperNet}(\mb{w}) = \mb{W} \in \mathbbm{R}^{n \times D}.
\end{equation}
To circumvent the introduction of additional parameters and learning costs, we can initialize a hypernet model randomly and then freeze its parameters. Subsequently, we utilize this frozen hypernet model to define the aforementioned parameter reconciliation function. In this approach, the parameter $l$ is not calculated automatically and needs to be manually set up as a hyper-parameter.

Compared to the aforementioned LoRR and Kronecker product-based reconciliation functions, hypernets offer greater flexibility in function definition but also entail higher computational costs, since the desired output parameter length $n \times D$ can be extremely high. We will delve into their performance through extensive experimental studies in the following sections.

%=================================================

\subsection{Remainder Functions}

The {remainder function} $\pi: \mathbbm{R}^m \to \mathbbm{R}^n$ plays a crucial role in ensuring the representation completeness of {\our}. This function provides complementary information that may not be encompassed by the expansion and reconciliation functions alone. In this subsection, we will introduce several different remainder functions that can be employed to construct the {\our} model. These remainder functions have also been summarized in Figure~\ref{fig:function_instances} as well.

%---------------------------------------------------

\subsubsection{Constant Remainder}

The constant remainder function $\pi: \mathbbm{R}^m \to \mathbbm{R}^n$ just projects all inputs to a constant vector, {\ie}
\begin{equation}\label{equ:constant_remainder}
\pi(\mb{x}) = \mb{c} \in \mathbbm{R}^n,
\end{equation}
where $\mb{c}$ is a constant vector. 

Specifically, when the output $\mathbf{c}$ is $\mathbf{0}$, it can be referred to as the \textbf{zero remainder}. This function represents the simplest form of remainder, assuming that the data expansion function and parameter reconciling function already perform well in capture all necessary information about the underly function already (a claim supported by forthcoming experimental results). Additionally, all constant remainder functions require no additional parameters.

%---------------------------------------------------

\subsubsection{Identity Remainder}

Similar to residual learning techniques employed in contemporary deep learning models \cite{He2015DeepRL}, we can define the remainder function $\pi: \mathbbm{R}^m \to \mathbbm{R}^n$ as an identity function based on the input. For instance, when $m = n$, we define the {identity remainder function} as follows:
\begin{equation}
\pi(\mb{x}) = \mb{x} \in \mathbbm{R}^n \text{, or } \pi(\mb{x}) = \sigma(\mb{x}) \in \mathbbm{R}^n,
\end{equation}
where notation $\sigma$ denotes the (optional) activation function, which can be {sigmoid}, {ReLU} and the recent {SiLU} \cite{Elfwing2017SigmoidWeightedLU}.

%---------------------------------------------------

\subsubsection{Linear Remainder}

Meanwhile, when the dimensions of the input and output spaces differ, {\ie} $m \neq n$, we must introduce additional parameters into the remainder function to adjust for this mismatch. Here, we introduce the {linear remainder function} as follows:
\begin{equation}
\pi(\mb{x}) = \mb{x} \mb{W}' \in \mathbbm{R}^n \text{, or } \pi(\mb{x}) = \sigma(\mb{x} \mb{W}') \in \mathbbm{R}^n.
\end{equation}
Similarly, the notation $\sigma$ denotes the optional activation function. Concurrently, the learnable parameter matrix $\mb{W}' \in \mathbbm{R}^{m \times n}$ is used here for vector dimension adjustment. For the parameter $\mb{W}'$, we add the prime symbol to differentiate it from the parameter used in the reconciliation function.

%---------------------------------------------------

\subsubsection{Complementary Expansion based Remainder}

%For most of the cases, the above remainder functions can build {\our} that generally meet our requirements. Meanwhile, adding learnable parameters into the remainder function actually provides {\our} with more model designing flexibility and capacity. In addition to the above remainder functions in simple forms, in this paper, we also allows {\our} to define the remainder function as an extra {\our} head coupled with a zero remainder.

For the majority of scenarios, the aforementioned remainder functions are sufficient for constructing {\our} models that generally fulfill our requirements. However, incorporating learnable parameters into the remainder function enhances the flexibility and capacity of {\our} models. In addition to the aforementioned simple forms of remainder functions, this paper also allows for the definition of the remainder function in an augmented manner, as a {\our} head coupled with a zero remainder.

Formally, the complementary expansion based remainder function can be defined as follows:
\begin{equation}
\pi(\mb{x} | \mb{w}') = \left\langle \kappa'(\mb{x}), \psi'(\mb{w}') \right\rangle + \underbrace{\pi'(\mb{x})}_{\pi'(\mb{x}) = \mb{0} \text{ by default}}.
\end{equation}

To distinguish the notations used in defining the original {\our} model, we will add the ``prime'' symbol to functions and parameters, indicating that they are defined within the complementary expansion. All the previously mentioned data expansion functions, parameter reconciliation functions, and remainder functions can be utilized to define the functions $\kappa'$, $\psi'$, and $\pi'$.

Since {\our} itself permits multi-head and multi-channel model architecture design within each layer, the performance of the complementary expansion-based remainder function can be equivalently represented through a multi-head {\our} layer assisted by the zero remainder. Meanwhile, this type of remainder function offers greater flexibility in model design, particularly for learning scenarios necessitating more potent remainders. At the same time, when employing the complementary expansion remainder function, it's customary to set the remainder function $\pi'$ as the zero remainder by default. This ensures that it doesn't introduce unnecessary redundancy in model design by serving as another complementary expansion-based remainder.

%--------------------------------------------------------------------------
\section{Unifying Existing Base Models with {\our} Canonical Representation}\label{sec:backbone_unification}

Building with the component functions outlined in the previous section, the {\our} model has versatile model architecture and attains superior modeling capability. Through strategic combinations of these component functions, we can establish a multi-head, multi-channel, and multi-layer framework, providing a unified basis for representing several influential base models such as Bayesian network, Markov network, kernel SVM, MLP, and KAN. 

In the previous Section~\ref{sec:function_learning}, we have already provided the brief comparisons of {\our} with these base models in terms of mathematical theorem foundation, formula and model architecture. This section further illustrates how, by selecting specific component functions, each of these models can be consolidated into {\our} canonical representation, characterized by the inner product of a data expansion function with a parameter reconciliation function, complemented by a remainder function. The following Figures~\ref{fig:backbone_representation_1} and~\ref{fig:backbone_representation_2} also demonstrate the unified representations of these base models with {\our} model.

%=================================================

\subsection{Unifying MLP with {\our}}\label{subsec:mlp}

%------------------------------------
\begin{figure*}[t]
    \begin{minipage}{\textwidth}
    \centering
    	\includegraphics[width=\linewidth]{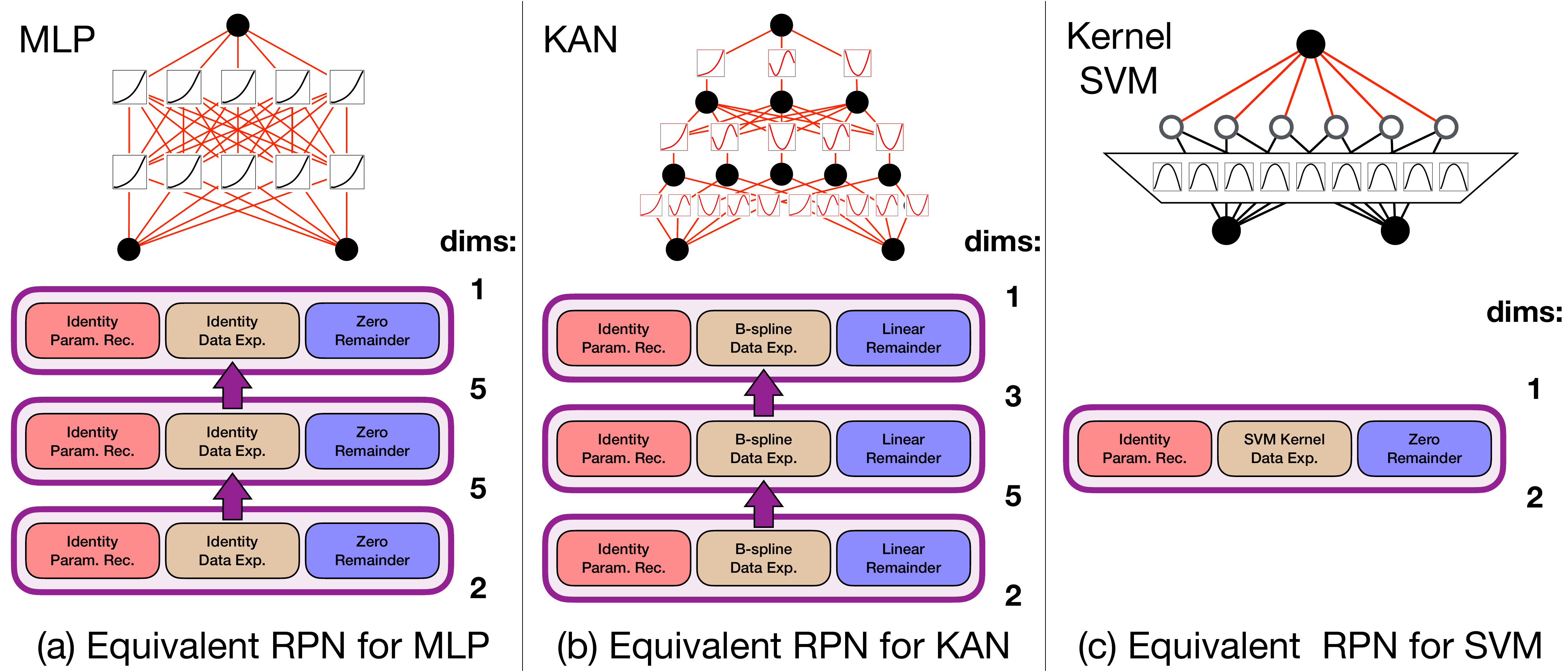}
    	\caption{An illustration of representing MLP, KAN and kernel SVM with {\our}.}
    	\label{fig:backbone_representation_1}
    \end{minipage}%
\end{figure*}
%------------------------------------

In this subsection, we will introduce the {Multi-Layer Perceptron (MLP)} model designed based on the {Universal Approximation Theorem}, and discuss how to represent MLP into the unified representation with the {\our} model.

%---------------------------------------------------

\subsubsection{Universal Approximation Theorem}

Before talking about the MLP model and representing MLP with {\our}, we will first introduce the Universal Approximation Theorem as follows.

\begin{theorem}
(Universal Approximation Theorem): Given a continuous multivariate $f: \mathbbm{R}^m \to \mathbbm{R}^n$, we can approximate $f$ with a series summation of function $\sigma: \mathbbm{R} \to \mathbbm{R}$. Function $\sigma$ is not polynomial if and only if for every $m, n \in \mathbbm{N}$ and $\epsilon > 0$, there exists $k \in \mathbbm{N}$, $\mb{A} \in \mathbbm{R}^{k \times m}$, $\mb{b} \in \mathbbm{R}^k$ and $\mb{C} \in \mathbbm{R}^{n \times k}$ defining function
\begin{equation}
g(\mb{x} | \mb{w}) = \mb{C} \cdot \sigma\left( \mb{A} \mb{x} + \mb{b} \right),
\end{equation}
that can approximate function $f$ with an error no greater than $\epsilon$, {\ie}
\begin{equation}
\sup_{\mb{x} \in \mathbbm{R}^m} \left\| f(\mb{x}) - g(\mb{x} | \mb{w}) \right\| < \epsilon.
\end{equation}

The parameter vector $\mb{w}$ covers all the aforementioned coefficient matrices $\mb{A}$, $\mb{C}$ and bias vector $\mb{b}$.
\end{theorem}

%---------------------------------------------------

\subsubsection{Multi-Layer Perceptron (MLP)}

Built upon the {Universal Approximation Theorem}, MLP proposes to approximate the function $f: \mathbbm{R}^m \to \mathbbm{R}^n$ by stacking neuron layers on top of each other, where each neuron sums up the accumulated inputs and generates the output through an activation function $\sigma: \mathbbm{R} \to \mathbbm{R}$ as follows:
\begin{equation}
g(\mb{x} | \mb{w}) = \mb{W}_1 \sigma \left( \mb{W}_2 \mb{x} + \mb{b} \right),
\end{equation}
where $\mb{w} = (\mb{W}_1, \mb{W}_2, \mb{b})$ covers the matrices $\mb{W}_1 \in \mathbbm{R}^{n \times k}$ and $\mb{W}_2 \in \mathbbm{R}^{k \times m}$ as the weights and $\mb{b} \in \mathbbm{R}^{k}$ as the bias. As to the activation function $\sigma$, many different types of activation functions have been proposed already, ranging from the classic binary-step and sigmoid function to the recent SiLU and dSiLU \cite{Elfwing2017SigmoidWeightedLU}.

%---------------------------------------------------

\subsubsection{Representing MLP with {\our}}

The MLP model can be easily represented with {\our} involving the \textit{identity data expansion function}, \textit{identity parameter reconciliation function} and \textit{zero remainder function} introduced in the previous Section~\ref{sec:functions}. 

Specifically, as depicted in Plot (a) of Figure~\ref{fig:backbone_representation_1}, we illustrate the representation of a three-layer MLP model at the top, and its corresponding representation with {\our} involving three {\our}-layers at the bottom, with the input and output dimensions indicated on the right-hand side.
\begin{itemize}
\item \textbf{{\our} Layer 1}: A single-head, single-channel {\our} layer consisting of (1) an identity data expansion function (with sigmoid as the optional output-processing function), (2) an identity parameter reconciliation function, and (3) a zero remainder function;
\item \textbf{{\our} Layer 2}: A single-head, single-channel {\our} layer consisting of (1) an identity data expansion function (with sigmoid as the optional output-processing function), (2) an identity parameter reconciliation function, and (3) a zero remainder function;
\item \textbf{{\our} Layer 3}: A single-head, single-channel {\our} layer consisting of (1) an identity data expansion function, (2) an identity parameter reconciliation function, and (3) a zero remainder function.
\end{itemize}

%By comparing {\our} with MLP, we observe that {\our} is more general and has greater learning capacity than MLP, which can be reduced to MLP by defining the \textit{data expansion function}, \textit{parameter reconciliation function} and \textit{remainder function} shown in Equation~(\ref{equ:deep_rpn}) as follows:
%\begin{itemize}
%\item Let $\kappa_k: \mathbbm{R}^{d_{k-1}} \to \mathbbm{R}^{d_k}$ for the $k_{th}$ layer with the expansion space with the identical dimension as the input space ({\ie} $d_{k-1} = d_k$), we can define
%\begin{equation}
%\kappa_1(\mb{h}_0) = \mb{h}_0 \text{, and } \kappa_k(\mb{h}_k) = \sigma(\mb{h}_k), \forall k \in \{1,2, \cdots, K\},
%\end{equation}
%where $\sigma$ denotes the activation function.
%\item Let $\psi_k: \mathbbm{R}^{d_k \times d_k} \to \mathbbm{R}^{d_k \times d_{k}}$, we can define
%\begin{equation}
%\psi_k(\mb{w}_k) = \mb{w}_k, \forall k \in \{1,2, \cdots, K\}.
%\end{equation}
%\item Let the remainder function with output to be a constant zero, {\ie}
%\begin{equation}
%\pi(\mb{x}) = \mb{0}.
%\end{equation}
%\end{itemize}

%=================================================

\subsection{Unifying KAN with {\our}}

The Kolmogorov-Arnold Network (KAN) \cite{Liu2024KANKN} is a new base model introduced recently in 2024, designed based on the Kolmogorov-Arnold RepresentationTheorem. Diverging from MLP's fixed activation functions, KAN suggests learning the activation functions for pairwise neuron connections using B-spline interpolation. Here, we will briefly introduce the Kolmogorov-Arnold Representation Theorem and subsequently discuss how to represent the KAN model architecture with {\our}.

%---------------------------------------------------

\subsubsection{Kolmogorov-Arnold Representation Theorem}

Kolmogorov-Arnold Representation Theorem posits that any multivariate continuous function can be expressed as a composition of the two-argument addition of continuous univariate functions. 

\begin{theorem}
(Kolmogorov-Arnold Representation Theorem): Formally, given a continuous multivariate function on a bounded domain, {\eg} $f: [0, 1]^m \to \mathbbm{R}$, the function $f$ can be written as a finite composition of continuous univariate functions and the binary operation of addition:
\begin{equation}
f(\mb{x}) = f([x_1, x_2, \cdots, x_m]^\top) = \sum_{q=0}^{2m} \phi_q (\sum_{p=1}^m \phi_{q,p}(x_p)),
\end{equation}
where $\phi_{q,p}: [0,1] \to \mathbbm{R}$ and $\phi_q: \mathbbm{R} \to \mathbbm{R}$.
\end{theorem}

%---------------------------------------------------

\subsubsection{Kolmogorov-Arnold Network (KAN)}

Based on the Kolmogorov-Arnold Representation Theorem mentioned above, a recent paper \cite{Liu2024KANKN} introduces the KAN model. The theorem imposes a specific constraint on the required function numbers, namely $2m+1$ and $m$. However, in the practical implementation of KAN, these constraints are relaxed, allowing the KAN model to employ a deep architecture by stacking multiple layers on top of each other to approximate the function $f: \mathbbm{R}^m \to \mathbbm{R}^n$, as indicated below:
\begin{equation}
g(\mb{x} | \mb{w}) = \text{KAN}(\mb{x}) = \Phi^K \circ \Phi^{K-1} \circ \cdots \circ \Phi^1 (\mb{x}).
\end{equation}
Here, the notation $\Phi^k = \{\phi^k_{i,j}\}_{i \in \{1, 2, \cdots, d_{in}\}, j \in \{1, 2, \cdots, d_{out}\}}$ represents a matrix of function $\phi^k_{i,j}$ with learnable parameters at the $k_{th}$ layer of the model, and $d_{in}$, $d_{out}$ are the corresponding input and output dimensions. Operator $\circ$ denotes the function composition of sequential layers in KAN.

In implementation, the function $\phi$ with learnable parameters is formally defined as
\begin{equation}
\phi(x) = b(x) + \text{spline}(x),
\text{ where }
\begin{cases}
&b(x) = \mbox{SiLU}(x) = \frac{x}{1 + \exp^{-x}},\\
&\text{spline}(x) = \sum_{i} w_i B_{i,d}(x).
\end{cases}
\end{equation}
In the above equation, the ``$\text{spline}(\cdot)$'' function is defined as a linear combination of B-splines of degree $d$, denoted by $\{ B_{i,d}(x) \}_i$, where $\{w_i\}_{i}$ represents the set of learnable parameters. Meanwhile, the base function $b(\cdot)$ is defined as the SiLU function based on the inputs. 

%---------------------------------------------------

\subsubsection{Representing KAN with {\our}}

Similar as MLP, the KAN model can also be easily represented with {\our} involving the \textit{B-spline data expansion function}, \textit{identity parameter reconciliation function} and \textit{linear remainder function} introduced in the previous Section~\ref{sec:functions}. 

Specifically, as depicted in Plot (b) of Figure~\ref{fig:backbone_representation_1}, for the three-layer KAN model illustrated at the top, its corresponding representation with {\our} involving three {\our}-layers at the bottom:
\begin{itemize}
\item \textbf{{\our} Layer 1}: A single-head, single-channel layer consisting of (1) a B-spline data expansion function, (2) an identity parameter reconciliation function, and (3) a linear remainder function (with SiLU as the activation function);
\item \textbf{{\our} Layer 2}: A single-head, single-channel layer consisting of (1) a B-spline data expansion function, (2) an identity parameter reconciliation function, and (3) a linear remainder function (with SiLU as the activation function);
\item \textbf{{\our} Layer 3}: A single-head, single-channel layer consisting of (1) a B-spline data expansion function, (2) an identity parameter reconciliation function, and (3) a linear remainder function (with SiLU as the activation function).
\end{itemize}

\subsection{Unifying Kernel SVM with {\our}}

Support vector machine (SVM) is a renowned supervised machine learning model introduced for data classification and regression analysis. While SVM is adept at linearly separating instances, it can also perform non-linear classifications through the utilization of kernel tricks.

%---------------------------------------------------

\subsubsection{Support Vector Machine (SVM)}\label{subsubsec:svm}

Support Vector Machine (SVM) is a non-probabilistic binary linear classifier model proposed based on statistical machine learning.  Various extensions enable SVM to operate in multi-class and multi-label classification scenarios.

Below, we will introduce SVM within the classic binary classification learning settings and employ it to approximate the underlying mapping, represented as $f: \mathbbm{R}^m \to \{-1, +1\}$ as follows:
\begin{equation}\label{equ:svm_model}
g(\mb{x} | \mb{w}, b) = \text{sign}(\mb{w}^\top \mb{x} + b),
\end{equation}
where $\text{sign}(\cdot)$ returns the polarity of the input term $\mb{w}^\top \mb{x} + b$, while $\mb{w}$, $b$ denote the learnable parameters.

\subsubsection{Kernel Tricks}

The SVM model described above is highly effective for linear classification tasks. However, for handling nonlinear tasks, techniques such as the kernel trick have been introduced. The kernel trick is widely used in regression, classification, and PCA. It facilitates the embedding of the problem into higher-dimensional spaces, often even infinite-dimensional ones, without the need for an infinite amount of computational effort.

Formally, given a vector $\mb{x} \in \mathbbm{R}^m$, the kernel trick introduces a feature mapping $\phi: \mathbbm{R}^m \to \mathbbm{R}^M$ to project the vector into a higher-dimensional space, denoted as $\phi(\mb{x}) \in \mathbbm{R}^M$, where $M > m$. An illustrative example of such a mapping function used in kernel tricks is presented below.

\begin{example}\label{exp:svm_example}

%%------------------------------------
%\begin{figure*}[t]
%    \begin{minipage}{\textwidth}
%    \centering
%    	\includegraphics[width=1.0\linewidth]{figures/method/svm.pdf}
%    	\caption{An example of the kernel tricks used in SVM for projecting data instances that are not linearly separable to a higher-order space for linear separation.}
%    	\label{fig:kernel_instance}
%    \end{minipage}%
%\end{figure*}
%%------------------------------------

For instance, given a vector $\mb{x} = [x_1, x_2]^\top \in \mathbbm{R}^2$, a very simple mapping shown below will project $\mb{x}$ from $\mathbbm{R}^2$ to a high-order dimension $\mathbbm{R}^3$:

\begin{equation}
\phi([x_1, x_2]^\top) = [x_1^2, \sqrt{2}x_1x_2, x_2^2]^\top.
\end{equation}

%As shown in Figure~\ref{fig:kernel_instance}, for the data instances that are not linearly separable in the original space, by projecting them to higher-order space via the above mapping, they will become linearly separable.
\end{example}

For data instances that are not linearly separable in the original space, projecting them using the mapping described above renders them linearly separable by the model represented in Equation~(\ref{equ:svm_model}) within the higher-dimensional space. The process of learning kernel SVM to obtain the parameters is beyond the scope of this paper, and will not be discussed here.

\subsubsection{Representing Kernel SVM with {\our}}

Unlike MLP and KAN, which employ deep architectures, the SVM model is typically proposed with a shallow architecture consisting of a single layer. Specifically, for the kernel SVM illustrated in Plot (c) of Figure~\ref{fig:backbone_representation_1}, it can be represented within {\our} with a single layer:
\begin{itemize}
\item \textbf{{\our} Layer}: A single-head, single-channel {\our} layer consisting of (1) a data expansion function corresponding to the kernel function ({\eg} linear or RBF), (2) an identity parameter reconciliation function, and (3) a zero remainder function.
\end{itemize}

%The {\our} may have a deep architecture, while SVM with the kernel tricks normally has no deep architecture. Furthermore, if we define the \textit{data expansion function} used in {\our} as the higher-order projection function used in SVM, the {\our} can also be reduced to SVM with kernel tricks by define the functions in the model as follows:
%\begin{itemize}
%\item Let $\kappa_1: \mathbbm{R}^{m} \to \mathbbm{R}^{D}$, we can define
%\begin{equation}
%\kappa_1(\mb{h}_0) = \phi(\mb{h}_0),
%\end{equation}
%where $\phi$ is the higher-order projection function mentioned above for SVM.
%\item Let $\psi_1: \mathbbm{R}^{l} \to \mathbbm{R}^{l}$, we can define
%\begin{equation}
%\psi_1(\mb{w}_1) = \mb{w}_1.
%\end{equation}
%\item Let the remainder function with output to be a constant zero, {\ie}
%\begin{equation}
%\pi(\mb{x}) = \mb{0}.
%\end{equation}
%\end{itemize}
%In other words, if we also define the decision function with the signs of the inner product between the reconciled parameter with the expanded data, similar kernel tricks used in SVM can also be applied to the learning {\our}.

%=================================================

\subsection{Unifying PMs with {\our}}\label{subsec:pm}

Probabilistic model denotes a broad family of statistical machine learning models build on probability theory. In this subsection, we will provide a brief introduction of the probabilistic models, followed by a discussion on representing these models using the {\our} model.

%---------------------------------------------------

\subsubsection{Probabilistic Models (PMs)}

% These models are inherently quantitative, capable of projecting not just a single outcome but a spectrum of possibilities. Representative examples of PMs include \textit{logistic regression}, \textit{Bayesian classifiers}, \textit{hidden markov models}, \textit{Bayesian networks} and even neural networks with \textit{softmax} can also be viewed as a PM. In this part, we will briefly introduce how PM works and then talk about how to represent the PMs with the {\our} model in the next subsection.

Probabilistic models (PMs) assume the relationships among the variables can be effectively modeled with probability distributions. In PMs, we use the upper case notations, such as $X_i$ and $Y_j$, to represent variables and lower case ones, such as $x_i$ and $y_j$, to represent their respective values.

Formally, to infer the underlying mapping $f: \mathbbm{R}^m \to \mathbbm{R}^n$, we represent the inputs and outputs as random variables $X_1, X_2, \cdots, X_m$ and $Y_1, Y_2, \cdots, Y_n$, respectively. The potential value spaces of the input and output are denoted as $\mc{X} \subset \mathbbm{R}^m$ and $\mc{Y} \subset \mathbbm{R}^n$. Given an input instance $[x_1, x_2, \cdots, x_m]^\top \in \mc{X}$, the underlying model $f$ will project it to the outputs $[y_1, \cdots, y_n]^\top \in \mc{Y}$ that maximize the following probability:
\begin{equation}\label{equ:probabilistic_models}
\max P\left(Y_1=y_1, \cdots, Y_n=y_n | X_1=x_1, \cdots, X_m=x_m\right),
\end{equation}
where $P(\cdot | \cdot)$ denotes the conditional probability defined based on certain distributions.

We can calculate the above conditional probability by dividing the joint probabilities of the random variables. Furthermore, by applying the logarithm operator to the probabilities, we can rewrite the above conditional probability calculation as follows (simplifying the notations to include only two variables, $X$ and $Y$):
\begin{equation}\label{equ:probabilistic_models_2}
\begin{aligned}
\log P({Y} = {y} | {X} = {x}) &= \log \left( \frac{P({Y} = {y} \land {X} = {x})}{P({X} = {x})} \right)\\
&= \log P({Y} = {y} \land {X} = {x}) - \log P({X} = {x}).
\end{aligned}
\end{equation}

The fundamental problem studied in PMs is how to calculate the joint probabilities $P(Y = y \land X = x)$ and $P(X = x)$, which involve multiple random variables. In PMs, we treat input and output random variables ({\ie} $Y$ and $X$) equally. To simplify notations, we will just illustrate how to calculate joint probabilities with $m$ random variables $X_1, X_2, \cdots, X_m$ below, noting that joint probabilities involving random variables from both $X$s and $Y$s can be calculated similarly.

\subsubsection{Naive Bayes and Probabilistic Graphical Models (PGMs)}

Numerous machine learning models fall under the category of PMs. Below, we will list three representative PMs: naive Bayes, Bayesian network, and Markov network, with Bayesian network and Markov network also commonly referred to as probabilistic graphical models. We will also demonstrate how to represent them using the {\our} model.

%Among the existing PMs, most of them can be classified into two major categories: \textit{Bayesian network} and \textit{Markov network}, both of which can represent and reason about the dependencies among a set of random variables, {\eg} $X_1, X_2, \cdots, X_m$. Below, we will briefly introduce how \textit{Bayesian network} and \textit{Markov network} model and calculate the joint probability to illustrate their differences.

\subsubsubsection{\textbf{Naive Bayes}}: The naive Bayes classifier operates under the assumption that, given the target class, the features of input data instances are conditionally independent. This assumption allows us to reformulate the above Equation~(\ref{equ:probabilistic_models}) and Equation~(\ref{equ:probabilistic_models_2}) as follows:
\begin{equation}
P (Y = y | X_1 = x_1, \cdots, X_m = x_m) \propto P(Y=y) \prod_{i=1}^m P(X_i=x_i | Y=y),
\end{equation}
and
\begin{equation}
\begin{aligned}
\log P (Y = y | X_1 = x_1, \cdots, X_m = x_m) \propto \log P(Y=y) + \sum_{i=1}^m \log P(X_i=x_i | Y=y).
\end{aligned}
\end{equation}

The stringent independence assumption restricts the applicability of naive Bayes to a narrow range of scenarios. In contrast, both Bayesian network and Markov network, which will be introduced below shortly, aim to capture the dependency relationships among the feature variables instead.

\subsubsubsection{\textbf{Bayesian network}}: Bayesian network assumes that the relationships between variables can be represented as a directed acyclic graph, wherein each node ({\ie} the variables) possesses directed edges pointing to its children, indicating a direct influence or causal relationship between them. These networks encode conditional dependencies using directed edges, while the joint probability distribution is factored as a product of conditional probabilities. 

Formally, based on Bayesian network, the joint log-probability distribution $P(X_1, X_2, \cdots, X_m)$ involving these $m$ random variables can be written as:
\begin{equation}\label{equ:bayesian_network}
\begin{aligned}
\log P(X_1, X_2, \cdots, X_m) &= \log \left(\prod_{i=1}^m P(X_i | \Gamma(X_i)) \right),\\
&= \sum_{i=1}^m \log P(X_i, \Gamma(X_i)) - \log P(\Gamma(X_i)),
\end{aligned}
\end{equation} 
where $\Gamma(X_i)$ denotes the set of parent nodes of $X_i$ in the probabilistic dependency DAG.

%Given the random variables $X_1, X_2, \cdots, X_m$, Bayesian networks assume the relationships between variables are represented as a Directed Acyclic Graph (DAG), where each node ({\ie} the variables) has directed edges pointing to its children, indicating a direct influence or causal relationship between them. Each node in the DAG can have multiple parents and children. For instance, if there is an edge from variable $X_i$ to variable $X_j$, it indicates that $X_i$ directly influences $X_j$, and we can say $X_i$ is a parent of $X_j$'s, which can also be represented as $X_i \in \Gamma(X_j)$, where $\Gamma(X_j)$ denotes the set of ``parents'' of $X_j$. If random variable $X_j$ doesn't depend on any other variables, its parent set will be $\Gamma(X_j) = \emptyset$ and we can write $P(X_j | \Gamma(X_j)) = P(X_j)$.

%Bayesian networks encode conditional dependencies using directed edges. The joint probability distribution is factored as a product of conditional probabilities. Specifically, the joint log-probability distribution $P(X_1, X_2, \cdots, X_m)$ involving these $m$ random variables can be written as:
%\begin{equation}\label{equ:bayesian_network}
%\begin{aligned}
%\log P(X_1, X_2, \cdots, X_m) &= \log \left(\prod_{i=1}^m P(X_i | \Gamma(X_i)) \right),\\
%&= \sum_{i=1}^m \log P(X_i, \Gamma(X_i)) - \log P(\Gamma(X_i)),
%\end{aligned}
%\end{equation} 
%where $\Gamma(X_i)$ denotes the set of parent nodes of $X_i$ in the DAG.

\subsubsubsection{\textbf{Markov network}}: Markov network, on the other hand, assumes that the relationships between variables can be represented as an undirected graph, where edges indicate a mutual dependency between nodes ({\ie} the variables), without implying any directionality or causation.

Formally, given the random variables $X_1, X_2, \cdots, X_m$, Markov network defines their relations as an undirected graph $G$, which can be divided into cliques (fully connected subsets of nodes) $\mc{C} = \left\{\mc{C}_1, \mc{C}_2, \cdots, \mc{C}_k \right\}$ and $\bigcup_{i=1}^k \mc{C}_i = \{X_1, X_2, \cdots, X_m\}$. Based on such cliques, the joint log-probability defined on all the variables can be factored as follows:
\begin{equation}\label{equ:markov_network}
\begin{aligned}
\log P(X_1, X_2, \cdots, X_m) &= \log \left( \frac{1}{Z} \prod_{i=1}^k \phi({X}_{\mc{C}_i}) \right),\\
&= \sum_{i=1}^k \log \phi( {X}_{\mc{C}_i} ) - \log Z,
\end{aligned}
\end{equation}
where $\log Z$ can be viewed as a constant. Regarding the \textit{factor potential functions} in $\{ \phi({X}_{\mc{C}_i}) \}_{\mc{C}_i \in \mc{C}}$, several different methods exist for defining them, such as the \textit{appearance count in the training set}, \textit{exponential potential function} and \textit{Gaussian potential function}.

\subsubsection{Representing PMs with {\our}}

%------------------------------------
\begin{figure*}[t]
    \begin{minipage}{\textwidth}
    \centering
    	\includegraphics[width=\linewidth]{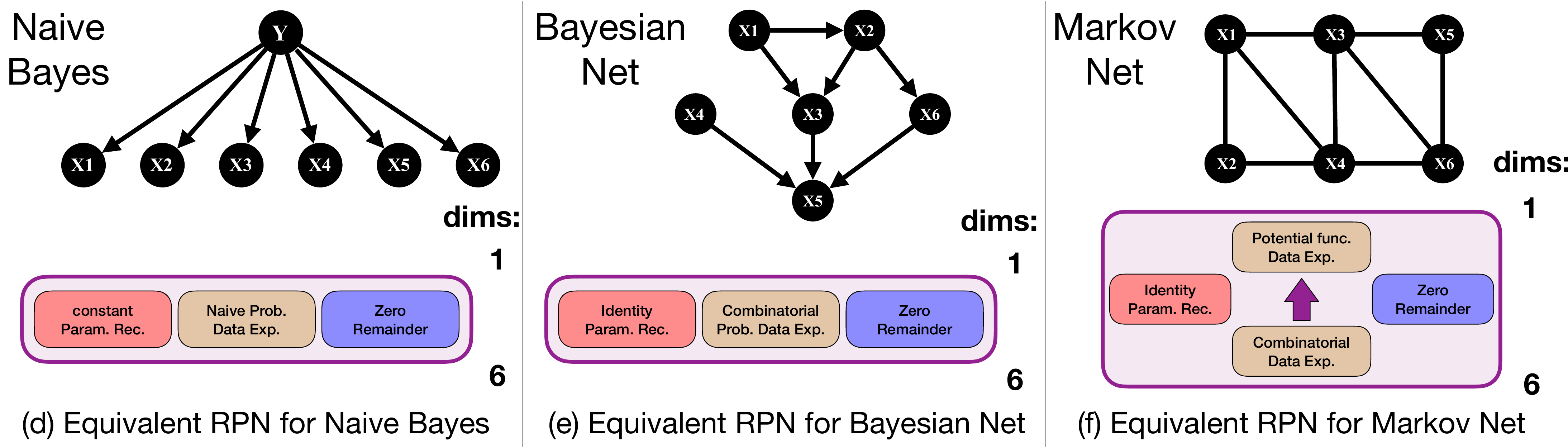}
    	\caption{An illustration of representing different probabilistic models with {\our}.}
    	\label{fig:backbone_representation_2}
    \end{minipage}%
\end{figure*}
%------------------------------------

Similar to SVM, representing naive Bayes, Bayesian network and Markov network with {\our} will also involve one single layer. However, because these probabilistic models are designed based on different assumptions, the component functions involved will also be distinct. As illustrated in Figure~\ref{fig:backbone_representation_2}, these probabilistic models can be depicted using {\our} as follows:
\begin{itemize}
\item \textbf{Naive Bayes}: A single-head, single-channel {\our} layer consisting of (1) a naive probabilistic data expansion function, (2) a constant parameter reconciliation function (containing constant ones), and (3) a zero remainder function.

\item \textbf{Bayesian Network}: A single-head, single-channel {\our} layer consisting of (1) a combinatorial probabilistic data expansion function, (2) an identity parameter reconciliation function, and (3) a zero remainder function.

\item \textbf{Markov Network}: A single-head, single-channel {\our} layer consisting of (1) a nested data expansion function composed of the combinatorial expansion followed by the ``factor potential function'' based expansion function, (2) an identity parameter reconciliation function, and (3) a zero remainder function.
\end{itemize}

\section{Empirical Evaluations of {\our}}\label{sec:experiments}

This section presents empirical evaluations of {\our} across various deep function learning tasks with extensive experiments on real-world benchmark datasets. We examine several key performance aspects of {\our} and organize our insightful findings as follows. In Section~\ref{subsec:continuous_function}, we provide experimental investigations of {\our} for continuous function learning on three datasets: elementary functions, composite functions, and Feynman functions. Section~\ref{subsec:discrete_classification} evaluates {\our} for discrete vision and language data classification, using image datasets (MNIST and CIFAR-10) and text datasets (IMDB, AGNews, and SST2). To assess {\our} for probabilistic dependency relationship inference, Section~\ref{subsec:dependency_inference} presents experiments on three classic tabular datasets: Iris Species, Pima Indians Diabetes, and Banknote. Throughout these subsections, we also provide experimental analysis of {\our} in terms of convergence, parameter sensitivity, ablation studies, interpretation, and visualization for the specific deep function learning tasks.

%=================================================

\subsection{Continuous Function Approximation}\label{subsec:continuous_function}

As previously described, {\our} can serve as a base model for effective continuous function approximation. In this section, we investigate the empirical effectiveness of {\our} using three continuous function datasets. We begin by introducing the datasets and experimental setups, followed by a detailed presentation of the experimental results and performance analysis.

%=======================================

\subsubsection{Dataset Descriptions and Experiment Setups}

%---------------------------------------------------------------
\begin{table}[h!]
\centering
\small
\caption{Statistics of continuous function datasets used in the experiments. For the Feynman function dataset, the input variable numbers can be different for different functions, which are not provided in the table. For each function in the dataset, we randomly generate $2,000$ input-output pairs according to the input variable value ranges, which are partitioned into the training and testing sets according to the $50:50$ ratio.}
\label{tab:continuous_dataset_statistics}
\begin{tabular}{|c|c|c|c|}
\hline
\multirow{2}{*}{} & \multicolumn{3}{c|}{\textbf{Continuous Function Datasets}} \\ \cline{2-4}
                          & \textbf{Elementary Functions} & \textbf{Composite Functions} & \textbf{Feynman Functions} \\ \hline
Equ. \#                & 17 & 17 &100 \\         \hline
Train \#                     & 1,000  & 1,000  & 1,000  \\ \hline
Test \#                     & 1,000  & 1,000  & 1,000  \\ \hline
Input Dim.                     & 2  & 2  & --   \\ \hline
Output Dim.                     & 1  & 1  & 1   \\ \hline
\end{tabular}
\end{table}
%---------------------------------------------------------------

%---------------------------------------------------------------
%---------------------------------------------------------------
\begin{table}[t]
    \caption{Experimental results of continuous function approximation on the elementary function dataset. All these models are trained with $2,000$ epochs, which guarantee their convergence, and the best testing scores achieved by all these methods within these $2,000$ epochs are cherry-picked, aiming to eliminate the impacts of epoch selection on the result evaluation. All these models are all trained with $5$ different random seeds, and the average scores together with the standard deviations are reported in the table. For the method names, {\our}-Ext denotes {\our} with extended expansion function (involving Taylor's expansion and Bspline expansion), low-rank reconciliation function and Zero remainder function; while {\our}-Nstd denotes {\our} with nested expansion function (involving Taylor's expansion and Bspline expansion), low-rank reconciliation function and Zero remainder function.}
    \label{tab:elementary_comparison}
    \tiny
    \centering
    \setlength{\tabcolsep}{3pt}
    \renewcommand{\arraystretch}{2}
    {\fontsize{6}{6}\selectfont
    \scalemath{0.8}{
    \begin{tabular}{|c|c|c||c|c||c|c||c|c||c|c|}
        \hline
        \thead{Eq.} & \thead{Formula} & \thead{Variables} & \thead{MLP\\ Architecture} & \thead{MLP\\ MSE} & \thead{KAN\\ Architecture} & \thead{KAN\\ MSE} & \thead{{\our}-Ext\\ Architecture} & \thead{{\our}-Ext\\ MSE} & \thead{{\our}-Nstd\\ Architecture} & \thead{{\our}-Nstd\\ MSE}\\ 
        \hline
        E.0 & $x+y$ & \makecell{$x, y\in$ \\ $ (0, 1)$} & \makecell{[2, 5, 5, 1]\\ param \#: 51} &\makecell{$6.25 \times 10^{-7}$\\ $ \pm 8.08 \times 10^{-7}$}  & \makecell{[2, 2, 1, 1]\\ param \#: 63}  &\makecell{ $4.23 \times 10^{-7}$ \\ $\pm$  $5.78 \times 10^{-7}$ } &\makecell{[2, 2, 1, 1]\\ param \#: 47}  &\makecell{\boldmath $8.40 \times 10^{-8}$ \\ \boldmath $\pm$  $1.12 \times 10^{-7}$ } &\makecell{[2, 2, 1, 1]\\ param \#: 547}  &\makecell{\textbf{\boldmath $1.93 \times 10^{-8}$} \\ \textbf{\boldmath$\pm$  $1.15 \times 10^{-8}$}} \\ 
        \hline
        E.1 & $\frac{1}{(x+y)}$ & \makecell{$x, y\in$ \\ $ (0, 1)$} &\makecell{[2, 5, 5, 1]\\ param \#: 51} &\makecell{$7.64 \times 10^{-1}$\\ $ \pm 6.05 \times 10^{-1}$}  &\makecell{[2, 2, 1, 1]\\ param \#: 63} &\makecell{\textbf{\boldmath $3.32 \times 10^{-2}$} \\ \textbf{\boldmath$\pm$  $5.47 \times 10^{-2}$} } &\makecell{[2, 2, 1, 1]\\ param \#: 47}  &\makecell{\boldmath $8.67 \times 10^{-2}$\boldmath \\ \boldmath$\pm$  $8.22 \times 10^{-2}$ } &\makecell{[2, 2, 1, 1]\\ param \#: 547}  &\makecell{ $1.03 \times 10^{-1}$ \\ $\pm$  $1.40 \times 10^{-1}$ } \\ 
        \hline
        E.2 & $(x+y)^2$ & \makecell{$x, y\in$ \\ $ (0, 1)$} &\makecell{[2, 5, 5, 1]\\ param \#: 51} &\makecell{$1.89 \times 10^{-3}$\\ $ \pm 1.81 \times 10^{-3}$}  &\makecell{[2, 2, 1, 1]\\ param \#: 63} &\makecell{ $1.32 \times 10^{-6}$ \\ $\pm$  $5.29 \times 10^{-7}$ } &\makecell{[2, 2, 1, 1]\\ param \#: 47}  &\makecell{\boldmath $2.56 \times 10^{-7}$ \\ \boldmath $\pm$  $1.48 \times 10^{-7}$ } &\makecell{[2, 2, 1, 1]\\ param \#: 547}  &\makecell{\textbf{\boldmath $5.06 \times 10^{-8}$} \\ \textbf{\boldmath$\pm$  $3.53 \times 10^{-8}$ }} \\ 
        \hline
        E.3 & $\exp(x+y)$ & \makecell{$x, y\in$ \\ $ (0, 1)$} &\makecell{[2, 5, 5, 1]\\ param \#: 51} &\makecell{$9.33 \times 10^{-3}$\\ $ \pm 1.14 \times 10^{-2}$}  &\makecell{[2, 2, 1, 1]\\ param \#: 63} &\makecell{ $1.22 \times 10^{-5}$ \\ $\pm$  $1.10 \times 10^{-5}$ } &\makecell{[2, 2, 1, 1]\\ param \#: 47}  &\makecell{\boldmath $3.81 \times 10^{-6}$ \\ \boldmath $\pm$  $6.56 \times 10^{-6}$ } &\makecell{[2, 2, 1, 1]\\ param \#: 547}  &\makecell{\textbf{\boldmath $1.26 \times 10^{-6}$} \\ \textbf{\boldmath$\pm$  $1.37 \times 10^{-6}$ }} \\
        \hline
        E.4 & $\ln(x+y)$ & \makecell{$x, y\in$ \\ $ (0, 1)$} &\makecell{[2, 5, 5, 1]\\ param \#: 51} &\makecell{$1.26 \times 10^{-3}$\\ $ \pm 8.24 \times 10^{-4}$}  &\makecell{[2, 2, 1, 1]\\ param \#: 63} &\makecell{\boldmath $3.95 \times 10^{-5}$ \\ \boldmath $\pm$  $3.86 \times 10^{-5}$ } &\makecell{[2, 2, 1, 1]\\ param \#: 47}  &\makecell{ $7.05 \times 10^{-5}$ \\ $\pm$  $3.12 \times 10^{-5}$ } &\makecell{[2, 2, 1, 1]\\ param \#: 547}  &\makecell{\textbf{\boldmath $1.80 \times 10^{-5}$} \\ \textbf{\boldmath $\pm$  $2.37 \times 10^{-5}$ }} \\ 
        \hline
        E.5 & $\sin(x+y)$ & \makecell{$x, y\in$ \\ $ (0, 1)$} &\makecell{[2, 5, 5, 1]\\ param \#: 51} &\makecell{$1.49 \times 10^{-3}$\\ $ \pm 2.50 \times 10^{-3}$}  &\makecell{[2, 2, 1, 1]\\ param \#: 63} &\makecell{\boldmath $2.14 \times 10^{-8}$ \\ \boldmath $\pm$  $9.06 \times 10^{-9}$ } &\makecell{[2, 2, 1, 1]\\ param \#: 47}  &\makecell{ $4.95 \times 10^{-8}$ \\ $\pm$  $4.39 \times 10^{-8}$ } &\makecell{[2, 2, 1, 1]\\ param \#: 547}  &\makecell{\textbf{\boldmath $5.67 \times 10^{-9}$} \\ \textbf{\boldmath$\pm$  $3.20 \times 10^{-9}$ }} \\ 
        \hline
        E.6 & $\cos(x+y)$ & \makecell{$x, y\in$ \\ $ (0, 1)$} &\makecell{[2, 5, 5, 1]\\ param \#: 51} &\makecell{$2.12 \times 10^{-2}$\\ $ \pm 4.23 \times 10^{-2}$}  &\makecell{[2, 2, 1, 1]\\ param \#: 63} &\makecell{ $2.20 \times 10^{-7}$ \\ $\pm$  $2.47 \times 10^{-7}$ } &\makecell{[2, 2, 1, 1]\\ param \#: 47}  &\makecell{\boldmath $1.25 \times 10^{-7}$ \\ \boldmath $\pm$  $1.34 \times 10^{-7}$ } &\makecell{[2, 2, 1, 1]\\ param \#: 547}  &\makecell{\textbf{\boldmath $5.93 \times 10^{-9}$} \\ \textbf{\boldmath $\pm$  $3.98 \times 10^{-9}$ }} \\ 
        \hline
        E.7 & $\tan(x+y)$ & \makecell{$x, y\in$ \\ $ (0, 1)$} &\makecell{[2, 5, 5, 1]\\ param \#: 51} &\makecell{ $2.87 \times 10^{-4}$ \\ $\pm$  $2.97 \times 10^{-4}$ }  &\makecell{[2, 2, 1, 1]\\ param \#: 63} &\makecell{ $3.26 \times 10^{-7}$ \\ $\pm$  $3.27 \times 10^{-7}$ } &\makecell{[2, 2, 1, 1]\\ param \#: 47}  &\makecell{\boldmath $6.02 \times 10^{-8}$ \\ \boldmath $\pm$  $4.46 \times 10^{-8}$ } &\makecell{[2, 2, 1, 1]\\ param \#: 547}  &\makecell{\textbf{\boldmath $1.67 \times 10^{-8}$} \\ \textbf{\boldmath $\pm$  $2.32 \times 10^{-8}$ }} \\ 
        \hline
        E.8 & $\arcsin(x+y)$ & \makecell{$x, y\in$ \\ $ (0, 0.5)$} &\makecell{[2, 5, 5, 1]\\ param \#: 51} &\makecell{$3.84 \times 10^{-4}$\\ $ \pm 2.92 \times 10^{-4}$}  &\makecell{[2, 2, 1, 1]\\ param \#: 63} &\makecell{\textbf{ \boldmath$2.61 \times 10^{-7}$} \\ \textbf{\boldmath$\pm$  $1.08 \times 10^{-7}$ }} &\makecell{[2, 2, 1, 1]\\ param \#: 47}  &\makecell{ $1.65 \times 10^{-6}$ \\ $\pm$  $2.06 \times 10^{-6}$ } &\makecell{[2, 2, 1, 1]\\ param \#: 547}  &\makecell{\boldmath $5.16 \times 10^{-7}$ \\\boldmath $\pm$  $2.97 \times 10^{-7}$ } \\ 
        \hline
        E.9 & $\arccos(x+y)$ & \makecell{$x, y\in$ \\ $ (0, 0.5)$} &\makecell{[2, 5, 5, 1]\\ param \#: 51} &\makecell{$1.26 \times 10^{-2}$\\ $ \pm 2.51 \times 10^{-2}$}  &\makecell{[2, 2, 1, 1]\\ param \#: 63} &\makecell{\boldmath $2.35 \times 10^{-6}$ \\ \boldmath $\pm$  $3.50 \times 10^{-6}$ } &\makecell{[2, 2, 1, 1]\\ param \#: 47}  &\makecell{ $4.49 \times 10^{-5}$ \\ $\pm$  $5.15 \times 10^{-5}$ } &\makecell{[2, 2, 1, 1]\\ param \#: 547}  &\makecell{\textbf{\boldmath $3.73 \times 10^{-7}$} \\ \textbf{\boldmath$\pm$  $2.32 \times 10^{-7}$ }} \\ 
        \hline
        E.10 & $\arctan(x+y)$ & \makecell{$x, y\in$ \\ $ (0, 0.5)$} &\makecell{[2, 5, 5, 1]\\ param \#: 51} &\makecell{$3.09 \times 10^{-5}$\\ $ \pm 4.88 \times 10^{-5}$}  &\makecell{[2, 2, 1, 1]\\ param \#: 63} &\makecell{ $6.09 \times 10^{-8}$ \\ $\pm$  $7.80 \times 10^{-8}$ } &\makecell{[2, 2, 1, 1]\\ param \#: 47}  &\makecell{\boldmath $4.86 \times 10^{-9}$ \\ \boldmath $\pm$  $2.17 \times 10^{-9}$ } &\makecell{[2, 2, 1, 1]\\ param \#: 547}  &\makecell{\textbf{\boldmath $3.02 \times 10^{-9}$} \\ \textbf{\boldmath$\pm$  $1.65 \times 10^{-9}$ }} \\ 
        \hline
        E.11 & $\sinh(x+y)$ & \makecell{$x, y\in$ \\ $ (0, 1)$} &\makecell{[2, 5, 5, 1]\\ param \#: 51} &\makecell{$2.42 \times 10^{-3}$\\ $ \pm 3.15 \times 10^{-3}$}  &\makecell{[2, 2, 1, 1]\\ param \#: 63} &\makecell{ $8.96 \times 10^{-7}$ \\ $\pm$  $7.13 \times 10^{-7}$ } &\makecell{[2, 2, 1, 1]\\ param \#: 47}  &\makecell{\boldmath $1.99 \times 10^{-7}$ \\ \boldmath $\pm$  $1.65 \times 10^{-7}$ } &\makecell{[2, 2, 1, 1]\\ param \#: 547}  &\makecell{\textbf{\boldmath $5.43 \times 10^{-8}$} \\ \textbf{\boldmath$\pm$  $5.25 \times 10^{-8}$ }} \\ 
        \hline
        E.12 & $\cosh(x+y)$ & \makecell{$x, y\in$ \\ $ (0, 1)$} &\makecell{[2, 5, 5, 1]\\ param \#: 51} &\makecell{$1.62 \times 10^{-3}$\\ $ \pm 8.37 \times 10^{-4}$ } &\makecell{[2, 2, 1, 1]\\ param \#: 63} &\makecell{ $8.77 \times 10^{-7}$ \\ $\pm$  $1.39 \times 10^{-7}$ } &\makecell{[2, 2, 1, 1]\\ param \#: 47}  &\makecell{\boldmath $7.00 \times 10^{-7}$ \\ \boldmath $\pm$  $4.65 \times 10^{-7}$ } &\makecell{[2, 2, 1, 1]\\ param \#: 547}  &\makecell{\textbf{\boldmath $4.47 \times 10^{-8}$} \\ \textbf{\boldmath $\pm$  $3.42 \times 10^{-8}$ }} \\ 
        \hline
        E.13 & $\tanh(x+y)$ & \makecell{$x, y\in$ \\ $ (0, 1)$} &\makecell{[2, 5, 5, 1]\\ param \#: 51} &\makecell{$8.35 \times 10^{-4}$\\ $ \pm 1.39 \times 10^{-3}$}  &\makecell{[2, 2, 1, 1]\\ param \#: 63} &\makecell{\boldmath $3.10 \times 10^{-8}$ \\ \boldmath $\pm$  $3.24 \times 10^{-8}$ } &\makecell{[2, 2, 1, 1]\\ param \#: 47}  &\makecell{ $5.73 \times 10^{-8}$ \\ $\pm$  $6.20 \times 10^{-8}$ } &\makecell{[2, 2, 1, 1]\\ param \#: 547}  &\makecell{\textbf{\boldmath $4.55 \times 10^{-9}$} \\ \textbf{\boldmath$\pm$  $2.73 \times 10^{-9}$ }} \\ 
        \hline
        E.14 & $\arcsinh(x+y)$ & \makecell{$x, y\in$ \\ $ (0, 0.5)$} &\makecell{[2, 5, 5, 1]\\ param \#: 51} &\makecell{$1.47 \times 10^{-5}$\\ $ \pm 2.21 \times 10^{-5}$}  &\makecell{[2, 2, 1, 1]\\ param \#: 63} &\makecell{ $1.98 \times 10^{-7}$ \\ $\pm$  $2.27 \times 10^{-7}$ } &\makecell{[2, 2, 1, 1]\\ param \#: 47}  &\makecell{\boldmath $5.13 \times 10^{-9}$ \\ \boldmath $\pm$  $3.94 \times 10^{-9}$ } &\makecell{[2, 2, 1, 1]\\ param \#: 547}  &\makecell{\textbf{\boldmath $4.34 \times 10^{-9}$} \\ \textbf{\boldmath$\pm$  $2.82 \times 10^{-9}$ }} \\ 
        \hline
        E.15 & $\arccosh(x+y)$ & \makecell{$x, y\in$ \\ $ (0.5, 1)$} &\makecell{[2, 5, 5, 1]\\ param \#: 51} &\makecell{$8.28 \times 10^{-3}$\\ $ \pm 1.57 \times 10^{-2}$}  &\makecell{[2, 2, 1, 1]\\ param \#: 63} &\makecell{\textbf{\boldmath $2.32 \times 10^{-7}$} \\ \textbf{\boldmath $\pm$  $7.08 \times 10^{-8}$ }} &\makecell{[2, 2, 1, 1]\\ param \#: 47}  &\makecell{ $8.94 \times 10^{-6}$ \\ $\pm$  $4.55 \times 10^{-6}$ } &\makecell{[2, 2, 1, 1]\\ param \#: 547}  &\makecell{\boldmath $5.38 \times 10^{-7}$ \\ \boldmath $\pm$  $5.62 \times 10^{-7}$ } \\ 
        \hline
        E.16 & $\arctanh(x+y)$ & \makecell{$x, y\in$ \\ $ (0, 0.5)$} &\makecell{[2, 5, 5, 1]\\ param \#: 51} & \makecell{$7.52 \times 10^{-4}$\\ $\pm 2.55 \times 10^{-4}$} &\makecell{[2, 2, 1, 1]\\ param \#: 63} &\makecell{\textbf{\boldmath $1.30 \times 10^{-5}$} \\ \textbf{\boldmath $\pm$  $1.64 \times 10^{-5}$ }} &\makecell{[2, 2, 1, 1]\\ param \#: 47}  &\makecell{ $3.00 \times 10^{-5}$ \\ $\pm$  $3.47 \times 10^{-5}$ } &\makecell{[2, 2, 1, 1]\\ param \#: 547}  &\makecell{\boldmath $2.51 \times 10^{-5}$ \\ \boldmath $\pm$  $3.04 \times 10^{-5}$ } \\ 
        \hline
    \end{tabular}
    }}
\end{table}
%---------------------------------------------------------------

%---------------------------------------------------------------

%---------------------------------------------------------------
%---------------------------------------------------------------
\begin{table}[t]
    \caption{Experimental results of continuous function approximation on the composite function dataset. The results are obtained in the same way and reported in the same format as the previous table on elementary functions.}
    \label{tab:composite_comparison}
    \tiny
    \centering
    \setlength{\tabcolsep}{3pt}
    \renewcommand{\arraystretch}{2}
    {\fontsize{6}{6}\selectfont
    \scalemath{0.8}{
    \begin{tabular}{|c|c|c||c|c||c|c||c|c||c|c|}
        \hline
        \thead{Eq.} & \thead{Formula} & \thead{Variables} & \thead{MLP\\ Architecture} & \thead{MLP\\ MSE} & \thead{KAN\\ Architecture} & \thead{KAN\\ MSE} & \thead{{\our}-Ext\\ Architecture} & \thead{{\our}-Ext\\ MSE} & \thead{{\our}-Nstd\\ Architecture} & \thead{{\our}-Nstd\\ MSE}\\ 
        \hline
	C.0 & \makecell{$(x+y)$\\ $+{1}/{(x+y)}$} & \makecell{$x, y\in$ \\ $ (0, 1)$} & \makecell{[2, 10, 10, 1]\\ param \#: 151}  &\makecell{ $2.18 \times 10^{-1}$ \\ $\pm$  $2.85 \times 10^{-1}$ } &\makecell{[2, 2, 2, 1]\\ param \#: 150} &\makecell{ $2.95 \times 10^{-1}$ \\ $\pm$  $4.85 \times 10^{-1}$ } &\makecell{[2, 2, 2, 1]\\ param \#: 71}  &\makecell{\boldmath $3.25 \times 10^{-2}$ \\ \boldmath $\pm$  $3.34 \times 10^{-2}$ } &\makecell{[2, 2, 2, 1]\\ param \#: 821}  &\makecell{\boldmath $2.73 \times 10^{-3}$ \\ \boldmath$\pm$  $3.32 \times 10^{-3}$ } \\ 
	\hline
	C.1 & \makecell{$(x+y)$\\ $+(x+y)^2$} & \makecell{$x, y\in$ \\ $ (0, 1)$} &\makecell{[2, 10, 10, 1]\\ param \#: 151} &\makecell{ $3.86 \times 10^{-4}$ \\ $\pm$  $2.40 \times 10^{-4}$ } &\makecell{[2, 2, 2, 1]\\ param \#: 150} &\makecell{\boldmath $5.04 \times 10^{-7}$ \\ \boldmath$\pm$  $2.07 \times 10^{-7}$ } &\makecell{[2, 2, 2, 1]\\ param \#: 71}  &\makecell{ $6.73 \times 10^{-7}$ \\ $\pm$  $3.04 \times 10^{-7}$ } &\makecell{[2, 2, 2, 1]\\ param \#: 821}  &\makecell{\boldmath $1.84 \times 10^{-7}$ \\ \boldmath$\pm$  $1.19 \times 10^{-7}$ } \\ 
	\hline
	C.2 & \makecell{$(x+y)^2$\\ $+\exp(x+y)$} & \makecell{$x, y\in$ \\ $ (0, 1)$} &\makecell{[2, 10, 10, 1]\\ param \#: 151} &\makecell{ $2.99 \times 10^{-3}$ \\ $\pm$  $1.62 \times 10^{-3}$ } &\makecell{[2, 2, 2, 1]\\ param \#: 150} &\makecell{ $1.61 \times 10^{-6}$ \\ $\pm$  $7.77 \times 10^{-7}$ } &\makecell{[2, 2, 2, 1]\\ param \#: 71}  &\makecell{\boldmath $1.33 \times 10^{-6}$ \\ \boldmath$\pm$  $1.29 \times 10^{-6}$ } &\makecell{[2, 2, 2, 1]\\ param \#: 821}  &\makecell{\boldmath $1.54 \times 10^{-6}$ \\\boldmath $\pm$  $9.85 \times 10^{-7}$ } \\ 
	\hline
	C.3 & \makecell{$\exp(x+y)$\\ $+\ln(x+y)$} & \makecell{$x, y\in$ \\ $ (0, 1)$} &\makecell{[2, 10, 10, 1]\\ param \#: 151} &\makecell{ $2.97 \times 10^{-2}$ \\ $\pm$  $3.03 \times 10^{-3}$ } &\makecell{[2, 2, 2, 1]\\ param \#: 150} &\makecell{\boldmath $4.53 \times 10^{-5}$ \\ \boldmath$\pm$  $6.66 \times 10^{-5}$ } &\makecell{[2, 2, 2, 1]\\ param \#: 71}  &\makecell{ $1.17 \times 10^{-4}$ \\ $\pm$  $1.18 \times 10^{-4}$ } &\makecell{[2, 2, 2, 1]\\ param \#: 821}  &\makecell{ \boldmath $3.86 \times 10^{-6}$ \\ \boldmath $\pm$  $2.99 \times 10^{-6}$ } \\ 
	\hline
	C.4 & \makecell{$(x+y)^2$\\ $+\sin(x+y)$} & \makecell{$x, y\in$ \\ $ (0, 1)$} &\makecell{[2, 10, 10, 1]\\ param \#: 151} &\makecell{ $1.35 \times 10^{-4}$ \\ $\pm$  $6.03 \times 10^{-5}$ } &\makecell{[2, 2, 2, 1]\\ param \#: 150}&\makecell{ $4.80 \times 10^{-7}$ \\ $\pm$  $6.21 \times 10^{-7}$ } &\makecell{[2, 2, 2, 1]\\ param \#: 71}  &\makecell{\boldmath $4.40 \times 10^{-7}$ \\\boldmath $\pm$  $4.59 \times 10^{-7}$ } &\makecell{[2, 2, 2, 1]\\ param \#: 821}  &\makecell{\boldmath $1.09 \times 10^{-7}$ \\ \boldmath $\pm$  $6.68 \times 10^{-8}$ } \\ 
	\hline
	C.5 & \makecell{$\cos(x+y)$\\ $+\arccos(x+y)$} & \makecell{$x, y\in$ \\ $ (0, 0.5)$} &\makecell{[2, 10, 10, 1]\\ param \#: 151} &\makecell{ $5.69 \times 10^{-4}$ \\ $\pm$  $7.87 \times 10^{-4}$ } &\makecell{[2, 2, 2, 1]\\ param \#: 150} &\makecell{\boldmath $2.28 \times 10^{-7}$ \\ \boldmath$\pm$  $1.46 \times 10^{-7}$ } &\makecell{[2, 2, 2, 1]\\ param \#: 71}  &\makecell{ $1.48 \times 10^{-6}$ \\ $\pm$  $1.69 \times 10^{-6}$ } &\makecell{[2, 2, 2, 1]\\ param \#: 821}  &\makecell{\boldmath $7.24 \times 10^{-8}$ \\ \boldmath$\pm$  $4.28 \times 10^{-8}$ } \\ 
	\hline
	C.6 & \makecell{$\exp(x+y)$\\ $\times {1}/{(x+y)}$} & \makecell{$x, y\in$ \\ $ (0, 1)$} &\makecell{[2, 10, 10, 1]\\ param \#: 151} &\makecell{ $2.25 \times 10^{-1}$ \\ $\pm$  $2.86 \times 10^{-1}$ } &\makecell{[2, 2, 2, 1]\\ param \#: 150} &\makecell{ $8.19 \times 10^{-2}$ \\ $\pm$  $1.46 \times 10^{-1}$ } &\makecell{[2, 2, 2, 1]\\ param \#: 71}  &\makecell{\boldmath $3.72 \times 10^{-2}$ \\ \boldmath$\pm$  $3.95 \times 10^{-2}$ } &\makecell{[2, 2, 2, 1]\\ param \#: 821}  &\makecell{\boldmath $2.25 \times 10^{-3}$ \\ \boldmath$\pm$  $2.49 \times 10^{-3}$ } \\ 
	\hline
	C.7 & \makecell{$(x+y)^2$\\ $ \times \ln(x+y)$} & \makecell{$x, y\in$ \\ $ (0, 1)$} &\makecell{[2, 10, 10, 1]\\ param \#: 151} &\makecell{ $9.74 \times 10^{-4}$ \\ $\pm$  $1.24 \times 10^{-3}$ } &\makecell{[2, 2, 2, 1]\\ param \#: 150} &\makecell{\boldmath $7.00 \times 10^{-8}$ \\ \boldmath$\pm$  $3.38 \times 10^{-8}$ } &\makecell{[2, 2, 2, 1]\\ param \#: 71}  &\makecell{ $4.97 \times 10^{-7}$ \\ $\pm$  $5.01 \times 10^{-7}$ } &\makecell{[2, 2, 2, 1]\\ param \#: 821}  &\makecell{\boldmath $2.94 \times 10^{-8}$ \\ \boldmath$\pm$  $1.37 \times 10^{-8}$ } \\ 
	\hline
	C.8 & \makecell{$(x+y) $\\ $\times \sin(x+y)$} & \makecell{$x, y\in$ \\ $ (0, 1)$} &\makecell{[2, 10, 10, 1]\\ param \#: 151} &\makecell{ $3.80 \times 10^{-5}$ \\ $\pm$  $3.60 \times 10^{-5}$ } &\makecell{[2, 2, 2, 1]\\ param \#: 150} &\makecell{\boldmath $3.03 \times 10^{-8}$ \\ \boldmath$\pm$  $1.79 \times 10^{-8}$ } &\makecell{[2, 2, 2, 1]\\ param \#: 71}  &\makecell{ $5.70 \times 10^{-8}$ \\ $\pm$  $1.98 \times 10^{-8}$ } &\makecell{[2, 2, 2, 1]\\ param \#: 821}  &\makecell{\boldmath $1.64 \times 10^{-8}$ \\ \boldmath$\pm$  $9.49 \times 10^{-9}$ } \\ 
	\hline
	C.9 & \makecell{$\exp(x+y) $\\ $\times \ln(x+y)$} & \makecell{$x, y\in$ \\ $ (0, 1)$} &\makecell{[2, 10, 10, 1]\\ param \#: 151} &\makecell{ $4.78 \times 10^{-3}$ \\ $\pm$  $2.58 \times 10^{-3}$ } &\makecell{[2, 2, 2, 1]\\ param \#: 150} &\makecell{\boldmath $3.10 \times 10^{-5}$ \\ \boldmath$\pm$  $2.39 \times 10^{-5}$ } &\makecell{[2, 2, 2, 1]\\ param \#: 71}  &\makecell{ $1.88 \times 10^{-4}$ \\ $\pm$  $2.20 \times 10^{-4}$ } &\makecell{[2, 2, 2, 1]\\ param \#: 821}  &\makecell{\boldmath $4.55 \times 10^{-6}$ \\ \boldmath$\pm$  $3.99 \times 10^{-6}$ } \\ 
	\hline
	C.10 & \makecell{$\sin(x+y) $\\ $\times \sinh(x+y)$} & \makecell{$x, y\in$ \\ $ (0, 1)$} &\makecell{[2, 10, 10, 1]\\ param \#: 151} &\makecell{ $1.82 \times 10^{-4}$ \\ $\pm$  $6.47 \times 10^{-5}$ } &\makecell{[2, 2, 2, 1]\\ param \#: 150} & \makecell{\boldmath $1.19 \times 10^{-7}$ \\ \boldmath$\pm$  $9.75 \times 10^{-8}$ } &\makecell{[2, 2, 2, 1]\\ param \#: 71}  &\makecell{ $3.40 \times 10^{-7}$ \\ $\pm$  $4.77 \times 10^{-7}$ } &\makecell{[2, 2, 2, 1]\\ param \#: 821}  &\makecell{\boldmath $3.58 \times 10^{-8}$ \\ \boldmath$\pm$  $1.18 \times 10^{-8}$ } \\ 
	\hline
	C.11 & \makecell{$\arccos(x+y) $\\ $\times \arctanh(x+y)$} & \makecell{$x, y\in$ \\ $ (0, 0.5)$} &\makecell{[2, 10, 10, 1]\\ param \#: 151} &\makecell{ $1.08 \times 10^{-4}$ \\ $\pm$  $9.94 \times 10^{-5}$ } &\makecell{[2, 2, 2, 1]\\ param \#: 150} &\makecell{\boldmath $2.75 \times 10^{-7}$ \\ \boldmath $\pm$  $2.34 \times 10^{-7}$ } &\makecell{[2, 2, 2, 1]\\ param \#: 71}  &\makecell{ \boldmath $3.57 \times 10^{-7}$ \\ \boldmath $\pm$  $2.56 \times 10^{-7}$ } &\makecell{[2, 2, 2, 1]\\ param \#: 821}  &\makecell{ $4.87 \times 10^{-7}$ \\ $\pm$  $8.71 \times 10^{-7}$ } \\ 
	\hline
	C.12 & \makecell{$\exp(\frac{1}{(x+y)} $\\ $+\exp(x+y))$} & \makecell{$x, y\in$ \\ $ (0, 0.5)$} &\makecell{[2, 10, 10, 1]\\ param \#: 151} &\makecell{ $1.07 \times 10^{-1}$ \\ $\pm$  $1.52 \times 10^{-1}$ } &\makecell{[2, 2, 2, 1]\\ param \#: 150} &\makecell{ $3.81 \times 10^{-4}$ \\ $\pm$  $4.55 \times 10^{-4}$ } &\makecell{[2, 2, 2, 1]\\ param \#: 71}  &\makecell{\boldmath $3.74 \times 10^{-5}$ \\ \boldmath$\pm$  $1.88 \times 10^{-5}$ } &\makecell{[2, 2, 2, 1]\\ param \#: 821}  &\makecell{\boldmath $7.17 \times 10^{-5}$ \\ \boldmath$\pm$  $8.87 \times 10^{-5}$ } \\ 
	\hline
	C.13 & \makecell{$\exp(\sin(x+y)$\\ $+\cos(x+y))$} & \makecell{$x, y\in$ \\ $ (0, 1)$} &\makecell{[2, 10, 10, 1]\\ param \#: 151} &\makecell{ $2.35 \times 10^{-3}$ \\ $\pm$  $2.65 \times 10^{-3}$ } &\makecell{[2, 2, 2, 1]\\ param \#: 150} &\makecell{\boldmath $3.62 \times 10^{-7}$ \\ \boldmath$\pm$  $1.65 \times 10^{-7}$ } &\makecell{[2, 2, 2, 1]\\ param \#: 71}  &\makecell{ $6.97 \times 10^{-7}$ \\ $\pm$  $2.29 \times 10^{-7}$ } &\makecell{[2, 2, 2, 1]\\ param \#: 821}  &\makecell{\boldmath $8.56 \times 10^{-8}$ \\ \boldmath$\pm$  $3.81 \times 10^{-8}$ } \\ 
	\hline
	C.14 & \makecell{$\ln((x+y)^2$\\ $+\exp(x+y))$} & \makecell{$x, y\in$ \\ $ (0.5, 1)$} &\makecell{[2, 10, 10, 1]\\ param \#: 151} &\makecell{ $1.26 \times 10^{-5}$ \\ $\pm$  $1.72 \times 10^{-5}$ } &\makecell{[2, 2, 2, 1]\\ param \#: 150} & \makecell{\boldmath $6.30 \times 10^{-8}$ \\ \boldmath$\pm$  $7.82 \times 10^{-8}$ } &\makecell{[2, 2, 2, 1]\\ param \#: 71}  &\makecell{ $3.50 \times 10^{-7}$ \\ $\pm$  $3.89 \times 10^{-7}$ } &\makecell{[2, 2, 2, 1]\\ param \#: 821}  &\makecell{\boldmath $1.69 \times 10^{-8}$ \\ \boldmath$\pm$  $2.42 \times 10^{-8}$ } \\ 
	\hline
	C.15 & \makecell{$\tan(\exp(x+y)$\\ $+\ln(x+y))$} & \makecell{$x, y\in$ \\ $ (0, 1)$} &\makecell{[2, 10, 10, 1]\\ param \#: 151} &\makecell{\boldmath $6.48 \times 10^{2}$ \\ \boldmath$\pm$  $4.50 \times 10^{2}$ } &\makecell{[2, 2, 2, 1]\\ param \#: 150} &\makecell{\boldmath $1.74 \times 10^{3}$ \\ \boldmath$\pm$  $1.72 \times 10^{3}$ } &\makecell{[2, 2, 2, 1]\\ param \#: 71}  &\makecell{ $2.47 \times 10^{3}$ \\ $\pm$  $4.39 \times 10^{3}$ } &\makecell{[2, 2, 2, 1]\\ param \#: 821}  &\makecell{ $1.17 \times 10^{4}$ \\ $\pm$  $2.11 \times 10^{4}$ } \\ 
	\hline
	C.16 & \makecell{$\frac{1}{1+\exp(-x-y)}$} & \makecell{$x, y\in$ \\ $ (0, 1)$} &\makecell{[2, 10, 10, 1]\\ param \#: 151} &\makecell{ $1.48 \times 10^{-6}$ \\ $\pm$  $1.69 \times 10^{-6}$ } &\makecell{[2, 2, 2, 1]\\ param \#: 150} &\makecell{ $2.86 \times 10^{-9}$ \\ $\pm$  $1.75 \times 10^{-9}$ } &\makecell{[2, 2, 2, 1]\\ param \#: 71}  &\makecell{\boldmath $1.74 \times 10^{-9}$ \\ \boldmath$\pm$  $1.09 \times 10^{-9}$ } &\makecell{[2, 2, 2, 1]\\ param \#: 821}  &\makecell{\boldmath $1.48 \times 10^{-9}$ \\ \boldmath$\pm$  $1.32 \times 10^{-9}$ } \\ 
        \hline
    \end{tabular}
    }}
\end{table}
%---------------------------------------------------------------

%---------------------------------------------------------------

%---------------------------------------------------------------
%---------------------------------------------------------------
\begin{table}[t]
    \caption{Experimental results of continuous function approximation on the Feynman function dataset. Method {\our}-Ext uses extended expansion, low-rank reconciliation and linear remainder.}
    \label{tab:feynman_comparison}
    \tiny
    \centering
    \setlength{\tabcolsep}{8pt}
    %\setlength{\extrarowheight}{-2pt}
    %\renewcommand{\arraystretch}{0}
    %\setcellgapes{2pt}
    {\fontsize{7}{7}\selectfont
    \scalemath{0.72}{
    \begin{tabular}{|c|c|c||c||c||c|}
        \hline
        \thead{Eq.} & \thead{Formula} & \thead{Variables} & \thead{MLP\\ MSE} & \thead{KAN\\ MSE} & \thead{{\our}-Ext\\ MSE} \\ 
        \hline
        I.6.2 & $\exp\left(-\frac{\theta^2}{2{\sigma}^2}\right)/\sqrt{2\pi}\sigma$ & $\theta, \sigma \in [1, 3]$ &\makecell{ $7.17 \times 10^{-5}$ \\ $\pm$  $8.37 \times 10^{-5}$ } & \makecell{ \boldmath$3.20 \times 10^{-7}$ \\ \boldmath$\pm$  $2.02 \times 10^{-7}$ } &\makecell{ $1.63 \times 10^{-6}$ \\ $\pm$  $1.31 \times 10^{-6}$ } \\%1
        \hline
        I.6.2b & $\exp\left(-\frac{(\theta-\theta_1)^2}{2 \sigma^2} \right)/\sqrt{2 \pi} \sigma$ & $\sigma$, $\theta$, $\theta_1 \in [1, 3]$ &\makecell{ $4.52 \times 10^{-5}$ \\ $\pm$  $1.90 \times 10^{-5}$ } &\makecell{ $8.97 \times 10^{-5}$ \\ $\pm$  $1.77 \times 10^{-4}$ } &\makecell{\boldmath $1.60 \times 10^{-5}$ \\ \boldmath$\pm$  $1.25 \times 10^{-5}$ } \\%2
        \hline
        I.9.18 & $\frac{G \cdot m_1 \cdot m_2}{(x_2 - x_1)^2+(y_2 - y_1)^2+(z_2 - z_1)^2}$ & \makecell{$G$, $m_1$, $m_2$, $x_2$, $y_2$, $z_2 \in [1, 2]$\\ $x_1$, $y_1$, $z_1 \in [3, 4]$} &\makecell{ $2.17 \times 10^{-4}$ \\ $\pm$  $7.42 \times 10^{-5}$ } &\makecell{ $1.60 \times 10^{-4}$ \\ $\pm$  $9.60 \times 10^{-5}$ } &\makecell{\boldmath $6.92 \times 10^{-5}$ \\ \boldmath$\pm$  $2.42 \times 10^{-5}$ } \\%4
        \hline
        I.12.11 & $q \cdot (E_f+B \cdot v \cdot \sin(\theta))$ & $q$, $E_f$, $B$, $v$, $\theta \in [1, 5]$ &\makecell{\boldmath $9.81 \times 10^{0}$ \\ \boldmath$\pm$  $2.36 \times 10^{0}$ } &\makecell{ $4.78 \times 10^{1}$ \\ $\pm$  $2.05 \times 10^{1}$ }  &\makecell{ $1.36 \times 10^{1}$ \\ $\pm$  $8.35 \times 10^{0}$ } \\%12
        \hline
        I.13.12 & $G \cdot m_1 \cdot m_2 \left(\frac{1}{r_2}-\frac{1}{r_1} \right)$ &\makecell{ $G$, $m_1$, $m_2$, $r_1$, $r_2 \in [1, 5]$ }&\makecell{ $1.24 \times 10^{0}$ \\ $\pm$  $2.83 \times 10^{-1}$ } &\makecell{ $7.07 \times 10^{0}$ \\ $\pm$  $3.25 \times 10^{0}$ } &\makecell{\boldmath $8.47 \times 10^{-1}$ \\ \boldmath$\pm$  $1.21 \times 10^{-1}$ } \\%13
        \hline
        I.15.3x & $\frac{(x-u \cdot t)}{\sqrt{1-u^2/c^2}}$ & \makecell{$x \in [5, 10]$, $u \in [1,2]$\\ $c \in [3, 20]$, $t \in [1,2]$} &\makecell{ $1.07 \times 10^{-2}$ \\ $\pm$  $9.00 \times 10^{-3}$ } &\makecell{ $1.49 \times 10^{-2}$ \\ $\pm$  $4.94 \times 10^{-3}$ } &\makecell{\boldmath $4.94 \times 10^{-3}$ \\ \boldmath$\pm$  $1.57 \times 10^{-3}$ } \\%16
        \hline
        I.16.6 & $\frac{(u+v)}{(1+u \cdot v/c^2)}$ & $c$, $v$, $u \in [1, 5]$ &\makecell{ $3.39 \times 10^{-3}$ \\ $\pm$  $3.31 \times 10^{-3}$ } &\makecell{ $8.47 \times 10^{-3}$ \\ $\pm$  $2.67 \times 10^{-3}$ } &\makecell{\boldmath $8.68 \times 10^{-4}$ \\ \boldmath$\pm$  $3.84 \times 10^{-4}$ } \\%19
        \hline
        I.18.4 & $(m_1 \cdot r_1+m_2 \cdot r_2)/(m_1+m_2)$ & $m_1$, $m_2$, $r_1$, $r_2 \in [1, 5]$ &\makecell{ $1.16 \times 10^{-2}$ \\ $\pm$  $1.13 \times 10^{-2}$ } &\makecell{ $2.00 \times 10^{-2}$ \\ $\pm$  $5.02 \times 10^{-3}$ } &\makecell{\boldmath $9.92 \times 10^{-4}$ \\ \boldmath$\pm$  $4.82 \times 10^{-4}$ } \\%20
        \hline
        I.26.2 & $\arcsin(n \cdot \sin(\theta_2))$ & $n \in [0, 1]$, $\theta_2 \in [1, 5]$ &\makecell{ $9.59 \times 10^{-4}$ \\ $\pm$  $3.33 \times 10^{-4}$ } &\makecell{ $7.64 \times 10^{-5}$ \\ $\pm$  $1.12 \times 10^{-4}$ } &\makecell{\boldmath $6.59 \times 10^{-5}$ \\ \boldmath$\pm$  $5.13 \times 10^{-5}$ } \\%25
        \hline
        I.27.6 & $\frac{1}{\frac{1}{d_1}+\frac{n}{d_2}}$ & $d_1$, $d_2$, $n \in [1, 5]$ &\makecell{ $5.40 \times 10^{-4}$ \\ $\pm$  $3.52 \times 10^{-4}$ } &\makecell{ $8.22 \times 10^{-4}$ \\ $\pm$  $1.24 \times 10^{-4}$ } &\makecell{\boldmath $4.98 \times 10^{-4}$ \\ \boldmath$\pm$  $6.95 \times 10^{-4}$ } \\%26
        \hline
        
        I.29.16 & $\sqrt{x_1^2+x_2^2-2 x_1 x_2 \cos(\theta_1-\theta_2)}$ & $x_1$, $x_2$, $\theta_1$, $\theta_2 \in [1, 5]$ &\makecell{ $1.16 \times 10^{-1}$ \\ $\pm$  $1.41 \times 10^{-1}$ } &\makecell{ $1.40 \times 10^{0}$ \\ $\pm$  $6.16 \times 10^{-1}$ } &\makecell{\boldmath $7.27 \times 10^{-2}$ \\ \boldmath$\pm$  $7.00 \times 10^{-2}$ } \\%28
        \hline
        
        I.30.3 & ${Int}_0 \cdot \sin \left(n \frac{\theta}{2} \right)^2/\sin \left(\frac{\theta}{2} \right)^2$ & ${Int}_0$, $\theta$, $n \in [1, 5]$ &\makecell{ $2.17 \times 10^{0}$ \\ $\pm$  $2.42 \times 10^{-1}$ } &\makecell{ $2.13 \times 10^{0}$ \\ $\pm$  $6.78 \times 10^{-1}$ } &\makecell{\boldmath $1.58 \times 10^{0}$ \\ \boldmath$\pm$  $8.47 \times 10^{-1}$ } \\%29
        \hline
        
        I.30.5 & $\arcsin(\frac{\lambda}{n \cdot d})$ & \makecell{$\lambda \in [1, 2]$ \\$d \in [2, 5]$, $n \in [1, 5]$} &\makecell{ $2.45 \times 10^{-4}$ \\ $\pm$  $2.14 \times 10^{-4}$ } &\makecell{ $6.26 \times 10^{-5}$ \\ $\pm$  $5.11 \times 10^{-5}$ } &\makecell{\boldmath $6.83 \times 10^{-6}$ \\ \boldmath$\pm$  $2.53 \times 10^{-6}$ } \\%30
        
        \hline
        I.37.4 & $I_1+I_2+2 \sqrt{I_1 \cdot I_2} \cdot \cos(\delta)$ & $I_1$, $I_2$, $\delta \in [1, 5]$ &\makecell{ $9.14 \times 10^{-2}$ \\ $\pm$  $5.89 \times 10^{-2}$ } &\makecell{ $1.74 \times 10^{-1}$ \\ $\pm$  $1.43 \times 10^{-1}$ } &\makecell{\boldmath $3.89 \times 10^{-2}$ \\ \boldmath$\pm$  $3.06 \times 10^{-2}$ } \\%37
        \hline
        
        I.40.1 & $n_0 \exp\left(-\frac{m \cdot g \cdot x}{(k_b \cdot T)} \right)$ & $n_0$, $m$, $g$, $x$, $k_b$, $T \in [1, 5]$ &\makecell{ $8.90 \times 10^{-3}$ \\ $\pm$  $6.01 \times 10^{-3}$ } &\makecell{ $1.09 \times 10^{-2}$ \\ $\pm$  $6.98 \times 10^{-3}$ } &\makecell{\boldmath $2.13 \times 10^{-3}$ \\ \boldmath$\pm$  $4.51 \times 10^{-4}$ } \\%42
        \hline
        
        I.44.4 & $n \cdot k_b \cdot T \cdot \ln \left(\frac{V_2}{V_1} \right)$ & $n$, $k_b$, $T$, $V_1$, $V_2 \in [1, 5]$ &\makecell{ $5.69 \times 10^{0}$ \\ $\pm$  $1.06 \times 10^{0}$ } &\makecell{ $2.58 \times 10^{1}$ \\ $\pm$  $1.76 \times 10^{1}$ } &\makecell{\boldmath $2.99 \times 10^{0}$ \\ \boldmath$\pm$  $5.68 \times 10^{-1}$ } \\%47
        \hline
        
        I.50.26 & $x_1 \cdot (\cos(\omega t)+ \alpha \cdot \cos(\omega t)^2)$ & $x_1$, $\omega$, $t$, $\alpha$ &\makecell{ $1.42 \times 10^{0}$ \\ $\pm$  $1.07 \times 10^{0}$ } &\makecell{\boldmath $7.10 \times 10^{-1}$ \\ \boldmath$\pm$  $2.03 \times 10^{-1}$ } &\makecell{ $1.03 \times 10^{0}$ \\ $\pm$  $7.47 \times 10^{-1}$ } \\%50
        \hline
        
        II.2.42 & $\frac{\kappa \cdot (T_2-T_1) \cdot A}{d}$ & $\kappa$, $T_1$, $T_2$, $A$, $d \in [1, 3]$ &\makecell{ $8.73 \times 10^{-1}$ \\ $\pm$  $3.55 \times 10^{-1}$ } &\makecell{ $1.09 \times 10^{0}$ \\ $\pm$  $7.20 \times 10^{-1}$ } &\makecell{ \boldmath $6.98 \times 10^{-1}$ \\ \boldmath $\pm$  $3.27 \times 10^{-1}$ } \\%51
        \hline
        
        II.6.15a & $\frac{3 z p_d}{(4 \pi \epsilon) r^5} \sqrt{x^2+y^2}$ & $\epsilon$, $p_d$, $r$, $x$, $y$, $z \in [1, 3]$ &\makecell{ $2.07 \times 10^{-3}$ \\ $\pm$  $1.61 \times 10^{-3}$ } &\makecell{ $6.36 \times 10^{-4}$ \\ $\pm$  $3.78 \times 10^{-4}$ } &\makecell{\boldmath $9.29 \times 10^{-5}$ \\ \boldmath $\pm$  $3.06 \times 10^{-5}$ } \\%55
        \hline
        
        II.11.17 & $n_0  \left(1+\frac{p_d \cdot E_f \cos \theta}{k_b \cdot T} \right)$ & \makecell{$n_0$, $k_b$, $T$, $\theta$,$p_d$, $E_f \in [1, 3]$} &\makecell{ $1.10 \times 10^{-1}$ \\ $\pm$  $1.35 \times 10^{-1}$ } &\makecell{ $1.01 \times 10^{-1}$ \\ $\pm$  $7.75 \times 10^{-2}$ } &\makecell{\boldmath $2.85 \times 10^{-2}$ \\ \boldmath$\pm$  $3.35 \times 10^{-3}$ } \\%61
        \hline
        
        II.11.27 & $\frac{n \cdot \alpha}{1- \frac{n \cdot \alpha}{3}} \epsilon E_f$ &\makecell{ $n$, $\alpha \in [0, 1]$,\\ $\epsilon$, $E_f \in [1, 2]$} &\makecell{ $2.23 \times 10^{-2}$ \\ $\pm$  $2.30 \times 10^{-2}$ } &\makecell{\boldmath $8.43 \times 10^{-5}$ \\ \boldmath$\pm$  $6.55 \times 10^{-5}$ } &\makecell{ $9.93 \times 10^{-5}$ \\ $\pm$  $2.70 \times 10^{-5}$ } \\%63
        \hline
        
        II.35.18 & $\frac{n_0}{\exp\left(\frac{\mu_m \cdot B}{k_b \cdot T}\right)+\exp\left(- \frac{\mu_m \cdot B}{k_b \cdot T} \right)}$ & $n_0$, $k_b$, $T$, $\mu_m$, $B \in [1, 3]$ &\makecell{ $7.56 \times 10^{-4}$ \\ $\pm$  $2.87 \times 10^{-4}$ } &\makecell{ $1.39 \times 10^{-3}$ \\ $\pm$  $2.38 \times 10^{-3}$ } &\makecell{\boldmath $1.59 \times 10^{-4}$ \\ \boldmath$\pm$  $1.08 \times 10^{-4}$ } \\%79
        \hline
        
        II.36.38 & $\frac{\mu_m \cdot H}{k_b \cdot T}+\frac{\mu_m \cdot \alpha \cdot M}{\epsilon \cdot c^2  \cdot k_b \cdot T}$ & \makecell{$\mu_m$, $H$, $k_b$, $T$, $\alpha$,\\ $\epsilon$, $c$, $M \in [1, 3]$}  &\makecell{ $5.44 \times 10^{-2}$ \\ $\pm$  $3.07 \times 10^{-2}$ } &\makecell{ $3.49 \times 10^{-2}$ \\ $\pm$  $9.15 \times 10^{-3}$ } &\makecell{\boldmath $4.89 \times 10^{-3}$ \\ \boldmath$\pm$  $2.59 \times 10^{-3}$ } \\%81
        \hline
        
        II.38.3 & $\frac{Y \cdot A \cdot x}{d}$ & $Y$, $A$, $d$, $x \in [1, 5]$  &\makecell{ $5.44 \times 10^{0}$ \\ $\pm$  $8.81 \times 10^{0}$ } &\makecell{ $1.38 \times 10^{0}$ \\ $\pm$  $4.98 \times 10^{-1}$ } &\makecell{\boldmath $1.24 \times 10^{-1}$ \\ \boldmath$\pm$  $1.02 \times 10^{-1}$ } \\%83
        \hline
        
        III.9.52 & $\frac{p_d \cdot E_f \cdot t}{\frac{h}{ 2 \pi}} \frac{\sin\left(\frac{(\omega-\omega_0)t}{2} \right)^2}{\left(\frac{(\omega-\omega_0) t}{2} \right)^2}$ & \makecell{$p_d$, $E_f$, $t$, $h \in [1, 3]$\\ $\omega$, $\omega_0 \in [1, 5]$}  &\makecell{ $9.20 \times 10^{0}$ \\ $\pm$  $1.59 \times 10^{0}$ } &\makecell{ $1.88 \times 10^{1}$ \\ $\pm$  $3.65 \times 10^{0}$ } &\makecell{\boldmath $7.10 \times 10^{0}$ \\ \boldmath$\pm$  $2.92 \times 10^{0}$ } \\%89
        \hline
        
        III.10.19 & $\mu_m \cdot \sqrt{B_x^2+B_y^2+B_z^2}$ & $\mu_m$, $B_x$, $B_y$, $B_z \in [1, 5]$ &\makecell{ $4.17 \times 10^{-1}$ \\ $\pm$  $2.59 \times 10^{-1}$ } &\makecell{ $2.69 \times 10^{-1}$ \\ $\pm$  $8.30 \times 10^{-2}$ } &\makecell{\boldmath $3.08 \times 10^{-2}$ \\ \boldmath$\pm$  $9.13 \times 10^{-3}$ } \\%90
        \hline
        
        III.17.37 & $\beta \cdot (1+\alpha \cdot \cos(\theta))$ & $\beta$, $\alpha$, $\theta \in [1, 5]$ &\makecell{ $8.06 \times 10^{-1}$ \\ $\pm$  $5.30 \times 10^{-1}$ } &\makecell{ $1.62 \times 10^{0}$ \\ $\pm$  $1.12 \times 10^{0}$ } &\makecell{\boldmath $2.66 \times 10^{-1}$ \\ \boldmath$\pm$  $1.64 \times 10^{-1}$ } \\%97
        \hline
    \end{tabular}
    }}
\end{table}
%---------------------------------------------------------------

%---------------------------------------------------------------

\subsubsubsection{\textbf{Dataset Descriptions}}: Three continuous function datasets are used in our experiments to evaluate the performance of {\our} against comparison methods MLP and KAN. The datasets are described below, with basic statistical information provided in Table~\ref{tab:continuous_dataset_statistics}.
\begin{itemize}
\item \textbf{Elementary Function Dataset}: We compose an elementary function dataset in this paper. The elementary function dataset comprises 17 elementary functions, each representing the simplest form of a multivariate function defined by two variables, $x$ and $y$, with specific value ranges. These function ids, formulas and their corresponding input value ranges are provided in the first three columns of Table~\ref{tab:elementary_comparison}.

\item \textbf{Composite Function Dataset}: Building upon these elementary functions, we created the composite function dataset by combining them through addition, multiplication, and nesting to form more complex functions. The 17 created composite functions and their input variable value ranges are presented in the first three columns of Table~\ref{tab:composite_comparison}.

\item \textbf{Feynman Function Dataset}: To evaluate {\our}'s effectiveness on real-world complex functions relevant to scientific research, we utilize the Feynman function dataset from \cite{Udrescu2019AIFA}. Unlike \cite{Liu2024KANKN}, which simplifies the functions to dimensionless forms, we use the original Feynman functions\footnote{https://space.mit.edu/home/tegmark/aifeynman.html} with their provided value ranges. The 27 functions used in our experiments, along with their input variable value ranges, are shown in the first three columns of Table~\ref{tab:feynman_comparison}.
\end{itemize}

All datasets used in these experiments have been made available in the {\toolkit} toolkit, allowing readers to conduct follow-up experimental testing and result replication.

\subsubsubsection{\textbf{Experiment Setups}}: For the continuous function approximation task, we randomly generate $2000$ input-output pairs for each function in the dataset. These pairs are divided into training and testing sets, with a $50:50$ ratio. To ensure fair comparisons, each model is trained with five different random seeds over $2000$ epochs. The best results encountered during training are selected to mitigate biases from epoch hyper-parameter selection. We use Mean Squared Error (MSE) as the evaluation metric. To account for variability due to random seed selection, we report the final evaluation results as ``mean $\pm$ std'' of MSE scores obtained from the five random seeds. It allows us to provide a comprehensive and unbiased assessment of model performance across multiple runs.

%=======================================

\subsubsection{The Main Results of {\our} on Continuous Function Approximation}

Tables~\ref{tab:elementary_comparison}, \ref{tab:composite_comparison}, and \ref{tab:feynman_comparison} present the main results of {\our} compared to MLP and KAN on the elementary, composite, and Feynman function datasets, respectively. Each table shows MSE (mean $\pm$ std) scores for each method, obtained using five random seeds. For the elementary and composite function datasets, we also provide the architecture and parameter counts of the compared methods.

For elementary functions, MLP uses a $[2, 5, 5, 1]$ architecture with $51$ parameters (including bias), while KAN uses $[2, 2, 1, 1]$ with $5$ input intervals divided by the knots, b-splines of order $3$, and $63$ parameters. For composite and Feynman functions, MLP uses $[2, 10, 10, 1]$ with $151$ parameters, and KAN uses $[2, 2, 2, 1]$ with $150$ parameters, $10$ intervals and order $4$. We compare two variants of {\our}: (1) {\our}-Ext using extended expansions (B-spline and Taylor's), LoRR reconciliation (rank 1), and zero remainder; and (2) {\our}-Nstd using nested expansions (B-spline and Taylor's), LoRR reconciliation (rank 1), and zero remainder. Both variants create higher-dimensional intermediate expanded vectors. {\our}-Ext has fewer learnable parameters than MLP and KAN for elementary functions and less than half for composite and Feynman functions.

The results in Tables~\ref{tab:elementary_comparison}-\ref{tab:feynman_comparison} show that {\our}-Ext outperforms MLP, reducing MSE by at least $\times 10^{-1}$ across almost all these functions, with improvements up to $\times 10^{-5}$ for some elementary functions (E.5, E.6) and $\times 10^{-4}$ for some composite functions (C.7, C.13). {\our}-Ext with fewer parameters still achieves comparable or slightly better performance than KAN. Meanwhile, {\our}-Nstd, with more parameters, significantly outperforms KAN, reducing MSE by at least $\times 10^{-1}$ and even $\times 10^{-2}$ for most elementary and composite functions. On the Feynman dataset, {\our}-Ext with a linear remainder and half the parameters of MLP and KAN consistently performs well. It outperforms the other methods on 23 out of 27 functions, with improvements of at least $\times 10^{-1}$ and up to $\times 10^{-2}$ in some cases (e.g., Eq. I.18.4). MLP only performs best on Eq. I.12.11, while KAN excels with slightly more advantages on Eq. I.6.2, I.50.26, and II.11.27.

These results demonstrate {\our}'s effectiveness in approximating both simple and complex continuous functions used in real-world scientific research discoveries. Additional experiments with different extension and remainder functions of {\our} are also available in Tables~\ref{tab:elementary_rpn_ablation}, \ref{tab:composite_rpn_ablation}, and \ref{tab:feynman_rpn_ablation} in the Appendix Section~\ref{subsec:appendix_continuous_function _fitting}.

%=======================================

\subsubsection{Model Learning Analysis}

%------------------------------
\begin{figure}[t]
    \centering
    \begin{subfigure}{0.48\textwidth}
        \centering
        \includegraphics[width=\textwidth]{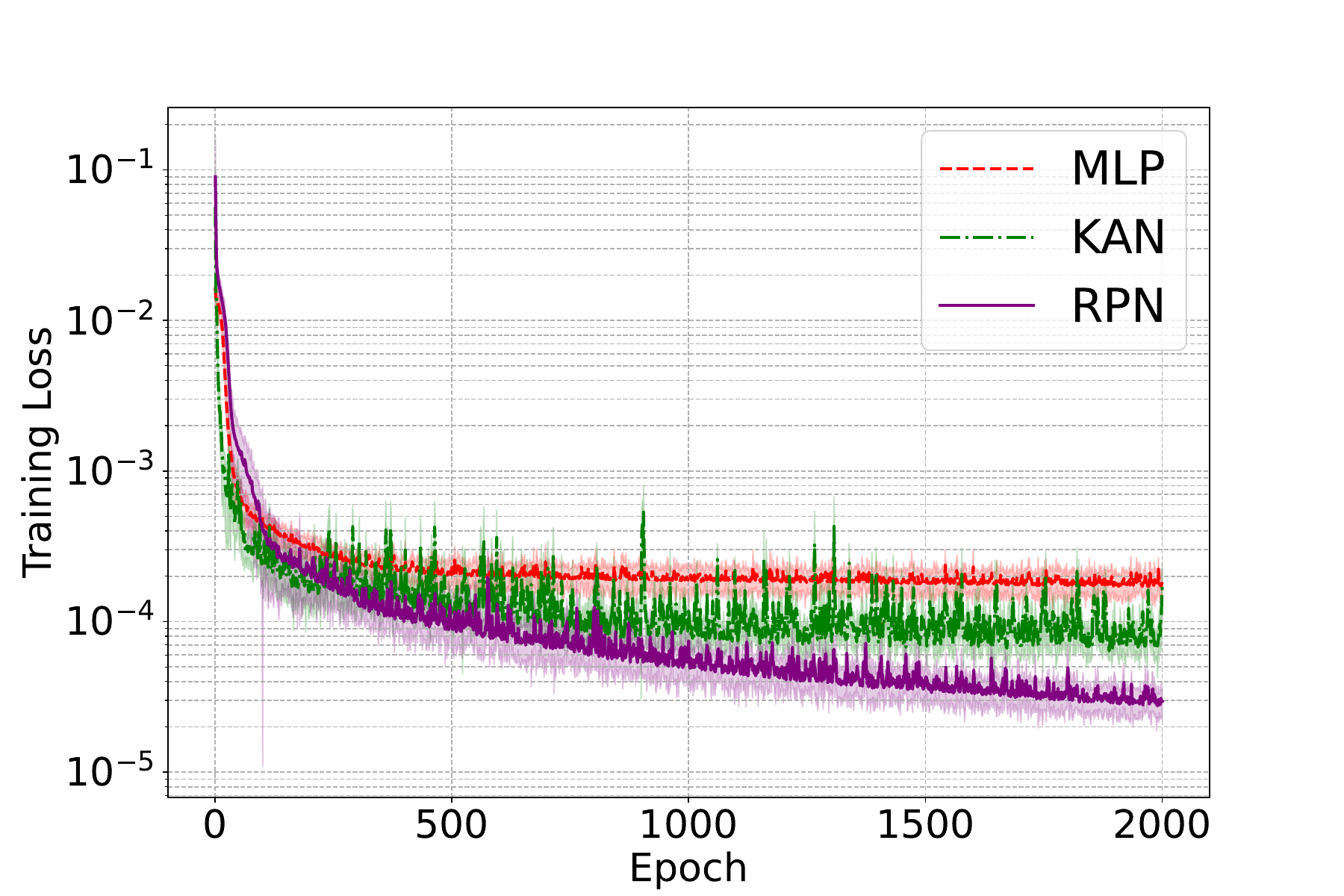}
        \caption{Training MSE on Feynman Eq. I.9.18}
        \label{fig:feynman_4_train}
    \end{subfigure}
    \hfill
    \begin{subfigure}{0.48\textwidth}
        \centering
        \includegraphics[width=\textwidth]{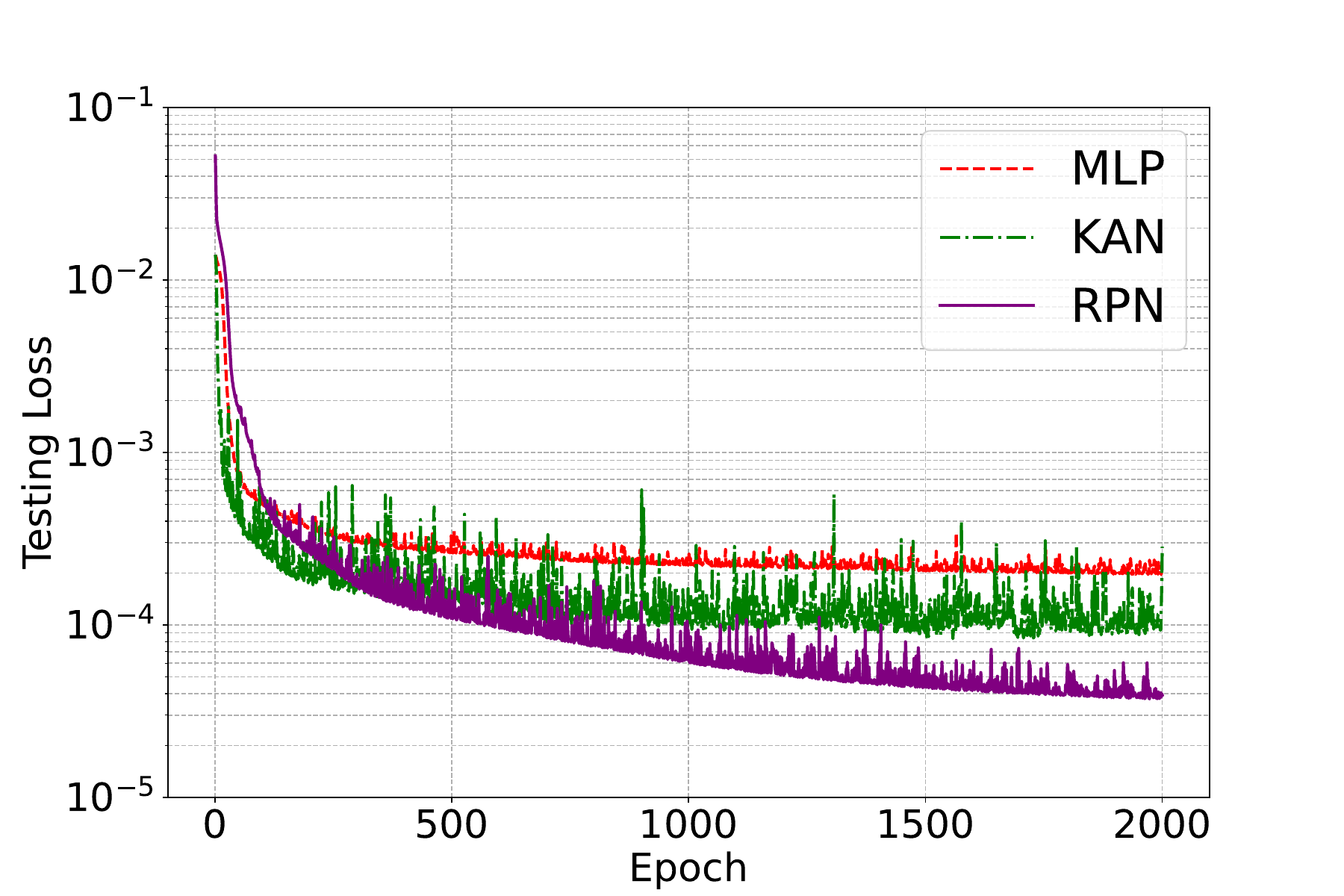}
        \caption{Testing MSE on Feynman Eq. I.9.18}
        \label{fig:feynman_4_test}
    \end{subfigure}
    \begin{subfigure}{0.48\textwidth}
        \centering
        \includegraphics[width=\textwidth]{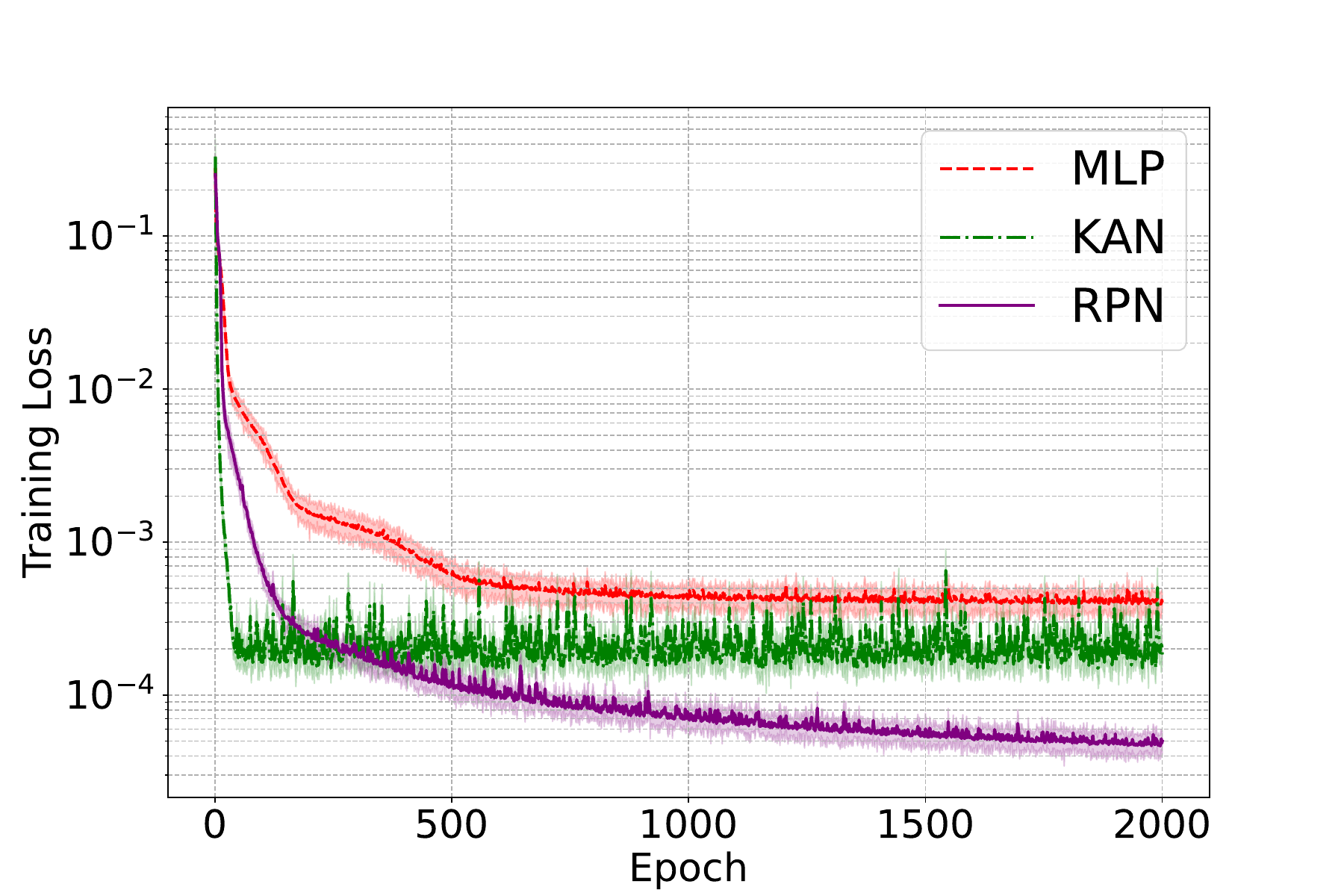}
        \caption{Training MSE on Feynman II.35.18}
        \label{fig:feynman_79_train}
    \end{subfigure}
    \hfill
    \begin{subfigure}{0.48\textwidth}
        \centering
        \includegraphics[width=\textwidth]{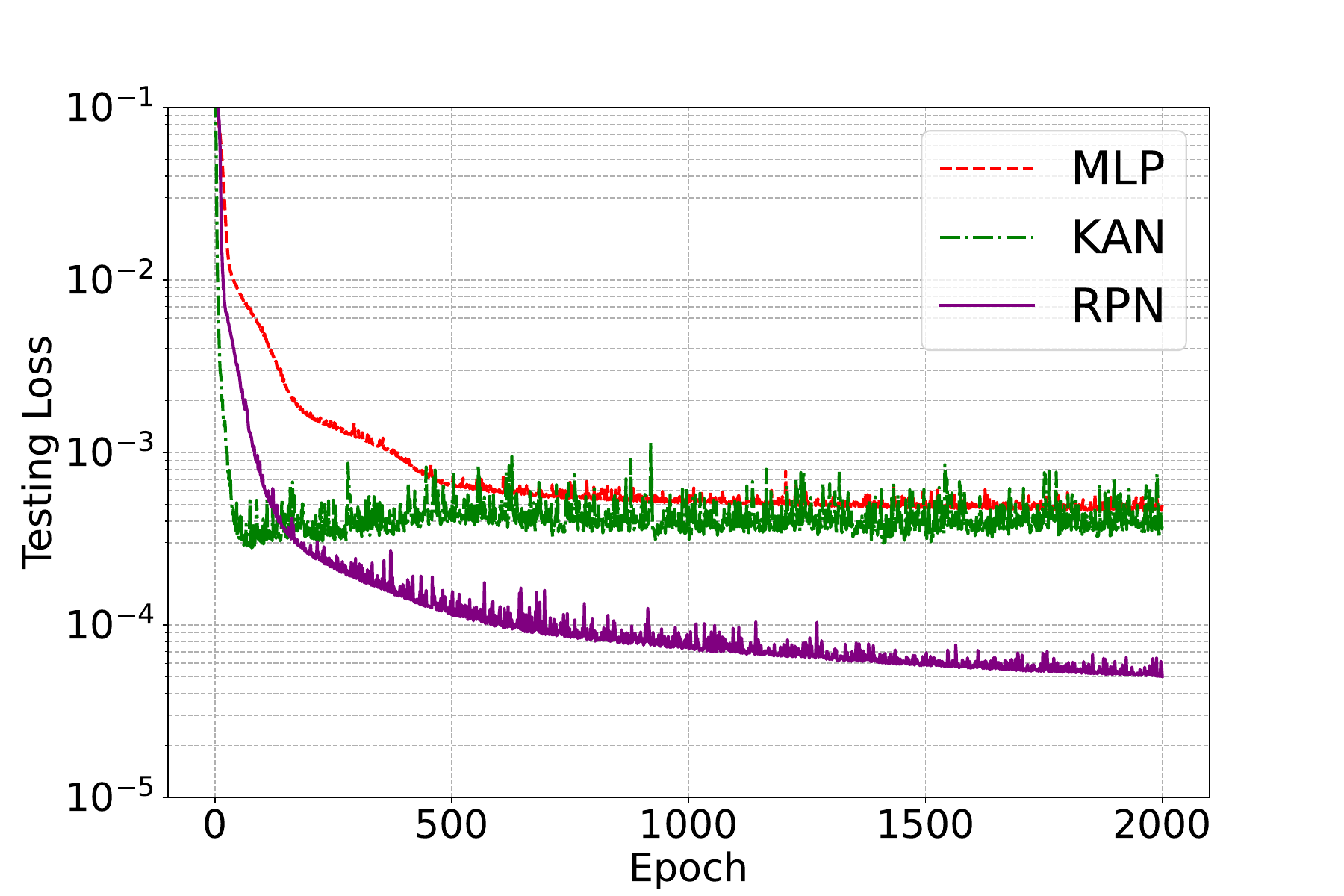}
        \caption{Testing MSE on Feynman Eq. II.35.18}
        \label{fig:feynman_79_test}
    \end{subfigure}
    \caption{Training and testing MSE curves of MLP, KAN and {\our} on fitting Feynman functions Eq. I.9.18 and Eq. II.35.18. The x axis denotes the training epochs and the y axes denote the training and testing MSE of {\our} on these two functions, respectively.}
    \label{fig:feynman_eq_curves}
\end{figure}
%------------------------------

To further illustrate the training process of MLP, KAN, and {\our} in approximating continuous functions, we randomly selected two Feynman functions: Eq. I.9.18 and Eq. II.35.18 (formulas available in Table~\ref{tab:feynman_comparison}). Figure~\ref{fig:feynman_eq_curves} presents the model training and testing curves for each epoch.
The {\our} method shown in the plots uses extended expansion, low-rank reconciliation (rank=1), and a linear remainder function. Notably, this configuration of {\our} has only half the learnable parameters of MLP and KAN.

The plots reveal that both MLP and KAN can approximate these functions with MSE on the scale of $10^{-4}$. However, {\our}, only using less than half of the learnable parameters, significantly outperforms MLP and KAN, achieving MSE on the scale of $10^{-5}$. Moreover, {\our}'s training and testing curves consistently descend with increasing epochs, in contrast to MLP and KAN. Remarkably, even without explicit parameter regularization, {\our} does not exhibit overfitting problems. These results further demonstrate {\our}'s superior performance and stability in continuous function approximation tasks.

%=======================================

\subsubsection{Low-Rank Reconciliation Function Rank Parameter Selection}

\begin{figure}[t]
    \centering
    \begin{subfigure}{0.9\textwidth}
        \centering
        \includegraphics[width=\textwidth]{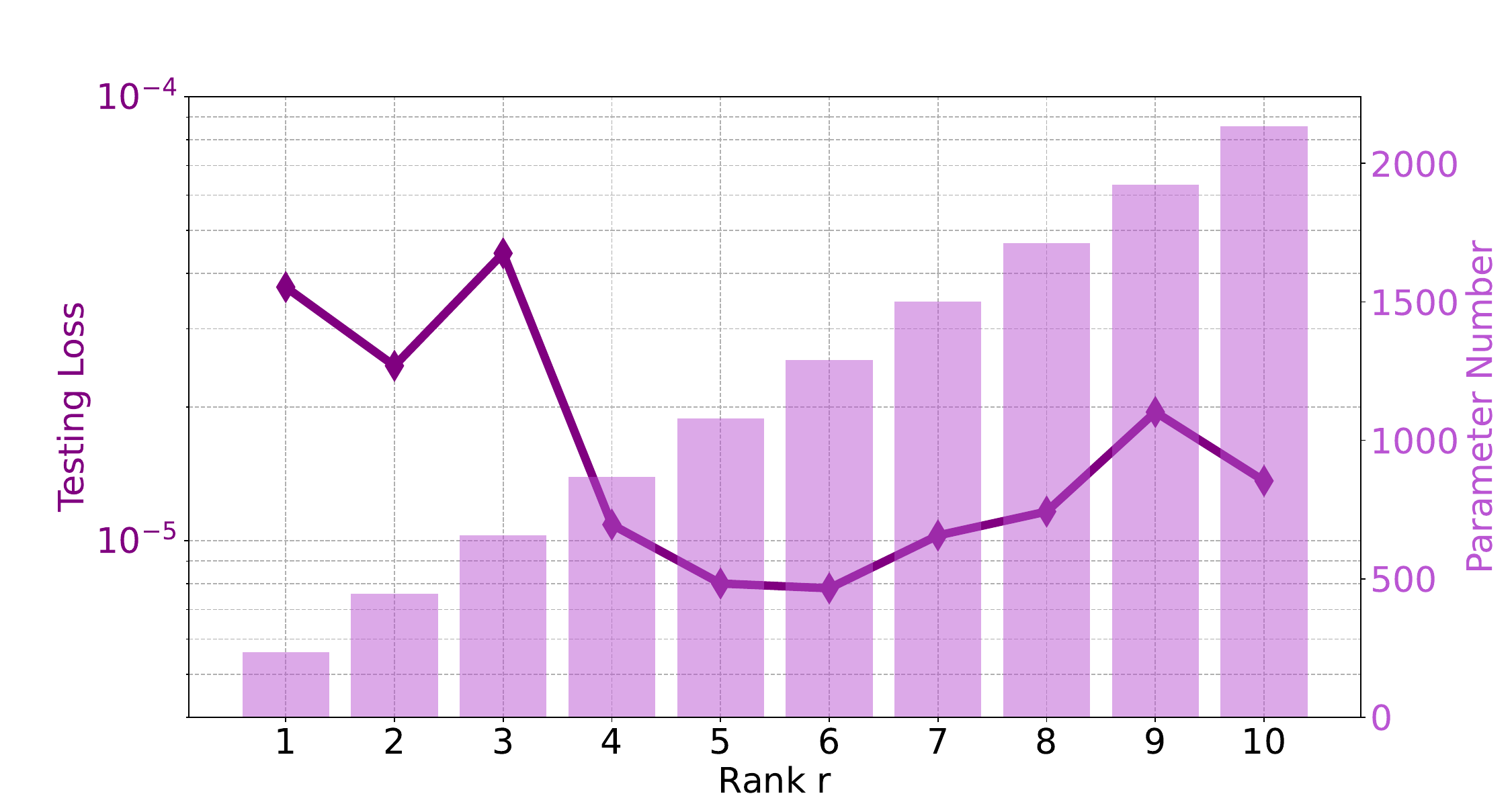}
    \end{subfigure}
        \caption{Analysis of rank parameter $r$ for low-rank reconciliation function used in {\our} on Feynman function Eq. I.9.18. The x axis: rank parameter $r$ value; left y axis: testing MSE loss; and right y axis: learnable parameter number.}
    \label{fig:feynman_r_analysis}
\end{figure}

The {\our} model analyzed previously used a low-rank reconciliation function with rank $r=1$ by default, which reduced the number of parameters to half that of MLP and KAN for both composite and Feynman function datasets. However, in real-world applications where parameter constraints are less stringent, the rank parameter in the LoRR reconciliation function can be fine-tuned to further enhance {\our}'s learning performance beyond what was reported in the previous tables.

Figure~\ref{fig:feynman_r_analysis} illustrates the learning performance and parameter count of {\our} when approximating Feynman Eq. I.9.18 with varying rank values. The testing loss curve shows that as the rank parameter $r$ increases, the model's testing loss initially decreases steadily, then increases, reaching its lowest point of $7.81 \times 10^{-6}$ at rank $r=6$. Concurrently, the number of parameters in {\our} grows consistently from $235$ at $r=1$ to $2,134$ at $r=10$.

The following subsection on discrete data classifications will provide more detailed information on ablation studies of {\our} with different expansion, reconciliation, and remainder functions.
    
%=================================================

\subsection{Discrete Image and Text Classification}\label{subsec:discrete_classification}

In addition to continuous function approximation tasks, this subsection examines the effectiveness of {\our} for discrete data classification, encompassing both image and text classification. We organize this subsection as follows: First, we introduce the dataset descriptions and experimental setups. Next, we investigate the effectiveness of different data expansion techniques, parameter reconciliation methods, and remainder functions, as well as conduct detailed analyses of {\our} in terms of model and reconciliation function hyper-parameters. Based on these insights, we apply {\our} to classification tasks on benchmark datasets for both images and text. Finally, we visualize the data expansions and parameter reconciliation process to help interpret the learning process and results of {\our}.

%---------------------------------------------------------------

\subsubsection{Dataset Descriptions and Experiment Setups}

%---------------------------------------------------------------
\begin{table}[h]
\centering
\small
\caption{Statistics of discrete image and text datasets. For the text datasets, we convert the each text data instance to a bag-of-word vector rescaled by TF-IDF, whose dimensions are also provided in the table.}
\label{tab:discrete_dataset_statistics}
\begin{tabular}{|c|c|c|c|c|c|}
\hline
\multirow{2}{*}{} & \multicolumn{2}{c|}{\textbf{Image Datasets}} & \multicolumn{3}{c|}{\textbf{Text Datasets}} \\ \cline{2-6}
                          & \textbf{MNIST} & \textbf{CIFAR-10} & \textbf{IMDB} & \textbf{AGNews} & \textbf{SST2} \\ \hline
Train \#                     & 60,000  & 50,000  & 25,000  & 120,000  & 67,349  \\ \hline
Test \#                     & 10,000  & 10,000  & 25,000  & 7,600  & 872  \\ \hline
Input Dim.                     & 28 $\times$ 28  & 32 $\times$ 32 $\times$ 3  & 26,964  & 25,985  & 10,325  \\ \hline
Output Dim.                     & 10  & 10  & 2  & 4  & 2  \\ \hline
\end{tabular}
\end{table}

%& \textbf{ImageNet} 
%& 1,281,167  
%& 50,000  
%& 224 $\times$ 224 $\times$ 3  
%& 1,000  
%---------------------------------------------------------------

%ImageNet (ILSVRC2012) \cite{ILSVRC15}

\subsubsubsection{\textbf{Dataset Descriptions}}: To demonstrate the generalizability and effectiveness of {\our}, in this part, we will provide the experimental investigations of {\our} for discrete data classification. Specifically, this subsection will focus on the experimental investigations on two categories of discrete datasets described below, whose basic statistical information is also provided in Table~\ref{tab:discrete_dataset_statistics}.
\begin{itemize}
\item \textbf{Image Datasets}: We use two benchmark datasets, MNIST and CIFAR-10, to investigate the performance of {\our} for image classification. Images in MNIST are all in the grayscale, while those in CIFAR-10 are colored instead.

\item \textbf{Text Datasets}: To examine the performance of {\our} for text classification, we employ three text benchmark datasets: IMDB, AGNews, and SST2. The AGNews is a multi-class dataset, while IMDB and SST2 are both binary-class datasets.

\end{itemize}

\subsubsubsection{\textbf{Experiment Setups}}: These image and text benchmark datasets have been pre-partitioned into training and testing sets, which will be used for all methods in our experiments. We preprocess the images by flattening and normalizing them before model training. For the text datasets, we use the sklearn TF-IDF vectorizer to preprocess the text inputs. The performance of all comparison methods, including {\our}, on these datasets are evaluated by using Accuracy as the default metric.

%---------------------------------------------------------------

\subsubsection{Component Function Composition Analysis}

%---------------------------------------------------------------
\begin{figure}[h!]
    \centering
    \begin{subfigure}{1.0\textwidth}
        \centering
        \includegraphics[width=\textwidth]{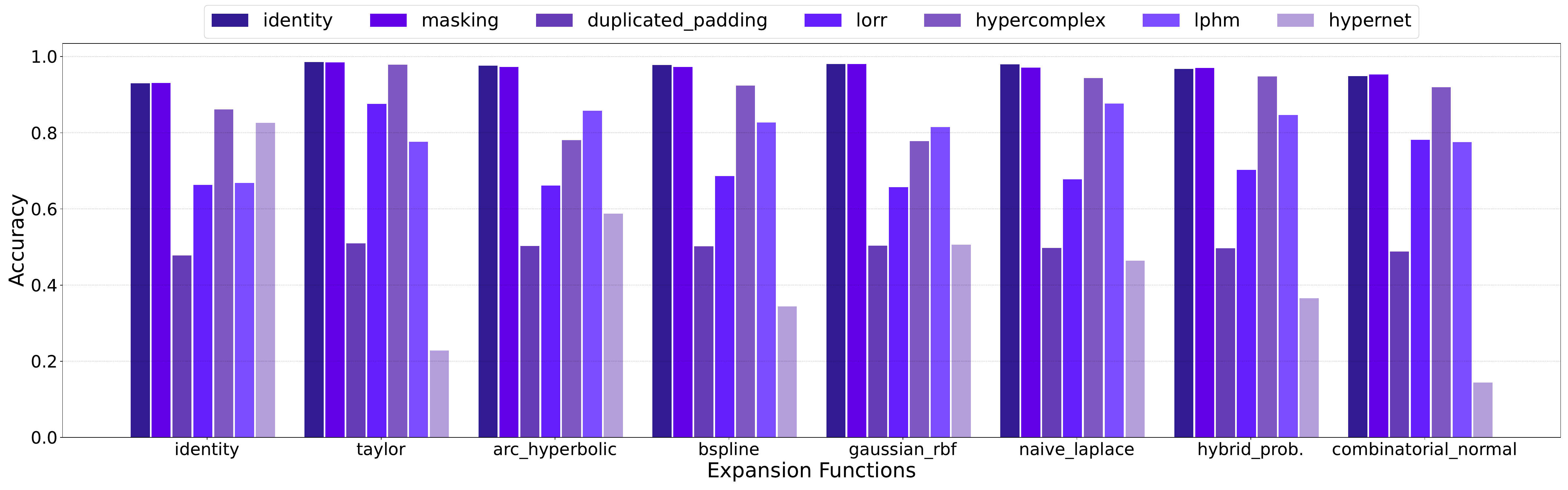}
        \caption{Ablation Studies on Expansion and Reconciliation Functions of {\our} with Zero Remainder}
        \label{fig:mnist_ablation_studies_zero_remainder}
    \end{subfigure}
    
    \vspace{15pt}
    
    \begin{subfigure}{1.0\textwidth}
        \centering
        \includegraphics[width=\textwidth]{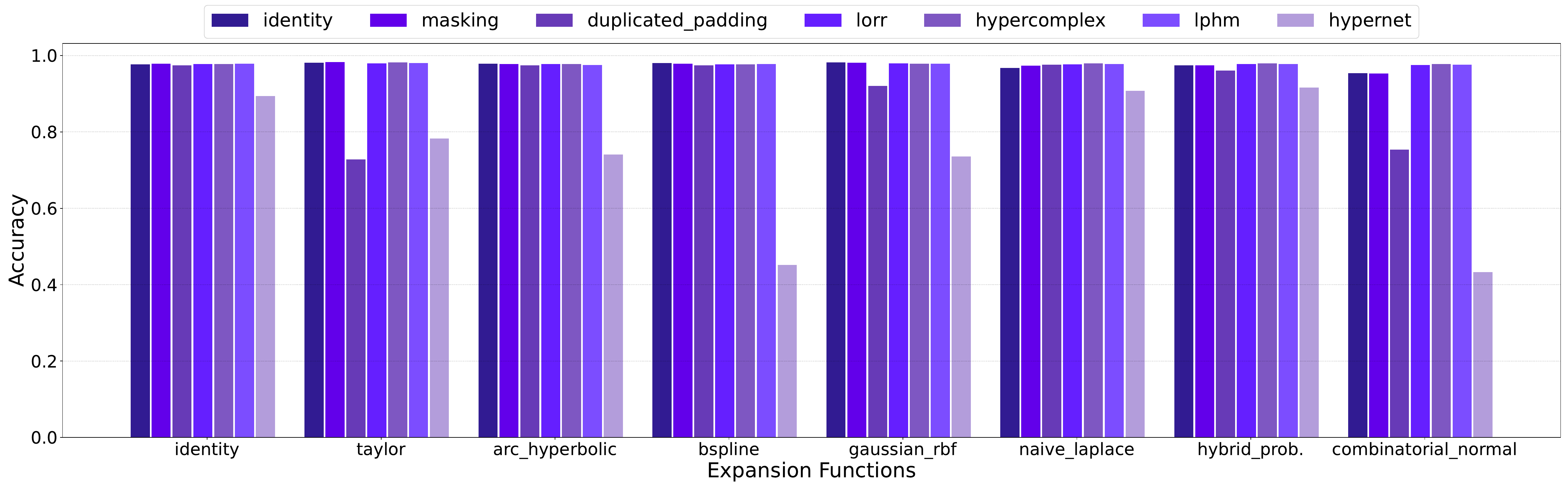}
        \caption{Ablation Studies on Expansion and Reconciliation Functions of {\our} with Linear Remainder}
        \label{fig:mnist_ablation_studies_linear_remainder}
    \end{subfigure}
    \caption{An illustration of the learning performance of {\our} with different expansion, reconciliation and remainder functions based on the MNIST dataset. The x axis denotes different expansion functions, and the y axis denotes the testing accuracy obtained by these methods. The bars with different colors denote different reconciliation functions with their names indicated in the legend. Plot (a): {\our} with zero remainder function; and Plot (b): {\our} with linear remainder function. For all the reconciliation functions, we just use their default hyper-parameters in the {\toolkit} toolkit without any tuning.}
    \label{fig:mnist_ablation_studies}
\end{figure}
%---------------------------------------------------------------

Before presenting the performance of {\our} and other comparison baselines, we will first analyze the component functions involved in {\our} to inform the design of an architecture that can achieve better performance. Figure~\ref{fig:mnist_ablation_studies} illustrates various combinations of the expansion, reconciliation, and remainder functions, as well as their performance on the MNIST dataset for image classification. Plot (a) corresponds to the zero remainder function, while Plot (b) corresponds to the linear remainder function. For both plots, the x axis denotes the expansion functions, and the bars in different colors denote different reconciliation functions, whose names are indicated in the plot legends.

%---------------------------------------------------------------
\begin{figure}[h!]
    \centering
    \begin{subfigure}{0.32\textwidth}
        \centering
        \includegraphics[width=\textwidth]{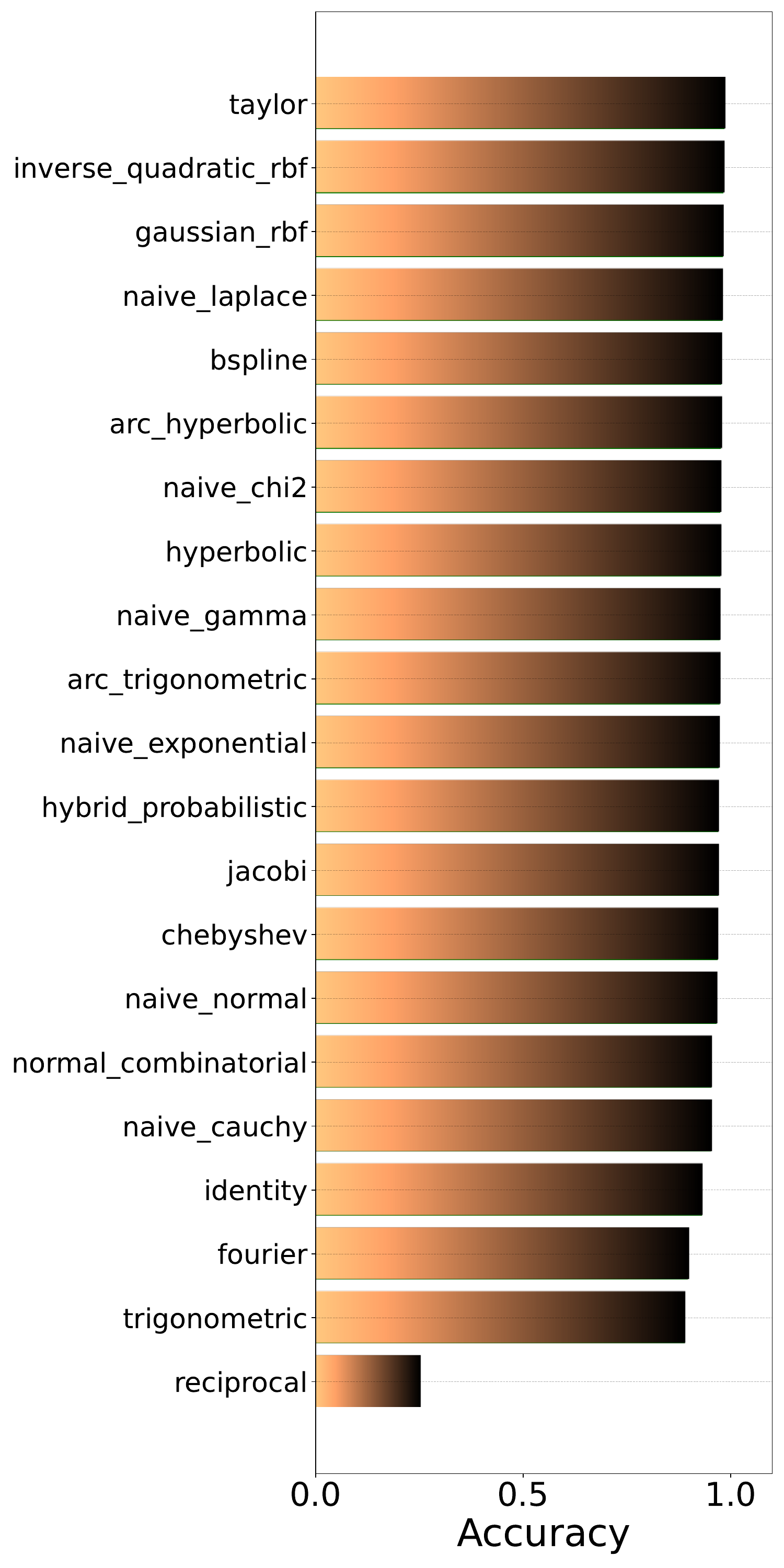}
        \caption{Expansion (Max Acc.)}
    \end{subfigure}
    \hfill
    \begin{subfigure}{0.32\textwidth}
        \centering
        \includegraphics[width=\textwidth]{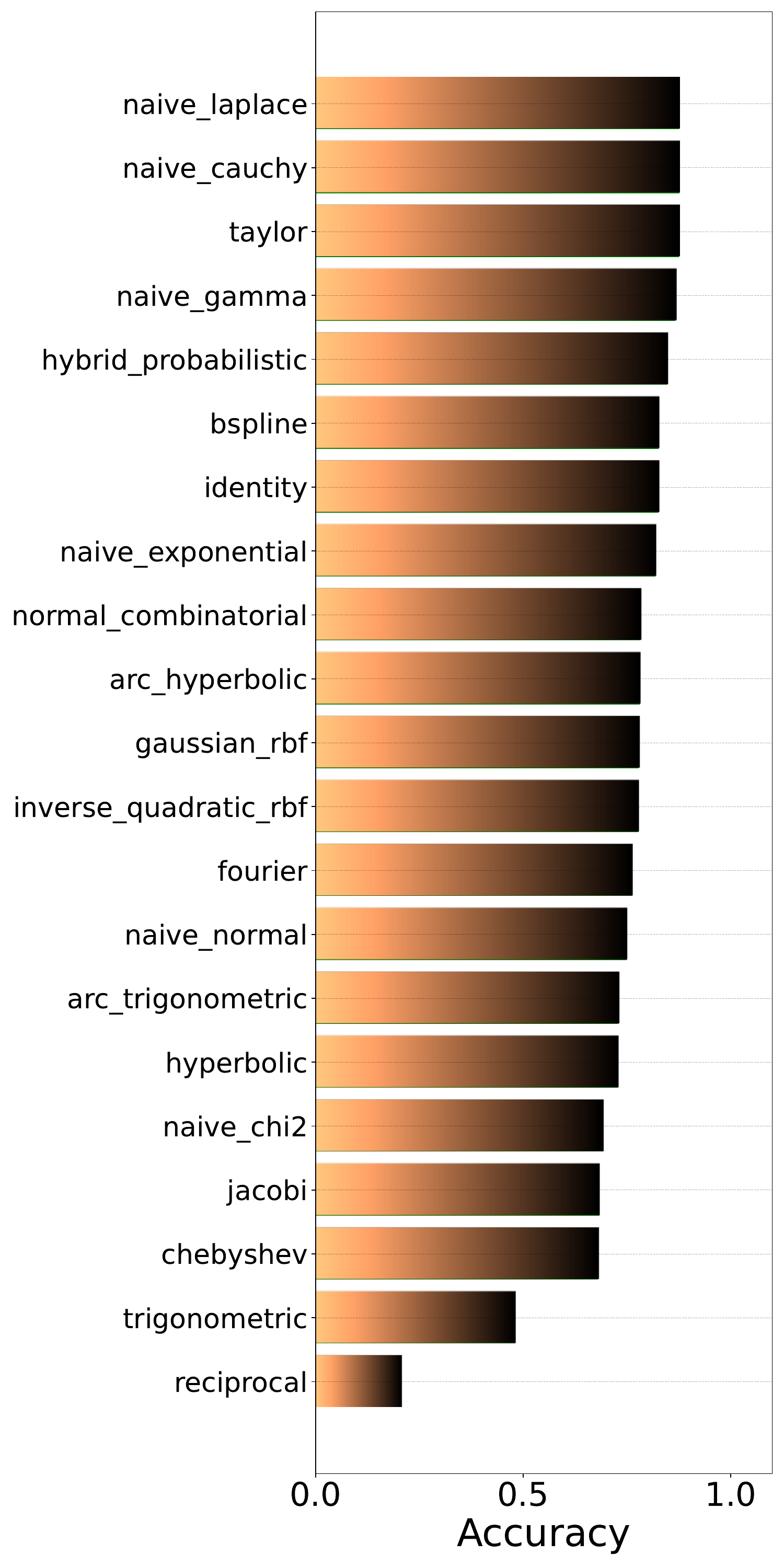}
        \caption{Expansion (Median Acc.)}
    \end{subfigure}
    \hfill
    \begin{subfigure}{0.32\textwidth}
        \centering
        \includegraphics[width=\textwidth]{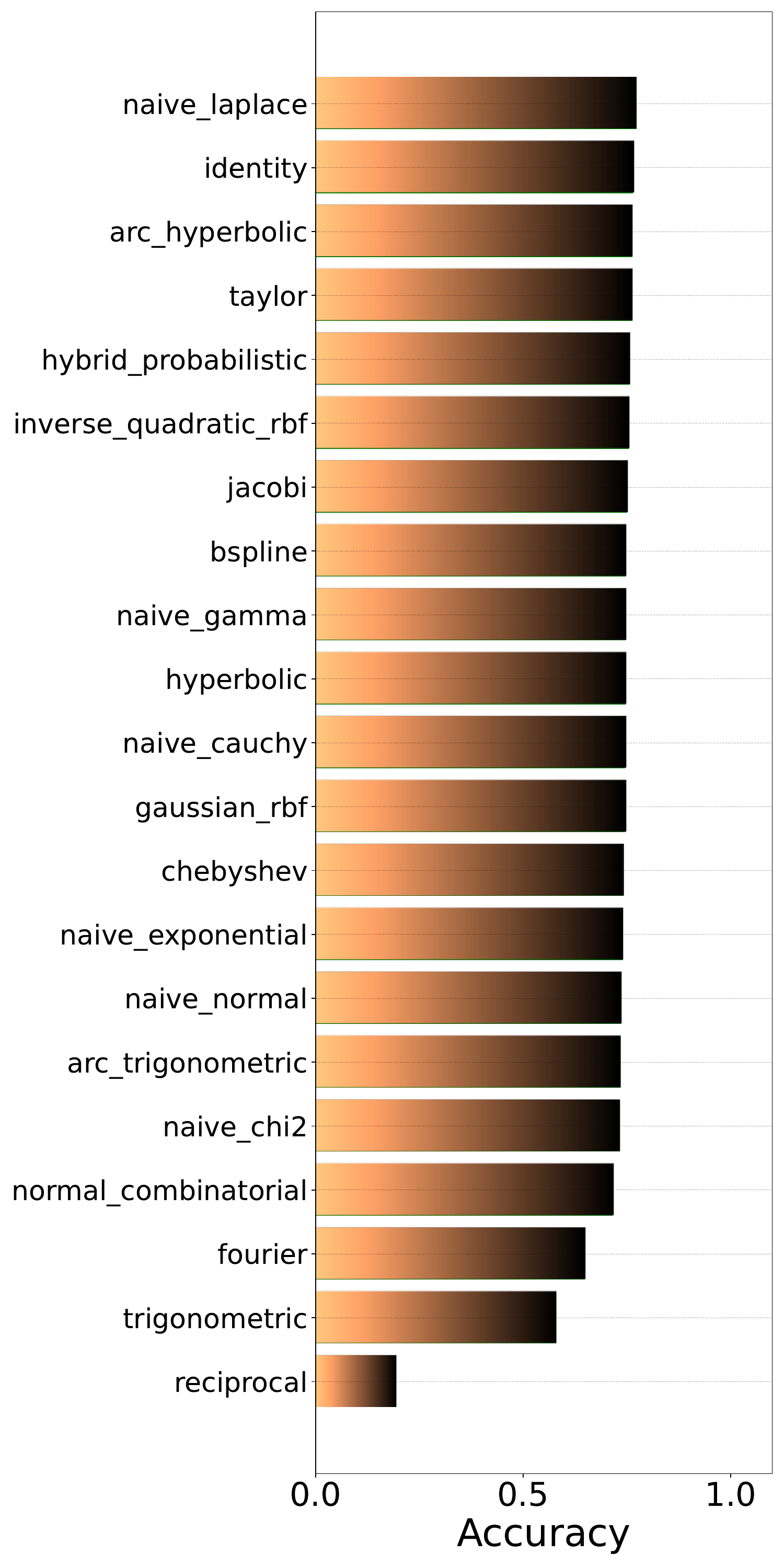}
        \caption{Expansion (Mean Acc.)}
    \end{subfigure}
    
    \vspace{15pt}
    
    \begin{subfigure}{0.32\textwidth}
        \centering
        \includegraphics[width=\textwidth]{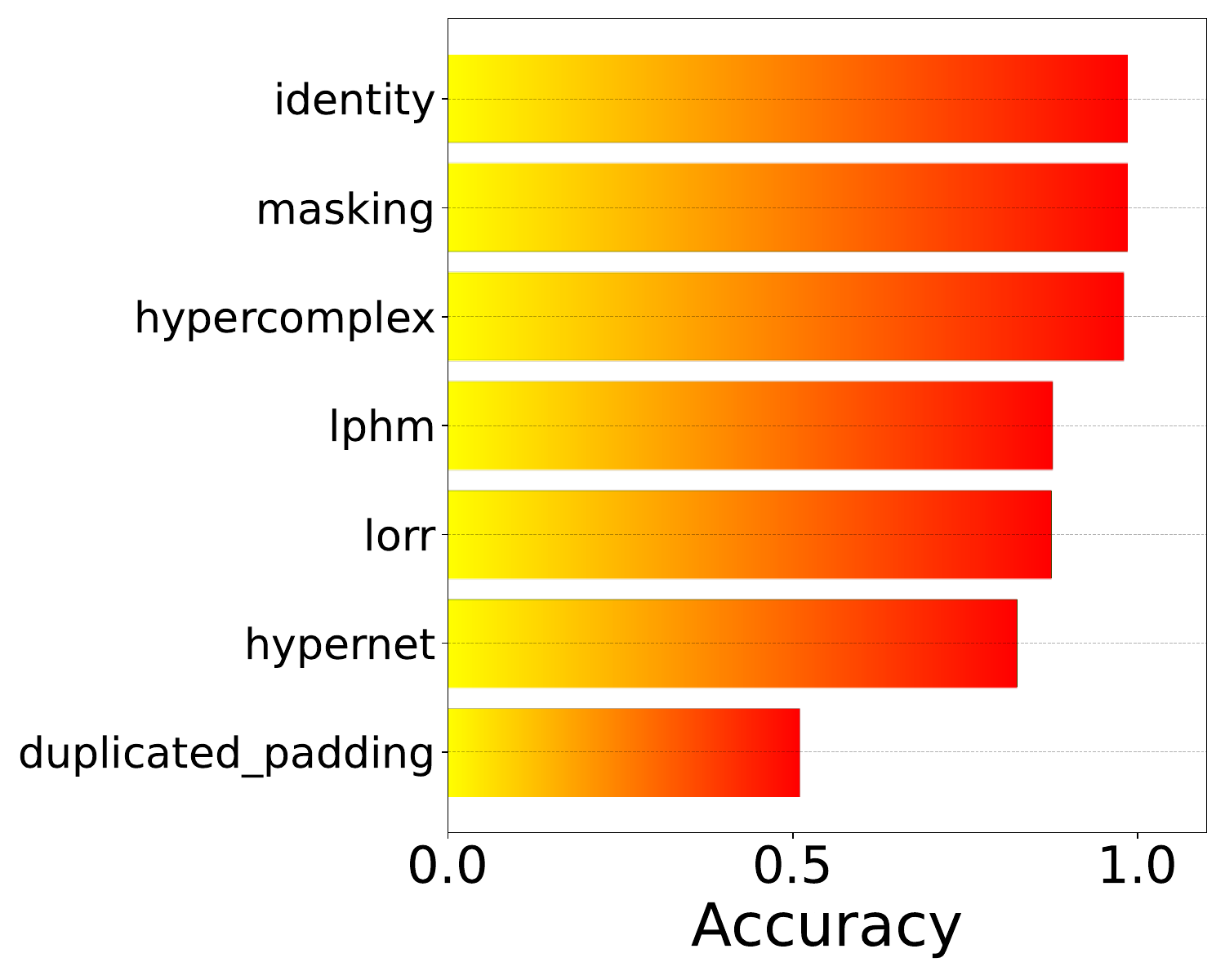}
        \caption{Reconciliation (Max Acc.)}
    \end{subfigure}
    \hfill
    \begin{subfigure}{0.32\textwidth}
        \centering
        \includegraphics[width=\textwidth]{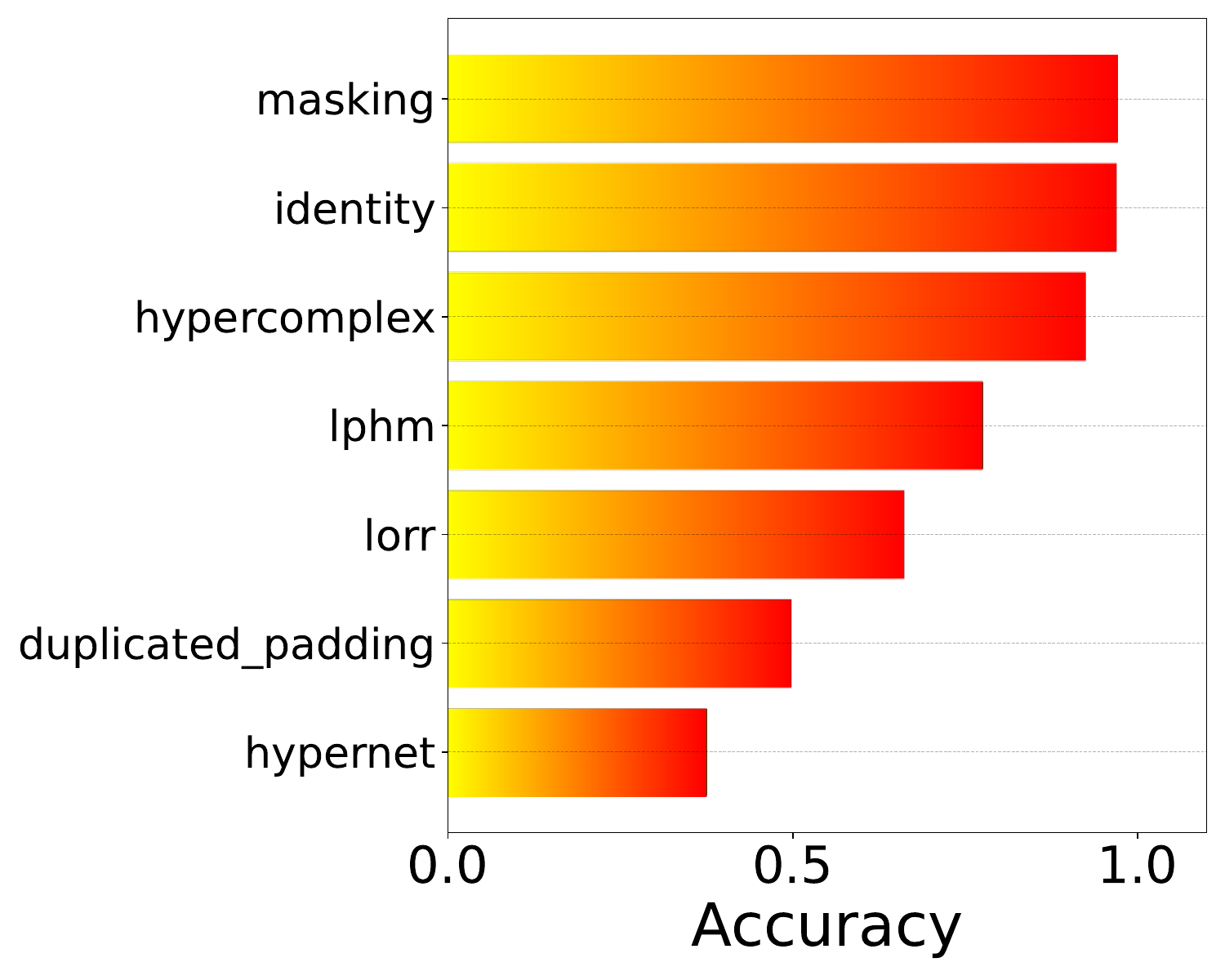}
        \caption{Reconciliation (Median Acc.)}
    \end{subfigure}
    \hfill
    \begin{subfigure}{0.32\textwidth}
        \centering
        \includegraphics[width=\textwidth]{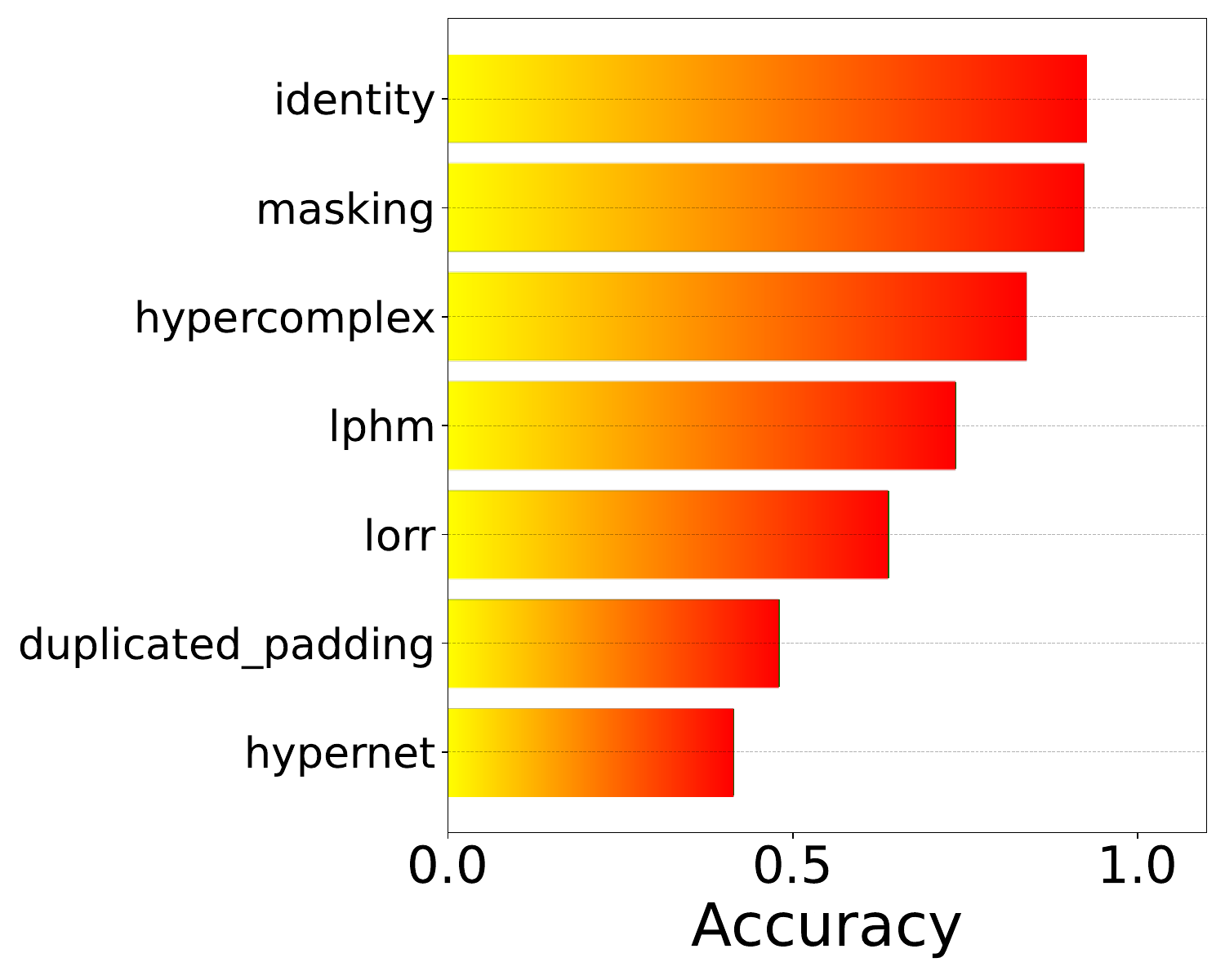}
        \caption{Reconciliation (Mean Acc.)}
    \end{subfigure}

    \vspace{15pt}
    
    \begin{subfigure}{0.32\textwidth}
        \centering
        \includegraphics[width=\textwidth]{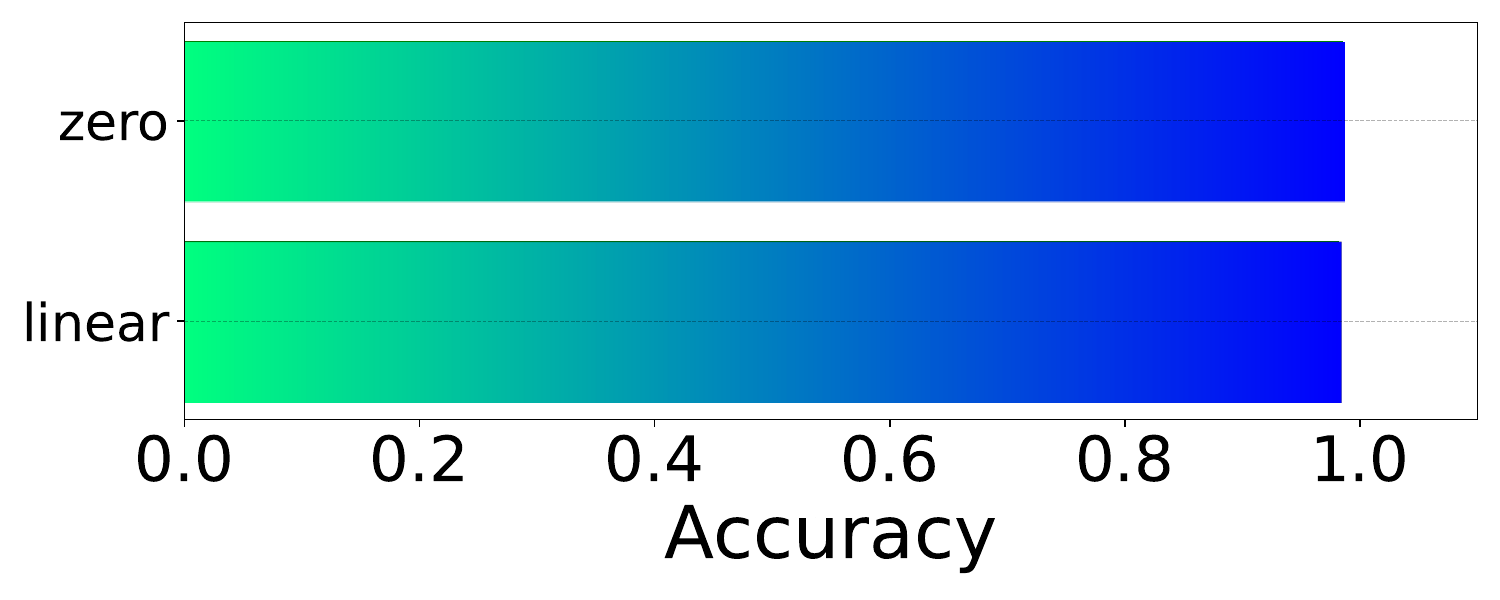}
        \caption{Remainder (Max Acc.)}
    \end{subfigure}
    \hfill
    \begin{subfigure}{0.32\textwidth}
        \centering
        \includegraphics[width=\textwidth]{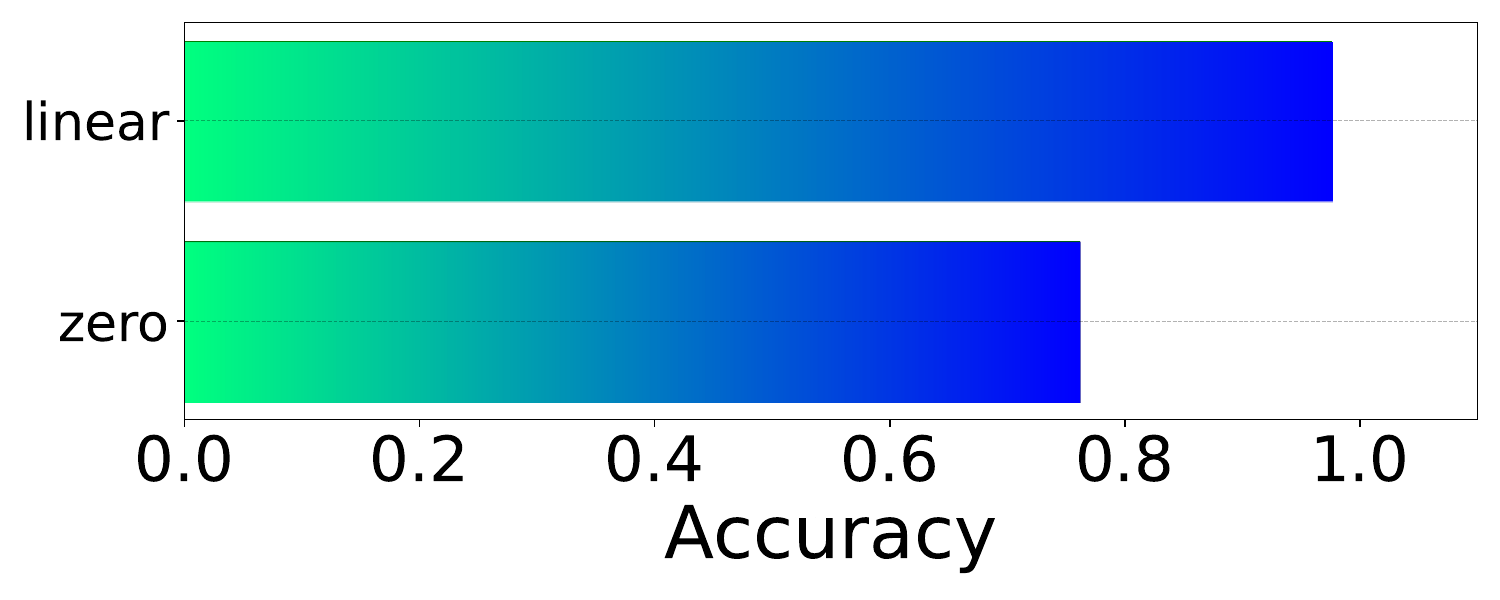}
        \caption{Remainder (Median Acc.)}
    \end{subfigure}
    \hfill
    \begin{subfigure}{0.32\textwidth}
        \centering
        \includegraphics[width=\textwidth]{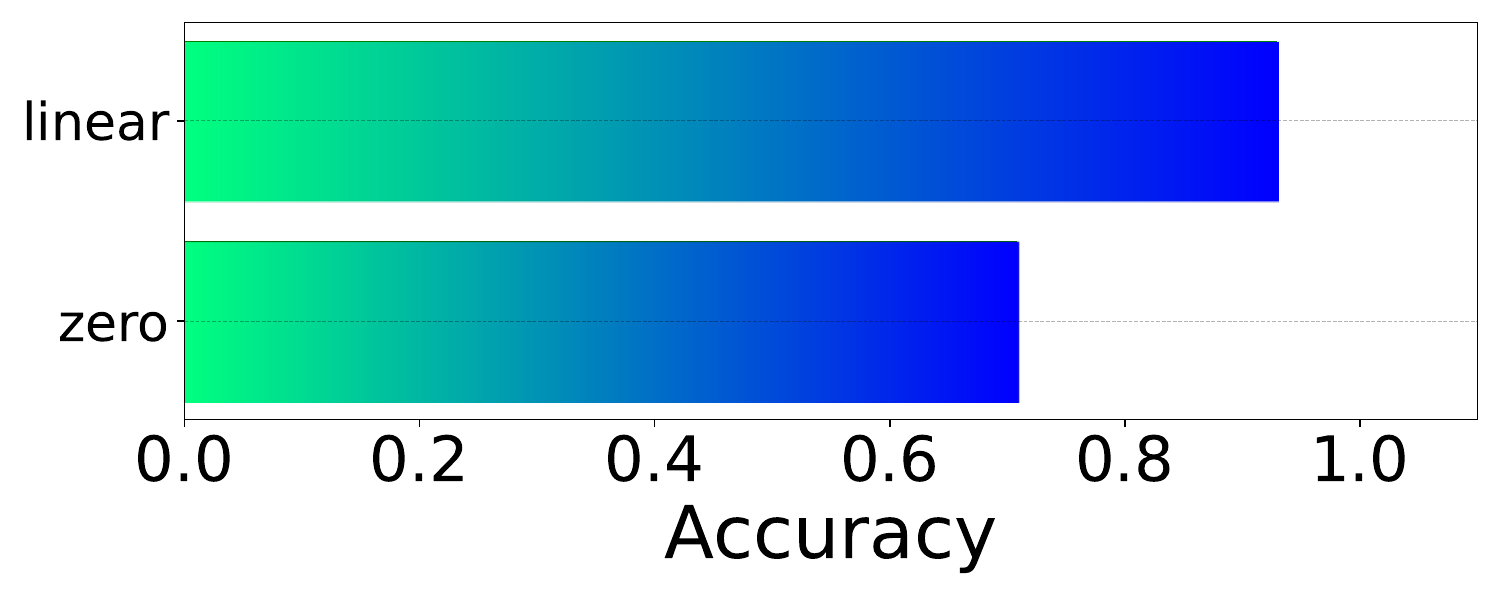}
        \caption{Remainder (Mean Acc.)}
    \end{subfigure}
    
    \caption{An illustration of the effectiveness of individual component functions in {\our} based on the MNIST dataset. Plots (a)-(c): performance of expansion functions; Plots (d)-(f): performance of reconciliation functions; and Plots (g)-(i): performance of remainder functions. For each individual function, we calculate the max, median, and mean accuracy scores obtained by the compositions involving them in the plots.}
    \label{fig:mnist_ablation_studies_individual_component_functions}
\end{figure}
%---------------------------------------------------------------

The plots reveal several interesting observations for {\our}'s performance. First of all, for {\our} with the zero remainder function as illustrated in Plot (a), its performance will solely depend on the expansion and reconciliation functions. Meanwhile, among all these reconciliation functions, identity and masking reconciliation outperform others, while hypernet and duplicated padding-based reconciliation underperform expectations. For expansion functions, Taylor's expansion, Gaussian RBF expansion, and naive Laplace distribution-based expansion slightly outperform the others. Notably, by comparing Plot (a) with Plot (b), we can observe that the linear remainder function dramatically improves performance for most expansion-reconciliation function pairs.

These observations provide valuable insights about the effectiveness of different component function combinations in {\our}. They will also help guid the selection of optimal functions for improved performance of {\our} to address the discrete data classification tasks studied in this paper.

%---------------------------------------------------------------

%---------------------------------------------------------------

\subsubsection{Individual Component Function Analysis}

Furthermore, to investigate the effectiveness of individual component functions, we also enumerate all potential combinations of these functions to build the {\our} model and apply it to the MNIST dataset. For each individual function, we extract all these potential combinations involving them and obtain their testing accuracy scores. The median, mean, and max performance are provided in Figure~\ref{fig:mnist_ablation_studies_individual_component_functions}. To exclude the performance boost created by the linear remainder, we use zero remainder by default for Plots (a)-(f), which will illustrate their expansion and reconciliation functions' effectiveness without remainders.

In Figure~\ref{fig:mnist_ablation_studies_individual_component_functions}, we sort these component functions according to their max, median, and mean accuracy scores, respectively. Among the 21 expansion functions investigated, those obtaining the highest accuracy scores in Plot (a) are Taylor's expansion, Inverse Quadratic RBF-based expansion, Gaussian RBF-based expansion, Naive Laplace-based expansion, and B-spline-based expansion. In addition to the maximum accuracy scores, the median and mean accuracy-based ranking of the expansion functions in Plots (b)-(c) illustrates the robustness of their performance when combined with various reconciliation functions.

Regarding reconciliation functions, as illustrated in Plot (d), those achieving the highest accuracy scores include Identity reconciliation, Masking-based reconciliation, and Hypercomplex Multiplication-based reconciliation. As shown in Plots (d)-(f), consistent with Figure~\ref{fig:mnist_ablation_studies}, the performance of the hypernet and duplicated padding-based reconciliation is slightly below the expectations, especially compared with the other reconciliation functions. It's important to note that we use default hyper-parameters for all these reconciliation functions, and the lower scores for these functions reported here don't necessarily indicate their ineffectiveness. Below, we will also provide a tuning of the hyper-parameters for some of the reconciliation functions to demonstrate that they can achieve comparable performance to the identity reconciliation functions with minor hyper-parameter tuning, while having far fewer learnable parameters.

For the remainder functions, according to Plots (g)-(i), linear remainder consistently outperforms zero remainder, demonstrating that remainder functions can provide complementary information to the learning results. In addition to these plots, more comprehensive results of the performance scores, time cost, and parameter number of these different component function compositions are provided in Tables~\ref{tab:mnist_comparison_identity}-\ref{tab:mnist_comparison_normal_combinatorial} in the Appendix Section~\ref{subsec:appendix_mnist_ablation_studies}. Readers may also refer to those tables for more detailed investigation results of {\our} composed with different component functions.

%---------------------------------------------------------------
\subsubsection{Model Depth and Width Analysis}

%---------------------------------------------------------------
\begin{table}[h!]
\scriptsize
\centering
\caption{An investigation of model depth and width on the performance of {\our}.}
\label{tab:mnist_depth_width_analysis}
\begin{tabular}{|c|c|c|c|c|c|c|c|c|c|c|}
\hline
\multirow{2}{*}{} & \multicolumn{10}{c|}{\textbf{Model Layer Number} (based on {\our} with 1 head per layer)} \\ \cline{2-11}
                          & \textbf{1} & \textbf{2} & \textbf{3} & \textbf{4} & \textbf{5} & \textbf{6} & \textbf{7} & \textbf{8} & \textbf{9} & \textbf{10} \\ \hline
Accuracy                     & 0.9791 & 0.9849 & \textbf{0.9853} & \textbf{0.9855} & \textbf{0.9852} & 0.9849 & 0.9848 & 0.9842 & 0.9833 & 0.9830    \\ \hline
Param. \#                     & 6.15M & 39.43M & 39.7M & 39.96M & 40.23M & 40.49M & 40.76M & 41.03M & 41.29M & 41.56M    \\ 
\hline
\hline
\multirow{2}{*}{} & \multicolumn{10}{c|}{\textbf{Model Head Number} (based on {\our} with 4 layer)} \\ \cline{2-11}
                          & \textbf{1} & \textbf{2} & \textbf{3} & \textbf{4} & \textbf{5} & \textbf{6} & \textbf{7} & \textbf{8} & \textbf{9} & \textbf{10} \\ \hline
Accuracy                     & 0.9855 & 0.9853 & 0.9846 & \textbf{0.9867} & 0.9847 & 0.9852 & 0.9851 & \textbf{0.9862} & 0.9852 & \textbf{0.9856}   \\ \hline
Param. \#                     & 39.96M & 79.92M & 119.89M & 159.85M & 199.81M & 239.77M & 279.74M & 319.7M & 359.66M & 399.62M    \\ 
\hline
\end{tabular}
\end{table}
%---------------------------------------------------------------

Beyond the shallow and narrow models analyzed previously, we also investigate the impacts of model depth and width on the performance of {\our}. The results are illustrated in Table~\ref{tab:mnist_depth_width_analysis}. For this analysis, we use {\our} with Taylor's expansion (order 2), identity reconciliation, and zero remainder as the default model architecture on the MNIST dataset.

The top part of Table~\ref{tab:mnist_depth_width_analysis} shows results for models with 1 head per layer, while increasing the number of layers from $1$ to $10$. Besides the input and output dimensions, for the hidden dimensions of the middle layers, we all use the default dimension $64$ in this experiment. Except for {\our} with depth 1, which achieves a best recorded testing accuracy of $0.979$, all other depths obtain testing accuracy scores above $0.980$. The highest recorded testing accuracy of $0.986$ is achieved at depth $4$.

Furthermore, for the {\our} model with 4 layers, we investigate the impact of head number on model performance, as shown in the bottom part of the table. The performance of {\our} with different numbers of heads is notably consistent and stable, achieving the highest scores with $4$, $8$, and $10$ heads, respectively.

%---------------------------------------------------------------

%---------------------------------------------------------------

\subsubsection{Reconciliation Function Hyper-Parameter Tuning Analysis}

%---------------------------------------------------------------
\begin{figure}[h!]
    \centering
    \begin{subfigure}{0.9\textwidth}
        \centering
        \includegraphics[width=\textwidth]{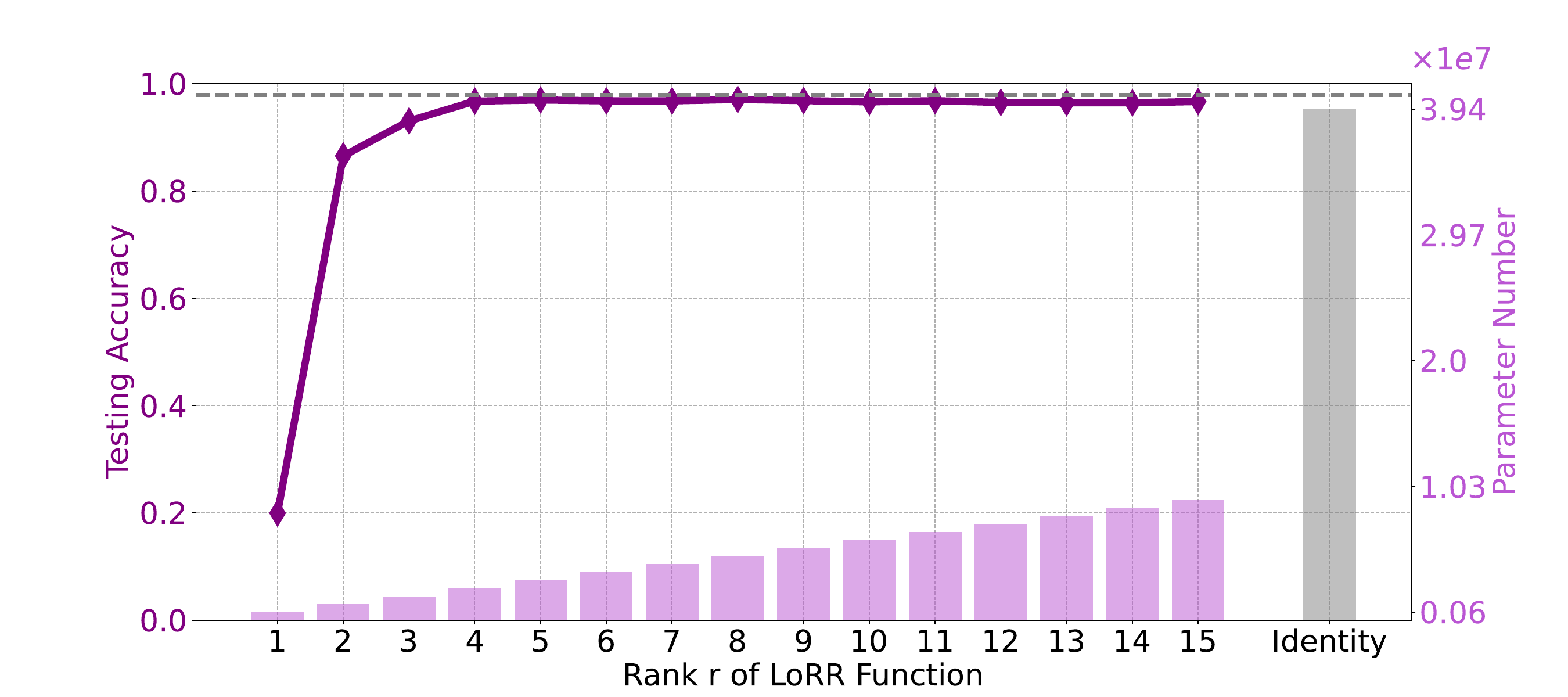}
    \end{subfigure}
        \caption{Analysis of rank parameter $r$ for LoRR reconciliation function for {\our} on MNIST dataset. Left y axis: Testing Accuracy; Right y axis: Learnable Parameter Number. For the dashed line and bar in gray color, they denote the testing accuracy (0.983) and parameter number ($3.94\times 10^7$) of the identity reconciliation function when applied to identical Taylor's expansions in {\our}.}
    \label{fig:mnist_r_analysis}
\end{figure}
%---------------------------------------------------------------

Prior to illustrating the main results of {\our} and other comparison methods, we further investigate the effectiveness of the low-rank reconciliation (LoRR) function, whose initial performance was not as good as expected in the previous Plots (d)-(f) of Figure~\ref{fig:mnist_ablation_studies_individual_component_functions}. Using the {\our} model with Taylor's expansion function (order 2) and zero remainder, we will tune the rank hyler-parameter in LoRR from $1$ to $15$. Figure~\ref{fig:mnist_r_analysis} presents both the obtained testing accuracy and the introduced model parameter numbers. For comparison, we also include the performance of the identity reconciliation function in the plot, which are represented by a gray dashed horizontal line for testing score and a gray bar for parameter number.

The curve and bar plot reveal that LoRR with ranks $1$, $2$, and $3$ yields lower performance than the identity reconciliation function. However, for ranks $4$ and above, LoRR's performance gradually approaches that of the identity reconciliation function while using significantly fewer parameters. This analysis demonstrates that with proper tuning, LoRR can achieve comparable accuracy to identity reconciliation. Similar performance have been observed for the other reconciliation functions as well. These insights will help in designing parameter-efficient {\our} models especially for the discrete data classification tasks presented below.

%---------------------------------------------------------------
%---------------------------------------------------------------

\subsubsection{The Main Results of {\our} on Discrete Data Classification}

%---------------------------------------------------------------
\begin{table}[h!]
    \caption{Classification results of discrete data classification on the image and text benchmark datasets. Both naive Bayes and SVM are implemented with sklearn, whose performance is very stable. For MLP, KAN and {\our}, we train them with $5$ randomly selected random seeds. For each random seed, they are trained with 50 epochs, and the best testing scores achieved by all these methods within these 50 epochs are cherry-picked. For MLP, KAN and {\our}, we report their average accuracy score together with the standard deviations in the table.}
    \label{tab:discrete_classification_benchmark_results}
    \small
    \centering
    \setlength{\tabcolsep}{2pt}
    \begin{tabular}{|c|c|c|c|c|c|}
        \hline
        \multirow{2}{*}{\textbf{Models}} & \multicolumn{2}{c|}{\textbf{Image Datasets}} & \multicolumn{3}{c|}{\textbf{Text Datasets}}  \\
        \cline{2-6}
         & \textbf{MNIST} & \textbf{CIFAR10} & \textbf{IMDB} & \textbf{AGNews} & \textbf{SST2}  \\
        \hline
         \hline
         \multirow{5}{*}{\makecell{Naive\\ Bayes}} & \makecell{$5.56 \times 10^{-1}$ \\ {- - - - - - - - - - - -} \\ (Gaussian)} & \makecell{$2.98 \times 10^{-1}$ \\ {- - - - - - - - - - - -} \\ (Gaussian)}  & \makecell{$6.30 \times 10^{-1}$ \\ {- - - - - - - - - - - -} \\ (Gaussian)} & \makecell{$8.14 \times 10^{-1}$ \\ {- - - - - - - - - - - -} \\ (Gaussian)}  & \makecell{$7.31 \times 10^{-1}$ \\ {- - - - - - - - - - - -} \\ (Gaussian)} \\
         \cline{2-6}
        & \makecell{$8.36 \times 10^{-1}$ \\ {- - - - - - - - - - - -} \\ (Multinomial)} & \makecell{$2.98 \times 10^{-1}$ \\ {- - - - - - - - - - - -} \\ (Multinomial)}  & \makecell{$8.37 \times 10^{-1}$ \\ {- - - - - - - - - - - -} \\ (Multinomial)} & \makecell{$9.03 \times 10^{-1}$ \\ {- - - - - - - - - - - -} \\ (Multinomial)}  & \makecell{$7.86 \times 10^{-1}$ \\ {- - - - - - - - - - - -} \\ (Multinomial)} \\
        \hline
        \hline
        \multirow{5}{*}{\makecell{SVM}} & \makecell{$9.31 \times 10^{-1}$ \\ {- - - - - - - - - - - -} \\ (Linear kernel)} & \makecell{$3.50 \times 10^{-1}$ \\ {- - - - - - - - - - - -} \\ (Linear kernel)}  & \makecell{$8.77 \times 10^{-1}$ \\ {- - - - - - - - - - - -} \\ (Linear kernel)} & \makecell{$9.18 \times 10^{-1}$ \\ {- - - - - - - - - - - -} \\ (Linear kernel)} & \makecell{$8.01 \times 10^{-1}$ \\ {- - - - - - - - - - - -} \\ (Linear kernel)} \\
        \cline{2-6}
        & \makecell{$9.79 \times 10^{-1}$ \\ {- - - - - - - - - - - -} \\ (RBF kernel)} & \makecell{$5.44 \times 10^{-1}$ \\ {- - - - - - - - - - - -} \\ (RBF kernel)}  & \makecell{$8.84 \times 10^{-1}$ \\ {- - - - - - - - - - - -} \\ (RBF kernel)} & \makecell{\boldmath $9.25 \times 10^{-1}$ \\ {- - - - - - - - - - - -} \\ (RBF kernel)} & \makecell{$8.01 \times 10^{-1}$ \\ {- - - - - - - - - - - -} \\ (RBF kernel)} \\
        \hline
        \hline
        \makecell{MLP} & \makecell{ $9.82 \times 10^{-1}$ \\ $\pm$  $6.79 \times 10^{-4}$ \\ {- - - - - - - - - - - -} \\ $[$784, 512, 256, 10$]$} & \makecell{\boldmath $5.63 \times 10^{-1}$ \\ \boldmath $\pm$  $1.67 \times 10^{-3}$ \\ {- - - - - - - - - - - -} \\ $[$784, 512, 256, 10$]$} & \makecell{$8.85 \times 10^{-1}$  \\ $\pm$ $5.54 \times 10^{-3}$ \\ {- - - - - - - - - - - -} \\ $[$26964, 128, 32, 2$]$} & \makecell{ $9.21 \times 10^{-1}$ \\ $\pm$ $8.22 \times 10^{-4}$ \\ {- - - - - - - - - - - -} \\ $[$25985, 128, 32, 4$]$} & \makecell{ $8.05 \times 10^{-1}$\\ $\pm$ $2.12 \times 10^{-3}$ \\ {- - - - - - - - - - - -} \\ $[$10325, 128, 32, 2$]$ } \\
        \hline
        \hline
        \makecell{KAN} & \makecell{ $9.75 \times 10^{-1}$ \\ $\pm$  $1.47 \times 10^{-3}$ \\ {- - - - - - - - - - - -} \\ $[$784, 64, 10$]$} & \makecell{ $5.27 \times 10^{-1}$ \\ $\pm$  $7.10 \times 10^{-4}$ \\ {- - - - - - - - - - - -} \\ $[$784, 64, 10$]$} & \makecell{5.00 $\times 10^{-1}$ \\ $\pm$  $0.00 \times 10^{0}$  \\ {- - - - - - - - - - - -} \\ $[$26964, 128, 32, 2$]$} & \makecell{2.50 $\times 10^{-1}$ \\ $\pm$  $0.00 \times 10^{0}$ \\ {- - - - - - - - - - - -} \\ $[$25985, 128, 32, 4$]$} & \makecell{4.91 $\times 10^{-1}$ \\ $\pm$  $0.00 \times 10^{0}$ \\ {- - - - - - - - - - - -} \\ $[$10325, 128, 32, 2$]$} \\
        \hline
        \hline
        \makecell{{\our}} & \makecell{\boldmath $9.86 \times 10^{-1}$ \\ \boldmath $\pm$  $8.70 \times 10^{-4}$ \\ {- - - - - - - - - - - -} \\ $[$784, 64, 64, 10$]$ } & \makecell{ $5.61 \times 10^{-1}$ \\  $\pm$  $1.66 \times 10^{-3}$ \\ {- - - - - - - - - - - -} \\ $[$3072, 512, 256, 10$]$ } & \makecell{\boldmath$8.86 \times 10^{-1}$  \\ \boldmath$\pm$ $4.59 \times 10^{-4}$ \\ {- - - - - - - - - - - -} \\ $[$26964, 128, 32, 2$]$ } & \makecell{$9.19 \times 10^{-1}$  \\ $\pm$ $2.55 \times 10^{-3}$ \\ {- - - - - - - - - - - -} \\ $[$25985, 128, 4$]$} & \makecell{\boldmath $8.07 \times 10^{-1}$  \\ \boldmath $\pm$ $1.72 \times 10^{-3}$ \\ {- - - - - - - - - - - -} \\ $[$10325, 128, 32, 2$]$}\\
        \hline
    \end{tabular}
\end{table}

%MNIST: [784, 64, 64, 10]	taylor, identity, zero	1 head per layer	[2, 2, 2]	2e-3, wd 2e-4, gamma 0.9	9875
%\makecell{{\our}} & \makecell{\boldmath $9.86 \times 10^{-1}$ \\ \boldmath $\pm$  $8.70 \times 10^{-4}$ \\ {- - - - - - - - - - - -} \\ $[$784, 64, 64, 10$]$ \\ (taylor, identity, zero)} & \makecell{ $5.61 \times 10^{-1}$ \\  $\pm$  $1.66 \times 10^{-3}$ \\ {- - - - - - - - - - - -} \\ $[$3072, 512, 256, 10$]$ \\ (identity, masking, zero)} & \makecell{\boldmath$8.86 \times 10^{-1}$  \\ \boldmath$\pm$ $4.59 \times 10^{-4}$ \\ {- - - - - - - - - - - -} \\ $[$26964, 128, 32, 2$]$ \\ (identity, lorr, zero)} & \makecell{$9.19 \times 10^{-1}$  \\ $\pm$ $2.55 \times 10^{-3}$ \\ {- - - - - - - - - - - -} \\ $[$25985, 128, 4$]$\\ (identity, lorr, zero)} & \makecell{\boldmath $8.07 \times 10^{-1}$  \\ \boldmath $\pm$ $1.72 \times 10^{-3}$ \\ {- - - - - - - - - - - -} \\ $[$10325, 128, 32, 2$]$\\ (identity, lorr, zero)}\\
%---------------------------------------------------------------

Based on our above investigations and analyses, we present the main results of {\our} on image and text benchmark datasets in Table~\ref{tab:discrete_classification_benchmark_results}. We compare {\our}'s performance with several baseline models, including naive Bayes (based on the Gaussian and multinomial distributions), SVM (with Linear and RBF kernels), MLP, and KAN. Besides the accuracy scores obtained on the benchmark datasets, the table also specifies model configurations and architecture dimensions of these methods on different benchmark datasets.

In Table~\ref{tab:discrete_classification_benchmark_results}, the aforementioned Bayesian network and Markov network are excluded from this comparison due to their extremely time-consuming training process on larger-scale datasets. These models will be discussed and compared with {\our} on smaller-sized tabular datasets for probabilistic relationship inference in the following subsection. For MLP, KAN, and {\our}, we report ``mean $\pm$ std'' accuracy obtained with $5$ randomly picked random seeds. With their implementation via sklearn, the performance scores of naive Bayes and SVM are reported directly without standard deviations due to their stability across random states. {\our}'s configurations vary by dataset: on MNIST, {\our} uses Taylor's expansion (order 2), identity reconciliation, zero remainder; on CIFAR-10, {\our} uses identity expansion, masking-based reconciliation (p=0.6), zero remainder; and on IMDB, AGNews, SST, {\our} uses Taylor's expansion (order 2), low-rank reconciliation (r=2), zero remainder.

Except for MNIST, {\our} uses much fewer learnable parameters than SVM, MLP, and KAN on these benchmark datasets due to masking and low-rank reconciliation functions. {\our} outperforms baseline methods on most discrete benchmark datasets. On MNIST, it achieves $0.986$ accuracy, significantly higher than naive Bayes, SVM, MLP, and KAN, demonstrating the effectiveness of Taylor's expansions in feature extraction from image data. On CIFAR-10 and AGNews, {\our} achieves comparable performance to the best baseline methods using fewer learnable parameters. And on IMDB and SST2, the performance of {\our} slightly exceeds the other comparison methods. These obtained experimental results all showcase the superiority of {\our} on discrete data classification tasks.

In addition to evaluating the performance of {\our}, our experiments yielded several insightful observations about the comparison methods. Among all approaches, naive Bayes stands out as the fastest, generating results within seconds, while other methods may require hours for training. SVM (with RBF kernel) and MLP demonstrate more consistent performance across diverse datasets, showcasing their robustness. Notably, the recently proposed KAN exhibits significant limitations in the experiments. Not surprisingly, KAN uses substantially more parameters than other methods and requires much longer training time. More critically, KAN fails to train effectively on text datasets using sparse vectorized data instances rescaled via bag-of-words and TF-IDF. These observations reveal major deficiencies in KAN's model design not discovered nor reported in the previous paper \cite{Liu2024KANKN}, which may pose challenges for it in replacing MLP as a new base model for more complex learning scenarios.

All implementations of {\our} in the above experiments have been included in the {\toolkit} tutorials, allowing readers to rapidly reproduce the reported experimental scores.

%---------------------------------------------------------------
\subsubsection{Data Expansion and Reconciled Parameter Visualization}

%---------------------------------------------------------------
\begin{figure}[t]
    \centering
    \begin{subfigure}{0.48\textwidth}
        \centering
        \includegraphics[width=\textwidth]{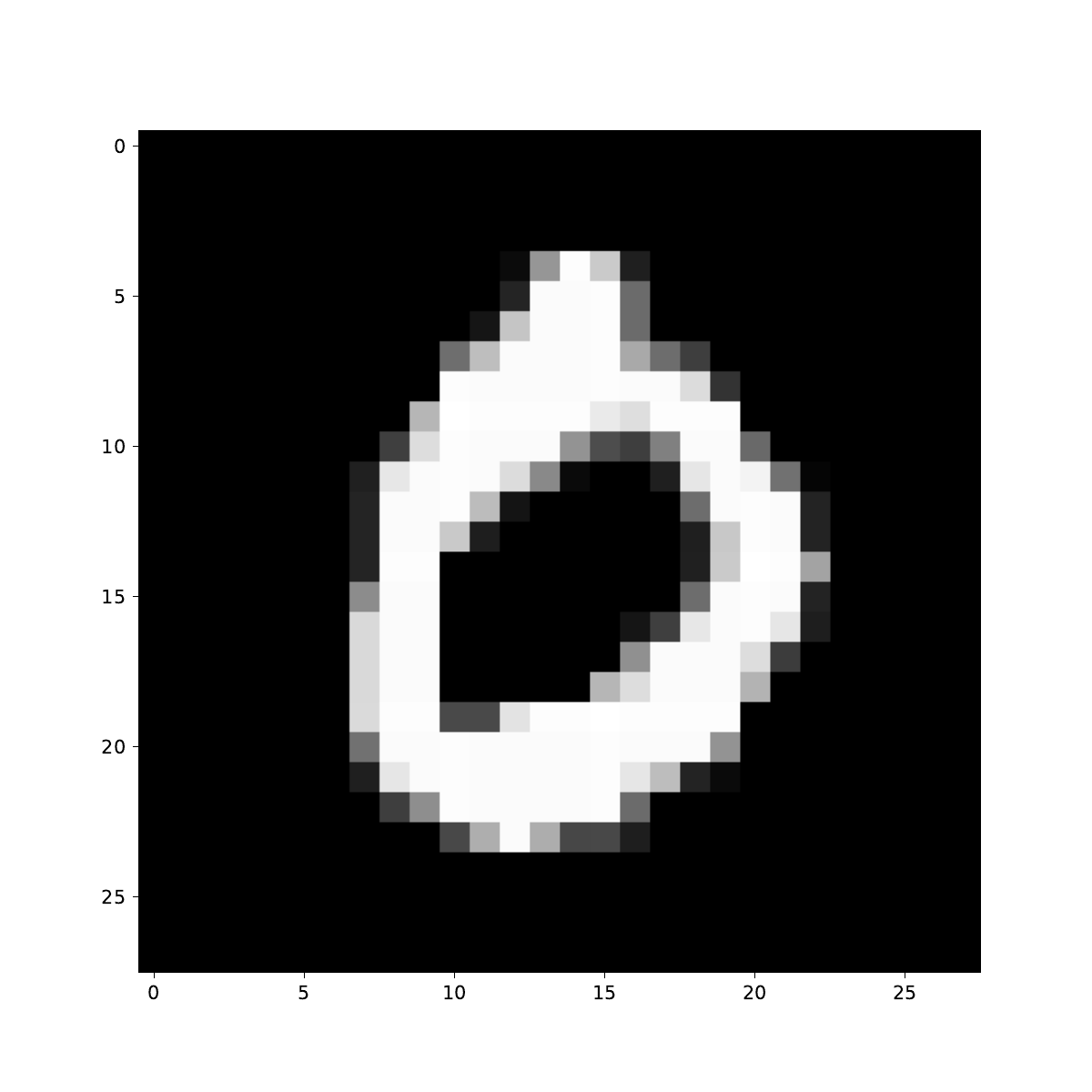}
        \caption{Taylor's Order 1 Expansion}
        \label{fig:mnist_visualization_raw_data}
    \end{subfigure}
    \hfill
    \begin{subfigure}{0.48\textwidth}
        \centering
        \includegraphics[width=\textwidth]{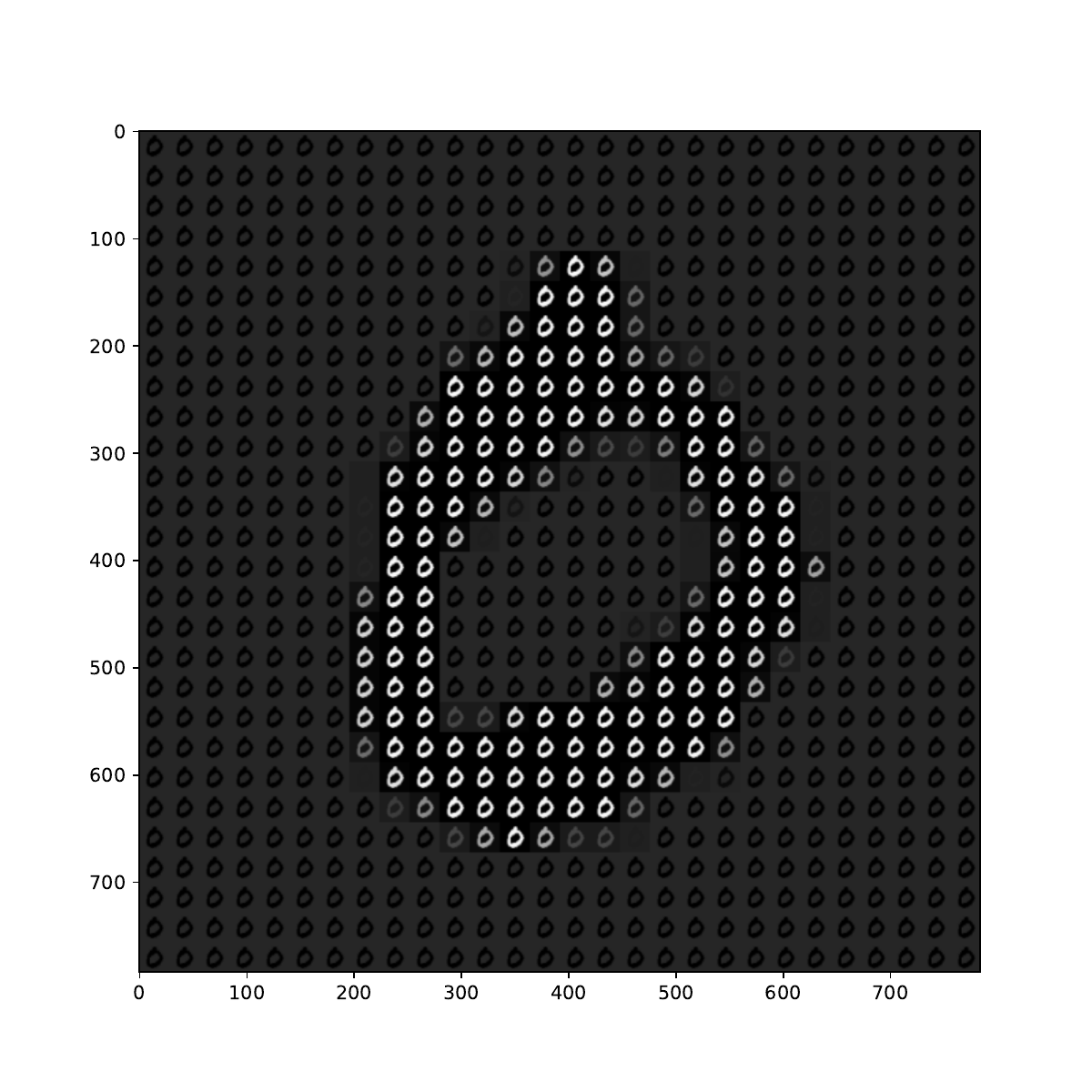}
        \caption{Taylor's Order 2 Expansion}
        \label{fig:mnist_visualization_expanded_data}
    \end{subfigure}
    \begin{subfigure}{0.48\textwidth}
        \centering
        \includegraphics[width=\textwidth]{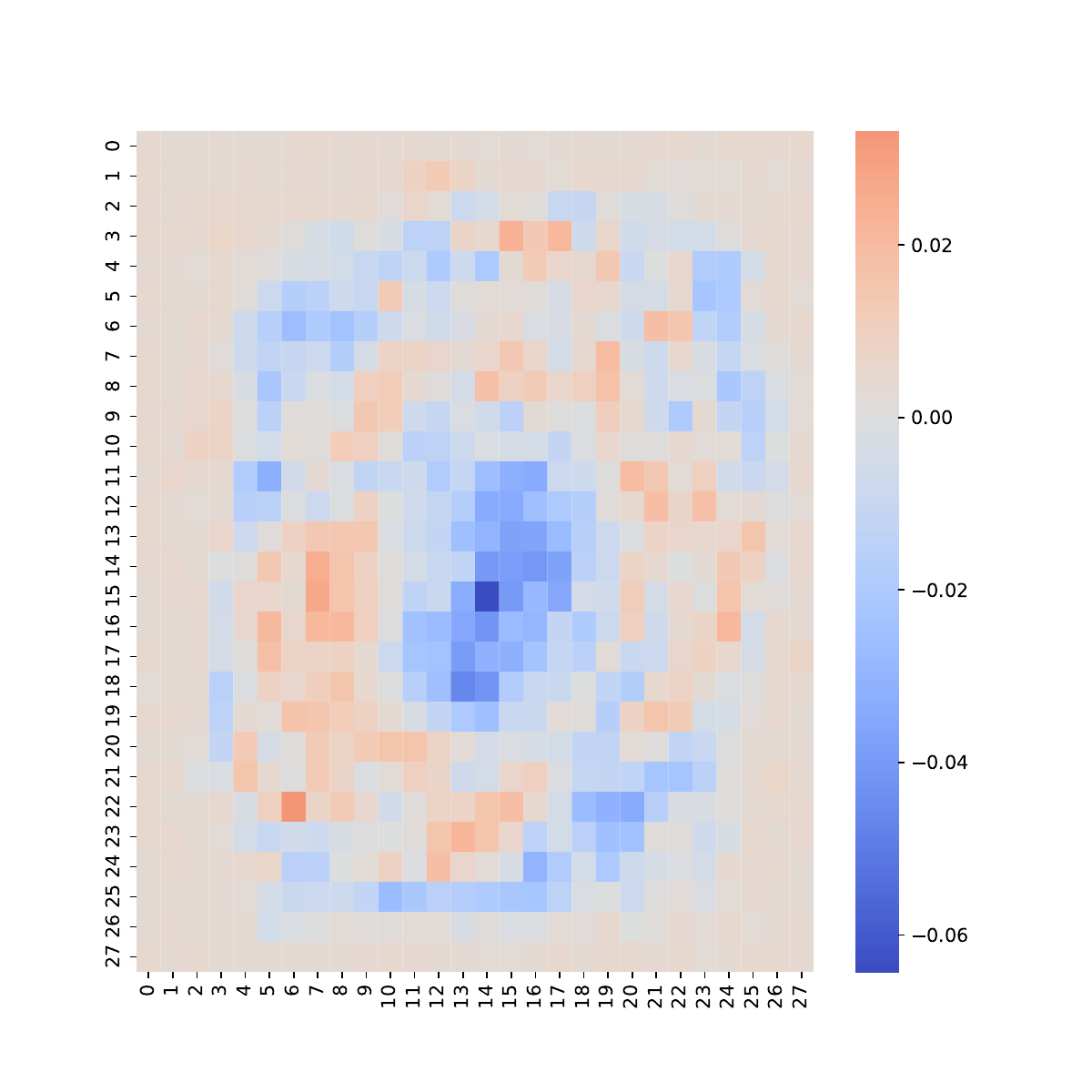}
        \caption{Parameters for Order 1 Expansion}
        \label{fig:mnist_visualization_raw_data_parameter}
    \end{subfigure}
    \hfill
    \begin{subfigure}{0.48\textwidth}
        \centering
        \includegraphics[width=\textwidth]{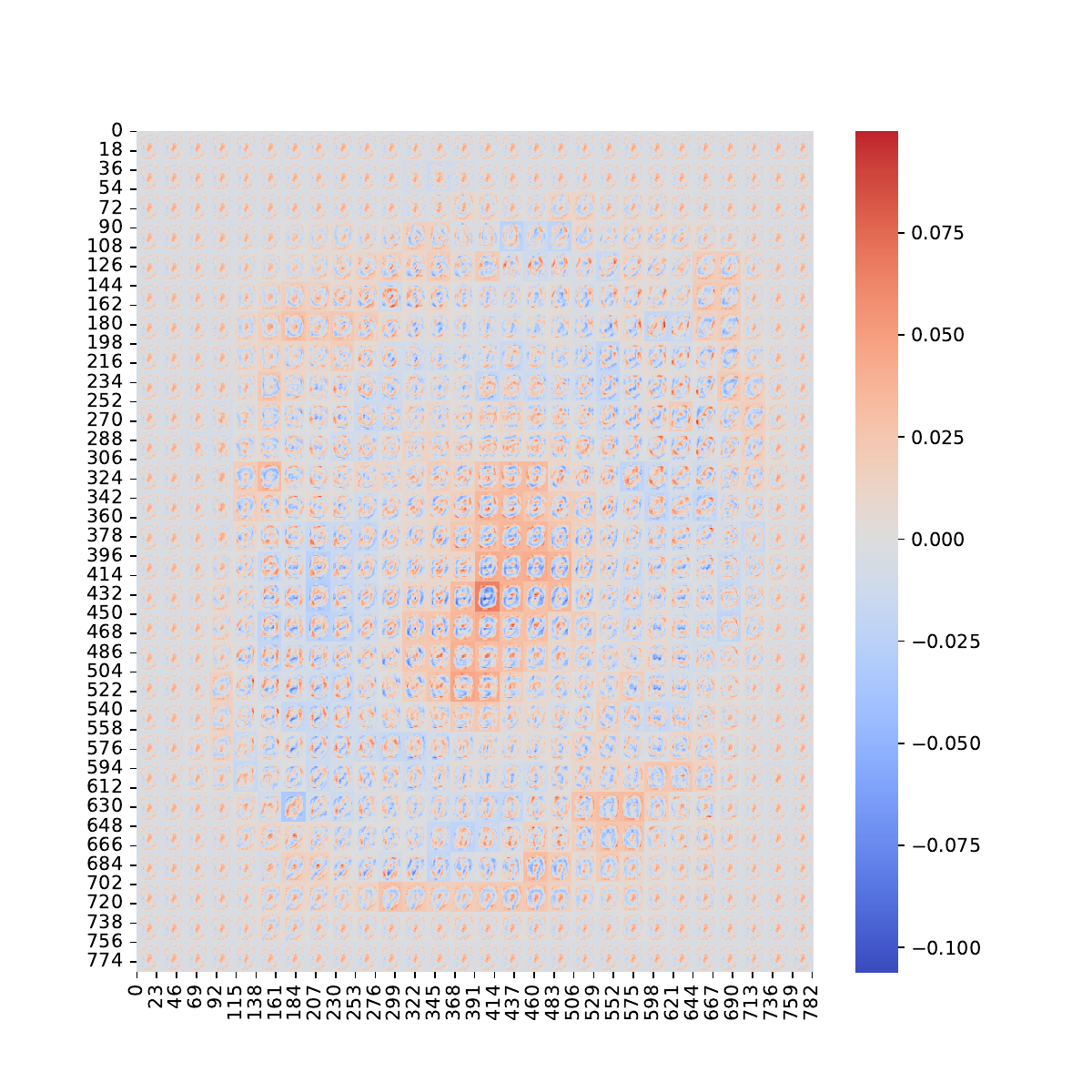}
        \caption{Parameters for Order 2 Expansion}
        \label{fig:mnist_visualization_expanded_data_parameter}
    \end{subfigure}
    \caption{An illustration of image with label $0$ from MNIST dataset. Plat (a): Taylor's order 1 expansion; Plot (b): Taylor's order 2 expansion; Plot (c): parameter corresponding to output neuron of label $0$ for order 1 expansion; and Plot (d): parameter corresponding to output of label $0$ for order 2 expansion.}
    \label{fig:mnist_expansion_visualization}
\end{figure}
%---------------------------------------------------------------

To interpret the learning process and obtained results, we illustrate the data expansion results along with the learned parameters of an image from MNIST in Figure~\ref{fig:mnist_expansion_visualization}. For this visualization, we build a 1-layered {\our} with Taylor's expansion (order 2), identity reconciliation, and zero remainder functions. From the testing set, we randomly pick one image with label $0$. Prior to expansion, {\our} flattens and normalizes the input image into a vector of length $784$, using mean-std normalization (with mean: $0.1307$ and std: $0.3081$).

The Taylor's expansion function extends the input image vector from length $784$ to a much longer vector of length $784 + 784 \times 784$, representing Taylor's order 1 and order 2 expansions as shown in Plots (a) and (b), respectively. The {\our} model correctly classifies this expanded vector. Plots (c) and (d) display the extracted parameters that project the expanded vector (including both order $1$ and $2$ expansions) to the output neuron corresponding to label $0$.

Plots (a) and (b) demonstrate that the expansion output contains multiple copies of the input image, re-scaled by each pixel value. After mean-std normalization, pixels in the images may have both positive and negative values. As illustrated in Plot (b), pixels with large positive values will increase the image grayscale values, while negative values will invert the image grayscales from light to dark and vice versa. The parameters indicate that both order $1$ and order $2$ expansions contribute effectively to generating the correct output predictions.

Typically, we expect pixels denoting object location and shapes to be more important and assigned larger weights, which aligns with the parameters observed in Plot (c). However, Plot (d) reveals an unexpected observations: for pixel values in the order 2 expansions, particularly for image copies re-scaled by pixels in the central dark regions, they are assigned much larger weights. This suggests that, after expansions, the model can capture and utilize high-order signals from seemingly less important pixels which may deliver the correct classification results.

In addition to the images with label 0 illustrated in Figure~\ref{fig:mnist_expansion_visualization}, we also randomly selected images with labels 1 through 9 and visualized their expansions and learned parameters. These visualizations are presented in Figures~\ref{fig:mnist_visualization_appendix_1}-\ref{fig:mnist_visualization_appendix_9} in Appendix Section~\ref{subsec:appendix_mnist_visualization_studies}. Consistent patterns are observed across images with different labels. Readers may refer to these visualizations in the Appendix for more comprehensive information.

%=================================================

\subsection{Probabilistic Dependency Inference}\label{subsec:dependency_inference}

According to the descriptions introduced in the previous sections, equipped with the probabilistic expansion functions, {\our} is capable to learn the probabilistic dependency relationships among the input variables (of both input features and output labels). In addition to the above continuous function approximation and discrete data classification tasks, in this part, we will investigate the effectiveness of {\our} for the probabilistic dependency relationship inference task. 

%=================================================

\subsubsection{Datasets Description and Experiment Setups}\label{subsec:experiment_setups}

%---------------------------------------------------------------
\begin{table}[h]
\centering
\caption{Statistics of classic machine learning tabular datasets. The datasets are partitioned into training and testing sets with 10-fold cross validation. The training and testing instance numbers for each fold are also reported in the table.}
\label{tab:tabular_dataset_statistics}
\begin{tabular}{|c|c|c|c|}
\hline
\multirow{2}{*}{} & \multicolumn{3}{c|}{\textbf{Classic Machine Learning Tabular Datasets}} \\ \cline{2-4}
                          & \textbf{Iris Species} & \textbf{Pima Indians Diabetes} & \textbf{Banknote} \\ \hline
Instance \# 	& 150 	& 768 	&1,372 	\\ \hline              %& 210 	
Train \#		&135		& 691	& 1,234	\\ \hline		%&189 	
Test \# 		&15		& 77		& 137	\\ \hline        %&21 		
Feature \# 	&4		& 8		& 4		\\ \hline		%&7 		
Class \# 		&3		& 2		& 2		\\ \hline		%&3 		
\end{tabular}
\end{table}
%---------------------------------------------------------------

In this subsection, we will investigate the effectiveness of {\our} in inferring probabilistic dependency relationships using three canonical tabular datasets. The basic statistical information of these datasets are summarized in Table~\ref{tab:tabular_dataset_statistics}. The Iris Species is a multi-class dataset, while the other two involve binary class instead. For all three datasets, we employ 10-fold cross-validation, allocating 90\% of the instances to the training set and the remaining 10\% to the testing set. To assess the classification performance of all comparative methods, we utilize Accuracy as our quantitative evaluation metric.

In the following sections, we will first present the quantitative results of {\our} and other benchmark methods in classifying these tabular datasets. Subsequently, we will elucidate the learned probabilistic dependency relationships among feature and label variables, as inferred by these comparative methods, focusing on the Iris Species dataset as an illustrative example.

%=================================================

\subsubsection{The Main Results of {\our} on Classic Tabular Data Classification}

%---------------------------------------------------------------
\begin{table}[h]
\centering
\caption{Classification results of classic tabular data classification with probabilistic models. For each method, we run them with 10 iterations based on the training and testing set partitioned via the 10-fold cross validation. The average and standard deviation of the obtained accuracy scores are reported in the table.}
\label{tab:tabular_dataset_results}
\begin{tabular}{|c|c|c|c|}
\hline
\multirow{2}{*}{} & \multicolumn{3}{c|}{\textbf{Classic Machine Learning Tabular Datasets}} \\ \cline{2-4} 
                          & \textbf{Iris Species} & \textbf{Pima Indians Diabetes} & \textbf{Banknote} \\
	\hline
	\hline
	% wheat &\makecell{$8.28 \times 10^{-1}$\\ $\pm$ $9.33 \times 10^{-2}$}  
        \makecell{Naive Bayes\\ (Multinomial)} &\makecell{$9.00 \times 10^{-1}$\\ $\pm$ $1.12 \times 10^{-1}$}  &\makecell{$6.51 \times 10^{-1}$\\ $\pm$ $4.17 \times 10^{-2}$}  &\makecell{$6.36 \times 10^{-1}$\\ $\pm$ $4.61 \times 10^{-2}$}  \\
        \hline
        % &\makecell{$9.00 \times 10^{-1}$\\ $\pm$ $ 4.97 \times 10^{-2}$}  
        \makecell{Naive Bayes\\ (Gaussian)} &\makecell{$9.53 \times 10^{-1}$\\ $\pm$ $ 4.26 \times 10^{-2}$}  &\makecell{$7.56 \times 10^{-1}$\\ $\pm$ $ 4.22 \times 10^{-2}$}  &\makecell{$8.39 \times 10^{-1}$\\ $\pm$ $ 4.72 \times 10^{-2}$}  \\
        \hline
%        \hline
        %\makecell{SVM\\ (Linear)} &\makecell{$9.60 \times 10^{-1}$\\ $\pm$ $ 5.30 \times 10^{-2}$}  &\makecell{$9.38 \times 10^{-1}$\\ $\pm$ $ 4.30  \times 10^{-2}$}  &\makecell{$7.67 \times 10^{-1}$\\ $\pm$ $ 4.30  \times 10^{-2}$}  &\makecell{$9.80 \times 10^{-1}$\\ $\pm$ $ 9.00  \times 10^{-3}$}  \\
%        \hline
        %\makecell{SVM\\ (RBF)} &\makecell{$9.60 \times 10^{-1}$\\ $\pm$ $ 5.30  \times 10^{-2}$}  &\makecell{$9.29 \times 10^{-1}$\\ $\pm$ $ 4.40  \times 10^{-2}$}  &\makecell{$7.68 \times 10^{-1}$\\ $\pm$ $ 4.10  \times 10^{-2}$}  &\makecell{$1.00 \times 10^{0}$\\ $\pm$ $ 0.00  \times 10^{0}$}  \\
        \hline
        %&\makecell{$8.69 \times 10^{-1}$\\ $\pm$ $ 5.20 \times 10^{-2}$} 
        \makecell{Bayesian\\ Network} &\makecell{$9.53 \times 10^{-1}$\\ $\pm$ $ 7.91 \times 10^{-2}$}  &\makecell{$6.64 \times 10^{-1}$\\ $\pm$ $ 5.47 \times 10^{-2}$} &\makecell{$9.24 \times 10^{-1}$\\ $\pm$ $ 2.42 \times 10^{-2}$}  \\
        \hline
        \hline
        %&\makecell{$4.19 \times 10^{-1}$\\ $\pm$ $ 1.08 \times 10^{-1}$} 
        \makecell{Markov\\ Network} &\makecell{$9.20 \times 10^{-1}$\\ $\pm$ $ 7.18 \times 10^{-2}$}  &\makecell{$7.09 \times 10^{-1}$\\ $\pm$ $ 4.17 \times 10^{-2}$} &\makecell{$9.02 \times 10^{-1}$\\ $\pm$ $ 4.07 \times 10^{-2}$}  \\
        \hline
        \hline
        \makecell{{\our}\\ (Naive Prob. Exp.)} & \makecell{\boldmath $9.73 \times 10^{-1}$\\ \boldmath $\pm$ $ 3.26 \times 10^{-2}$}  & \makecell{$7.74 \times 10^{-1}$\\ $\pm$ $ 3.91 \times 10^{-2}$} & \makecell{$9.77 \times 10^{-1}$\\ $\pm$ $ 2.27 \times 10^{-2}$} \\
        \hline
        \makecell{{\our}\\ (Comb. Prob. Exp)} & \makecell{$9.67 \times 10^{-1}$\\ $\pm$ $ 4.47 \times 10^{-2}$}  & \makecell{\boldmath $7.80 \times 10^{-1}$\\ \boldmath $\pm$ $ 3.36 \times 10^{-2}$} & \makecell{\boldmath $9.79 \times 10^{-1}$\\ \boldmath $\pm$ $ 2.18 \times 10^{-2}$} \\
        \hline
\end{tabular}
\end{table}
%---------------------------------------------------------------

Table~\ref{tab:tabular_dataset_results} presents a comparative analysis of {\our} against several established probabilistic methods, including naive Bayes (utilizing both Multinomial and Gaussian distributions), Bayesian networks, and Markov networks. We evaluate {\our} using two distinct architectures: (1) {\our} with a naive Laplace probabilistic expansion function, identity reconciliation function, and linear remainder function; and (2) {\our} with a combinatorial Gaussian probabilistic expansion function, identity reconciliation function, and linear remainder function. The dataset is partitioned using the 10-fold cross-validation, and the Accuracy scores for each method across these 10 iterations are reported as ``mean $\pm$ std'' in the table.

The results indicate that {\our}, employing both naive and combinatorial probabilistic expansion functions, consistently outperforms the other methods. Specifically, {\our} with naive Laplace expansion and identity reconciliation functions demonstrates superior parameter learning and performance compared to naive Bayes methods. Furthermore, {\our} with combinatorial Gaussian expansion and identity reconciliation functions exhibits enhanced capability in learning dependency relationships among feature and label variables, consistently surpassing the performance of Bayesian and Markov networks. These findings substantiate the efficacy of {\our} in inferring the probabilistic dependency relationships of the feature variables and classifying the data instances.

%=================================================

\subsubsection{Probabilistic Dependency Relationship Visualization}

%---------------------------------------------------------------
\begin{figure}[h!]
    \centering
    \begin{subfigure}{0.9\textwidth}
        \centering
        \includegraphics[width=\textwidth]{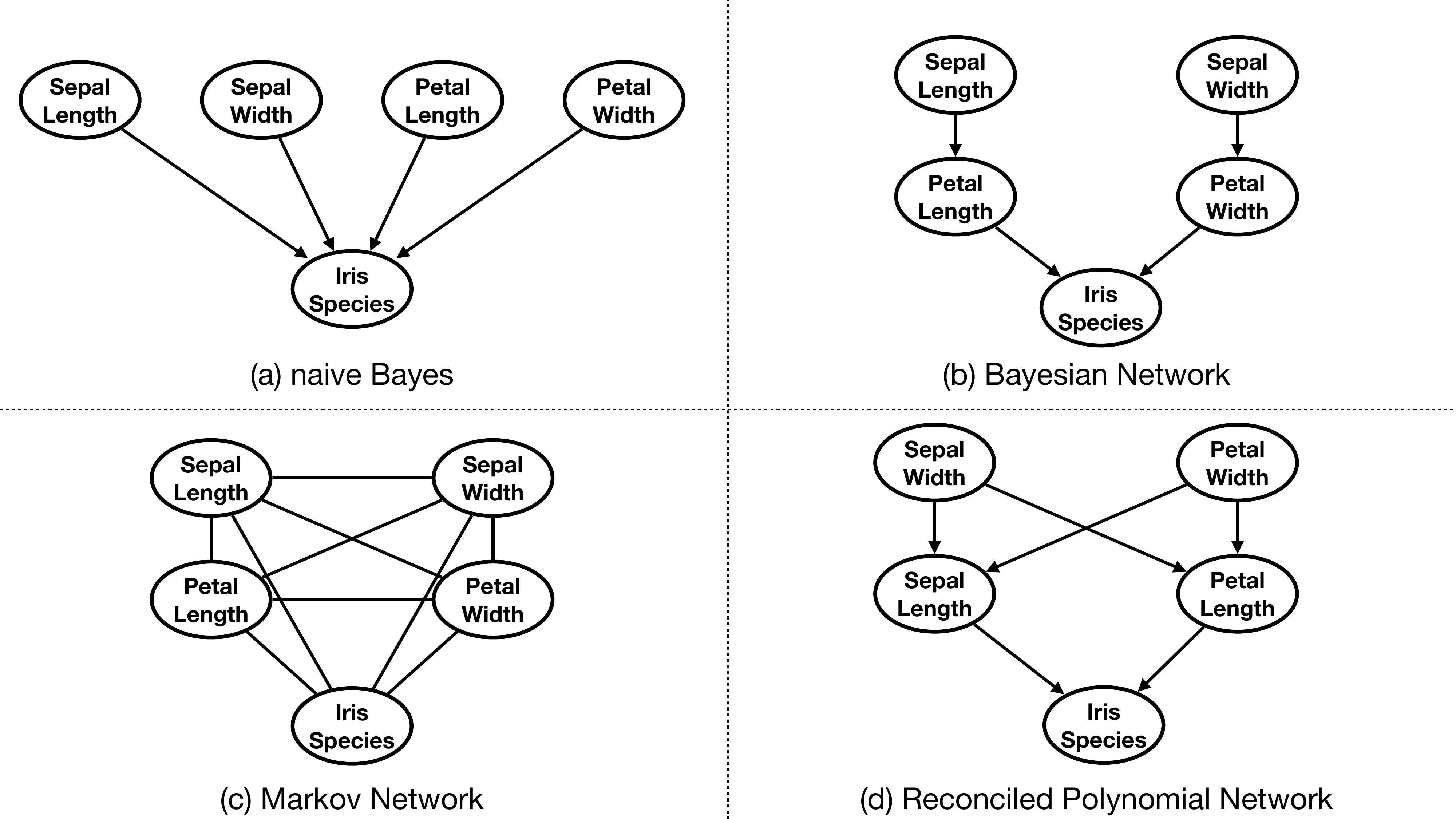}
    \end{subfigure}
        \caption{An illustration of dependency relationships infered by different methods on the Iris Species dataset. In the plots, Sepal Length, Sepal Width, Petal Length and Petal Width denote the feature variables, and Iris Species denotes the label variable of the Iris dataset.}
    \label{fig:dependency_relationship_inference}
\end{figure}
%---------------------------------------------------------------

In addition to the quantitative effectiveness evaluations, we also visualize the relationships learned by the these different probabilistic methods on the Iris Species dataset, as illustrated in Figure~\ref{fig:dependency_relationship_inference}. For the Bayesian network, we employ Hill Climb Search to learn the network structure from the training data, utilizing the Bayesian Information Criterion as the scoring function. For the Markov network, we need to manually define the variable clique and the factorization approach. In this experiment, all feature variables in the dataset are designated as the clique, with the factorization function based on data instance appearance counts. To facilitate this approach, float features are pre-partitioned and categorized into discrete bins. For {\our}, we extract the elements with the $k$ largest positive coefficients and $k$ smallest negative coefficients to define the dependency graphs.

As shown in Plots (a) and (c) of Figure~\ref{fig:dependency_relationship_inference}, the manually crafted variable relationships indicate that ``Iris Species'' (the label) is determined by all feature variables (``Sepal Length'', ``Sepal Width'', ``Petal Length'', and ``Petal Width'') for both naive Bayes and Markov network models. The Markov network additionally considers relationships among the feature variables. In contrast, Bayesian network and {\our} method, based on learned results from training data, extract different dependency relationships. In the Bayesian network, the ``Iris Species'' label is determined by ``Petal Length $|$ Sepal Length'' and ``Petal Width $|$ Sepal Width''. For {\our} method, the label is determined by ``Petal Length $|$ (Petal Width, Sepal Width)'' and ``Sepal Length $|$ (Petal Width, Sepal Width)''. We use the notation ``A $|$ B'' to represent the dependency of variable A on B, borrowing from conditional probability notation. The probabilistic dependency relationships learned by {\our} method align more closely with the feature variable correlations for different label instances in the Iris dataset, as illustrated in Figure~\ref{fig:iris_correlation}.

The visualized plots in Figure~\ref{fig:dependency_relationship_inference} reveal that {\our} can learn more complex probabilistic dependency relationships among variables, effectively elucidating its learning process and classification results. However, we also observe certain limitations in the current {\our} model. The static nature of the current probabilistic expansion functions, which rely on manually defined distributions and pre-provided distribution hyper-parameters, may present challenges when applied to more intricate probabilistic dependency relationships. Addressing this limitation will be a key focus in future work of {\our}.

%---------------------------------------------------------------
\begin{figure}[t]
    \centering
    \begin{subfigure}{0.9\textwidth}
        \centering
        \includegraphics[width=\textwidth]{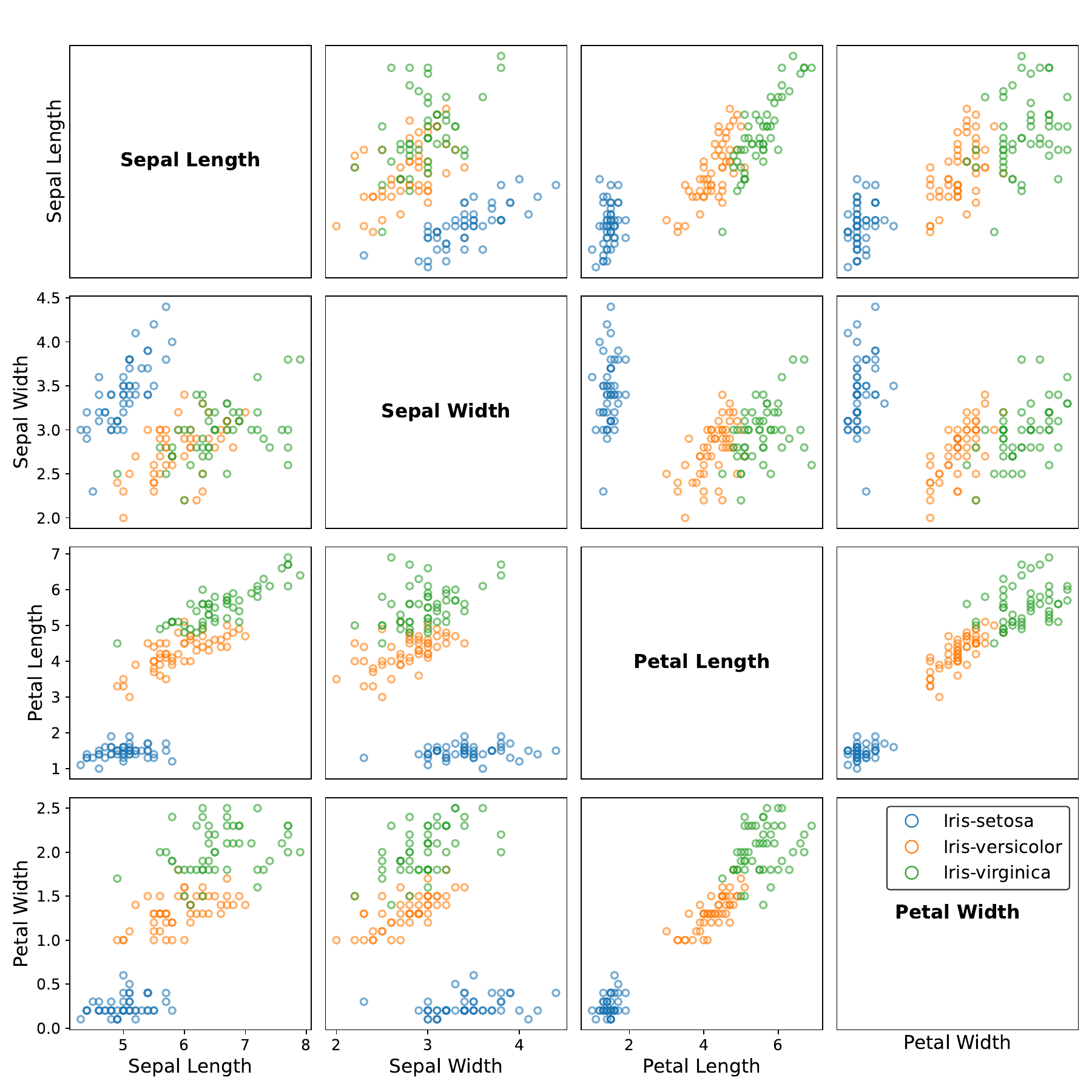}
    \end{subfigure}
        \caption{An illustration about the correlations of the pairwise features in the Iris dataset. The scatter dots in different colors denote the data instances of different classes.}
    \label{fig:iris_correlation}
\end{figure}
%---------------------------------------------------------------

%--------------------------------------------------------------------------
%\input{sec_models.tex}
%--------------------------------------------------------------------------
%\input{sec_learning.tex}
%--------------------------------------------------------------------------
%=================================================
%=================================================

\section{Interpretations of the {\our} Model Design}\label{sec:interpretation}

In addition to the above empirical evaluations, in this section, we will discuss the interpretations of our model design from various perspectives, encompassing technical machine learning, and biological neuroscience. Through these discussions, we aim to illustrate the motivations and advantages inherent in {\our} model design.

% and compare against other existing base models. 

%=================================================

\subsection{Theoretic Machine Learning Interpretations}

As introduced in the previous section, the {\our} model is composed of the {data expansion function}, {parameter reconciliation function} and the {remainder function}. Each of these components functions plays a crucial role in shaping the {\our} model, striking a balance among {model capacity}, {learning feasibility} and {performance robustness}.

%---------------------------------------------------

\subsubsection{Understanding {\our} from VC-Dimension}

According to the Vapnik-Chervonenkis theory \cite{doi:10.1137/1116025, 10.1145/76359.76371}, to evaluate the quality of the {\our} model $g(\mb{x} | \mb{w})$, we compute the introduced errors by comparing $g(\mb{x} | \mb{w})$ against the underlying function $f(\mb{x})$ with the input data space $\mc{D} \subset \mathbbm{R}^m$. This space encompasses all data instances, including those seen in the training set $\mc{T}$ and unseen instances in $\mc{D} \setminus \mc{T}$:
\begin{equation}
{\scriptsize 
\underbrace{\int_{\mb{x} \in \mc{D}} p(\mb{x}) \left\| g(\mb{x} | \mb{w}) - f(\mb{x}) \right\| \mathrm{d} \mb{x} }_{\text{overall error } \mc{L}}
=
\underbrace{\int_{\mb{x} \in \mc{T}} p(\mb{x}) \left\| g(\mb{x} | \mb{w}) - f(\mb{x}) \right\| \mathrm{d} \mb{x} }_{\text{empirical error } \mc{L}_{em}}
+
\underbrace{\int_{\mb{x} \in \mc{D} \setminus \mc{T}} p(\mb{x}) \left\| g(\mb{x} | \mb{w}) - f(\mb{x}) \right\| \mathrm{d} \mb{x} }_{\text{expected error } \mc{L}_{exp}}, }
\end{equation}
where $p(\mb{x})$ denotes the probability density function for drawing instance from the input data space $\mc{D}$ and $\left\| \cdot \right\|$ denotes a norm measuring the difference between the outputs of $f(\mb{x})$ and $g(\mb{x} | \mb{w})$. As indicated by the equation, the \textbf{overall error} $\mc{L}$ measuring the difference between the model $g(\mb{x} | \mb{w})$ and the underlying unknown mapping $f(\mb{x})$ consists of two parts: (1) the \textbf{empirical error} $\mc{L}_{em}$ calculated on the training set $\mc{T}$, and (2) the \textbf{expected error} $\mc{L}_{exp}$ calculated on unseen data instances from $\mc{D} \setminus \mc{T}$, both of which are closely related to the VC-dimension of the model.

%------------------------------------
%https://www.svms.org/srm/
\begin{figure*}[t]
    \begin{minipage}{\textwidth}
    \centering
    	\includegraphics[width=0.9\linewidth]{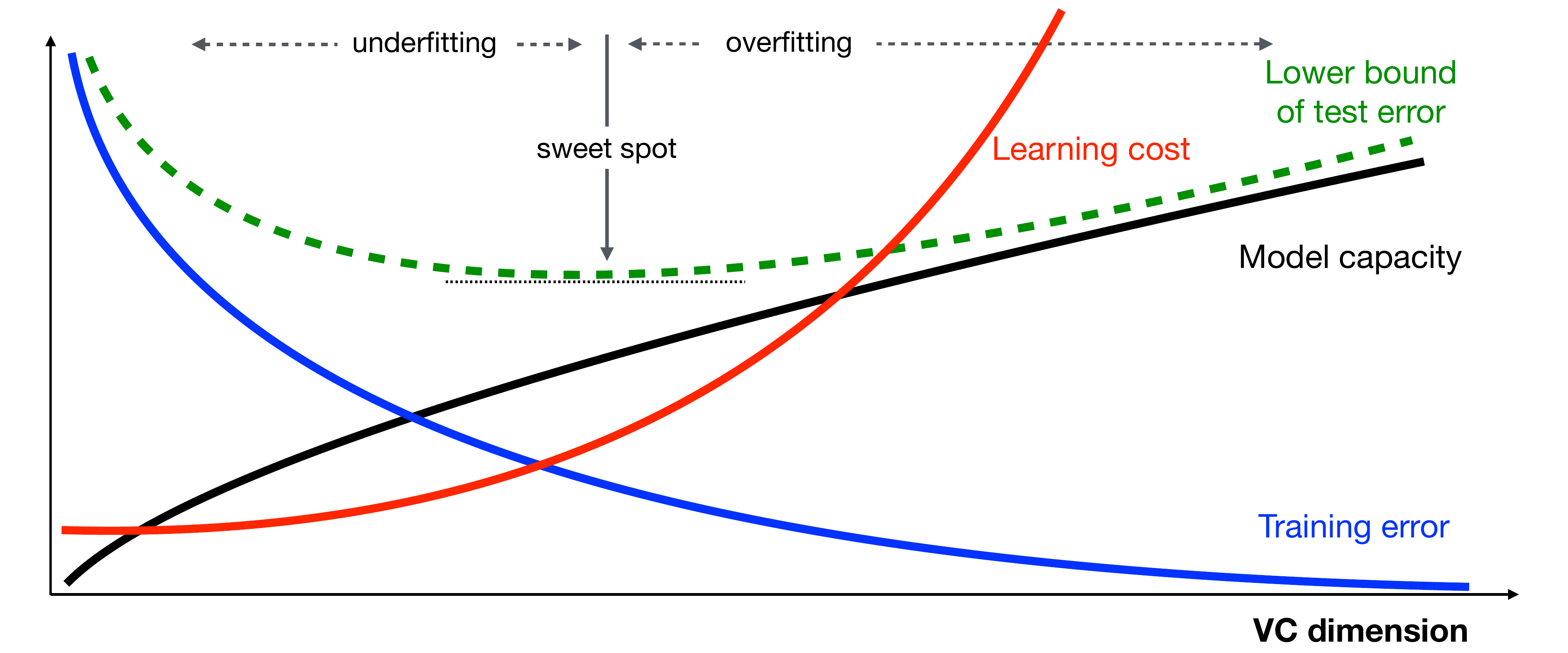}
    	\caption{An illustration of VC dimension and model learning: the x-axis represents the VC-dimension, where the blue curve denotes the model training errors, the black curve represents the model capacity, and the green dashed curve signifies the lower bound of test error. The red curve, not previously present in VC theory, represents the added aspect of model learning and operating costs ({\eg} required computational hardwares, time and space costs, and energy consumptions).}
    	\label{fig:vc_dimension}
    \end{minipage}%
\end{figure*}
%------------------------------------

As illustrated by the curves in Figure~\ref{fig:vc_dimension}, as the VC-dimension of a model increases, the model will have a greater capacity (and complexity) and the training error will gradually decrease. Meanwhile, for complex models in higher VC-dimensions, the testing errors of the model on unseen data will first decrease, reaching the harmonious ``sweet spot'', and then increase in a U-shape. In addition to the curves we borrowed from the VC-theory, we also add a red curve denoting the learning cost. As the VC-dimension increases and the model becomes more complex, without necessary regulations on the parameters, the learning cost of the model (in terms of time, space, data consumptions in the training phase, as well as subsequent manual tuning for alignment and safety) will increase dramatically.

For {\our} model, the VC-dimension is determined by both the data expansion function and parameter reconciliation function, and potentially the remainder function if it is non-zero. All these component functions collectively influence the model's complexity and learning capacity. By projecting data instances through the expansion function into higher-dimensional space, we augment the model's representational complexity. However, instead of directly training the model in this high-dimensional space, the parameter reconciliation function acts as a regulator of model complexity, enabling effective operation on data instances reduced from high-dimensional spaces while mitigating learning costs and guiding testing errors toward an optimal range. Of course, in real practice, the perfect balance between the expansion and reconciliation functions is hard to achieve, which may unnecessarily lead to learning errors and performance degradation, but the remainder function can compensate for potential deficiencies and improve the robustness of {\our}.

%=================================================

\subsubsection{Understanding {\our} from Vector Space Projection}

%\textcolor{red}{Shall we replace this example with concrete experiments on the half-moon and double circle dataset?}

The layers in {\our} model form a sequence of data projections across spaces. Initially, data is expanded from the input space to an intermediate higher-dimensional space via the expansion function, and subsequently, it is projected back to a lower-dimensional output space through the inner product with the reconciled parameters. This process enables the {\our} model to effectively handle data instances that pose classification challenges in the original input space. By undergoing these projections, separation of the data instances in the output space can be much easier compared to the original input space.

As illustrated in Figure~\ref{fig:vector_space}, we present a simplified example comprising data instances from two classes: blue circles represent the negative class, while red squares represent the positive class. The data expansion function projects these instances from the two-dimensional input space to an intermediate space, such as a three-dimensional space. Here, positive and negative instances are projected to the top and bottom regions of the newly created $x_3$ dimension, respectively. Rather than performing classification directly in this intermediate space, which would incur unnecessary high learning costs, our {\our} model further projects the data instances back to the output space by excluding the $x_2$ dimension. This is accomplished through an inner product with the reconciled parameters. Separation of such instances in the output space is considerably easier than in the original input space.

%------------------------------------
%https://www.svms.org/srm/
\begin{figure*}[t]
    \begin{minipage}{\textwidth}
    \centering
    	\includegraphics[width=0.9\linewidth]{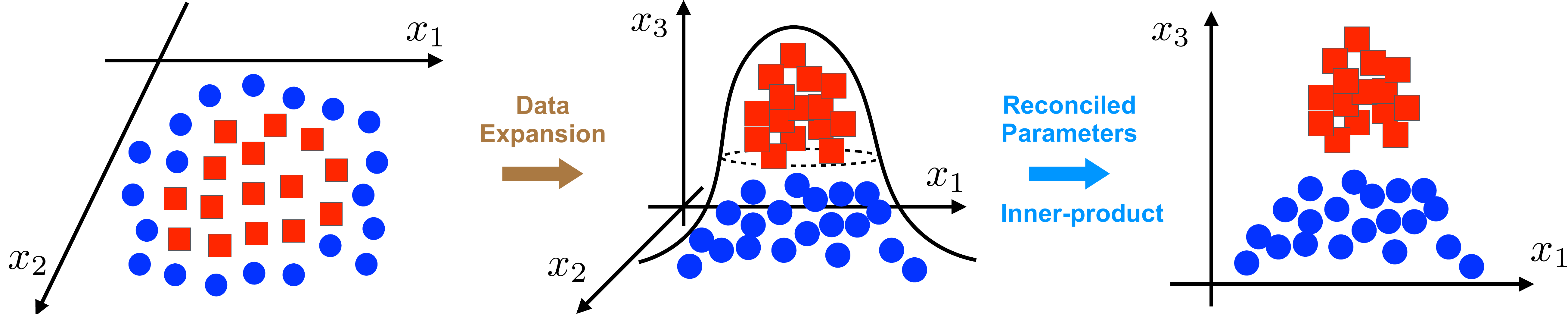}
    	\caption{An Interpretations of {\our} from vector space projection perspective.}
    	\label{fig:vector_space}
    \end{minipage}%
\end{figure*}
%------------------------------------

Compared to MLP, the expansion function provides {\our} with greater learning capacity to handle more complex tasks. Meanwhile, compared to KAN, the parameter reconciliation function allows {\our} to fit the dataset with fewer parameters, thus making it more robust to overfitting and feasible for handling complex learning tasks. {\our} and kernel SVM both operate within the expanded space. However, unlike kernel SVM, which directly trains the model with data instances in the expanded space, the reduction back to the output space for learning helps address the ``curse-of-dimensionality'' and high learning cost issues.

%=================================================

\subsection{Biological Neuroscience Interpretations}

Compared to the other existing base models mentioned in this paper, the {\our} model introduced in this paper provides a closer approximation to current biological nervous systems. The expansion, reconciliation and remainder functions used in the design of the {\our} model correspond to the functioning mechanisms of biological neurons.

%---------------------------------------------------

\subsubsection{Biological Neuron Structure and Membrane Potential}

Unlike other existing neural models (including MLP and KAN, as well as their derivatives), {\our} provides a more accurate modeling of the real-world biological neurons from the neuroscience perspective. Most readers interested in this paper may possess the basic background knowledge about neuroscience that supports the design of artificial neurons and neural network models. However, to help readers understand the {\our} model better, we will briefly describe the structure and working mechanisms of biological neurons in this section.

\begin{wrapfigure}{r}{0.5\textwidth}
\vspace{-20pt}
  \begin{center}
    \includegraphics[width=0.5\textwidth]{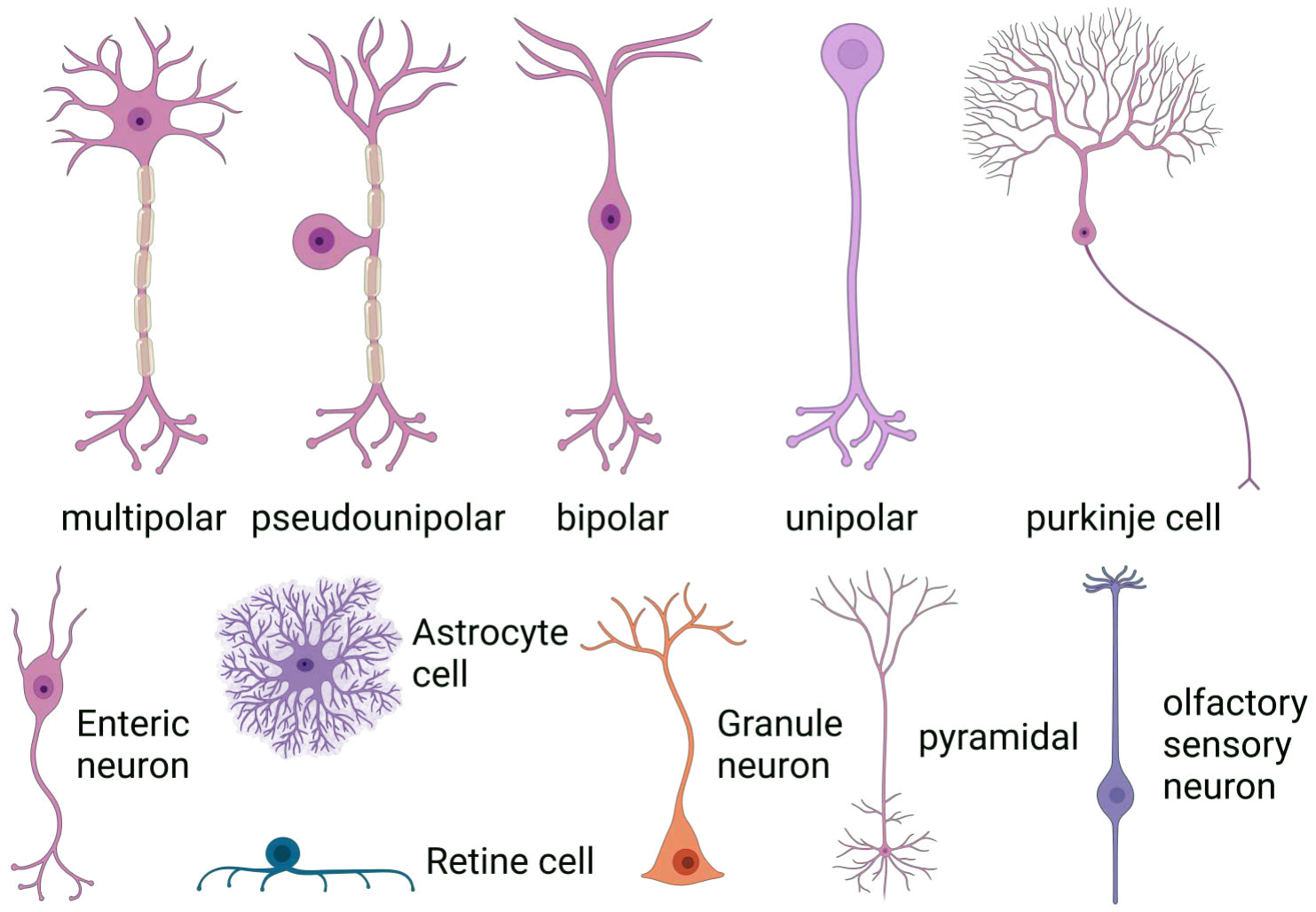}
  \end{center}
  \vspace{-20pt}
  \caption{Different neuron structures.}\label{fig:neurons}
\end{wrapfigure}

As illustrated in Figure~\ref{fig:neurons}, the biological nervous systems for animal creatures ({\eg} mammals, birds, fishes, insects and even worms) consists of a large number of inter-connected biological neurons, which are highly specialized for the processing and transmitting cellular signals. These neurons' cell body includes several important parts as shown in Figure~\ref{fig:action_potential} (a), including the \textit{soma}, \textit{nucleus}, \textit{dendrite}, \textit{axon}, \textit{myelin sheath}, \textit{Schwann cell}, \textit{node of Ranvier} and \textit{axon terminal}. The neuron's axon can branch out and connect to a large number of downstream neurons' dendrites at sites called \textit{synapses}, whose zoomed-in structure is illustrated in Figure~\ref{fig:action_potential} (c). The neuron structure we show in the plot is called the ``multiploar neuron'', which possesses a single axon and many dendrites. Besides this, as illustrated in Figure~\ref{fig:neurons}, there also exist many other neuron structures, such as ``bipolar neuron'', ``pseudo-unipolar neuron'' and ``unipolar neurons'', which will form neurons with different functions. Also neurons in real nervous systems are not arranged layer by layer in an orderly manner; their connections can be somewhat ``chaotic'', with many ``skip-layer'' connections for faster signal transmission.

These structures of biological neurons all contribute to the complex functioning of the creatures' nervous systems. Like all other animals' cells, the neuron cells are also enclosed in a plasma membrane, which has the structure of a lipid bilayer with many types of large protein molecules ({\eg} the ion channels and ion pumps) embedded in it as shown in Plot (b). Biological neurons are just like a ``salty banana'', with more sodium ions (purple circles in Plot (b)) outside the cell and more potassium ions (green triangles in Plot (b)) inside. The membrane serves as both an insulator and a diffusion barrier to the movement of ions. Ion pump proteins break down ATP for energy to actively push ions across the membrane and establish concentration gradients across the membrane; while the ion channels allow ions to move across the membrane down those concentration gradients. Ion pumps and ion channels are electrically equivalent to a set of batteries and resistors inserted in the membrane, creating a voltage between the two sides of the membrane, which is called the \textit{membrane potential}. As illustrated in Plot (e), at the resting state, a neuron's membrane potential is sitting in millivolts, ranging from $-40$mV to $-80$mV.

%------------------------------------
%https://www.svms.org/srm/
\begin{figure*}[t]
    \begin{minipage}{\textwidth}
    \centering
    	\includegraphics[width=1.0\linewidth]{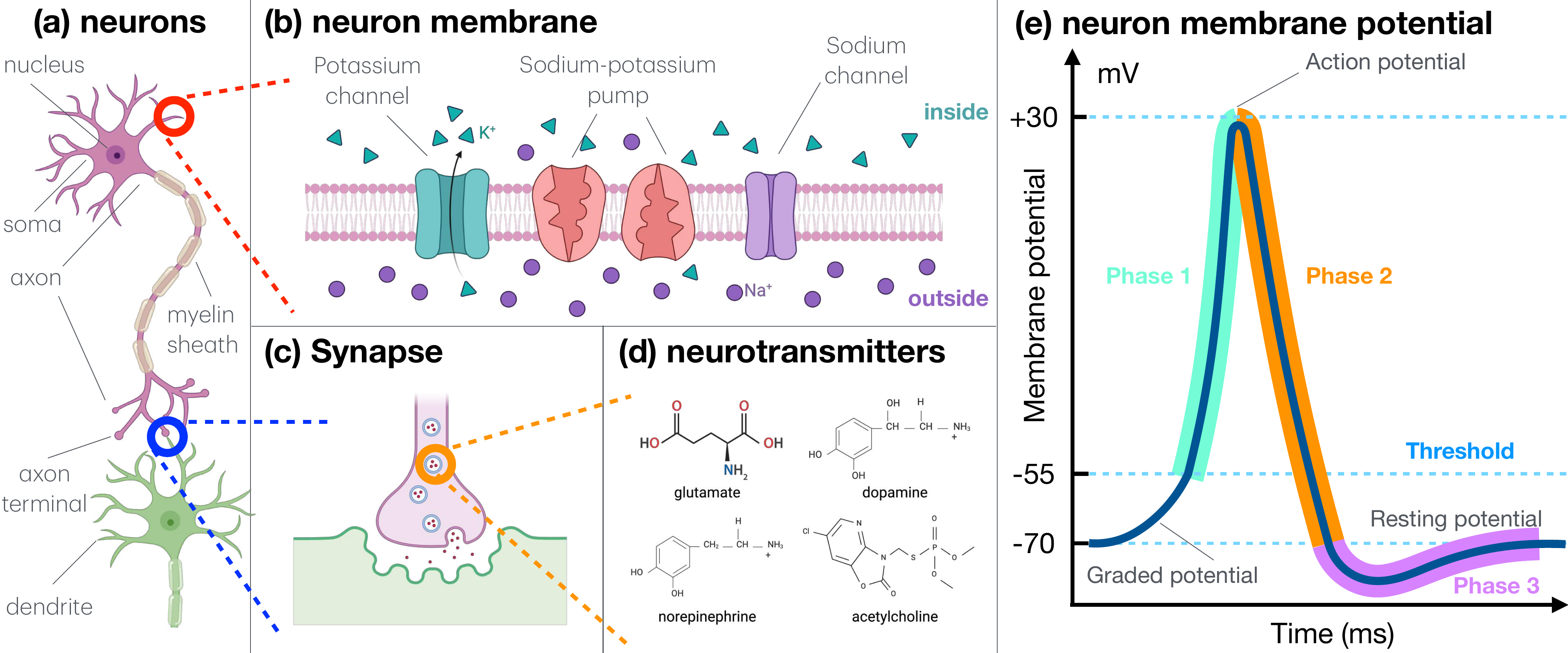}
    	\caption{An illustration of connected neuron cells and the neuron cell membrane potential. For the neuron cell, we also provide a zoomed-in view to depict both the neuron membrane with its various channels allowing the ion propagation and the synapse connections releasing neurotransmitters that enable neurotransmissions.}
    	\label{fig:action_potential}
    \end{minipage}%
\end{figure*}
%------------------------------------

The ion channels embedded in the neuron membrane are integral membrane proteins with a pore, which can be either open or closed for ion passage. Ion channels can be classified by how they respond to their environment: (a) ligand-gated channels open once receiving some certain types of chemical neurotransmitters, (b) voltage-dependent channels change their permeability according to the membrane potential, and (c) mechanically gated channels open due to physical distortion of the cell membrane ({\eg} sense of touch). When a channel is open, ions permeate through the channel pore down the transmembrane concentration gradient for that particular ion, such as sodium moves in and potassium moves out, causing the neuron membrane to change.

We all know that the ``neurons receiving inputs greater than their inherent thresholds will be activated to transmit signals to surround neurons connected to them'', but the details of such a process can be more complicated than readers may imagine. As illustrated in Plots (c) and (d), through the synapses, the neuron's axon terminal will releases synaptic vesicles, each containing numerous chemical neurotransmitters that act on different connected dendrite branches to open ligand-gated channels. There exist various categories of neurotransmitters, {\eg} \textit{amino acids}, \textit{gasotransmitters}, \textit{monoamines}, \textit{trace amines}, \textit{peptides}, \textit{purines}, \textit{catecholamines} and many others. Each category of neurotransmitter delivers different types of cellular signals to other connected neurons. If a neuron's dendrite receives sufficient chemical neurotransmitters from the connected axon terminals via the synapses, it will gradually open more ligand-gated channels, causing the nearby membrane potential to change.
 
Furthermore, once the membrane potential near the dendrites exceeds a certain threshold, {\eg} $-55$mV, the voltage-dependent sodium channels in the membrane will be abruptly triggered to open, allowing a large influx of sodium ions that rapidly increases the membrane potential near the dendrites from $-55$mV to $+30$mV, {\ie} phase 1 as shown in Plot (e). This triggers more sequential channels to open, producing a greater electric current  across the cell membrane from dendrites to the axon, thereby activating the neuron to transmit signals to other neurons. As the membrane potential reaches its peak, voltage-sensitive potassium channels will open, which will dramatically increase the membrane’s potassium permeability and drive the membrane voltage back down to $-70$mV, {\ie} phase 2 in Plot (e). Afterwards, these voltage-dependent sodium/potassium channels shut down, and ion pumps will take the control to gradually restore neurons to their resting potential, as shown in phase 3 of Plot (e).

%---------------------------------------------------

\subsubsection{Understanding {\our} from Biological Neuroscience}

We provide the detailed descriptions of biological neurons and their working mechanisms above to highlight the following key points to our readers:

\begin{itemize}

\item \textbf{Membrane Potentials of Neuron}: Biological neurons' membrane potential defines the neuron cell state, which is much more complex than the simple 0/1 (active/inactive) states previously understood. Neurons maintain their membrane potential states through a sophisticated biological mechanism (see Plot (b) in Figure~\ref{fig:action_potential}). Neurons' membrane potentials typically form a periodic spiking curve (see Plot (e) in Figure~\ref{fig:action_potential}).

\item \textbf{Synapses of Neuron}: Biological neuron synapses transmit signals through physical media, with axon terminals releasing neurotransmitters of various categories (see Plots (c) and (d) in Figure~\ref{fig:action_potential}), which act only on specific ion gates located on the dendrite branches. The combination of neurotransmitters and ligand-gated ion channels determines the effectiveness of signal transmission between neurons.

\item \textbf{Connections of Neuron}: Biological neurons' structure are vey complex and diverse, and each of them may have specific functions depending on their positions in the nervous system (see Figure~\ref{fig:neurons}). Biological neuron connections are extensive and often appear ``chaotic''. Besides being arranged in an orderly layer-by-layer manner, ``skip-layer'' neuron connections are common in nervous systems for faster and robust signal transmission.

\end{itemize}

The working mechanisms of biological neurons summarized above actually inspire the design of the {\our} model in its current representation form. The data expansion function models the neuron's cell states by projecting them into a high-dimensional space rather than the original space. This approach allows us to model the complex state, including the spiking curve of neurons, moving beyond the simple binary 0/1 state or an integer in the range $[0, 1]$. The parameter expansion function adapts the parameters to accommodate the high-dimensional states of neurons, facilitating signal transmission to connected neurons. This models the fabrication of neurotransmitter-ligand-gated ion channels in forming synapses for signal transmission. Furthermore, the remainder function allows skip-layer data projection, mirroring the extensive connections of biological neurons for faster signal transmission.

%\textcolor{red}{Shall we add some experimental testing of {\our} to fit the spiking curve?}

%%---------------------------------------------------
%
%\subsubsection{Biological Neuron Modeling with {\our}}
%
%In this part, we will try to build the {\our} to model the spiking curve shown in Figure~\ref{fig:action_potential} (e) during activation via their high-dimensional representations.
%

%=================================================

%\subsection{Electronic Circuits Signal Processing Interpretations}

%https://www.electricaltechnology.org/2013/12/the-main-difference-between-linear-and-nonlinear-circuits.html

%https://electronics.stackexchange.com/questions/94849/electrical-circuit-with-fourier-transform

%https://www.hwe.design/theories-concepts/foundation-of-the-study-of-linear-algebra-and-functional-analysis/vector-space-signal-processing

%---------------------------------------------------

%--------------------------------------------------------------------------
%=================================================
%=================================================

\section{Intellectual Merits, Limitations and Future Work of {\our}}\label{sec:merits_limitations}

In this section, we will briefly discuss the merits and limitations of the {\our} model in various aspects, which may also illustrate some potential future research opportunities on the {\our} model.

\subsection{Intellectual Merits of {\our}}

In this paper, we introduce the {\our} model that can unify several machine learning and deep learning base models with a canonical representation. The {\our} model has transformative impacts on the current and future academic research and industrial development of machine learning, deep learning and artificial intelligence. We summarize the intellectual merits of {\our} from several different perspectives as follows:

\textbf{Theoretical Merits}: The {\our} model disentangles data from parameters and formalizes several machine learning and deep learning base model architectures as the inner product of expanded data instances with the reconciled parameters, derived according to the Taylor's theorem. From the machine learning theory perspective, the data expansion functions are responsible for input data vector projection from one base space to another; and the parameter reconciliation functions will accommodate the parameters dimensions and regularize them to avoid overfitting according to the VC-theory. Meanwhile, from the biological neuroscience perspective, the data expansion functions models the complex states of neurons before, during and after the activations; and the parameter reconciliation functions fabricate the neurotransmitters and their ligand-gated ion channels to compose the synapses. The {\our} model architecture makes it much easier to interpret the physical meanings of both the results and the learning process from both theoretic machine learning and biological neuroscience perspectives.

\textbf{Technical Merits}: The {\our} model provides a unified representation for both classic machine learning models and the family of deep models under one framework. We use non-deep machine learning models, such as SVM and probabilistic models (including naive Bayes, Bayesian network and Markov network), and deep models, such as MLP and KAN, as examples to illustrate how they can be represented with {\our}. Additionally, the multi-layer, multi-head, and multi-channel design of {\our} provides significant flexibility for future model design and deployment in various applications. The unified representations in {\our} greatly simplify the design and development of future machine learning, deep learning, and artificial intelligence models and systems.

\textbf{Computational Merits}: The disentanglement of data from parameters at each layer in {\our} allows their computations to be separately routed to different chips, machines, and cloud platforms, protecting both data privacy and model parameter security. Additionally, {\our}'s reliance on inner product computations facilitates easy parallelization across various computational platforms, systems, and hardware. The {\our} model can be effectively trained using the conventional error backpropagation. Gradient computation and parameter updates can be performed locally with minimal message passing across different computational facilities, enabling the deployment of {\our} in real-world large-scale systems and models with significantly reduced computational resources and energy consumption.

\subsection{Limitations and Future Work of {\our}}

We have observed several limitations with the current {\our} model based on our experiences in both framework development and experimental testing. These limitations can also highlight potential future research opportunities and suggest development directions for readers. We summarize these limitations from various perspectives as follows:

\textbf{Modeling Limitations}: While we aim to unify existing machine and deep learning base models with {\our}, it's important to note that there still exist a large number of models we haven't covered or investigated regarding how to represent them with {\our}. Examples include tree family models ({\eg} decision tree, gradient boosted trees, and random forest), ensemble learning models ({\eg} bagging, boosting, and stacking-based models), unsupervised learning models ({\eg} k-means, principal component analysis), and reinforcement learning models ({\eg} q-learning, policy gradient methods). Many of these models can potentially be represented with {\our}; for instance, decision trees can be interpreted with conditional probabilities, and ensemble learning models can be modeled in {\our} with the multi-head multi-channel mechanism. However, we leave the exploration of these models for future research endeavors.

\textbf{Efficiency Limitations}: The data expansion functions in {\our} expand data instances to a higher dimension to better approximate the desired target function. Although we can utilize deep architectures to constrain the dimensionality of the expansion space at each layer, the intermediate representations of data instances may still consume significant computational resources, and deep architectures can introduce higher calculation time overhead. For instance, if we plan to use {\our} to build language models with longer context widths, certain expansion functions ({\eg} combinatorial expansion and those defined with the Kronecker product) may not be applicable. Furthermore, in the design of data expansion functions, parameter reconciliation functions, and remainder functions, we did not consider any techniques for optimizing performance on high-performance computing (HPC) systems, which could potentially address efficiency issues. We intend to investigate these areas in future work.

\textbf{Learning Limitations}: The training of the {\our} model in this paper still relies on the error backpropagation algorithm, which has faced criticism in recent years due to its high computational costs, slow convergence, and sensitivity to hyperparameters and initializations. Additionally, the higher-dimensional expansions may lead to wider outputs for input data instances, potentially causing learning problems such as overfitting and performance robustness issues. However, the disentanglement of parameters from data into different functions at each layer presents an opportunity to explore new model training algorithms, a discussion which is not included in this paper. We consider to explore and design new learning algorithms for {\our} as one of the most important directions for future research.

%=================================================
%=================================================
%
%\section{What's the Next?}\label{sec:future_work}
%
%\subsection{Explainable {\our}}
%
%Theoretic performance bound of {\our}
%
%\subsection{Generalized {\our}}
%
%Define more diverse functions for diverse applications
%
%Applications: we will also investigate to build larger models with {\our} for real-world applications, which will greatly enrich the technological ecosystem of the current and future AI thrive development.
%
%\subsection{Reusable {\our}}
%
%Transfer learning of {\our}
%
%\subsection{Efficient {\our}}
%
%HPC
%
%Distribution expansion function: to learning the distribution parameters dynamically for better data expansions.
%
%\subsection{Effectiveness {\our}}
%
%Learning algorithms.
%
%We will investigate whether the dual solutions proposed for SVM with kernel tricks can also be used for the learning of {\our} without the need to calculate the high-dimensional representations.
%

%--------------------------------------------------------------------------
\section{Related Work}\label{sec:related_work}

This section briefly discusses the other existing work related to the {\our} model introduced in this paper, including the existing machine learning and deep learning base models, and the techniques utilized for designing the component functions in {\our}.

\subsection{Machine Learning and Deep Learning Base Models}

In this paper, we compare {\our} with two main types of machine learning base models, {\ie} probabilistic models and support vector machine, and two main types of deep learning base models, {\ie} multi-layer perceptron and Kolmogorov-Arnold network. We will briefly introduce their development history and some important papers that made critical technical breakthroughs about them below.

\subsubsection{Probabilistic Models}

In classic machine learning, lots of models have been proposed based on probability theory. Among them, naive Bayes \cite{10.5555/2074158.2074196}, based on the assumption that the features are conditionally independent given the target class, is one of the most well-known linear probabilistic classifiers designed based on the Bayes' theorem. Different from naivey Bayes, Markov network \cite{besag1974spatial} and Bayesian network \cite{10.5555/534975} models the dependency relationships among the variables as an inter-connected variable network. A random field can be said to be a Markov random field if it satisfy the Markov property \cite{article}. Markov network, also known as Markov random field and Markov model, was proposed by Besag \cite{besag1974spatial} to represent the spatial interactions of lattice systems based on the Hammersley-Clifford Theorem. As one of the pioneers of Bayesian networks and probabilistic models, in his book \cite{10.5555/534975}, Pearl introduced the networks for plausible inference for probabilistic reasoning in intelligent systems. For calculating and updating the likelihood in the Bayesian network, Pearl proposed the Belief Propagation algorithm in \cite{10.5555/2876686.2876719}. Expectation-maximization algorithm initially introduced in \cite{DEMP1977} was used for probabilistic graphical model training afterwards as proposed in \cite{emalgorithm}. All these aforementioned probabilistic models have comprehensively introduced in the book written by Koller and Friedman \cite{10.5555/1795555}.

\subsubsection{Support Vector Machines}

Support vector machine (SVM), as a supervised max-margin model, was mainly developed by Vladimir Vapnik and his colleagues. In \cite{10.1023/A:1022627411411, 10.1007/BFb0020166}, Cortes and Vapnik introduced support-vector network as a binary classification model. With the kernel tricks \cite{10.1145/130385.130401}, Boser, Guyon and Vapnik proposed to project input vectors that are non-linearly separable into a high-dimensional space, within which the data instances can be separated with a linear decision surface. For the training of SVM with kernel tricks, \cite{10.1145/130385.130401} addressed the optimization problem within the dual space instead to identify the support vectors closet to the decision boundary. Meanwhile, as the training data size increases, the costs of training SVM will increase dramatically. Platt \cite{platt1998sequential} proposed the Sequential Minimal Optimization (SMO) algorithm which breaks the quadratic programming (QP) problem in SVM training into a series of smallest possible QP problems, which made fast SVM training possible. Besides classification tasks, Vapnik, Golowich and Smola also reported the performance of SVM for function approximation, regression estimation and signal processing in \cite{10.5555/2998981.2999021}. Another follow-up work \cite{10.5555/944790.944807} used support vector machine for the clustering task. All these model and training algorithms related to SVM were implemented by Chang and Lin in the LIBSVM package \cite{libsvm}.

\subsubsection{Multi-layer Perceptron}

Multi-layer perceptron (MLP) denotes the perceptron model \cite{rosenblatt1958perceptron} with a multi-layered architecture built based on the MP-neurons \cite{mcculloch43a}. To address the non-linear XOR problem proposed by Minsky \cite{Min69}, hidden layers were added to the single-layered perceptron to compose a multi-layered network with the feedforward architecture by Rosenblatt \cite{Ivakhnenko:209675}. The deep MLP learned with stochastic gradient descent was proposed by Amari in \cite{4039068}, which introduced a five-layered feedforward network with two learning layers. As to the theoretic foundation, Cybenko introduced the universal approximation theorem in \cite{citeulike:3561150} and the multilayer perceptron with hidden layers was demonstrated to be a universal approximator \cite{HORNIK1989359}. For the training of deep MLP models, the modern version of backpropagation algorithm designed based on the chain-rule was published by Linnainmaa in his master thesis \cite{Linnainmaa}, and was later evaluated with experimental analysis by Rumelhart, Hinton and Williams in \cite{10.5555/104279.104293}. Nowadays, MLP still serves as the base model for most of the current deep learning models.

\subsubsection{Kolmogorov-Arnold Network}

Unlike MLPs with predefined, static activation functions, the recent Kolmogorov-Arnold network (KAN) model \cite{Liu2024KANKN} proposes to learn the activation functions attached to neuron-neuron connections. KAN was designed based on the Kolmogorov-Arnold superposition theorem \cite{kolmogorov:superposition, Arnold2009} and models multivariate continuous functions as a superposition of the two-argument addition of continuous functions of one variable. Prior to the KAN mode, many other prior works \cite{10.1162/neco.1993.5.1.18, 10.1016/S0893-6080(01)00107-1, 10.1007/3-540-46084-5_77} have proposed to build neural networks based on the Kolmogorov-Arnold theorem already. Based on the Kolmogorov-Arnold superposition theorem, \cite{Montanelli2019ErrorBF} investigates the error bounds of deep ReLU networks; \cite{He2023OnTO} studies the expressive power of ReLU deep neural networks in function approximation; and \cite{Fakhoury2022ExSpliNetAI} investigates the interpretability of spline-based neural networks. Besides the continuous function approximation, \cite{inbook_kolmogorov} studied the Kolmogorov spline network for image processing. In addition, the Kolmogorov-Arnold superposition theorem has also been demonstrated to be capable to address the curse of dimension when approximating high-dimensional functions \cite{Lai2021TheKS}.

\subsection{Component Function Design}

In this paper, we introduce a tripartite set of compositional component functions - data expansion, parameter reconciliation, and remainder functions - that serve as the building blocks for the {\our} model. These component functions' designing and definitions are closely related to the techniques used for data augmentation, parameter-efficient learning and residual learning. We will briefly introduce the related work of these topics below.

\subsubsection{Data Augmentation}

%https://arxiv.org/pdf/2401.15422

In addition to the kernel tricks \cite{10.1145/130385.130401} and Taylor's expansion based machine learning preliminary work introduced in the previous Section~\ref{subsec:taylor_theorem_machine_learning}, some data expansion functions introduced in this paper are also inspired by the data augmentation techniques proposed for deep learning model training. Data augmentation \cite{Zhou2024ASO} is a frequently used approach for model learning with incomplete or imbalanced data, which creates new data instances via minor modifications on existing data. Some frequently used data augmentation techniques used in current deep models include geometric transformation \cite{10.5555/2999134.2999257, Yang2022ImageDA}, space transformation \cite{Devries2017DatasetAI, Liu2018FeatureST}, noise injection \cite{Devries2017DatasetAI} and adversarial instance creation \cite{Ganin2015DomainAdversarialTO}. Frequently used geometric transformation techniques include flipping, rotation, scaling, translation and reflection in the input space, which may improve the models in combating against the overfitting issues \cite{10.5555/2999134.2999257}. The space transformation and noise injection techniques introduced in \cite{Devries2017DatasetAI} performances the data augmentation not in the input space but in a learned feature space instead \cite{Liu2018FeatureST}. Specifically, via adding noise, interpolation and extrapolation, the data augmentation can improve model learning performance. In addition, in \cite{Ganin2015DomainAdversarialTO} and other follow-up works \cite{Sinha2017CertifyingSD, Volpi2018GeneralizingTU, Zhao2020MaximumEntropyAD}, adversarial data augmentation was utilized to improve the model generalization and robustness. A comprehensive survey of related data augmentation techniques used for image and language data and models is has also been provided in \cite{Yang2022ImageDA, Zhou2024ASO}.

\subsubsection{Parameter-Efficient Learning}

Many of the parameter reconciliation functions introduced in this paper are actually defined based on the current parameter-efficient fine-tuning (PEFT) techniques proposed for language models. PEFT offers an effective solution by reducing the number of fine-tuning parameters and memory usage while achieving comparable performance to full fine-tuning. Current PEFT methods can be roughly categorized into several types, such as adapter-based methods \cite{Houlsby2019ParameterEfficientTL, Lin2020ExploringVG}, parameter masking based methods \cite{Zhao2020MaskingAA, Sung2021TrainingNN}, low-rank matrix decomposition based methods \cite{Hu2021LoRALA, Valipour2022DyLoRAPT}. Adapter based fine-tuning methods propose to freeze pre-trained model parameters and introduce new extra trainable parameters for task-specific fine-tuning \cite{Houlsby2019ParameterEfficientTL, Lin2020ExploringVG, Pfeiffer2020AdapterHubAF, Rckl2020AdapterDropOT}. Besides adding these new parameters as adapters, researchers also propose to pre-pend these new parameters to the input data as prompts instead, such as \cite{Li2021PrefixTuningOC, Lester2021ThePO}. The masking based fine-tuning methods reduce the number of fine-tuned parameters by selecting a subset of pre-trained parameters critical to downstream tasks while discarding unimportant ones. The parameter masking can be applied to either the pre-trained parameters \cite{Zhao2020MaskingAA, Sung2021TrainingNN} or the delta parameters \cite{Ansell2021ComposableSF, Xu2021RaiseAC}. Based on the adapters, LoRA (Low-Rank Adaptation) \cite{Hu2021LoRALA} and its derivatives \cite{Valipour2022DyLoRAPT, Zhang2023AdaLoRAAB, Zi2023DeltaLoRAFH} propose to decompose the parameter matrix into the product of low-rank sub-matrices instead. Instead of composing the parameters into low-rank matrix products, \cite{Zhang2021BeyondFL} proposes to further reduce the parameters with the hypercomplex multiplication of small-sized sub-matrices instead; \cite{He2022ParameterEfficientMA} integrates both LoRA with the hypercomplex multiplication for parameter matrix decomposition. A recent survey \cite{Xu2023ParameterEfficientFM} provides a more comprehensive introduction about the latest parameter-efficient model learning methods for pre-trained language models.

\subsubsection{Residual Learning}

%file:///Users/jiaweizhang/Downloads/applsci-12-08972.pdf

The remainder function derived from Taylor's expansions plays a very similar role in the {\our} model as the skip-layer connections used in residual learning. Cross-layer residual connections \cite{Srivastava2015HighwayN, He2015DeepRL} have been extensively in building current deep learning models, which greatly improves the model stability, robustness and trainability. Prior to the ResNet, Schmidhuber et al. proposed the Highway networks \cite{Srivastava2015HighwayN} that allow unimpeded information flow across several layers for building deep models with gradient based training algorithms. Inspired by Highway network, He et al. introduced the ResNet \cite{He2015DeepRL} for building deep CNN models to achieve about $28\%$ improvement on the CoCo dataset. Several follow-up work \cite{He2016IdentityMI, Veit2016ResidualNA, Liao2016BridgingTG, Jastrzebski2017ResidualCE} investigate the interpretation and theoretic analysis of the ResNet and residual learning. In \cite{He2016IdentityMI}, He et al. analyze the propagation formulation behind the residual building block in both forward and backward propagations; \cite{Veit2016ResidualNA} interprets residual networks as a collection of many paths of differing length, where the shortest paths are leveraged during training; \cite{Liao2016BridgingTG} interprets ResNet layers as iterative refinement of learned representations; and \cite{Jastrzebski2017ResidualCE} investigates that residual connections naturally encourage features of residual blocks to move along the negative gradient of loss. In addition to CNN, residual learning and the skip-layer highway connections have become a standard paradigm in deep learning. It has been extensively used in the follow-up deep models, including Transformer \cite{Vaswani2017AttentionIA}, BERT \cite{Devlin2019BERTPO}, Graph-BERT \cite{Zhang2020GraphBertOA}, ViT \cite{Dosovitskiy2020AnII}, Stable Diffusion \cite{Ho2020DenoisingDP} and almost all the recent large language models.

%--------------------------------------------------------------------------
%=================================================
%=================================================

\section{Conclusion}\label{sec:conclusion}

In this paper, we introduce the Reconciled Polynomial Network ({\our}) as a novel base model for deep function learning tasks. By incorporating multi-layers, multi-heads and multi-channels, {\our} has a versatile model architecture and attains superior modeling capability for diverse deep function learning tasks on various multi-modality data. What's more, through specific selections of component functions - including data expansion, parameter reconciliation and remainder functions - {\our} provides a unifying framework for several influential base models into a canonical representation. To demonstrate the effectiveness of {\our}, this paper presents the extensive empirical experiments conducted across numerous benchmark datasets for various deep function learning tasks. The results consistently demonstrate {\our}'s superior performance compared to other base models. 

Compared with other existing machine learning and deep learning base models, {\our} presents significant advantages across several critical dimensions, including generalizability, interpretability, and reusability. Without any prior assumptions on data modalities, {\our} serves as a general model applicable to multi-modal data from the outset. Furthermore, through the disentanglement of data from parameters using the expansion, reconciliation, and remainder functions, {\our} achieves greater interpretability compared to existing base models. Moreover, the component functions designed and learned in {\our} is inherently well-suited for reusability and continual learning. In addition to the empirical evaluations via experiments on benchmark datasets, this paper also discusses the interpretations of the {\our} model design from both technical machine learning and biological neuroscience perspectives.

To facilitate the adoption, implementation and experimentation of {\our}, we have released a comprehensive toolkit named {\toolkit} in this paper. The {\toolkit} toolkit provides a rich library of pre-implemented component functions introduced in this paper, which allows researchers to rapidly design, customize, and effectively deploy {\our} models across various learning tasks.
%--------------------------------------------------------------------------
%=================================================
%=================================================

\acksection

This work is partially supported by NSF through grants IIS-1763365 and IIS-2106972.

%--------------------------------------------------------------------------
\newpage
\bibliography{reference}
\bibliographystyle{plainnat}
%--------------------------------------------------------------------------
}

{
\newpage
\appendix
\section{Appendix}

%==============================================================
\subsection{Performance of {\our} Variants for Continuous Function Fitting and Approximation}\label{subsec:appendix_continuous_function _fitting}

\noindent \textbf{Descriptions}: In the following Tables~\ref{tab:elementary_rpn_ablation}-\ref{tab:feynman_rpn_ablation}, we provide the experimental results of {\our} on the elementary function, composite function and Feynman function datasets, respectively. Several different variants of {\our} are compared:
\begin{itemize}
\item \textbf{{\our}-Exd-LoRR-Linear}: The {\our} model with extended expansion function, involving Taylor's expansion function and Bspline expansion function, low-rank reconciliation function and linear remainder function.

\item \textbf{{\our}-Exd-LoRR-Zero}: The {\our} model with similar component function as {\our}-Exd-LoRR-Linear, but replace the remainder with zero remainder function instead.

\item \textbf{{\our}-Nstd-LoRR-Linear}: The {\our} model with nested expansion function, involving Taylor's expansion function and Bspline expansion function, low-rank reconciliation function and linear remainder function. The nested expansion will dramatically increase the intermediate expansion space dimensions, it may involve more learnable parameters in the model.

\item \textbf{{\our}-Nstd-LoRR-Zero}: The {\our} model with similar component function as {\our}-Nstd-LoRR-Linear, but replace the remainder with zero remainder function instead.
\end{itemize}

%-----------------------------------------
%------------------------------------------------------------------------------------
\begin{table}[h!]
    \tiny
    \centering
    \caption{Elementary Function Fitting with {\our} Variants.}
    \label{tab:elementary_rpn_ablation}
    {\fontsize{7}{10}\selectfont
    \scalemath{0.9}{
    \begin{tabular}{|c|c|c|c|c|}
  \hline
  \thead{Eq.} & \thead{{\our}\\ (Extended Expansion-LoRR-Linear)} & \thead{{\our}\\ (Extended Expansion-LoRR-Zero)} & \thead{{\our}\\ (Nested Expansion-LoRR-Linear)} & \thead{{\our}\\ (Nested Expansion-LoRR-Zero)} \\ \hline
  E.0 &\makecell{ $1.92 \times 10^{-8}$ $\pm$  $1.30 \times 10^{-8}$ } &\makecell{ $8.40 \times 10^{-8}$ $\pm$  $1.12 \times 10^{-7}$ } &\makecell{ $1.35 \times 10^{-8}$ $\pm$  $6.69 \times 10^{-9}$ } &\makecell{ $1.93 \times 10^{-8}$ $\pm$  $1.15 \times 10^{-8}$ } \\ \hline
E.1 &\makecell{ $6.42 \times 10^{-2}$ $\pm$  $3.08 \times 10^{-2}$ } &\makecell{ $8.67 \times 10^{-2}$ $\pm$  $8.22 \times 10^{-2}$ } &\makecell{ $1.00 \times 10^{-1}$ $\pm$  $1.22 \times 10^{-1}$ } &\makecell{ $1.03 \times 10^{-1}$ $\pm$  $1.40 \times 10^{-1}$ } \\ \hline
E.2 &\makecell{ $4.08 \times 10^{-7}$ $\pm$  $4.65 \times 10^{-7}$ } &\makecell{ $2.56 \times 10^{-7}$ $\pm$  $1.48 \times 10^{-7}$ } &\makecell{ $9.90 \times 10^{-8}$ $\pm$  $5.91 \times 10^{-8}$ } &\makecell{ $5.06 \times 10^{-8}$ $\pm$  $3.53 \times 10^{-8}$ } \\ \hline
E.3 &\makecell{ $5.51 \times 10^{-7}$ $\pm$  $4.69 \times 10^{-7}$ } &\makecell{ $3.81 \times 10^{-6}$ $\pm$  $6.56 \times 10^{-6}$ } &\makecell{ $1.08 \times 10^{-6}$ $\pm$  $1.53 \times 10^{-6}$ } &\makecell{ $1.26 \times 10^{-6}$ $\pm$  $1.37 \times 10^{-6}$ } \\ \hline
E.4 &\makecell{ $4.64 \times 10^{-5}$ $\pm$  $4.69 \times 10^{-5}$ } &\makecell{ $7.05 \times 10^{-5}$ $\pm$  $3.12 \times 10^{-5}$ } &\makecell{ $1.87 \times 10^{-5}$ $\pm$  $2.34 \times 10^{-5}$ } &\makecell{ $1.80 \times 10^{-5}$ $\pm$  $2.37 \times 10^{-5}$ } \\ \hline
E.5 &\makecell{ $7.28 \times 10^{-8}$ $\pm$  $6.79 \times 10^{-8}$ } &\makecell{ $4.95 \times 10^{-8}$ $\pm$  $4.39 \times 10^{-8}$ } &\makecell{ $3.99 \times 10^{-9}$ $\pm$  $1.85 \times 10^{-9}$ } &\makecell{ $5.67 \times 10^{-9}$ $\pm$  $3.20 \times 10^{-9}$ } \\ \hline
E.6 &\makecell{ $4.32 \times 10^{-8}$ $\pm$  $2.43 \times 10^{-8}$ } &\makecell{ $1.25 \times 10^{-7}$ $\pm$  $1.34 \times 10^{-7}$ } &\makecell{ $1.41 \times 10^{-8}$ $\pm$  $8.39 \times 10^{-9}$ } &\makecell{ $5.93 \times 10^{-9}$ $\pm$  $3.98 \times 10^{-9}$ } \\ \hline
E.7 &\makecell{ $9.25 \times 10^{-8}$ $\pm$  $5.53 \times 10^{-8}$ } &\makecell{ $6.02 \times 10^{-8}$ $\pm$  $4.46 \times 10^{-8}$ } &\makecell{ $6.88 \times 10^{-9}$ $\pm$  $4.46 \times 10^{-9}$ } &\makecell{ $1.67 \times 10^{-8}$ $\pm$  $2.32 \times 10^{-8}$ } \\ \hline
E.8 &\makecell{ $2.05 \times 10^{-6}$ $\pm$  $2.22 \times 10^{-6}$ } &\makecell{ $1.65 \times 10^{-6}$ $\pm$  $2.06 \times 10^{-6}$ } &\makecell{ $1.26 \times 10^{-6}$ $\pm$  $6.89 \times 10^{-7}$ } &\makecell{ $5.16 \times 10^{-7}$ $\pm$  $2.97 \times 10^{-7}$ } \\ \hline
E.9 &\makecell{ $3.57 \times 10^{-6}$ $\pm$  $3.84 \times 10^{-6}$ } &\makecell{ $4.49 \times 10^{-5}$ $\pm$  $5.15 \times 10^{-5}$ } &\makecell{ $1.48 \times 10^{-6}$ $\pm$  $8.78 \times 10^{-7}$ } &\makecell{ $3.73 \times 10^{-7}$ $\pm$  $2.32 \times 10^{-7}$ } \\ \hline
E.10 &\makecell{ $8.42 \times 10^{-9}$ $\pm$  $7.09 \times 10^{-9}$ } &\makecell{ $4.86 \times 10^{-9}$ $\pm$  $2.17 \times 10^{-9}$ } &\makecell{ $1.13 \times 10^{-9}$ $\pm$  $2.69 \times 10^{-10}$ } &\makecell{ $3.02 \times 10^{-9}$ $\pm$  $1.65 \times 10^{-9}$ } \\ \hline
E.11 &\makecell{ $1.56 \times 10^{-7}$ $\pm$  $9.03 \times 10^{-8}$ } &\makecell{ $1.99 \times 10^{-7}$ $\pm$  $1.65 \times 10^{-7}$ } &\makecell{ $8.83 \times 10^{-8}$ $\pm$  $7.14 \times 10^{-8}$ } &\makecell{ $5.43 \times 10^{-8}$ $\pm$  $5.25 \times 10^{-8}$ } \\ \hline
E.12 &\makecell{ $5.28 \times 10^{-7}$ $\pm$  $6.84 \times 10^{-7}$ } &\makecell{ $7.00 \times 10^{-7}$ $\pm$  $4.65 \times 10^{-7}$ } &\makecell{ $1.11 \times 10^{-7}$ $\pm$  $9.78 \times 10^{-8}$ } &\makecell{ $4.47 \times 10^{-8}$ $\pm$  $3.42 \times 10^{-8}$ } \\ \hline
E.13 &\makecell{ $2.61 \times 10^{-8}$ $\pm$  $1.79 \times 10^{-8}$ } &\makecell{ $5.73 \times 10^{-8}$ $\pm$  $6.20 \times 10^{-8}$ } &\makecell{ $4.23 \times 10^{-9}$ $\pm$  $2.15 \times 10^{-9}$ } &\makecell{ $4.55 \times 10^{-9}$ $\pm$  $2.73 \times 10^{-9}$ } \\ \hline
E.14 &\makecell{ $4.94 \times 10^{-9}$ $\pm$  $5.74 \times 10^{-9}$ } &\makecell{ $5.13 \times 10^{-9}$ $\pm$  $3.94 \times 10^{-9}$ } &\makecell{ $1.32 \times 10^{-9}$ $\pm$  $3.41 \times 10^{-10}$ } &\makecell{ $4.34 \times 10^{-9}$ $\pm$  $2.82 \times 10^{-9}$ } \\ \hline
E.15 &\makecell{ $4.85 \times 10^{-6}$ $\pm$  $2.36 \times 10^{-6}$ } &\makecell{ $8.94 \times 10^{-6}$ $\pm$  $4.55 \times 10^{-6}$ } &\makecell{ $4.75 \times 10^{-7}$ $\pm$  $2.04 \times 10^{-7}$ } &\makecell{ $5.38 \times 10^{-7}$ $\pm$  $5.62 \times 10^{-7}$ } \\ \hline
E.16 &\makecell{ $2.61 \times 10^{-5}$ $\pm$  $2.20 \times 10^{-5}$ } &\makecell{ $3.00 \times 10^{-5}$ $\pm$  $3.47 \times 10^{-5}$ } &\makecell{ $4.21 \times 10^{-5}$ $\pm$  $4.52 \times 10^{-5}$ } &\makecell{ $2.51 \times 10^{-5}$ $\pm$  $3.04 \times 10^{-5}$ } \\ \hline
\end{tabular}
    }}
\end{table}
%------------------------------------------------------------------------------------

\newpage

%------------------------------------------------------------------------------------
\begin{table}[h!]
    \tiny
    \centering
    \caption{Composite Function Fitting with {\our} Variants.}
    \label{tab:composite_rpn_ablation}
    {\fontsize{7}{10}\selectfont
    \scalemath{0.9}{
    \begin{tabular}{|c|c|c|c|c|}
  \hline
  \thead{Eq.} & \thead{{\our}\\ (Extended Expansion-LoRR-Linear)} & \thead{{\our}\\ (Extended Expansion-LoRR-Zero)} & \thead{{\our}\\ (Nested Expansion-LoRR-Linear)} & \thead{{\our}\\ (Nested Expansion-LoRR-Zero)} \\ \hline
C.0 &\makecell{ $1.18 \times 10^{-1}$ $\pm$  $1.20 \times 10^{-1}$ } &\makecell{ $3.25 \times 10^{-2}$ $\pm$  $3.34 \times 10^{-2}$ } &\makecell{ $6.45 \times 10^{-3}$ $\pm$  $3.49 \times 10^{-3}$ } &\makecell{ $2.73 \times 10^{-3}$ $\pm$  $3.32 \times 10^{-3}$ } \\ \hline
C.1 &\makecell{ $3.71 \times 10^{-7}$ $\pm$  $3.64 \times 10^{-7}$ } &\makecell{ $6.73 \times 10^{-7}$ $\pm$  $3.04 \times 10^{-7}$ } &\makecell{ $1.76 \times 10^{-7}$ $\pm$  $1.20 \times 10^{-7}$ } &\makecell{ $1.84 \times 10^{-7}$ $\pm$  $1.19 \times 10^{-7}$ } \\ \hline
C.2 &\makecell{ $1.23 \times 10^{-6}$ $\pm$  $8.21 \times 10^{-7}$ } &\makecell{ $1.33 \times 10^{-6}$ $\pm$  $1.29 \times 10^{-6}$ } &\makecell{ $2.40 \times 10^{-6}$ $\pm$  $3.59 \times 10^{-6}$ } &\makecell{ $1.54 \times 10^{-6}$ $\pm$  $9.85 \times 10^{-7}$ } \\ \hline
C.3 &\makecell{ $9.36 \times 10^{-5}$ $\pm$  $7.99 \times 10^{-5}$ } &\makecell{ $1.17 \times 10^{-4}$ $\pm$  $1.18 \times 10^{-4}$ } &\makecell{ $1.03 \times 10^{-5}$ $\pm$  $1.57 \times 10^{-5}$ } &\makecell{ $3.86 \times 10^{-6}$ $\pm$  $2.99 \times 10^{-6}$ } \\ \hline
C.4 &\makecell{ $4.40 \times 10^{-7}$ $\pm$  $3.09 \times 10^{-7}$ } &\makecell{ $4.40 \times 10^{-7}$ $\pm$  $4.59 \times 10^{-7}$ } &\makecell{ $1.56 \times 10^{-7}$ $\pm$  $1.02 \times 10^{-7}$ } &\makecell{ $1.09 \times 10^{-7}$ $\pm$  $6.68 \times 10^{-8}$ } \\ \hline
C.5 &\makecell{ $4.95 \times 10^{-6}$ $\pm$  $5.14 \times 10^{-6}$ } &\makecell{ $1.48 \times 10^{-6}$ $\pm$  $1.69 \times 10^{-6}$ } &\makecell{ $3.70 \times 10^{-7}$ $\pm$  $3.73 \times 10^{-7}$ } &\makecell{ $7.24 \times 10^{-8}$ $\pm$  $4.28 \times 10^{-8}$ } \\ \hline
C.6 &\makecell{ $6.53 \times 10^{-2}$ $\pm$  $5.58 \times 10^{-2}$ } &\makecell{ $3.72 \times 10^{-2}$ $\pm$  $3.95 \times 10^{-2}$ } &\makecell{ $6.25 \times 10^{-3}$ $\pm$  $5.42 \times 10^{-3}$ } &\makecell{ $2.25 \times 10^{-3}$ $\pm$  $2.49 \times 10^{-3}$ } \\ \hline
C.7 &\makecell{ $6.67 \times 10^{-7}$ $\pm$  $3.73 \times 10^{-7}$ } &\makecell{ $4.97 \times 10^{-7}$ $\pm$  $5.01 \times 10^{-7}$ } &\makecell{ $7.72 \times 10^{-8}$ $\pm$  $7.22 \times 10^{-8}$ } &\makecell{ $2.94 \times 10^{-8}$ $\pm$  $1.37 \times 10^{-8}$ } \\ \hline
C.8 &\makecell{ $3.78 \times 10^{-7}$ $\pm$  $2.98 \times 10^{-7}$ } &\makecell{ $5.70 \times 10^{-8}$ $\pm$  $1.98 \times 10^{-8}$ } &\makecell{ $1.40 \times 10^{-8}$ $\pm$  $6.12 \times 10^{-9}$ } &\makecell{ $1.64 \times 10^{-8}$ $\pm$  $9.49 \times 10^{-9}$ } \\ \hline
C.9 &\makecell{ $8.77 \times 10^{-5}$ $\pm$  $5.73 \times 10^{-5}$ } &\makecell{ $1.88 \times 10^{-4}$ $\pm$  $2.20 \times 10^{-4}$ } &\makecell{ $5.59 \times 10^{-6}$ $\pm$  $3.63 \times 10^{-6}$ } &\makecell{ $4.55 \times 10^{-6}$ $\pm$  $3.99 \times 10^{-6}$ } \\ \hline
C.10 &\makecell{ $4.68 \times 10^{-7}$ $\pm$  $3.11 \times 10^{-7}$ } &\makecell{ $3.40 \times 10^{-7}$ $\pm$  $4.77 \times 10^{-7}$ } &\makecell{ $5.30 \times 10^{-8}$ $\pm$  $3.02 \times 10^{-8}$ } &\makecell{ $3.58 \times 10^{-8}$ $\pm$  $1.18 \times 10^{-8}$ } \\ \hline
C.11 &\makecell{ $9.17 \times 10^{-7}$ $\pm$  $9.75 \times 10^{-7}$ } &\makecell{ $3.57 \times 10^{-7}$ $\pm$  $2.56 \times 10^{-7}$ } &\makecell{ $2.02 \times 10^{-7}$ $\pm$  $1.09 \times 10^{-7}$ } &\makecell{ $4.87 \times 10^{-7}$ $\pm$  $8.71 \times 10^{-7}$ } \\ \hline
C.12 &\makecell{ $3.63 \times 10^{-5}$ $\pm$  $2.25 \times 10^{-5}$ } &\makecell{ $3.74 \times 10^{-5}$ $\pm$  $1.88 \times 10^{-5}$ } &\makecell{ $4.11 \times 10^{-5}$ $\pm$  $4.54 \times 10^{-5}$ } &\makecell{ $7.17 \times 10^{-5}$ $\pm$  $8.87 \times 10^{-5}$ } \\ \hline
C.13 &\makecell{ $2.83 \times 10^{-6}$ $\pm$  $2.03 \times 10^{-6}$ } &\makecell{ $6.97 \times 10^{-7}$ $\pm$  $2.29 \times 10^{-7}$ } &\makecell{ $2.45 \times 10^{-7}$ $\pm$  $1.35 \times 10^{-7}$ } &\makecell{ $8.56 \times 10^{-8}$ $\pm$  $3.81 \times 10^{-8}$ } \\ \hline
C.14 &\makecell{ $3.45 \times 10^{-7}$ $\pm$  $4.58 \times 10^{-7}$ } &\makecell{ $3.50 \times 10^{-7}$ $\pm$  $3.89 \times 10^{-7}$ } &\makecell{ $2.22 \times 10^{-8}$ $\pm$  $1.04 \times 10^{-8}$ } &\makecell{ $1.69 \times 10^{-8}$ $\pm$  $2.42 \times 10^{-8}$ } \\ \hline
C.15 &\makecell{ $2.56 \times 10^{3}$ $\pm$  $4.35 \times 10^{3}$ } &\makecell{ $2.47 \times 10^{3}$ $\pm$  $4.39 \times 10^{3}$ } &\makecell{ $1.20 \times 10^{4}$ $\pm$  $2.18 \times 10^{4}$ } &\makecell{ $1.17 \times 10^{4}$ $\pm$  $2.11 \times 10^{4}$ } \\ \hline
C.16 &\makecell{ $3.42 \times 10^{-8}$ $\pm$  $2.79 \times 10^{-8}$ } &\makecell{ $1.74 \times 10^{-9}$ $\pm$  $1.09 \times 10^{-9}$ } &\makecell{ $2.10 \times 10^{-9}$ $\pm$  $1.58 \times 10^{-9}$ } &\makecell{ $1.48 \times 10^{-9}$ $\pm$  $1.32 \times 10^{-9}$ } \\ \hline
\end{tabular}
    }}
\end{table}
%------------------------------------------------------------------------------------

\newpage

%------------------------------------------------------------------------------------
\begin{table}[h!]
    \tiny
    \centering
    \caption{Feynman Function Fitting with {\our} Variants.}
    \label{tab:feynman_rpn_ablation}
    {\fontsize{7}{10}\selectfont
    \scalemath{0.9}{
    \begin{tabular}{|c|c|c|c|c|}
  \hline
  \thead{Eq.} & \thead{{\our}\\ (Extended Expansion-LoRR-Linear)} & \thead{{\our}\\ (Extended Expansion-LoRR-Zero)} & \thead{{\our}\\ (Nested Expansion-LoRR-Linear)} & \thead{{\our}\\ (Nested Expansion-LoRR-Zero)} \\ \hline
F.0 &\makecell{ $1.63 \times 10^{-6}$ $\pm$  $1.31 \times 10^{-6}$ } &\makecell{ $8.48 \times 10^{-5}$ $\pm$  $6.93 \times 10^{-5}$ } &\makecell{ $5.45 \times 10^{-7}$ $\pm$  $5.88 \times 10^{-7}$ } &\makecell{ $1.41 \times 10^{-4}$ $\pm$  $4.19 \times 10^{-6}$ } \\ \hline
F.1 &\makecell{ $1.60 \times 10^{-5}$ $\pm$  $1.25 \times 10^{-5}$ } &\makecell{ $5.02 \times 10^{-5}$ $\pm$  $4.33 \times 10^{-6}$ } &\makecell{ $3.97 \times 10^{-5}$ $\pm$  $4.80 \times 10^{-5}$ } &\makecell{ $7.01 \times 10^{-4}$ $\pm$  $2.40 \times 10^{-5}$ } \\ \hline
F.2 &\makecell{ $6.92 \times 10^{-5}$ $\pm$  $2.42 \times 10^{-5}$ } &\makecell{ $6.59 \times 10^{-5}$ $\pm$  $1.01 \times 10^{-5}$ } &\makecell{ $9.99 \times 10^{-5}$ $\pm$  $1.32 \times 10^{-5}$ } &\makecell{ $3.29 \times 10^{-3}$ $\pm$  $1.69 \times 10^{-4}$ } \\ \hline
F.3 &\makecell{ $1.36 \times 10^{1}$ $\pm$  $8.35 \times 10^{0}$ } &\makecell{ $5.02 \times 10^{1}$ $\pm$  $1.31 \times 10^{1}$ } &\makecell{ $1.20 \times 10^{2}$ $\pm$  $9.14 \times 10^{1}$ } &\makecell{ $4.76 \times 10^{2}$ $\pm$  $4.21 \times 10^{1}$ } \\ \hline
F.4 &\makecell{ $8.47 \times 10^{-1}$ $\pm$  $1.21 \times 10^{-1}$ } &\makecell{ $5.66 \times 10^{0}$ $\pm$  $9.33 \times 10^{-1}$ } &\makecell{ $2.00 \times 10^{1}$ $\pm$  $7.81 \times 10^{0}$ } &\makecell{ $3.67 \times 10^{1}$ $\pm$  $5.25 \times 10^{0}$ } \\ \hline
F.5 &\makecell{ $4.94 \times 10^{-3}$ $\pm$  $1.57 \times 10^{-3}$ } &\makecell{ $7.72 \times 10^{-2}$ $\pm$  $1.29 \times 10^{-1}$ } &\makecell{ $1.28 \times 10^{-2}$ $\pm$  $6.23 \times 10^{-3}$ } &\makecell{ $2.20 \times 10^{0}$ $\pm$  $2.36 \times 10^{-2}$ } \\ \hline
F.6 &\makecell{ $8.68 \times 10^{-4}$ $\pm$  $3.84 \times 10^{-4}$ } &\makecell{ $3.34 \times 10^{-3}$ $\pm$  $3.00 \times 10^{-4}$ } &\makecell{ $2.37 \times 10^{-2}$ $\pm$  $2.07 \times 10^{-2}$ } &\makecell{ $4.16 \times 10^{-1}$ $\pm$  $1.81 \times 10^{-2}$ } \\ \hline
F.7 &\makecell{ $9.92 \times 10^{-4}$ $\pm$  $4.82 \times 10^{-4}$ } &\makecell{ $2.47 \times 10^{-3}$ $\pm$  $2.05 \times 10^{-4}$ } &\makecell{ $2.62 \times 10^{-2}$ $\pm$  $1.95 \times 10^{-2}$ } &\makecell{ $3.32 \times 10^{-1}$ $\pm$  $9.83 \times 10^{-3}$ } \\ \hline
F.8 &\makecell{ $6.59 \times 10^{-5}$ $\pm$  $5.13 \times 10^{-5}$ } &\makecell{ $2.75 \times 10^{-2}$ $\pm$  $1.90 \times 10^{-2}$ } &\makecell{ $4.91 \times 10^{-5}$ $\pm$  $7.83 \times 10^{-5}$ } &\makecell{ $1.42 \times 10^{-1}$ $\pm$  $2.49 \times 10^{-3}$ } \\ \hline
F.9 &\makecell{ $4.98 \times 10^{-4}$ $\pm$  $6.95 \times 10^{-4}$ } &\makecell{ $1.27 \times 10^{-2}$ $\pm$  $2.49 \times 10^{-2}$ } &\makecell{ $4.83 \times 10^{-4}$ $\pm$  $1.78 \times 10^{-4}$ } &\makecell{ $3.92 \times 10^{-2}$ $\pm$  $1.99 \times 10^{-3}$ } \\ \hline
F.10 &\makecell{ $7.27 \times 10^{-2}$ $\pm$  $7.00 \times 10^{-2}$ } &\makecell{ $7.13 \times 10^{-1}$ $\pm$  $2.95 \times 10^{-2}$ } &\makecell{ $1.23 \times 10^{0}$ $\pm$  $5.38 \times 10^{-1}$ } &\makecell{ $2.95 \times 10^{0}$ $\pm$  $2.31 \times 10^{-1}$ } \\ \hline
F.11 &\makecell{ $1.58 \times 10^{0}$ $\pm$  $8.47 \times 10^{-1}$ } &\makecell{ $2.61 \times 10^{0}$ $\pm$  $9.01 \times 10^{-1}$ } &\makecell{ $2.88 \times 10^{0}$ $\pm$  $4.33 \times 10^{-1}$ } &\makecell{ $4.33 \times 10^{0}$ $\pm$  $3.73 \times 10^{-1}$ } \\ \hline
F.12 &\makecell{ $6.83 \times 10^{-6}$ $\pm$  $2.53 \times 10^{-6}$ } &\makecell{ $1.08 \times 10^{-5}$ $\pm$  $6.83 \times 10^{-6}$ } &\makecell{ $7.39 \times 10^{-6}$ $\pm$  $6.04 \times 10^{-6}$ } &\makecell{ $5.45 \times 10^{-3}$ $\pm$  $6.94 \times 10^{-4}$ } \\ \hline
F.13 &\makecell{ $3.89 \times 10^{-2}$ $\pm$  $3.06 \times 10^{-2}$ } &\makecell{ $4.56 \times 10^{-1}$ $\pm$  $3.37 \times 10^{-1}$ } &\makecell{ $6.00 \times 10^{-1}$ $\pm$  $4.57 \times 10^{-1}$ } &\makecell{ $4.12 \times 10^{0}$ $\pm$  $2.04 \times 10^{-2}$ } \\ \hline
F.14 &\makecell{ $2.13 \times 10^{-3}$ $\pm$  $4.51 \times 10^{-4}$ } &\makecell{ $3.08 \times 10^{-3}$ $\pm$  $6.18 \times 10^{-4}$ } &\makecell{ $4.08 \times 10^{-2}$ $\pm$  $2.46 \times 10^{-2}$ } &\makecell{ $1.95 \times 10^{-1}$ $\pm$  $1.57 \times 10^{-2}$ } \\ \hline
F.15 &\makecell{ $2.99 \times 10^{0}$ $\pm$  $5.68 \times 10^{-1}$ } &\makecell{ $1.80 \times 10^{1}$ $\pm$  $1.06 \times 10^{0}$ } &\makecell{ $9.47 \times 10^{1}$ $\pm$  $4.85 \times 10^{1}$ } &\makecell{ $2.25 \times 10^{2}$ $\pm$  $1.42 \times 10^{1}$ } \\ \hline
F.16 &\makecell{ $1.03 \times 10^{0}$ $\pm$  $7.47 \times 10^{-1}$ } &\makecell{ $1.66 \times 10^{0}$ $\pm$  $5.41 \times 10^{-1}$ } &\makecell{ $9.40 \times 10^{-1}$ $\pm$  $6.44 \times 10^{-1}$ } &\makecell{ $2.33 \times 10^{0}$ $\pm$  $2.13 \times 10^{-1}$ } \\ \hline
F.17 &\makecell{ $6.98 \times 10^{-1}$ $\pm$  $3.27 \times 10^{-1}$ } &\makecell{ $3.71 \times 10^{0}$ $\pm$  $8.11 \times 10^{-1}$ } &\makecell{ $1.39 \times 10^{1}$ $\pm$  $9.53 \times 10^{0}$ } &\makecell{ $4.18 \times 10^{1}$ $\pm$  $6.45 \times 10^{0}$ } \\ \hline
F.18 &\makecell{ $9.29 \times 10^{-5}$ $\pm$  $3.06 \times 10^{-5}$ } &\makecell{ $3.20 \times 10^{-4}$ $\pm$  $1.19 \times 10^{-4}$ } &\makecell{ $4.08 \times 10^{-3}$ $\pm$  $4.96 \times 10^{-3}$ } &\makecell{ $1.20 \times 10^{-2}$ $\pm$  $2.10 \times 10^{-3}$ } \\ \hline
F.19 &\makecell{ $2.85 \times 10^{-2}$ $\pm$  $3.35 \times 10^{-3}$ } &\makecell{ $1.42 \times 10^{-1}$ $\pm$  $5.50 \times 10^{-3}$ } &\makecell{ $4.10 \times 10^{-2}$ $\pm$  $6.58 \times 10^{-3}$ } &\makecell{ $5.06 \times 10^{-1}$ $\pm$  $9.98 \times 10^{-2}$ } \\ \hline
F.20 &\makecell{ $9.93 \times 10^{-5}$ $\pm$  $2.70 \times 10^{-5}$ } &\makecell{ $1.18 \times 10^{-4}$ $\pm$  $2.03 \times 10^{-5}$ } &\makecell{ $6.98 \times 10^{-5}$ $\pm$  $3.37 \times 10^{-5}$ } &\makecell{ $8.50 \times 10^{-5}$ $\pm$  $8.81 \times 10^{-6}$ } \\ \hline
F.21 &\makecell{ $1.59 \times 10^{-4}$ $\pm$  $1.08 \times 10^{-4}$ } &\makecell{ $3.90 \times 10^{-4}$ $\pm$  $5.99 \times 10^{-5}$ } &\makecell{ $2.43 \times 10^{-4}$ $\pm$  $2.52 \times 10^{-4}$ } &\makecell{ $1.37 \times 10^{-2}$ $\pm$  $7.85 \times 10^{-4}$ } \\ \hline
F.22 &\makecell{ $4.89 \times 10^{-3}$ $\pm$  $2.59 \times 10^{-3}$ } &\makecell{ $6.61 \times 10^{-3}$ $\pm$  $1.97 \times 10^{-3}$ } &\makecell{ $2.18 \times 10^{-2}$ $\pm$  $1.05 \times 10^{-2}$ } &\makecell{ $1.89 \times 10^{-1}$ $\pm$  $1.87 \times 10^{-2}$ } \\ \hline
F.23 &\makecell{ $1.24 \times 10^{-1}$ $\pm$  $1.02 \times 10^{-1}$ } &\makecell{ $3.98 \times 10^{-1}$ $\pm$  $7.39 \times 10^{-2}$ } &\makecell{ $1.11 \times 10^{0}$ $\pm$  $7.73 \times 10^{-1}$ } &\makecell{ $4.58 \times 10^{1}$ $\pm$  $2.09 \times 10^{0}$ } \\ \hline
F.24 &\makecell{ $7.10 \times 10^{0}$ $\pm$  $2.92 \times 10^{0}$ } &\makecell{ $1.17 \times 10^{1}$ $\pm$  $7.65 \times 10^{-1}$ } &\makecell{ $1.02 \times 10^{2}$ $\pm$  $4.24 \times 10^{1}$ } &\makecell{ $1.31 \times 10^{2}$ $\pm$  $3.35 \times 10^{0}$ } \\ \hline
F.25 &\makecell{ $3.08 \times 10^{-2}$ $\pm$  $9.13 \times 10^{-3}$ } &\makecell{ $5.17 \times 10^{-2}$ $\pm$  $1.40 \times 10^{-3}$ } &\makecell{ $4.99 \times 10^{0}$ $\pm$  $6.20 \times 10^{0}$ } &\makecell{ $2.45 \times 10^{1}$ $\pm$  $1.98 \times 10^{0}$ } \\ \hline
F.26 &\makecell{ $2.66 \times 10^{-1}$ $\pm$  $1.64 \times 10^{-1}$ } &\makecell{ $6.78 \times 10^{0}$ $\pm$  $2.06 \times 10^{0}$ } &\makecell{ $8.10 \times 10^{-1}$ $\pm$  $3.09 \times 10^{-1}$ } &\makecell{ $1.33 \times 10^{1}$ $\pm$  $5.73 \times 10^{-1}$ } \\ \hline
\end{tabular}
    }}
\end{table}
%------------------------------------------------------------------------------------

%-----------------------------------------

%==============================================================
\newpage
\subsection{Ablation Studies of {\our} on MNIST Dataset}\label{subsec:appendix_mnist_ablation_studies}

\noindent \textbf{Descriptions}: In this section, we provide the performance of {\our} with different component function combinations on the MNIST dataset. We enumerate different combinations of the expansion, reconciliation and remainder functions studied in the previous Figures~\ref{fig:mnist_ablation_studies}-\ref{fig:mnist_ablation_studies_individual_component_functions} in Section~\ref{subsec:discrete_classification}. For each of these combinations, we provide the testing accuracy, model parameter number and time cost of training {\our}. In addition, since {\our} allows the optional pre-processing and post-processing functions to the expansion function, we also investigate the impacts of such optional functions in the table, where the layer-norm is analyzed as the default function.

Specifically, the following Tables~\ref{tab:mnist_comparison_identity}-\ref{tab:mnist_comparison_normal_combinatorial} will be organized according to the data expansion functions used in composing {\our}. Each table presents the results learned for each individual expansion function combined with different reconciliation and remainder functions. The reported results in these tables have also been summarized in the previous Figures~\ref{fig:mnist_ablation_studies}-\ref{fig:mnist_ablation_studies_individual_component_functions} already.

\newpage
%-----------------------------------------
%---------------------------------------------------------------------------------------------------------
{\tiny
\begin{longtable}{|c|c|c|c|c|c|c|c|}
    \caption{Performance analysis of {\our} with the identity expansion function and different reconciliation and remainder functions on the MNIST dataset.}\label{tab:mnist_comparison_identity}\\
    	\hline
    	\textbf{Expansion} & \textbf{Reconciliation} & \textbf{Remainder} & \textbf{Parameter\#} & \textbf{Pre-Layer-Norm} & \textbf{Post-Layer-Norm} & \textbf{Time Cost} & \textbf{Accuracy}\\
    	\hline
    	\hline
%------------------------------------------------------------------------------------------------------------------------------------------
\multirow{56}{*}{identity}
%------------------------------------------------------------------------------------------------------------------------------------------
&\multirow{8}{*}{identity} 	 &\multirow{4}{*}{zero} 	 &\multirow{4}{*}{50816} 	 &False	 &False	 & $1.44 \times 10^{3}$	 & $9.27 \times 10^{-1}$ \\ 	 \cline{5-8}
&	 & 	 & 	 &True 	 &False	 & $1.37 \times 10^{3}$	 & $9.30 \times 10^{-1}$ \\ 	 \cline{5-8}
&	 & 	 & 	 &False 	 &True	 & $1.39 \times 10^{3}$	 & $9.30 \times 10^{-1}$ \\ 	 \cline{5-8}
&	 & 	 & 	 &True 	 &True	 & $9.19 \times 10^{2}$	 & $9.30 \times 10^{-1}$ \\ 	 \cline{3-8}
& 	 &\multirow{4}{*}{linear} 	 &\multirow{4}{*}{101632} 	 &False	 &False	 & $7.66 \times 10^{2}$	 & $9.77 \times 10^{-1}$ \\ 	 \cline{5-8}
&	 & 	 & 	 &True 	 &False	 & $7.72 \times 10^{2}$	 & $9.76 \times 10^{-1}$ \\
\hline
\end{longtable}}
\newpage
%---------------------------------------------------------------------------------------------------------
{\tiny
\begin{longtable}{|c|c|c|c|c|c|c|c|}
    \caption{Performance analysis of {\our} with the reciprocal expansion function and different reconciliation and remainder functions on the MNIST dataset.}\label{tab:mnist_comparison_reciprocal}\\
    	\hline
    	\textbf{Expansion} & \textbf{Reconciliation} & \textbf{Remainder} & \textbf{Parameter\#} & \textbf{Pre-Layer-Norm} & \textbf{Post-Layer-Norm} & \textbf{Time Cost} & \textbf{Accuracy}\\
    	\hline
    	\hline
%------------------------------------------------------------------------------------------------------------------------------------------
\multirow{56}{*}{reciprocal}
%------------------------------------------------------------------------------------------------------------------------------------------
&\multirow{8}{*}{identity} 	 &\multirow{4}{*}{zero} 	 &\multirow{4}{*}{50816} 	 &False	 &False	 & $1.51 \times 10^{3}$	 & $2.51 \times 10^{-1}$ \\ 	 \cline{5-8}
&	 & 	 & 	 &True 	 &False	 & $1.62 \times 10^{3}$	 & $1.73 \times 10^{-1}$ \\ 	 \cline{5-8}
&	 & 	 & 	 &False 	 &True	 & $1.56 \times 10^{3}$	 & $1.24 \times 10^{-1}$ \\ 	 \cline{5-8}
&	 & 	 & 	 &True 	 &True	 & $1.65 \times 10^{3}$	 & $1.30 \times 10^{-1}$ \\ 	 \cline{3-8}
& 	 &\multirow{4}{*}{linear} 	 &\multirow{4}{*}{101632} 	 &False	 &False	 & $1.58 \times 10^{3}$	 & $6.59 \times 10^{-1}$ \\ 	 \cline{5-8}
&	 & 	 & 	 &True 	 &False	 & $1.59 \times 10^{3}$	 & $3.09 \times 10^{-1}$ \\
\hline
\end{longtable}}
\newpage
%---------------------------------------------------------------------------------------------------------
{\tiny
\begin{longtable}{|c|c|c|c|c|c|c|c|}
    \caption{Performance analysis of {\our} with the taylor expansion function and different reconciliation and remainder functions on the MNIST dataset.}\label{tab:mnist_comparison_taylor}\\
    	\hline
    	\textbf{Expansion} & \textbf{Reconciliation} & \textbf{Remainder} & \textbf{Parameter\#} & \textbf{Pre-Layer-Norm} & \textbf{Post-Layer-Norm} & \textbf{Time Cost} & \textbf{Accuracy}\\
    	\hline
    	\hline
%------------------------------------------------------------------------------------------------------------------------------------------
\multirow{56}{*}{taylor}
%------------------------------------------------------------------------------------------------------------------------------------------
&\multirow{8}{*}{identity} 	 &\multirow{4}{*}{zero} 	 &\multirow{4}{*}{39429760} 	 &False	 &False	 & $2.28 \times 10^{3}$	 & $9.83 \times 10^{-1}$ \\ 	 \cline{5-8}
&	 & 	 & 	 &True 	 &False	 & $2.35 \times 10^{3}$	 & $9.85 \times 10^{-1}$ \\ 	 \cline{5-8}
&	 & 	 & 	 &False 	 &True	 & $2.54 \times 10^{3}$	 & $9.85 \times 10^{-1}$ \\ 	 \cline{5-8}
&	 & 	 & 	 &True 	 &True	 & $2.57 \times 10^{3}$	 & $9.85 \times 10^{-1}$ \\ 	 \cline{3-8}
& 	 &\multirow{4}{*}{linear} 	 &\multirow{4}{*}{39480576} 	 &False	 &False	 & $2.26 \times 10^{3}$	 & $9.81 \times 10^{-1}$ \\ 	 \cline{5-8}
&	 & 	 & 	 &True 	 &False	 & $2.26 \times 10^{3}$	 & $9.81 \times 10^{-1}$ \\
\hline
\end{longtable}}
\newpage
%---------------------------------------------------------------------------------------------------------
{\tiny
\begin{longtable}{|c|c|c|c|c|c|c|c|}
    \caption{Performance analysis of {\our} with the fourier expansion function and different reconciliation and remainder functions on the MNIST dataset.}\label{tab:mnist_comparison_fourier}\\
    	\hline
    	\textbf{Expansion} & \textbf{Reconciliation} & \textbf{Remainder} & \textbf{Parameter\#} & \textbf{Pre-Layer-Norm} & \textbf{Post-Layer-Norm} & \textbf{Time Cost} & \textbf{Accuracy}\\
    	\hline
    	\hline
%------------------------------------------------------------------------------------------------------------------------------------------
\multirow{56}{*}{fourier}
%------------------------------------------------------------------------------------------------------------------------------------------
&\multirow{8}{*}{identity} 	 &\multirow{4}{*}{zero} 	 &\multirow{4}{*}{508160} 	 &False	 &False	 & $1.85 \times 10^{3}$	 & $1.06 \times 10^{-1}$ \\ 	 \cline{5-8}
&	 & 	 & 	 &True 	 &False	 & $1.85 \times 10^{3}$	 & $8.95 \times 10^{-1}$ \\ 	 \cline{5-8}
&	 & 	 & 	 &False 	 &True	 & $1.74 \times 10^{3}$	 & $1.07 \times 10^{-1}$ \\ 	 \cline{5-8}
&	 & 	 & 	 &True 	 &True	 & $1.76 \times 10^{3}$	 & $8.97 \times 10^{-1}$ \\ 	 \cline{3-8}
& 	 &\multirow{4}{*}{linear} 	 &\multirow{4}{*}{558976} 	 &False	 &False	 & $1.71 \times 10^{3}$	 & $1.05 \times 10^{-1}$ \\ 	 \cline{5-8}
&	 & 	 & 	 &True 	 &False	 & $1.72 \times 10^{3}$	 & $9.47 \times 10^{-1}$ \\
\hline
\end{longtable}}
\newpage
%---------------------------------------------------------------------------------------------------------
{\tiny
\begin{longtable}{|c|c|c|c|c|c|c|c|}
    \caption{Performance analysis of {\our} with the hyperbolic expansion function and different reconciliation and remainder functions on the MNIST dataset.}\label{tab:mnist_comparison_hyperbolic}\\
    	\hline
    	\textbf{Expansion} & \textbf{Reconciliation} & \textbf{Remainder} & \textbf{Parameter\#} & \textbf{Pre-Layer-Norm} & \textbf{Post-Layer-Norm} & \textbf{Time Cost} & \textbf{Accuracy}\\
    	\hline
    	\hline
%------------------------------------------------------------------------------------------------------------------------------------------
\multirow{56}{*}{hyperbolic}
%------------------------------------------------------------------------------------------------------------------------------------------
&\multirow{8}{*}{identity} 	 &\multirow{4}{*}{zero} 	 &\multirow{4}{*}{152448} 	 &False	 &False	 & $1.86 \times 10^{3}$	 & $2.55 \times 10^{-1}$ \\ 	 \cline{5-8}
&	 & 	 & 	 &True 	 &False	 & $1.88 \times 10^{3}$	 & $9.72 \times 10^{-1}$ \\ 	 \cline{5-8}
&	 & 	 & 	 &False 	 &True	 & $1.85 \times 10^{3}$	 & $9.39 \times 10^{-1}$ \\ 	 \cline{5-8}
&	 & 	 & 	 &True 	 &True	 & $1.86 \times 10^{3}$	 & $9.75 \times 10^{-1}$ \\ 	 \cline{3-8}
& 	 &\multirow{4}{*}{linear} 	 &\multirow{4}{*}{203264} 	 &False	 &False	 & $1.74 \times 10^{3}$	 & $3.11 \times 10^{-1}$ \\ 	 \cline{5-8}
&	 & 	 & 	 &True 	 &False	 & $1.74 \times 10^{3}$	 & $9.73 \times 10^{-1}$ \\
\hline
\end{longtable}}
\newpage
%---------------------------------------------------------------------------------------------------------
{\tiny
\begin{longtable}{|c|c|c|c|c|c|c|c|}
    \caption{Performance analysis of {\our} with the arc-hyperbolic expansion function and different reconciliation and remainder functions on the MNIST dataset.}\label{tab:mnist_comparison_arc_hyperbolic}\\
    	\hline
    	\textbf{Expansion} & \textbf{Reconciliation} & \textbf{Remainder} & \textbf{Parameter\#} & \textbf{Pre-Layer-Norm} & \textbf{Post-Layer-Norm} & \textbf{Time Cost} & \textbf{Accuracy}\\
    	\hline
    	\hline
%------------------------------------------------------------------------------------------------------------------------------------------
\multirow{56}{*}{arc-hyperbolic}
%------------------------------------------------------------------------------------------------------------------------------------------
&\multirow{8}{*}{identity} 	 &\multirow{4}{*}{zero} 	 &\multirow{4}{*}{152448} 	 &False	 &False	 & $1.44 \times 10^{3}$	 & $9.72 \times 10^{-1}$ \\ 	 \cline{5-8}
&	 & 	 & 	 &True 	 &False	 & $1.59 \times 10^{3}$	 & $9.61 \times 10^{-1}$ \\ 	 \cline{5-8}
&	 & 	 & 	 &False 	 &True	 & $1.51 \times 10^{3}$	 & $9.74 \times 10^{-1}$ \\ 	 \cline{5-8}
&	 & 	 & 	 &True 	 &True	 & $1.63 \times 10^{3}$	 & $9.76 \times 10^{-1}$ \\ 	 \cline{3-8}
& 	 &\multirow{4}{*}{linear} 	 &\multirow{4}{*}{203264} 	 &False	 &False	 & $1.59 \times 10^{3}$	 & $9.70 \times 10^{-1}$ \\ 	 \cline{5-8}
&	 & 	 & 	 &True 	 &False	 & $1.61 \times 10^{3}$	 & $9.79 \times 10^{-1}$ \\
\hline
\end{longtable}}
\newpage
%---------------------------------------------------------------------------------------------------------
{\tiny
\begin{longtable}{|c|c|c|c|c|c|c|c|}
    \caption{Performance analysis of {\our} with the trigonometric expansion function and different reconciliation and remainder functions on the MNIST dataset.}\label{tab:mnist_comparison_trigonometric}\\
    	\hline
    	\textbf{Expansion} & \textbf{Reconciliation} & \textbf{Remainder} & \textbf{Parameter\#} & \textbf{Pre-Layer-Norm} & \textbf{Post-Layer-Norm} & \textbf{Time Cost} & \textbf{Accuracy}\\
    	\hline
    	\hline
%------------------------------------------------------------------------------------------------------------------------------------------
\multirow{56}{*}{trigonometric}
%------------------------------------------------------------------------------------------------------------------------------------------
&\multirow{8}{*}{identity} 	 &\multirow{4}{*}{zero} 	 &\multirow{4}{*}{152448} 	 &False	 &False	 & $1.80 \times 10^{3}$	 & $2.42 \times 10^{-1}$ \\ 	 \cline{5-8}
&	 & 	 & 	 &True 	 &False	 & $1.86 \times 10^{3}$	 & $4.87 \times 10^{-1}$ \\ 	 \cline{5-8}
&	 & 	 & 	 &False 	 &True	 & $1.84 \times 10^{3}$	 & $1.08 \times 10^{-1}$ \\ 	 \cline{5-8}
&	 & 	 & 	 &True 	 &True	 & $1.89 \times 10^{3}$	 & $8.89 \times 10^{-1}$ \\ 	 \cline{3-8}
& 	 &\multirow{4}{*}{linear} 	 &\multirow{4}{*}{203264} 	 &False	 &False	 & $1.78 \times 10^{3}$	 & $1.85 \times 10^{-1}$ \\ 	 \cline{5-8}
&	 & 	 & 	 &True 	 &False	 & $1.80 \times 10^{3}$	 & $5.95 \times 10^{-1}$ \\
\hline
\end{longtable}}
\newpage
%---------------------------------------------------------------------------------------------------------
{\tiny
\begin{longtable}{|c|c|c|c|c|c|c|c|}
    \caption{Performance analysis of {\our} with the arc-trigonometric expansion function and different reconciliation and remainder functions on the MNIST dataset.}\label{tab:mnist_comparison_arc_trigonometric}\\
    	\hline
    	\textbf{Expansion} & \textbf{Reconciliation} & \textbf{Remainder} & \textbf{Parameter\#} & \textbf{Pre-Layer-Norm} & \textbf{Post-Layer-Norm} & \textbf{Time Cost} & \textbf{Accuracy}\\
    	\hline
    	\hline
%------------------------------------------------------------------------------------------------------------------------------------------
\multirow{56}{*}{arc-trigonometric}
%------------------------------------------------------------------------------------------------------------------------------------------
&\multirow{8}{*}{identity} 	 &\multirow{4}{*}{zero} 	 &\multirow{4}{*}{152448} 	 &False	 &False	 & $8.83 \times 10^{2}$	 & $9.66 \times 10^{-1}$ \\ 	 \cline{5-8}
&	 & 	 & 	 &True 	 &False	 & $1.45 \times 10^{3}$	 & $9.52 \times 10^{-1}$ \\ 	 \cline{5-8}
&	 & 	 & 	 &False 	 &True	 & $1.45 \times 10^{3}$	 & $9.68 \times 10^{-1}$ \\ 	 \cline{5-8}
&	 & 	 & 	 &True 	 &True	 & $1.42 \times 10^{3}$	 & $9.72 \times 10^{-1}$ \\ 	 \cline{3-8}
& 	 &\multirow{4}{*}{linear} 	 &\multirow{4}{*}{203264} 	 &False	 &False	 & $8.00 \times 10^{2}$	 & $9.66 \times 10^{-1}$ \\ 	 \cline{5-8}
&	 & 	 & 	 &True 	 &False	 & $8.02 \times 10^{2}$	 & $9.78 \times 10^{-1}$ \\
\hline
\end{longtable}}
\newpage
%---------------------------------------------------------------------------------------------------------
{\tiny
\begin{longtable}{|c|c|c|c|c|c|c|c|}
    \caption{Performance analysis of {\our} with the bspline expansion function and different reconciliation and remainder functions on the MNIST dataset.}\label{tab:mnist_comparison_bspline}\\
    	\hline
    	\textbf{Expansion} & \textbf{Reconciliation} & \textbf{Remainder} & \textbf{Parameter\#} & \textbf{Pre-Layer-Norm} & \textbf{Post-Layer-Norm} & \textbf{Time Cost} & \textbf{Accuracy}\\
    	\hline
    	\hline
%------------------------------------------------------------------------------------------------------------------------------------------
\multirow{56}{*}{bspline}
%------------------------------------------------------------------------------------------------------------------------------------------
&\multirow{8}{*}{identity} 	 &\multirow{4}{*}{zero} 	 &\multirow{4}{*}{406528} 	 &False	 &False	 & $2.01 \times 10^{3}$	 & $9.28 \times 10^{-1}$ \\ 	 \cline{5-8}
&	 & 	 & 	 &True 	 &False	 & $2.05 \times 10^{3}$	 & $9.72 \times 10^{-1}$ \\ 	 \cline{5-8}
&	 & 	 & 	 &False 	 &True	 & $2.02 \times 10^{3}$	 & $9.80 \times 10^{-2}$ \\ 	 \cline{5-8}
&	 & 	 & 	 &True 	 &True	 & $2.07 \times 10^{3}$	 & $9.78 \times 10^{-1}$ \\ 	 \cline{3-8}
& 	 &\multirow{4}{*}{linear} 	 &\multirow{4}{*}{457344} 	 &False	 &False	 & $1.95 \times 10^{3}$	 & $9.64 \times 10^{-1}$ \\ 	 \cline{5-8}
&	 & 	 & 	 &True 	 &False	 & $1.94 \times 10^{3}$	 & $9.76 \times 10^{-1}$ \\
\hline
\end{longtable}}
\newpage
%---------------------------------------------------------------------------------------------------------
{\tiny
\begin{longtable}{|c|c|c|c|c|c|c|c|}
    \caption{Performance analysis of {\our} with the chebyshev expansion function and different reconciliation and remainder functions on the MNIST dataset.}\label{tab:mnist_comparison_chebyshev}\\
    	\hline
    	\textbf{Expansion} & \textbf{Reconciliation} & \textbf{Remainder} & \textbf{Parameter\#} & \textbf{Pre-Layer-Norm} & \textbf{Post-Layer-Norm} & \textbf{Time Cost} & \textbf{Accuracy}\\
    	\hline
    	\hline
%------------------------------------------------------------------------------------------------------------------------------------------
\multirow{56}{*}{chebyshev}
%------------------------------------------------------------------------------------------------------------------------------------------
&\multirow{8}{*}{identity} 	 &\multirow{4}{*}{zero} 	 &\multirow{4}{*}{254080} 	 &False	 &False	 & $3.68 \times 10^{3}$	 & $8.45 \times 10^{-1}$ \\ 	 \cline{5-8}
&	 & 	 & 	 &True 	 &False	 & $3.70 \times 10^{3}$	 & $9.68 \times 10^{-1}$ \\ 	 \cline{5-8}
&	 & 	 & 	 &False 	 &True	 & $3.69 \times 10^{3}$	 & $9.65 \times 10^{-1}$ \\ 	 \cline{5-8}
&	 & 	 & 	 &True 	 &True	 & $3.71 \times 10^{3}$	 & $9.68 \times 10^{-1}$ \\ 	 \cline{3-8}
& 	 &\multirow{4}{*}{linear} 	 &\multirow{4}{*}{304896} 	 &False	 &False	 & $3.50 \times 10^{3}$	 & $8.47 \times 10^{-1}$ \\ 	 \cline{5-8}
&	 & 	 & 	 &True 	 &False	 & $3.51 \times 10^{3}$	 & $9.67 \times 10^{-1}$ \\
\hline
\end{longtable}}
\newpage
%---------------------------------------------------------------------------------------------------------
{\tiny
\begin{longtable}{|c|c|c|c|c|c|c|c|}
    \caption{Performance analysis of {\our} with the jacobi expansion function and different reconciliation and remainder functions on the MNIST dataset.}\label{tab:mnist_comparison_jacobi}\\
    	\hline
    	\textbf{Expansion} & \textbf{Reconciliation} & \textbf{Remainder} & \textbf{Parameter\#} & \textbf{Pre-Layer-Norm} & \textbf{Post-Layer-Norm} & \textbf{Time Cost} & \textbf{Accuracy}\\
    	\hline
    	\hline
%------------------------------------------------------------------------------------------------------------------------------------------
\multirow{56}{*}{jacobi}
%------------------------------------------------------------------------------------------------------------------------------------------
&\multirow{8}{*}{identity} 	 &\multirow{4}{*}{zero} 	 &\multirow{4}{*}{254080} 	 &False	 &False	 & $2.00 \times 10^{3}$	 & $8.67 \times 10^{-1}$ \\ 	 \cline{5-8}
&	 & 	 & 	 &True 	 &False	 & $1.71 \times 10^{3}$	 & $9.70 \times 10^{-1}$ \\ 	 \cline{5-8}
&	 & 	 & 	 &False 	 &True	 & $1.87 \times 10^{3}$	 & $9.66 \times 10^{-1}$ \\ 	 \cline{5-8}
&	 & 	 & 	 &True 	 &True	 & $1.57 \times 10^{3}$	 & $9.70 \times 10^{-1}$ \\ 	 \cline{3-8}
& 	 &\multirow{4}{*}{linear} 	 &\multirow{4}{*}{304896} 	 &False	 &False	 & $1.40 \times 10^{3}$	 & $8.69 \times 10^{-1}$ \\ 	 \cline{5-8}
&	 & 	 & 	 &True 	 &False	 & $1.40 \times 10^{3}$	 & $9.67 \times 10^{-1}$ \\
\hline
\end{longtable}}
\newpage
%---------------------------------------------------------------------------------------------------------
{\tiny
\begin{longtable}{|c|c|c|c|c|c|c|c|}
    \caption{Performance analysis of {\our} with the gaussian-rbf expansion function and different reconciliation and remainder functions on the MNIST dataset.}\label{tab:mnist_comparison_gaussian_rbf}\\
    	\hline
    	\textbf{Expansion} & \textbf{Reconciliation} & \textbf{Remainder} & \textbf{Parameter\#} & \textbf{Pre-Layer-Norm} & \textbf{Post-Layer-Norm} & \textbf{Time Cost} & \textbf{Accuracy}\\
    	\hline
    	\hline
%------------------------------------------------------------------------------------------------------------------------------------------
\multirow{56}{*}{gaussian-rbf}
%------------------------------------------------------------------------------------------------------------------------------------------
&\multirow{8}{*}{identity} 	 &\multirow{4}{*}{zero} 	 &\multirow{4}{*}{508160} 	 &False	 &False	 & $8.74 \times 10^{2}$	 & $9.82 \times 10^{-2}$ \\ 	 \cline{5-8}
&	 & 	 & 	 &True 	 &False	 & $1.43 \times 10^{3}$	 & $9.79 \times 10^{-1}$ \\ 	 \cline{5-8}
&	 & 	 & 	 &False 	 &True	 & $1.45 \times 10^{3}$	 & $1.07 \times 10^{-1}$ \\ 	 \cline{5-8}
&	 & 	 & 	 &True 	 &True	 & $1.37 \times 10^{3}$	 & $9.81 \times 10^{-1}$ \\ 	 \cline{3-8}
& 	 &\multirow{4}{*}{linear} 	 &\multirow{4}{*}{558976} 	 &False	 &False	 & $7.90 \times 10^{2}$	 & $7.95 \times 10^{-1}$ \\ 	 \cline{5-8}
&	 & 	 & 	 &True 	 &False	 & $7.92 \times 10^{2}$	 & $9.82 \times 10^{-1}$ \\
\hline
\end{longtable}}
\newpage
%---------------------------------------------------------------------------------------------------------
{\tiny
\begin{longtable}{|c|c|c|c|c|c|c|c|}
    \caption{Performance analysis of {\our} with the inverse-quadratic-rbf expansion function and different reconciliation and remainder functions on the MNIST dataset.}\label{tab:mnist_comparison_inverse_quadratic_rbf}\\
    	\hline
    	\textbf{Expansion} & \textbf{Reconciliation} & \textbf{Remainder} & \textbf{Parameter\#} & \textbf{Pre-Layer-Norm} & \textbf{Post-Layer-Norm} & \textbf{Time Cost} & \textbf{Accuracy}\\
    	\hline
    	\hline
%------------------------------------------------------------------------------------------------------------------------------------------
\multirow{56}{*}{inverse-quadratic-rbf}
%------------------------------------------------------------------------------------------------------------------------------------------
&\multirow{8}{*}{identity} 	 &\multirow{4}{*}{zero} 	 &\multirow{4}{*}{508160} 	 &False	 &False	 & $1.46 \times 10^{3}$	 & $3.18 \times 10^{-1}$ \\ 	 \cline{5-8}
&	 & 	 & 	 &True 	 &False	 & $1.61 \times 10^{3}$	 & $9.81 \times 10^{-1}$ \\ 	 \cline{5-8}
&	 & 	 & 	 &False 	 &True	 & $1.55 \times 10^{3}$	 & $9.48 \times 10^{-1}$ \\ 	 \cline{5-8}
&	 & 	 & 	 &True 	 &True	 & $1.63 \times 10^{3}$	 & $9.80 \times 10^{-1}$ \\ 	 \cline{3-8}
& 	 &\multirow{4}{*}{linear} 	 &\multirow{4}{*}{558976} 	 &False	 &False	 & $1.60 \times 10^{3}$	 & $8.79 \times 10^{-1}$ \\ 	 \cline{5-8}
&	 & 	 & 	 &True 	 &False	 & $1.62 \times 10^{3}$	 & $9.80 \times 10^{-1}$ \\
\hline
\end{longtable}}
\newpage
%---------------------------------------------------------------------------------------------------------
{\tiny
\begin{longtable}{|c|c|c|c|c|c|c|c|}
    \caption{Performance analysis of {\our} with the normal expansion function and different reconciliation and remainder functions on the MNIST dataset.}\label{tab:mnist_comparison_normal}\\
    	\hline
    	\textbf{Expansion} & \textbf{Reconciliation} & \textbf{Remainder} & \textbf{Parameter\#} & \textbf{Pre-Layer-Norm} & \textbf{Post-Layer-Norm} & \textbf{Time Cost} & \textbf{Accuracy}\\
    	\hline
    	\hline
%------------------------------------------------------------------------------------------------------------------------------------------
\multirow{56}{*}{normal}
%------------------------------------------------------------------------------------------------------------------------------------------
&\multirow{8}{*}{identity} 	 &\multirow{4}{*}{zero} 	 &\multirow{4}{*}{304896} 	 &False	 &False	 & $1.17 \times 10^{3}$	 & $9.66 \times 10^{-1}$ \\ 	 \cline{5-8}
&	 & 	 & 	 &True 	 &False	 & $1.37 \times 10^{3}$	 & $1.14 \times 10^{-1}$ \\ 	 \cline{5-8}
&	 & 	 & 	 &False 	 &True	 & $1.40 \times 10^{3}$	 & $1.31 \times 10^{-1}$ \\ 	 \cline{5-8}
&	 & 	 & 	 &True 	 &True	 & $1.42 \times 10^{3}$	 & $8.73 \times 10^{-1}$ \\ 	 \cline{3-8}
& 	 &\multirow{4}{*}{linear} 	 &\multirow{4}{*}{355712} 	 &False	 &False	 & $1.12 \times 10^{3}$	 & $9.69 \times 10^{-1}$ \\ 	 \cline{5-8}
&	 & 	 & 	 &True 	 &False	 & $1.37 \times 10^{3}$	 & $3.42 \times 10^{-1}$ \\
\hline
\end{longtable}}
\newpage
%---------------------------------------------------------------------------------------------------------
{\tiny
\begin{longtable}{|c|c|c|c|c|c|c|c|}
    \caption{Performance analysis of {\our} with the cauchy expansion function and different reconciliation and remainder functions on the MNIST dataset.}\label{tab:mnist_comparison_cauchy}\\
    	\hline
    	\textbf{Expansion} & \textbf{Reconciliation} & \textbf{Remainder} & \textbf{Parameter\#} & \textbf{Pre-Layer-Norm} & \textbf{Post-Layer-Norm} & \textbf{Time Cost} & \textbf{Accuracy}\\
    	\hline
    	\hline
%------------------------------------------------------------------------------------------------------------------------------------------
\multirow{56}{*}{cauchy}
%------------------------------------------------------------------------------------------------------------------------------------------
&\multirow{8}{*}{identity} 	 &\multirow{4}{*}{zero} 	 &\multirow{4}{*}{304896} 	 &False	 &False	 & $8.42 \times 10^{4}$	 & $9.26 \times 10^{-1}$ \\ 	 \cline{5-8}
&	 & 	 & 	 &True 	 &False	 & $1.98 \times 10^{3}$	 & $1.14 \times 10^{-1}$ \\ 	 \cline{5-8}
&	 & 	 & 	 &False 	 &True	 & $1.91 \times 10^{3}$	 & $9.33 \times 10^{-1}$ \\ 	 \cline{5-8}
&	 & 	 & 	 &True 	 &True	 & $1.83 \times 10^{3}$	 & $9.45 \times 10^{-1}$ \\ 	 \cline{3-8}
& 	 &\multirow{4}{*}{linear} 	 &\multirow{4}{*}{355712} 	 &False	 &False	 & $6.79 \times 10^{3}$	 & $9.48 \times 10^{-1}$ \\ 	 \cline{5-8}
&	 & 	 & 	 &True 	 &False	 & $1.99 \times 10^{3}$	 & $9.16 \times 10^{-1}$ \\
\hline
\end{longtable}}
\newpage
%---------------------------------------------------------------------------------------------------------
{\tiny
\begin{longtable}{|c|c|c|c|c|c|c|c|}
    \caption{Performance analysis of {\our} with the chi2 expansion function and different reconciliation and remainder functions on the MNIST dataset.}\label{tab:mnist_comparison_chi2}\\
    	\hline
    	\textbf{Expansion} & \textbf{Reconciliation} & \textbf{Remainder} & \textbf{Parameter\#} & \textbf{Pre-Layer-Norm} & \textbf{Post-Layer-Norm} & \textbf{Time Cost} & \textbf{Accuracy}\\
    	\hline
    	\hline
%------------------------------------------------------------------------------------------------------------------------------------------
\multirow{56}{*}{chi2}
%------------------------------------------------------------------------------------------------------------------------------------------
&\multirow{8}{*}{identity} 	 &\multirow{4}{*}{zero} 	 &\multirow{4}{*}{101632} 	 &False	 &False	 & $1.36 \times 10^{3}$	 & $9.72 \times 10^{-1}$ \\ 	 \cline{5-8}
&	 & 	 & 	 &True 	 &False	 & $1.32 \times 10^{3}$	 & $9.57 \times 10^{-1}$ \\ 	 \cline{5-8}
&	 & 	 & 	 &False 	 &True	 & $1.31 \times 10^{3}$	 & $9.75 \times 10^{-1}$ \\ 	 \cline{5-8}
&	 & 	 & 	 &True 	 &True	 & $1.28 \times 10^{3}$	 & $9.75 \times 10^{-1}$ \\ 	 \cline{3-8}
& 	 &\multirow{4}{*}{linear} 	 &\multirow{4}{*}{152448} 	 &False	 &False	 & $1.26 \times 10^{3}$	 & $9.74 \times 10^{-1}$ \\ 	 \cline{5-8}
&	 & 	 & 	 &True 	 &False	 & $1.30 \times 10^{3}$	 & $9.72 \times 10^{-1}$ \\
\hline
\end{longtable}}
\newpage
%---------------------------------------------------------------------------------------------------------
{\tiny
\begin{longtable}{|c|c|c|c|c|c|c|c|}
    \caption{Performance analysis of {\our} with the gamma expansion function and different reconciliation and remainder functions on the MNIST dataset.}\label{tab:mnist_comparison_gamma}\\
    	\hline
    	\textbf{Expansion} & \textbf{Reconciliation} & \textbf{Remainder} & \textbf{Parameter\#} & \textbf{Pre-Layer-Norm} & \textbf{Post-Layer-Norm} & \textbf{Time Cost} & \textbf{Accuracy}\\
    	\hline
    	\hline
%------------------------------------------------------------------------------------------------------------------------------------------
\multirow{56}{*}{gamma}
%------------------------------------------------------------------------------------------------------------------------------------------
&\multirow{8}{*}{identity} 	 &\multirow{4}{*}{zero} 	 &\multirow{4}{*}{304896} 	 &False	 &False	 & $1.58 \times 10^{3}$	 & $9.55 \times 10^{-1}$ \\ 	 \cline{5-8}
&	 & 	 & 	 &True 	 &False	 & $1.65 \times 10^{3}$	 & $9.32 \times 10^{-1}$ \\ 	 \cline{5-8}
&	 & 	 & 	 &False 	 &True	 & $1.73 \times 10^{3}$	 & $9.47 \times 10^{-1}$ \\ 	 \cline{5-8}
&	 & 	 & 	 &True 	 &True	 & $1.59 \times 10^{3}$	 & $9.14 \times 10^{-1}$ \\ 	 \cline{3-8}
& 	 &\multirow{4}{*}{linear} 	 &\multirow{4}{*}{355712} 	 &False	 &False	 & $1.71 \times 10^{3}$	 & $9.61 \times 10^{-1}$ \\ 	 \cline{5-8}
&	 & 	 & 	 &True 	 &False	 & $1.67 \times 10^{3}$	 & $9.45 \times 10^{-1}$ \\
\hline
\end{longtable}}
\newpage
%---------------------------------------------------------------------------------------------------------
{\tiny
\begin{longtable}{|c|c|c|c|c|c|c|c|}
    \caption{Performance analysis of {\our} with the laplace expansion function and different reconciliation and remainder functions on the MNIST dataset.}\label{tab:mnist_comparison_laplace}\\
    	\hline
    	\textbf{Expansion} & \textbf{Reconciliation} & \textbf{Remainder} & \textbf{Parameter\#} & \textbf{Pre-Layer-Norm} & \textbf{Post-Layer-Norm} & \textbf{Time Cost} & \textbf{Accuracy}\\
    	\hline
    	\hline
%------------------------------------------------------------------------------------------------------------------------------------------
\multirow{56}{*}{laplace}
%------------------------------------------------------------------------------------------------------------------------------------------
&\multirow{8}{*}{identity} 	 &\multirow{4}{*}{zero} 	 &\multirow{4}{*}{304896} 	 &False	 &False	 & $1.15 \times 10^{3}$	 & $9.65 \times 10^{-1}$ \\ 	 \cline{5-8}
&	 & 	 & 	 &True 	 &False	 & $7.45 \times 10^{4}$	 & $9.77 \times 10^{-1}$ \\ 	 \cline{5-8}
&	 & 	 & 	 &False 	 &True	 & $7.91 \times 10^{4}$	 & $9.77 \times 10^{-1}$ \\ 	 \cline{5-8}
&	 & 	 & 	 &True 	 &True	 & $7.94 \times 10^{4}$	 & $9.80 \times 10^{-1}$ \\ 	 \cline{3-8}
& 	 &\multirow{4}{*}{linear} 	 &\multirow{4}{*}{355712} 	 &False	 &False	 & $1.10 \times 10^{3}$	 & $9.68 \times 10^{-1}$ \\ 	 \cline{5-8}
&	 & 	 & 	 &True 	 &False	 & $1.51 \times 10^{3}$	 & $8.88 \times 10^{-1}$ \\
\hline
\end{longtable}}
\newpage
%---------------------------------------------------------------------------------------------------------
{\tiny
\begin{longtable}{|c|c|c|c|c|c|c|c|}
    \caption{Performance analysis of {\our} with the exponential expansion function and different reconciliation and remainder functions on the MNIST dataset.}\label{tab:mnist_comparison_exponential}\\
    	\hline
    	\textbf{Expansion} & \textbf{Reconciliation} & \textbf{Remainder} & \textbf{Parameter\#} & \textbf{Pre-Layer-Norm} & \textbf{Post-Layer-Norm} & \textbf{Time Cost} & \textbf{Accuracy}\\
    	\hline
    	\hline
%------------------------------------------------------------------------------------------------------------------------------------------
\multirow{56}{*}{exponential}
%------------------------------------------------------------------------------------------------------------------------------------------
&\multirow{8}{*}{identity} 	 &\multirow{4}{*}{zero} 	 &\multirow{4}{*}{304896} 	 &False	 &False	 & $5.36 \times 10^{4}$	 & $9.53 \times 10^{-1}$ \\ 	 \cline{5-8}
&	 & 	 & 	 &True 	 &False	 & $1.36 \times 10^{3}$	 & $9.54 \times 10^{-1}$ \\ 	 \cline{5-8}
&	 & 	 & 	 &False 	 &True	 & $1.30 \times 10^{3}$	 & $9.61 \times 10^{-1}$ \\ 	 \cline{5-8}
&	 & 	 & 	 &True 	 &True	 & $1.16 \times 10^{3}$	 & $9.54 \times 10^{-1}$ \\ 	 \cline{3-8}
& 	 &\multirow{4}{*}{linear} 	 &\multirow{4}{*}{355712} 	 &False	 &False	 & $3.28 \times 10^{4}$	 & $9.57 \times 10^{-1}$ \\ 	 \cline{5-8}
&	 & 	 & 	 &True 	 &False	 & $1.37 \times 10^{3}$	 & $9.56 \times 10^{-1}$ \\
\hline
\end{longtable}}
\newpage
%---------------------------------------------------------------------------------------------------------
{\tiny
\begin{longtable}{|c|c|c|c|c|c|c|c|}
    \caption{Performance analysis of {\our} with the hybrid-probabilistic expansion function and different reconciliation and remainder functions on the MNIST dataset.}\label{tab:mnist_comparison_hybrid_probabilistic}\\
    	\hline
    	\textbf{Expansion} & \textbf{Reconciliation} & \textbf{Remainder} & \textbf{Parameter\#} & \textbf{Pre-Layer-Norm} & \textbf{Post-Layer-Norm} & \textbf{Time Cost} & \textbf{Accuracy}\\
    	\hline
    	\hline
%------------------------------------------------------------------------------------------------------------------------------------------
\multirow{56}{*}{hybrid-probabilistic}
%------------------------------------------------------------------------------------------------------------------------------------------
&\multirow{8}{*}{identity} 	 &\multirow{4}{*}{zero} 	 &\multirow{4}{*}{304896} 	 &False	 &False	 & $1.47 \times 10^{3}$	 & $9.67 \times 10^{-1}$ \\ 	 \cline{5-8}
&	 & 	 & 	 &True 	 &False	 & $2.26 \times 10^{3}$	 & $8.08 \times 10^{-1}$ \\ 	 \cline{5-8}
&	 & 	 & 	 &False 	 &True	 & $2.17 \times 10^{3}$	 & $9.53 \times 10^{-1}$ \\ 	 \cline{5-8}
&	 & 	 & 	 &True 	 &True	 & $2.39 \times 10^{3}$	 & $9.42 \times 10^{-1}$ \\ 	 \cline{3-8}
& 	 &\multirow{4}{*}{linear} 	 &\multirow{4}{*}{355712} 	 &False	 &False	 & $1.61 \times 10^{3}$	 & $9.74 \times 10^{-1}$ \\ 	 \cline{5-8}
&	 & 	 & 	 &True 	 &False	 & $3.51 \times 10^{3}$	 & $9.29 \times 10^{-1}$ \\
\hline
\end{longtable}}
\newpage
%---------------------------------------------------------------------------------------------------------
{\tiny
\begin{longtable}{|c|c|c|c|c|c|c|c|}
    \caption{Performance analysis of {\our} with the normal-combinatorial expansion function and different reconciliation and remainder functions on the MNIST dataset.}\label{tab:mnist_comparison_normal_combinatorial}\\
    	\hline
    	\textbf{Expansion} & \textbf{Reconciliation} & \textbf{Remainder} & \textbf{Parameter\#} & \textbf{Pre-Layer-Norm} & \textbf{Post-Layer-Norm} & \textbf{Time Cost} & \textbf{Accuracy}\\
    	\hline
    	\hline
%------------------------------------------------------------------------------------------------------------------------------------------
\multirow{56}{*}{normal-combinatorial}
%------------------------------------------------------------------------------------------------------------------------------------------
&\multirow{8}{*}{identity} 	 &\multirow{4}{*}{zero} 	 &\multirow{4}{*}{19714880} 	 &False	 &False	 & $4.31 \times 10^{4}$	 & $9.49 \times 10^{-1}$ \\ 	 \cline{5-8}
&	 & 	 & 	 &True 	 &False	 & $4.30 \times 10^{4}$	 & $1.14 \times 10^{-1}$ \\ 	 \cline{5-8}
&	 & 	 & 	 &False 	 &True	 & $7.14 \times 10^{4}$	 & $1.00 \times 10^{-1}$ \\ 	 \cline{5-8}
&	 & 	 & 	 &True 	 &True	 & $7.13 \times 10^{4}$	 & $6.96 \times 10^{-1}$ \\ 	 \cline{3-8}
& 	 &\multirow{4}{*}{linear} 	 &\multirow{4}{*}{19765696} 	 &False	 &False	 & $4.30 \times 10^{4}$	 & $9.54 \times 10^{-1}$ \\ 	 \cline{5-8}
&	 & 	 & 	 &True 	 &False	 & $4.30 \times 10^{4}$	 & $1.14 \times 10^{-1}$ \\
\hline
\end{longtable}}

%-----------------------------------------
%==============================================================
\newpage
\subsection{Visualization of {\our} Data Expansion and Parameter Reconciliation on MNIST Dataset}\label{subsec:appendix_mnist_visualization_studies}

\noindent \textbf{Descriptions}: The following Figures~\ref{fig:mnist_visualization_appendix_1}-\ref{fig:mnist_visualization_appendix_9} present the visualizations of the expanded images with labels 1 through 9. Besides the image data, we also illustrate the learned parameters of {\our} corresponding to these expansions as well.

%-----------------------------------------
%------------------------------------------------------------------------------------------
\begin{figure}[h]
    \centering
    \begin{subfigure}{0.48\textwidth}
        \centering
        \includegraphics[width=\textwidth]{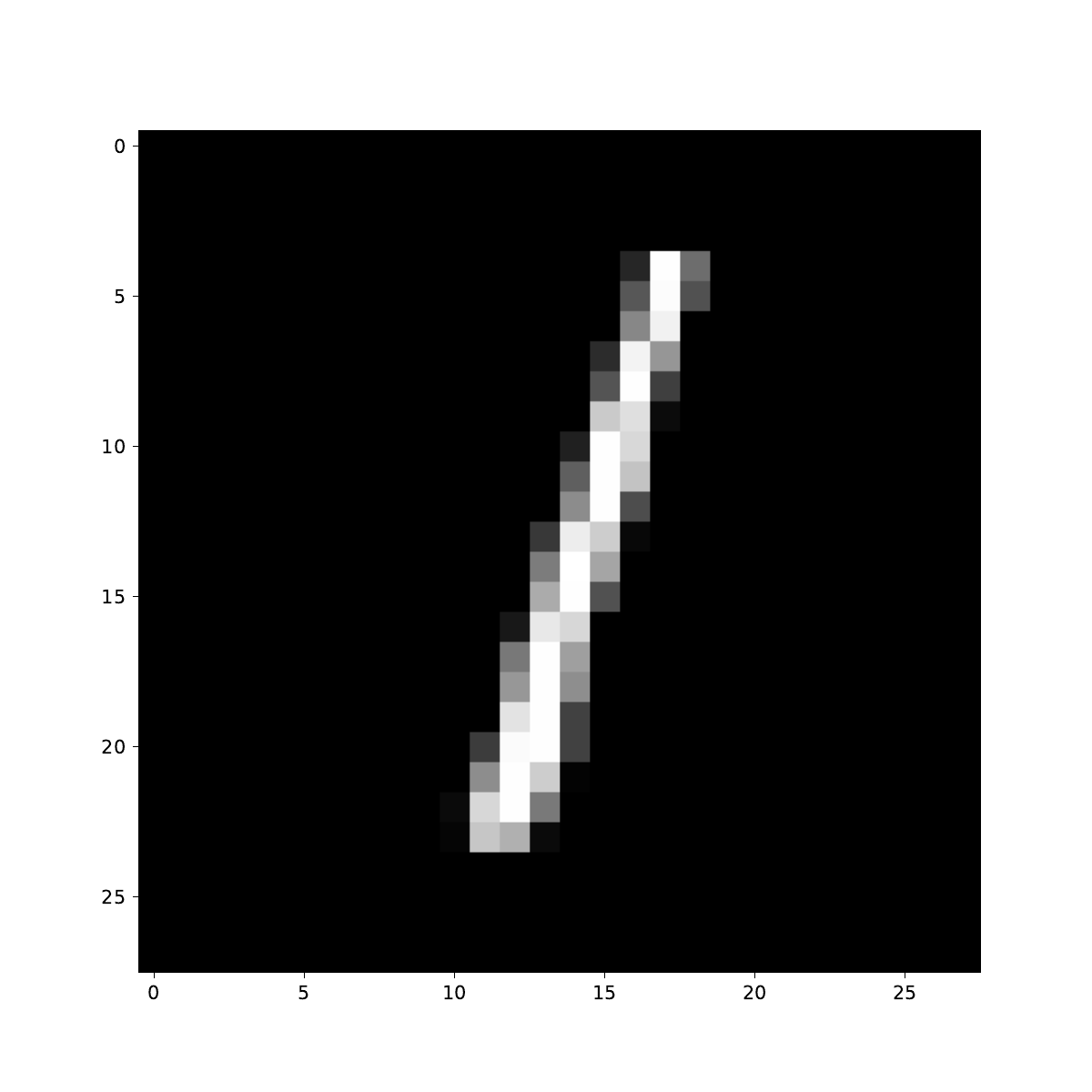}
        \caption{Input Raw Data}
    \end{subfigure}
    \hfill
    \begin{subfigure}{0.48\textwidth}
        \centering
        \includegraphics[width=\textwidth]{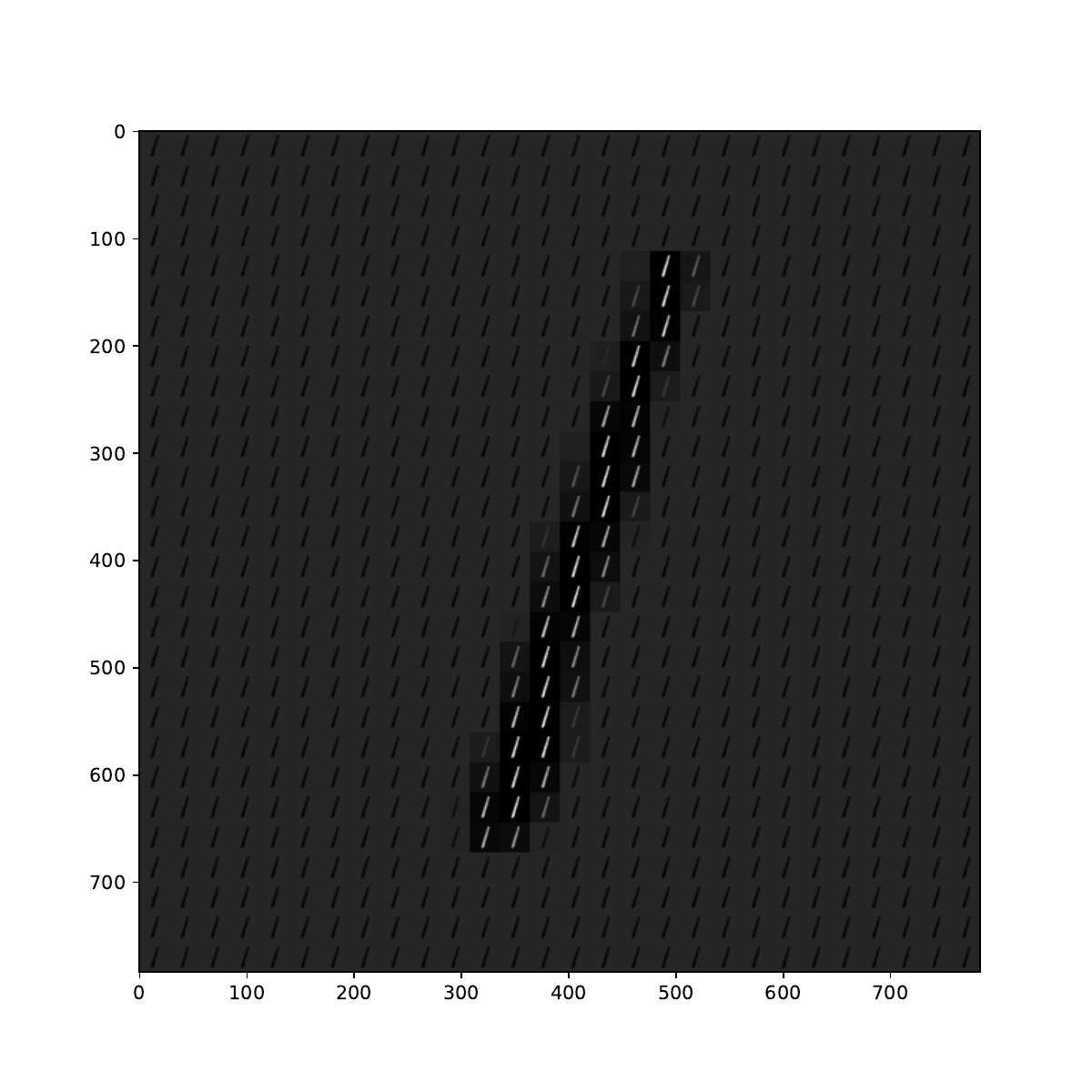}
        \caption{Expanded Data}
    \end{subfigure}
    \begin{subfigure}{0.48\textwidth}
        \centering
        \includegraphics[width=\textwidth]{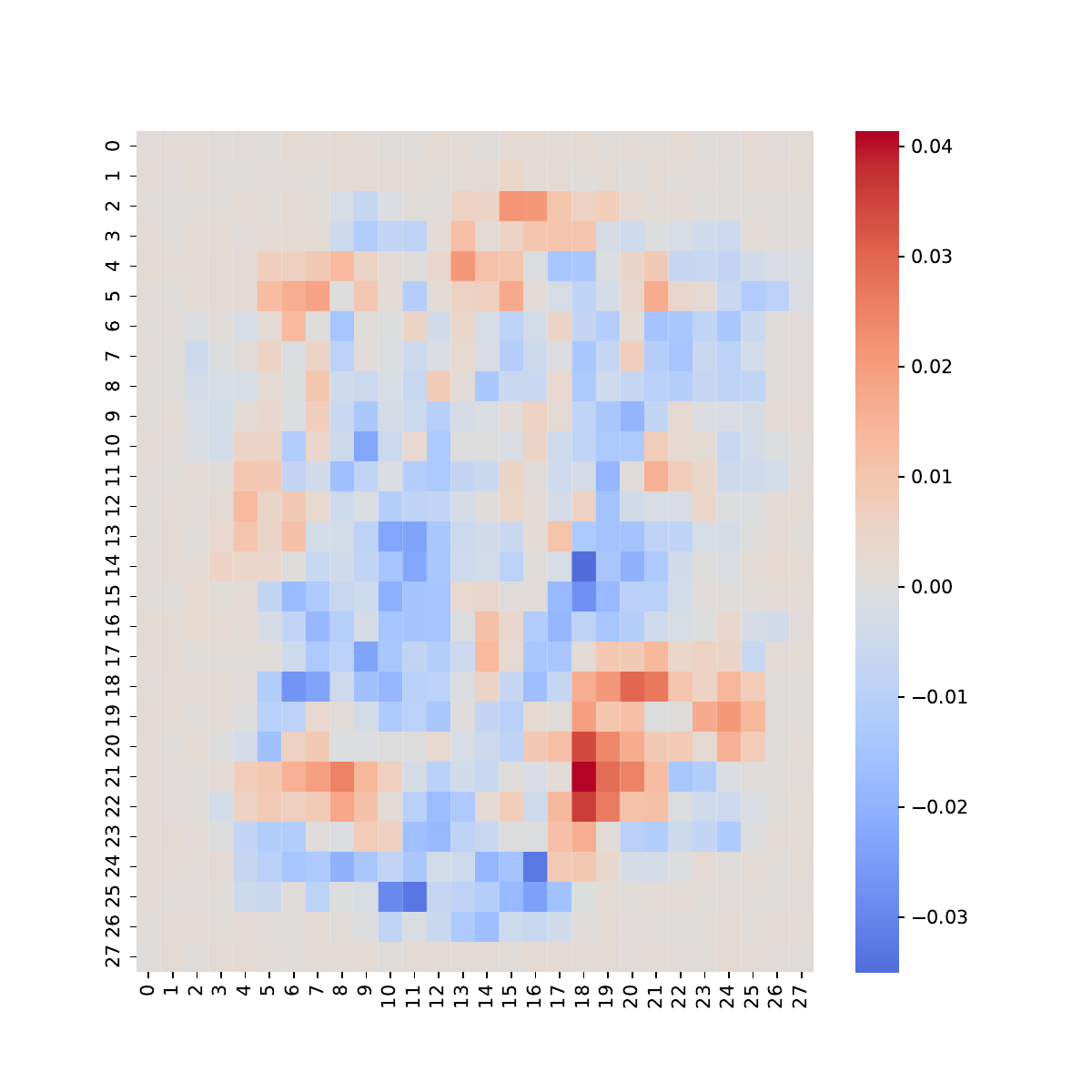}
        \caption{Parameters for Raw Data}
    \end{subfigure}
    \hfill
    \begin{subfigure}{0.48\textwidth}
        \centering
        \includegraphics[width=\textwidth]{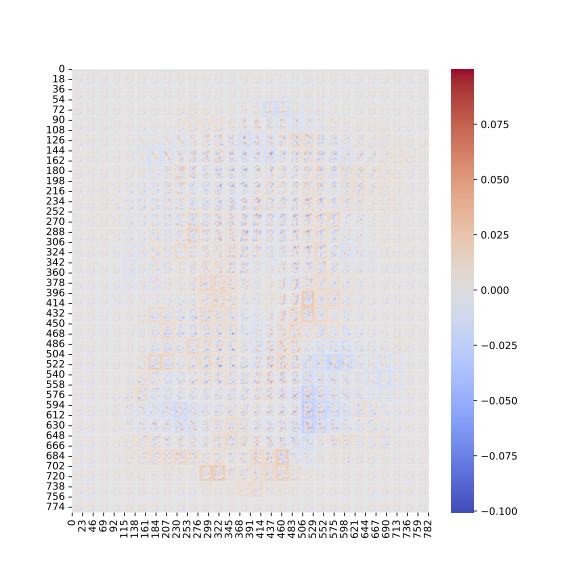}
        \caption{Parameters for Expanded Data}
    \end{subfigure}
    \caption{An illustration of an image with label $1$ randomly selected from the MNIST dataset. Plat (a): raw image data; Plot (b): expanded data; Plot (c): parameter corresponding to output neuron of label $1$ for raw image data; and Plot (d): parameter corresponding to output neuron of label $1$ for expanded image data.}
    \label{fig:mnist_visualization_appendix_1}
\end{figure}
%------------------------------------------------------------------------------------------

%------------------------------------------------------------------------------------------
\newpage
\begin{figure}[h]
    \centering
    \begin{subfigure}{0.48\textwidth}
        \centering
        \includegraphics[width=\textwidth]{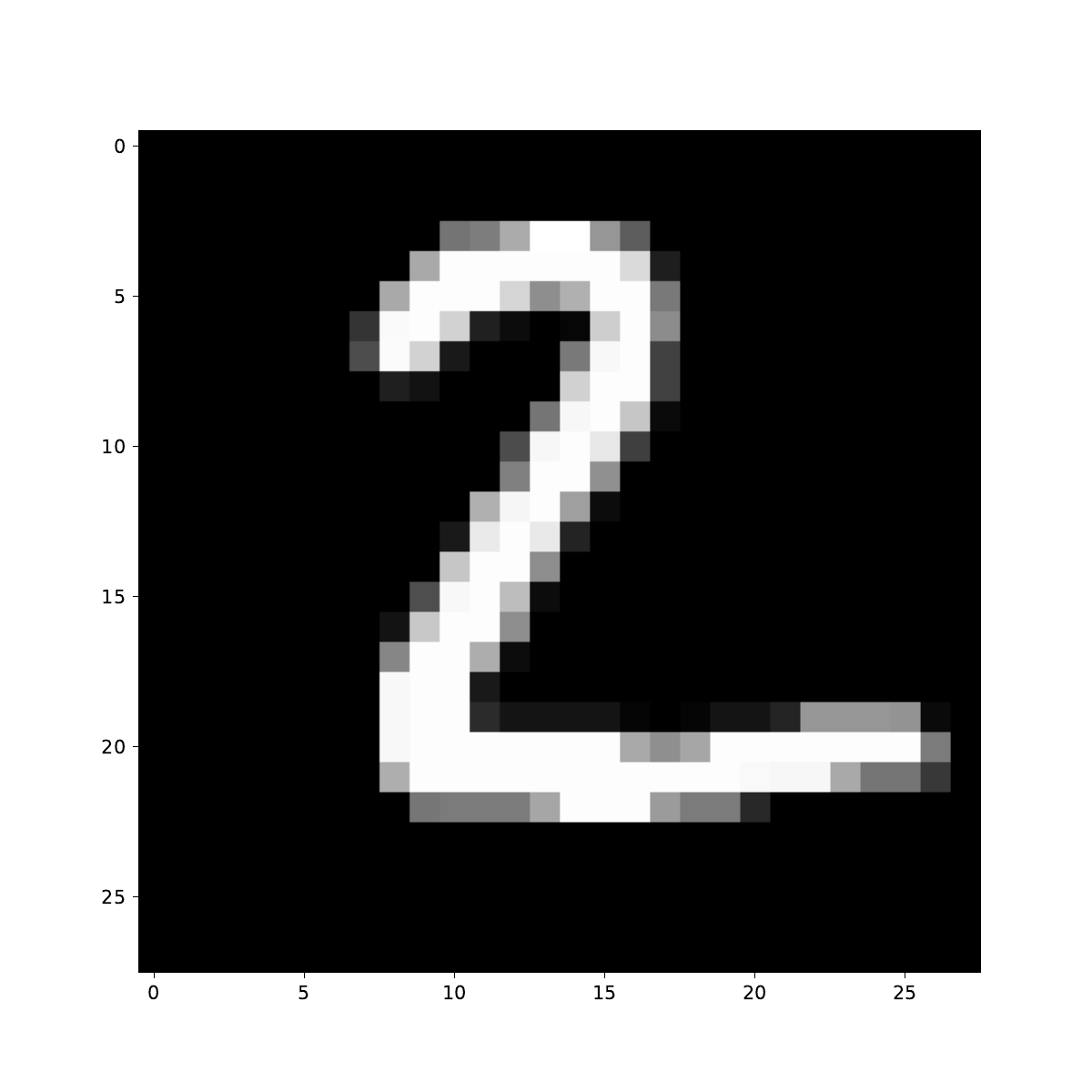}
        \caption{Input Raw Data}
    \end{subfigure}
    \hfill
    \begin{subfigure}{0.48\textwidth}
        \centering
        \includegraphics[width=\textwidth]{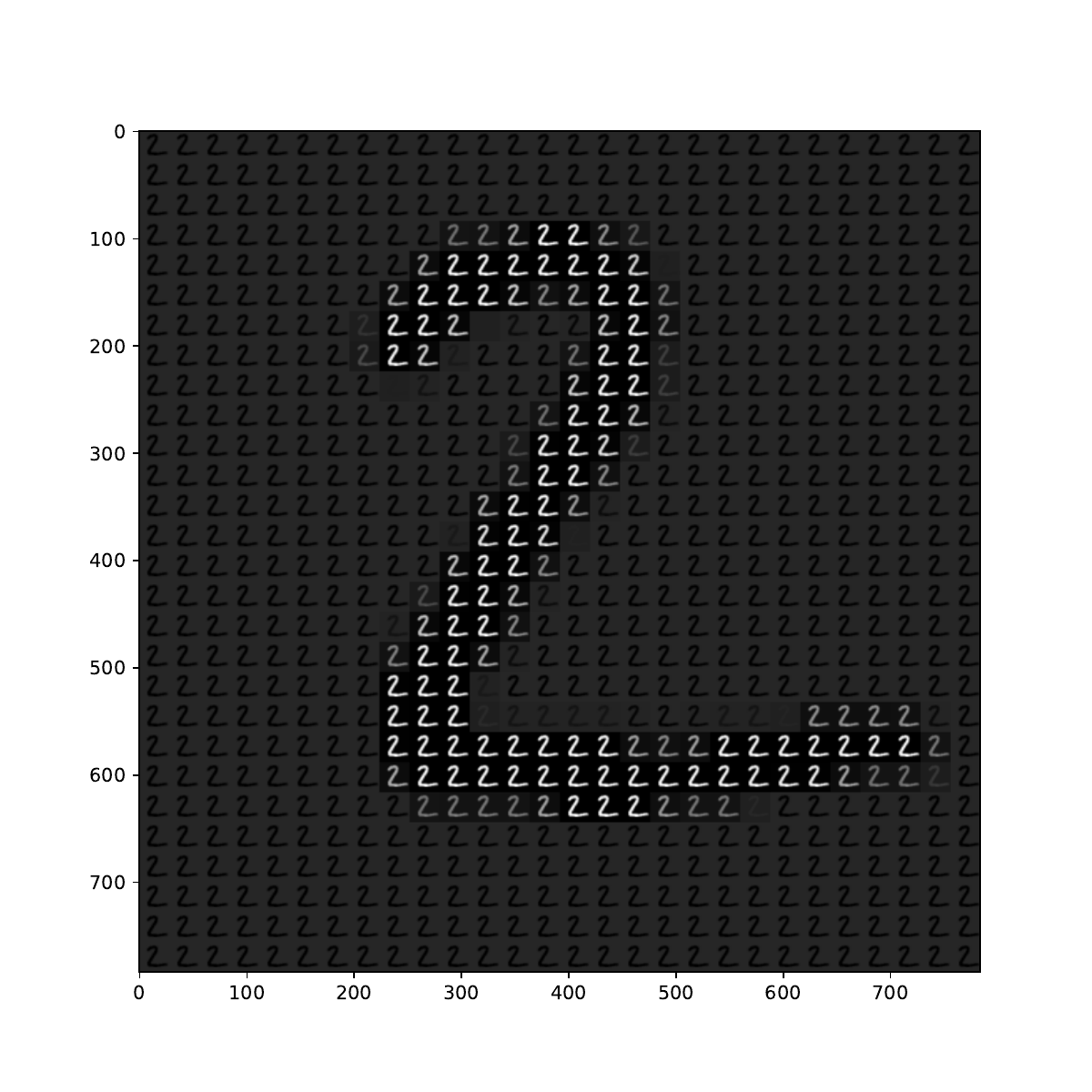}
        \caption{Expanded Data}
    \end{subfigure}
    \begin{subfigure}{0.48\textwidth}
        \centering
        \includegraphics[width=\textwidth]{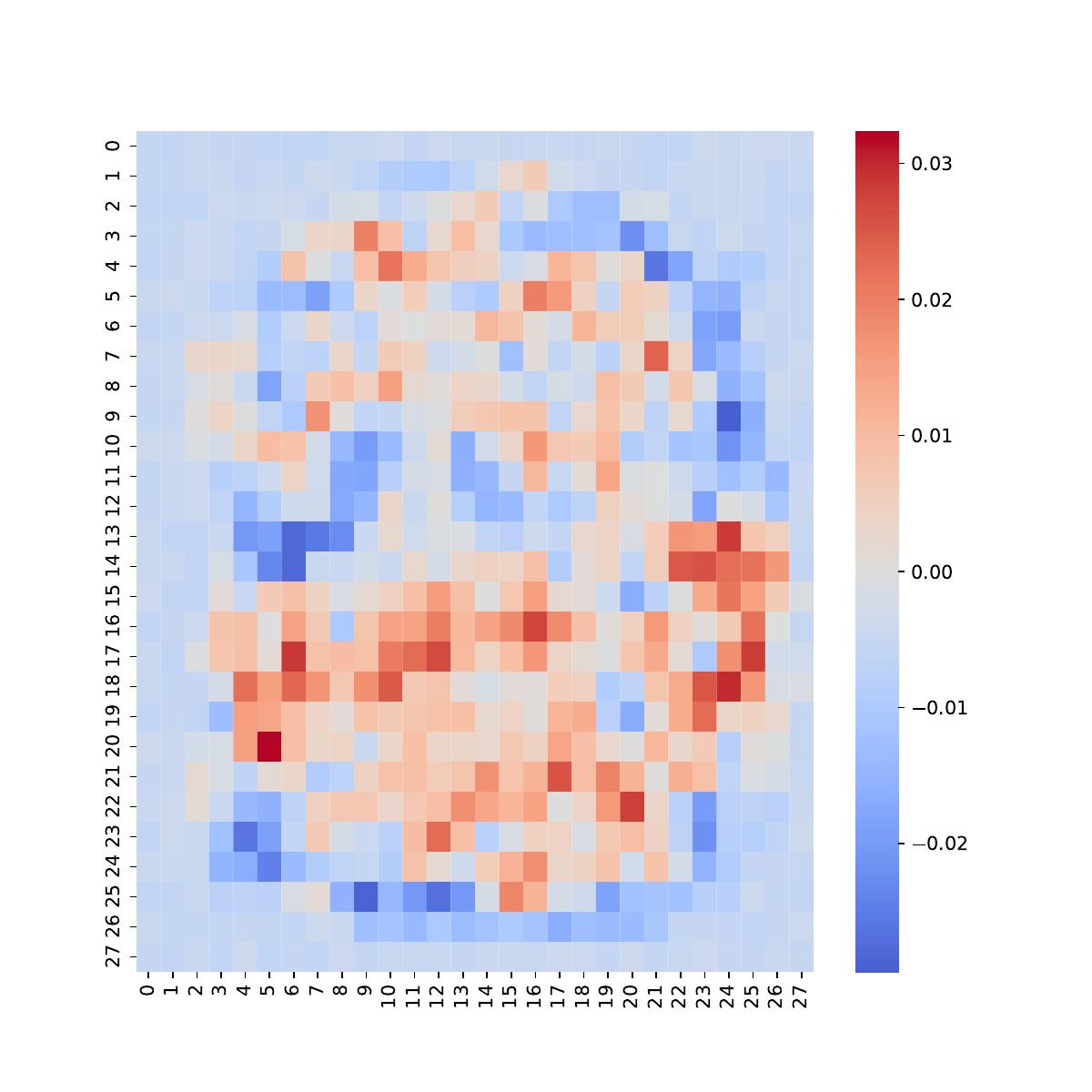}
        \caption{Parameters for Raw Data}
    \end{subfigure}
    \hfill
    \begin{subfigure}{0.48\textwidth}
        \centering
        \includegraphics[width=\textwidth]{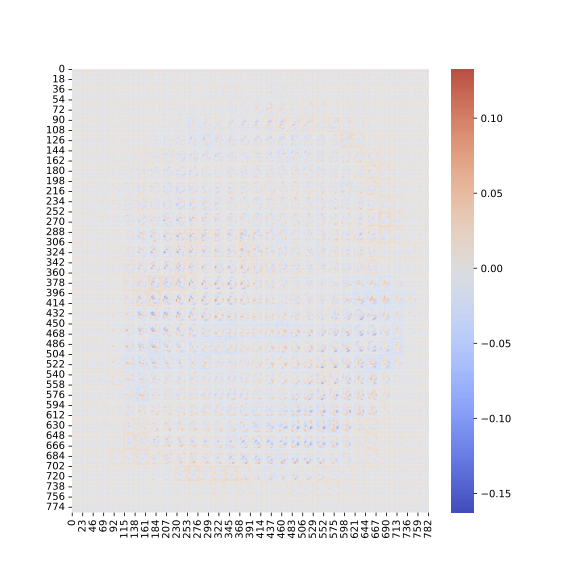}
        \caption{Parameters for Expanded Data}
    \end{subfigure}
    \caption{An illustration of an image with label $2$ randomly selected from the MNIST dataset. Plat (a): raw image data; Plot (b): expanded data; Plot (c): parameter corresponding to output neuron of label $2$ for raw image data; and Plot (d): parameter corresponding to output neuron of label $2$ for expanded image data.}
    \label{fig:mnist_visualization_appendix_2}
\end{figure}
%------------------------------------------------------------------------------------------

%------------------------------------------------------------------------------------------
\newpage
\begin{figure}[h]
    \centering
    \begin{subfigure}{0.48\textwidth}
        \centering
        \includegraphics[width=\textwidth]{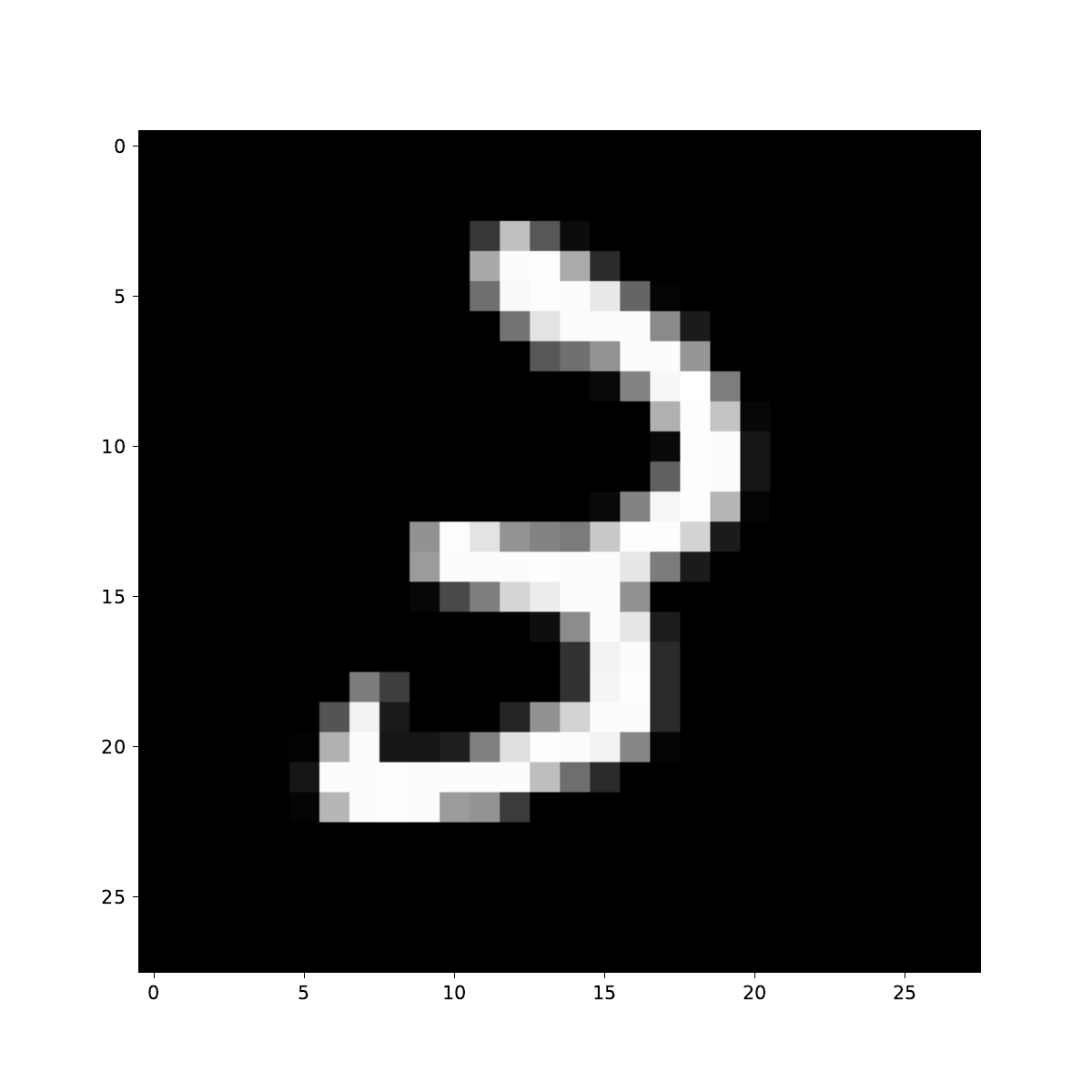}
        \caption{Input Raw Data}
    \end{subfigure}
    \hfill
    \begin{subfigure}{0.48\textwidth}
        \centering
        \includegraphics[width=\textwidth]{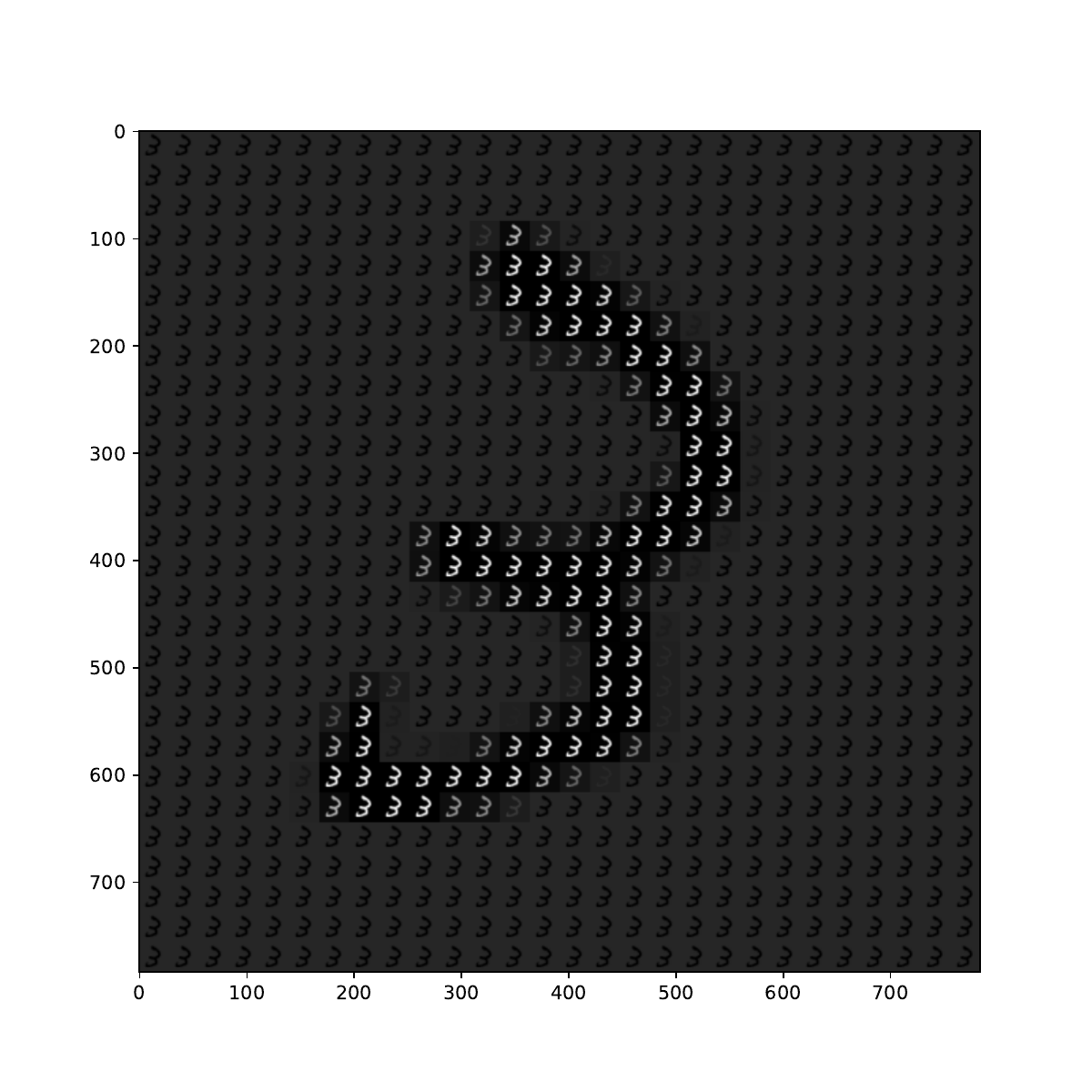}
        \caption{Expanded Data}
    \end{subfigure}
    \begin{subfigure}{0.48\textwidth}
        \centering
        \includegraphics[width=\textwidth]{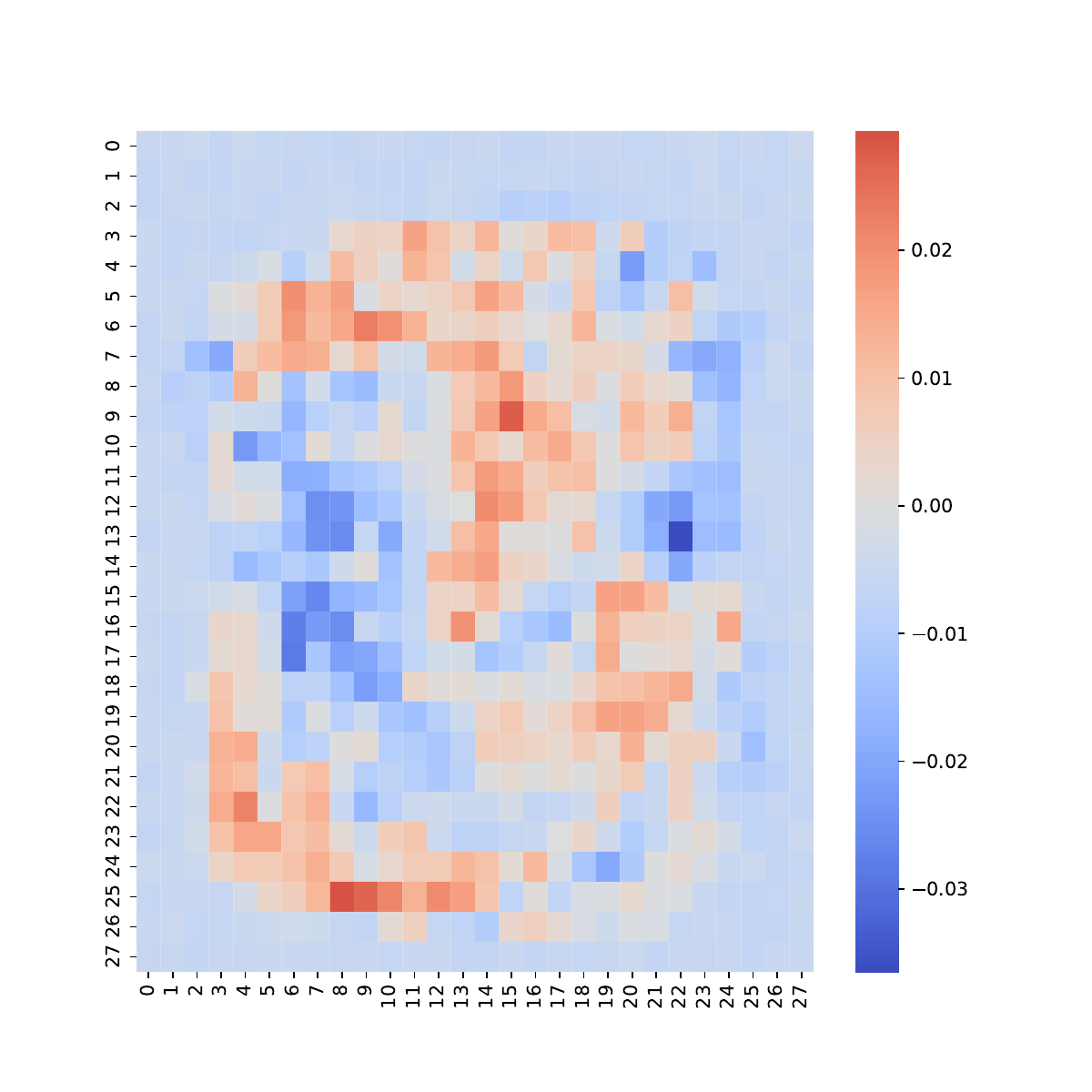}
        \caption{Parameters for Raw Data}
    \end{subfigure}
    \hfill
    \begin{subfigure}{0.48\textwidth}
        \centering
        \includegraphics[width=\textwidth]{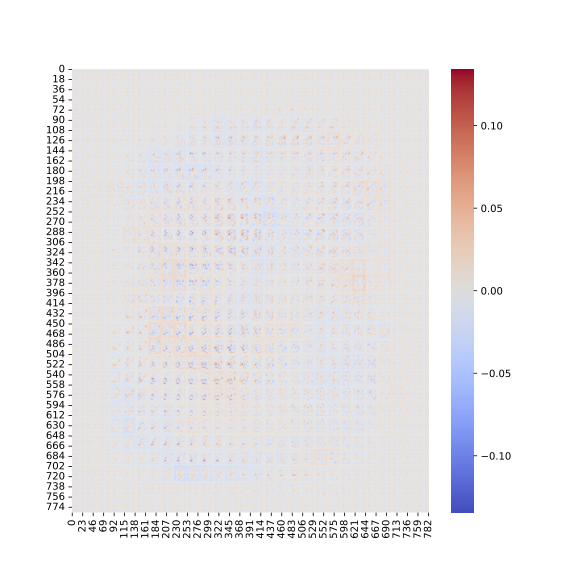}
        \caption{Parameters for Expanded Data}
    \end{subfigure}
    \caption{An illustration of an image with label $3$ randomly selected from the MNIST dataset. Plat (a): raw image data; Plot (b): expanded data; Plot (c): parameter corresponding to output neuron of label $3$ for raw image data; and Plot (d): parameter corresponding to output neuron of label $3$ for expanded image data.}
    \label{fig:mnist_visualization_appendix_3}
\end{figure}
%------------------------------------------------------------------------------------------

%------------------------------------------------------------------------------------------
\newpage
\begin{figure}[h]
    \centering
    \begin{subfigure}{0.48\textwidth}
        \centering
        \includegraphics[width=\textwidth]{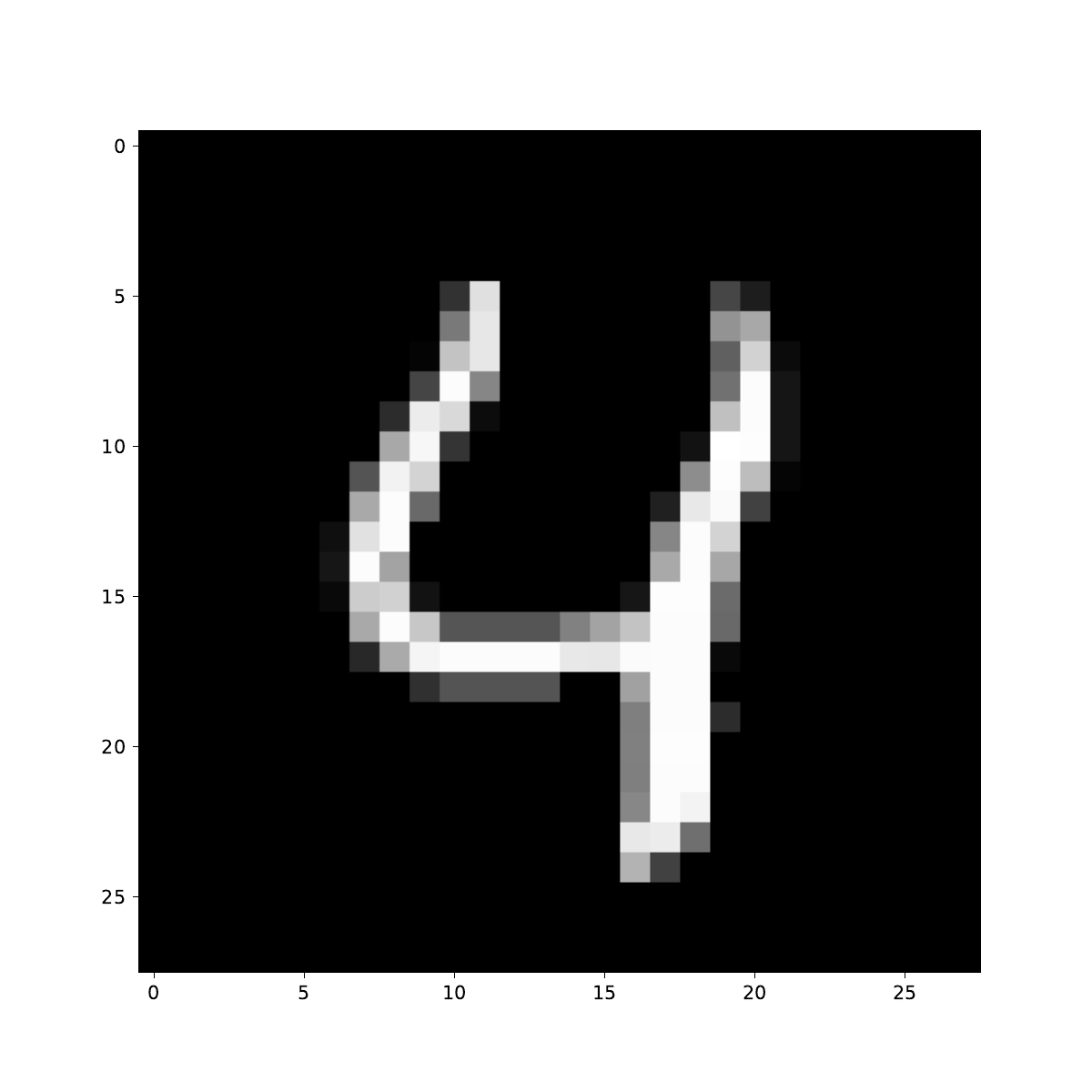}
        \caption{Input Raw Data}
    \end{subfigure}
    \hfill
    \begin{subfigure}{0.48\textwidth}
        \centering
        \includegraphics[width=\textwidth]{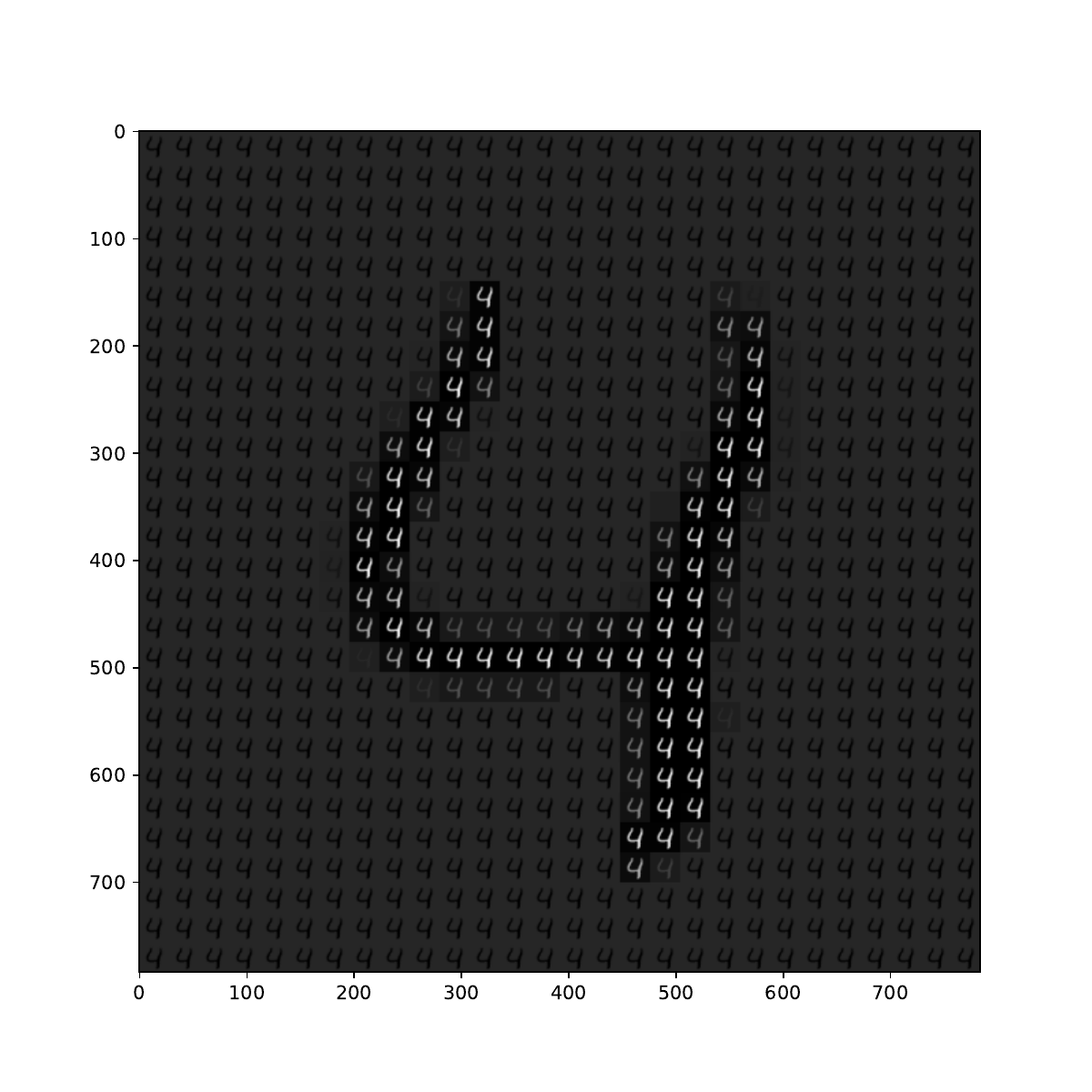}
        \caption{Expanded Data}
    \end{subfigure}
    \begin{subfigure}{0.48\textwidth}
        \centering
        \includegraphics[width=\textwidth]{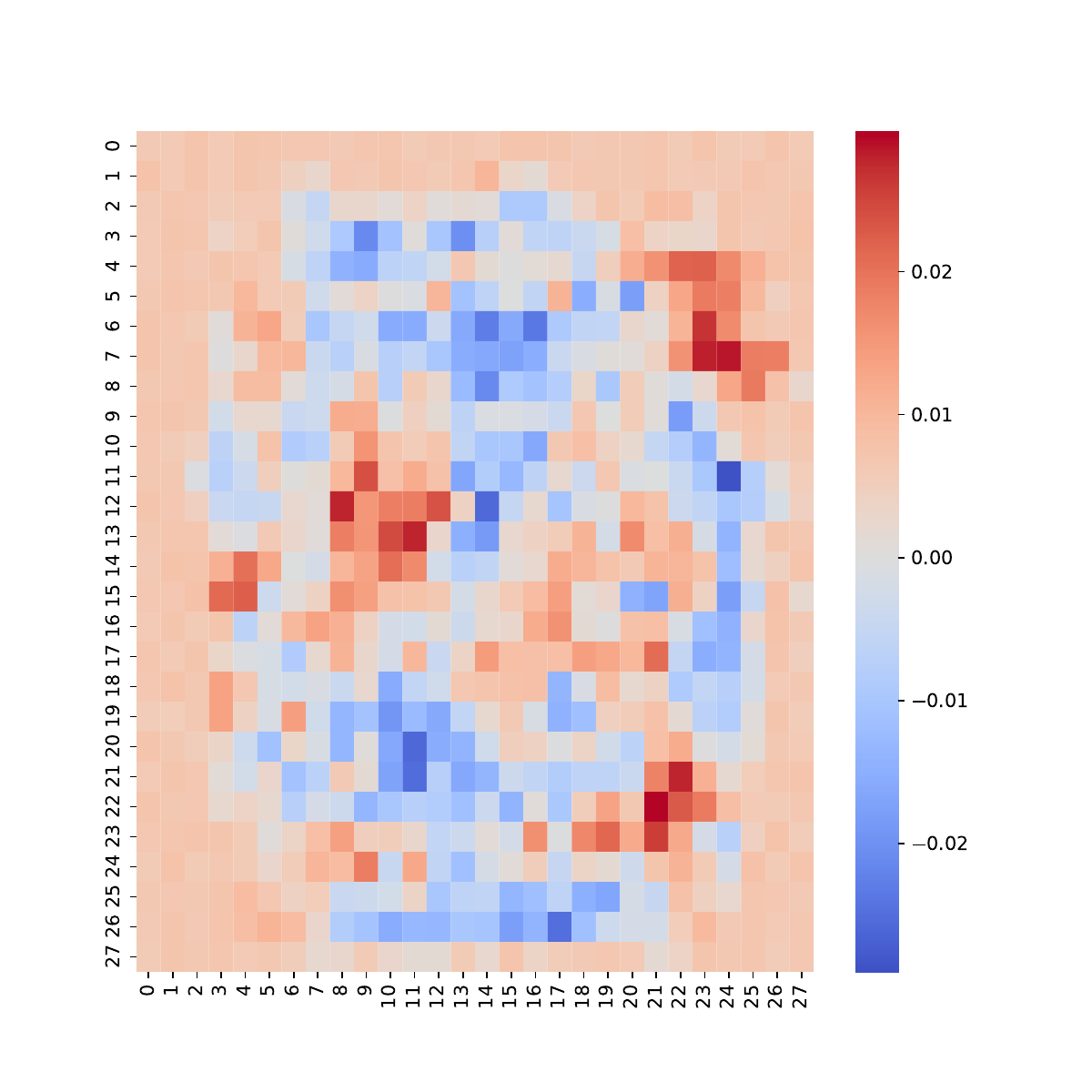}
        \caption{Parameters for Raw Data}
    \end{subfigure}
    \hfill
    \begin{subfigure}{0.48\textwidth}
        \centering
        \includegraphics[width=\textwidth]{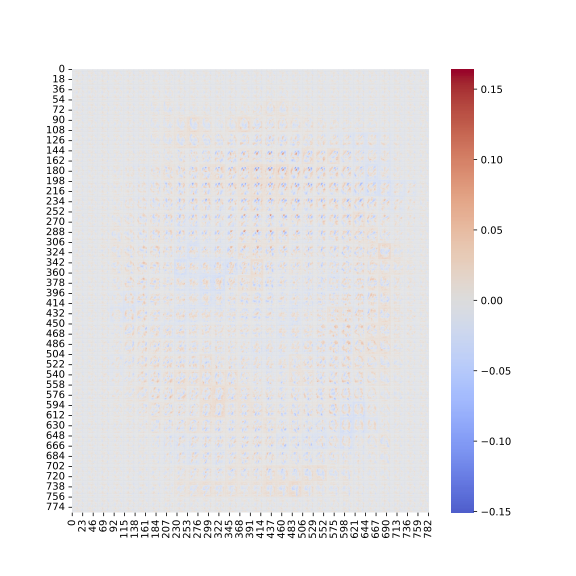}
        \caption{Parameters for Expanded Data}
    \end{subfigure}
    \caption{An illustration of an image with label $4$ randomly selected from the MNIST dataset. Plat (a): raw image data; Plot (b): expanded data; Plot (c): parameter corresponding to output neuron of label $4$ for raw image data; and Plot (d): parameter corresponding to output neuron of label $4$ for expanded image data.}
    \label{fig:mnist_visualization_appendix_4}
\end{figure}
%------------------------------------------------------------------------------------------

%------------------------------------------------------------------------------------------
\newpage
\begin{figure}[h]
    \centering
    \begin{subfigure}{0.48\textwidth}
        \centering
        \includegraphics[width=\textwidth]{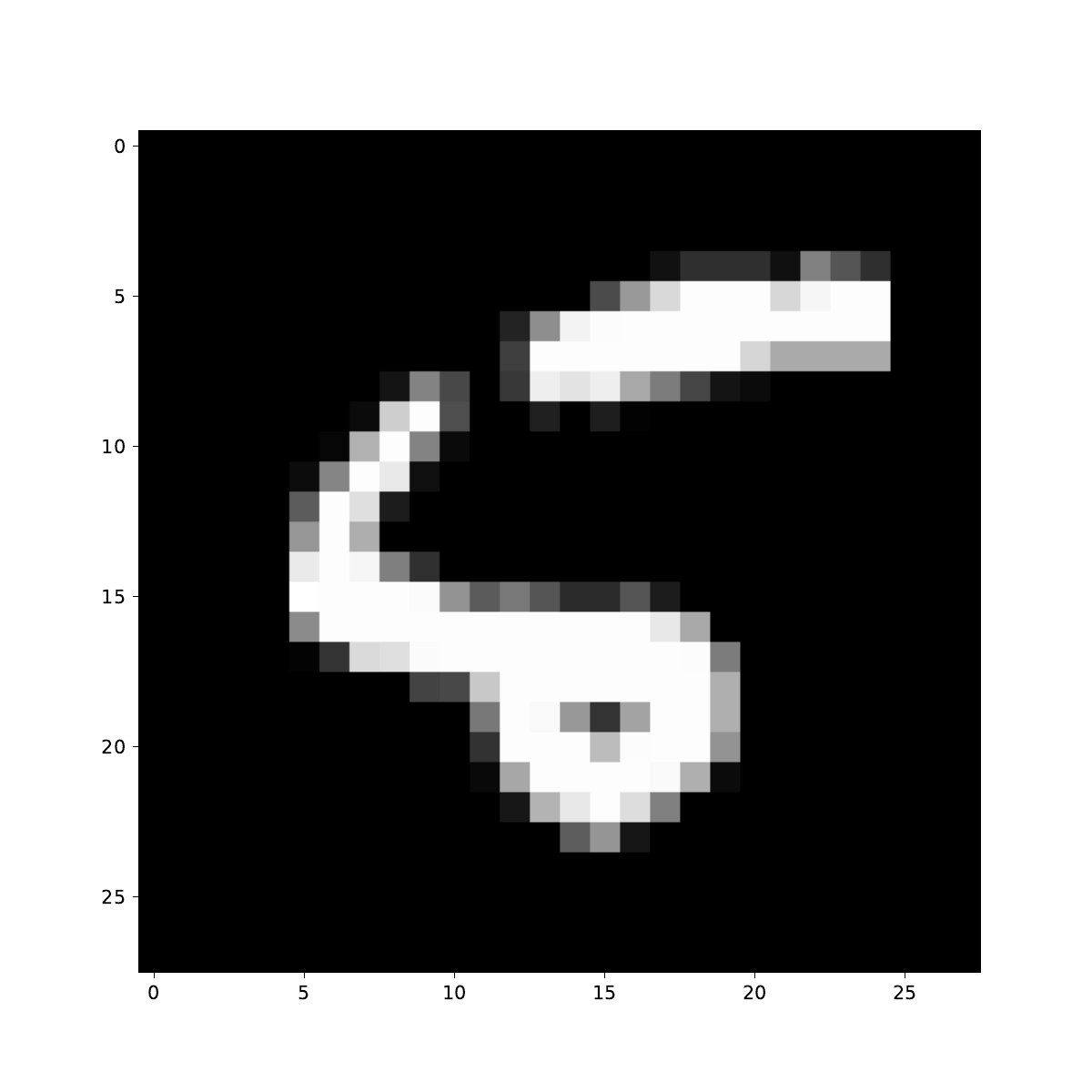}
        \caption{Input Raw Data}
    \end{subfigure}
    \hfill
    \begin{subfigure}{0.48\textwidth}
        \centering
        \includegraphics[width=\textwidth]{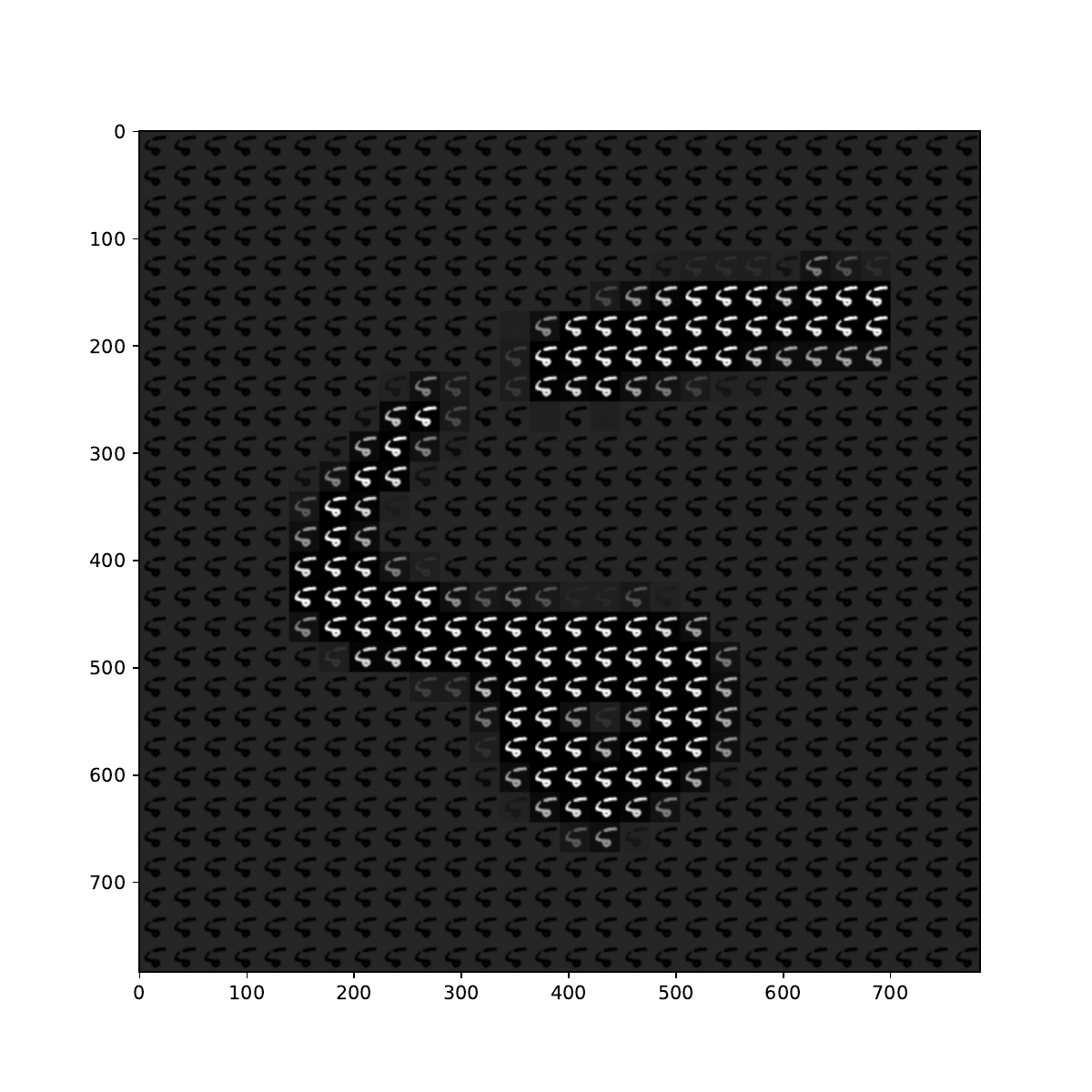}
        \caption{Expanded Data}
    \end{subfigure}
    \begin{subfigure}{0.48\textwidth}
        \centering
        \includegraphics[width=\textwidth]{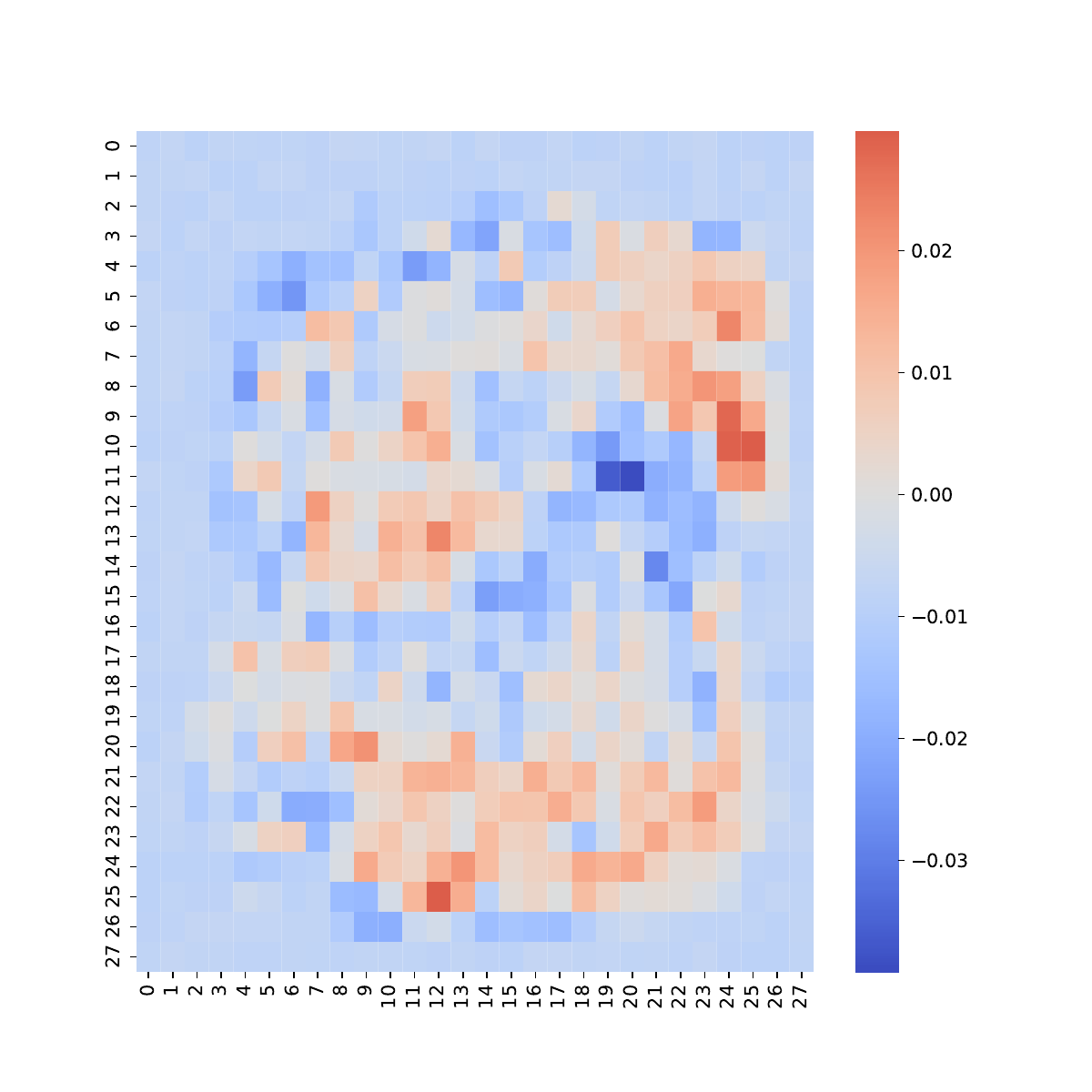}
        \caption{Parameters for Raw Data}
    \end{subfigure}
    \hfill
    \begin{subfigure}{0.48\textwidth}
        \centering
        \includegraphics[width=\textwidth]{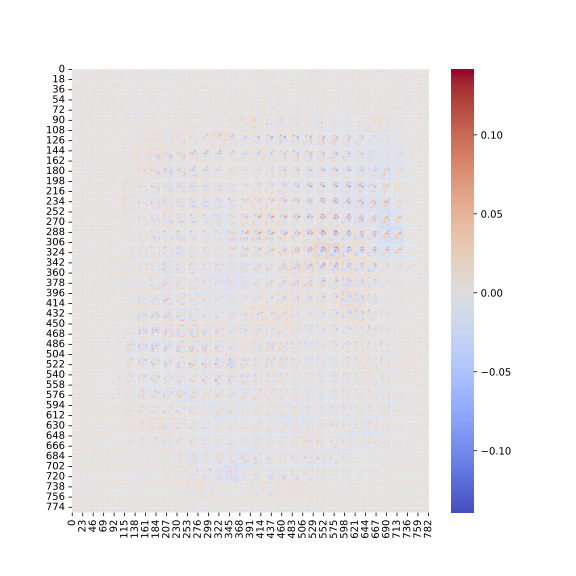}
        \caption{Parameters for Expanded Data}
    \end{subfigure}
    \caption{An illustration of an image with label $5$ randomly selected from the MNIST dataset. Plat (a): raw image data; Plot (b): expanded data; Plot (c): parameter corresponding to output neuron of label $5$ for raw image data; and Plot (d): parameter corresponding to output neuron of label $5$ for expanded image data.}
    \label{fig:mnist_visualization_appendix_5}
\end{figure}
%------------------------------------------------------------------------------------------

%------------------------------------------------------------------------------------------
\newpage
\begin{figure}[h]
    \centering
    \begin{subfigure}{0.48\textwidth}
        \centering
        \includegraphics[width=\textwidth]{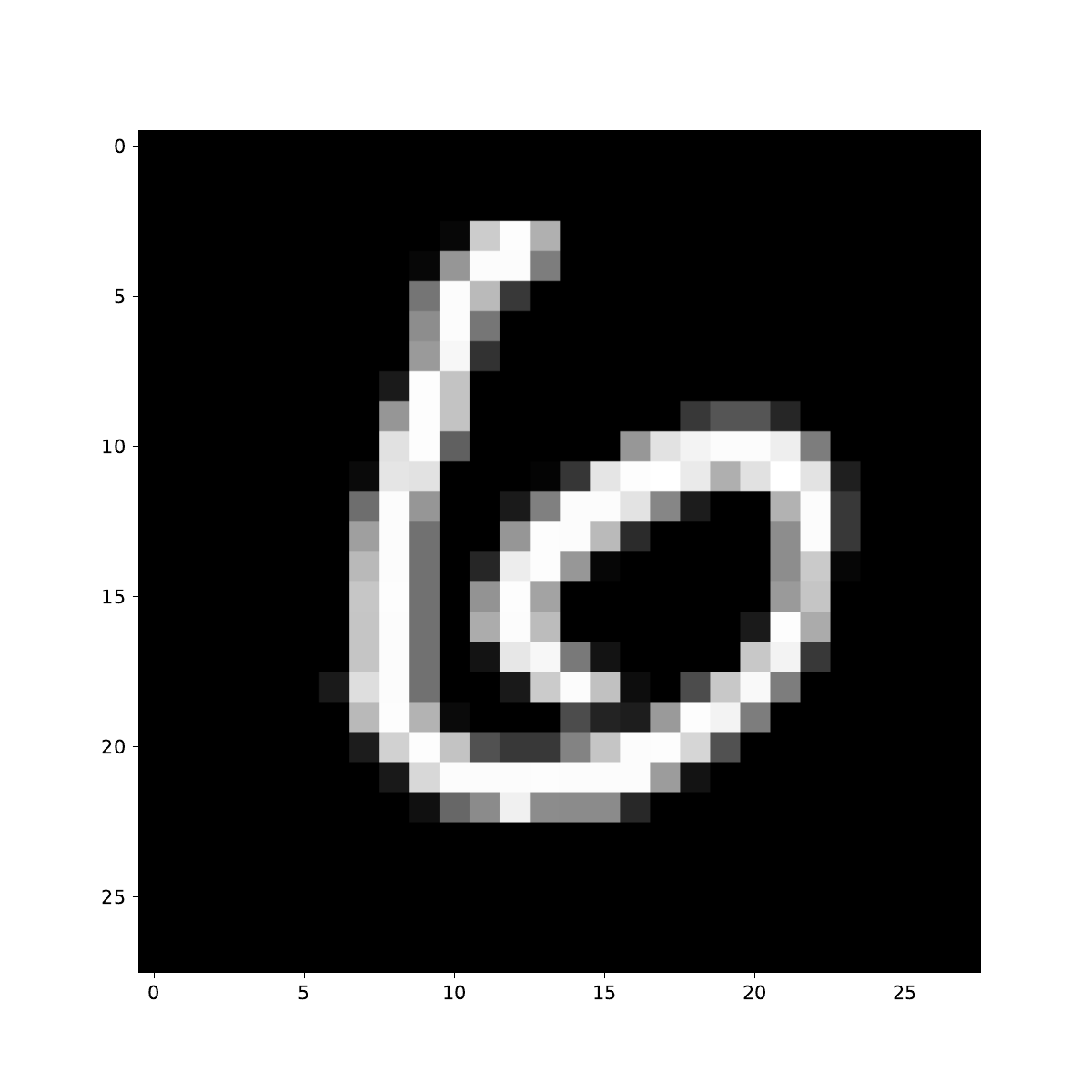}
        \caption{Input Raw Data}
    \end{subfigure}
    \hfill
    \begin{subfigure}{0.48\textwidth}
        \centering
        \includegraphics[width=\textwidth]{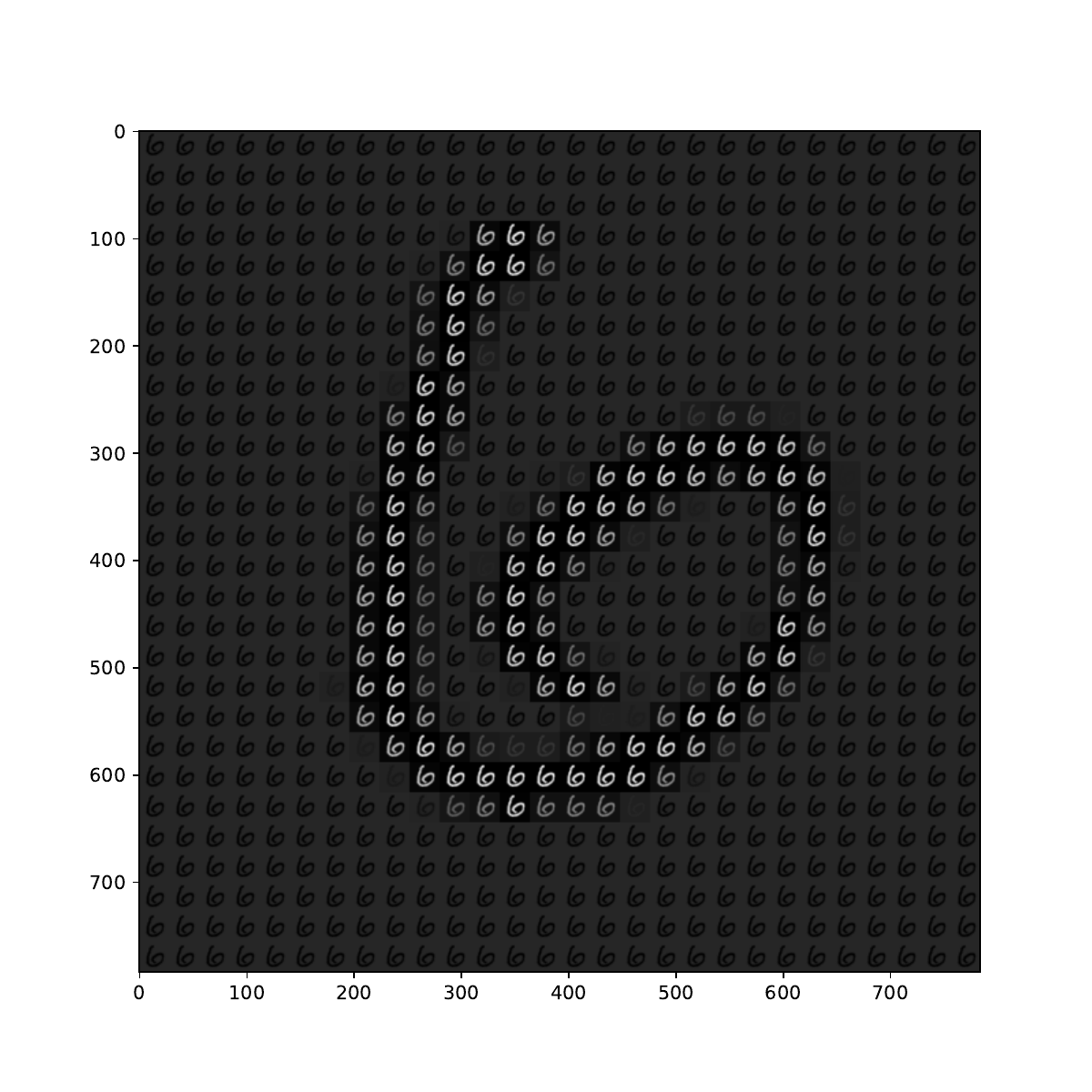}
        \caption{Expanded Data}
    \end{subfigure}
    \begin{subfigure}{0.48\textwidth}
        \centering
        \includegraphics[width=\textwidth]{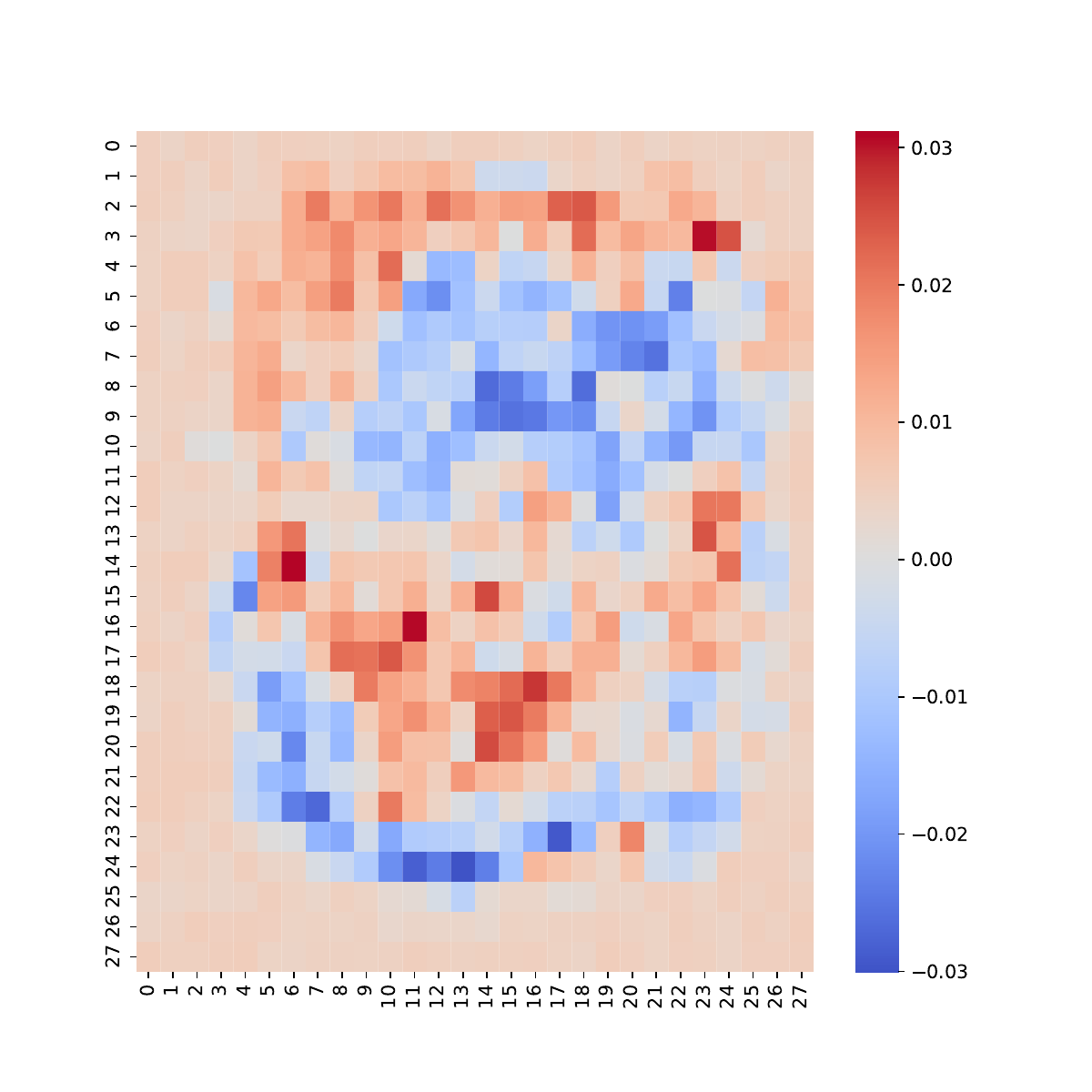}
        \caption{Parameters for Raw Data}
    \end{subfigure}
    \hfill
    \begin{subfigure}{0.48\textwidth}
        \centering
        \includegraphics[width=\textwidth]{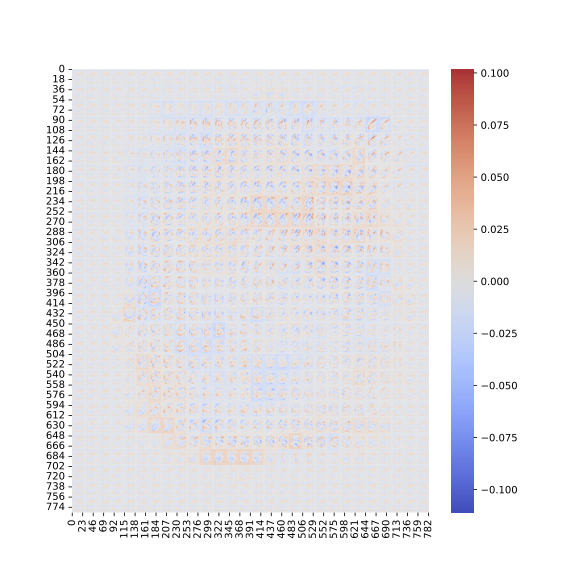}
        \caption{Parameters for Expanded Data}
    \end{subfigure}
    \caption{An illustration of an image with label $6$ randomly selected from the MNIST dataset. Plat (a): raw image data; Plot (b): expanded data; Plot (c): parameter corresponding to output neuron of label $6$ for raw image data; and Plot (d): parameter corresponding to output neuron of label $6$ for expanded image data.}
    \label{fig:mnist_visualization_appendix_6}
\end{figure}
%------------------------------------------------------------------------------------------

%------------------------------------------------------------------------------------------
\newpage
\begin{figure}[h]
    \centering
    \begin{subfigure}{0.48\textwidth}
        \centering
        \includegraphics[width=\textwidth]{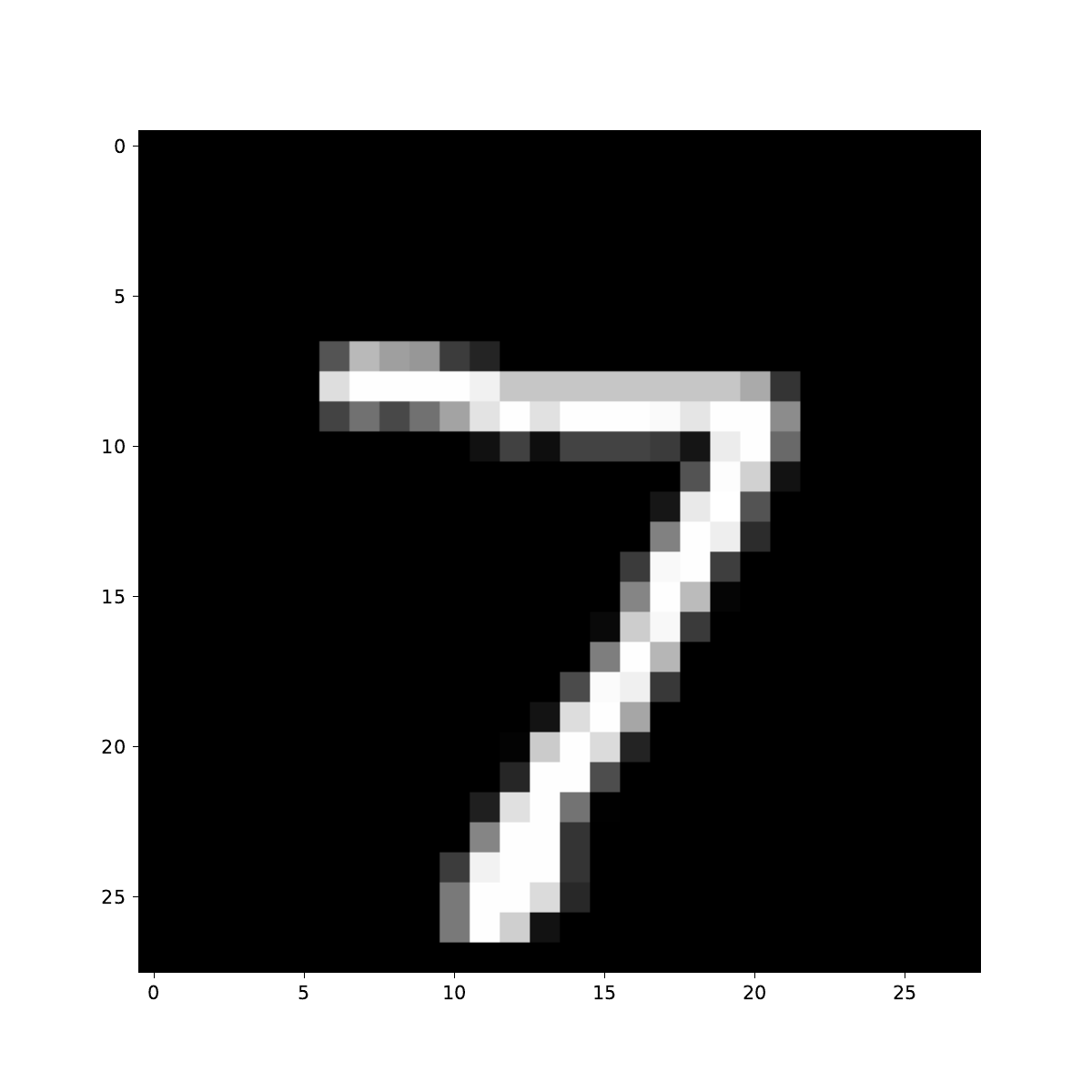}
        \caption{Input Raw Data}
    \end{subfigure}
    \hfill
    \begin{subfigure}{0.48\textwidth}
        \centering
        \includegraphics[width=\textwidth]{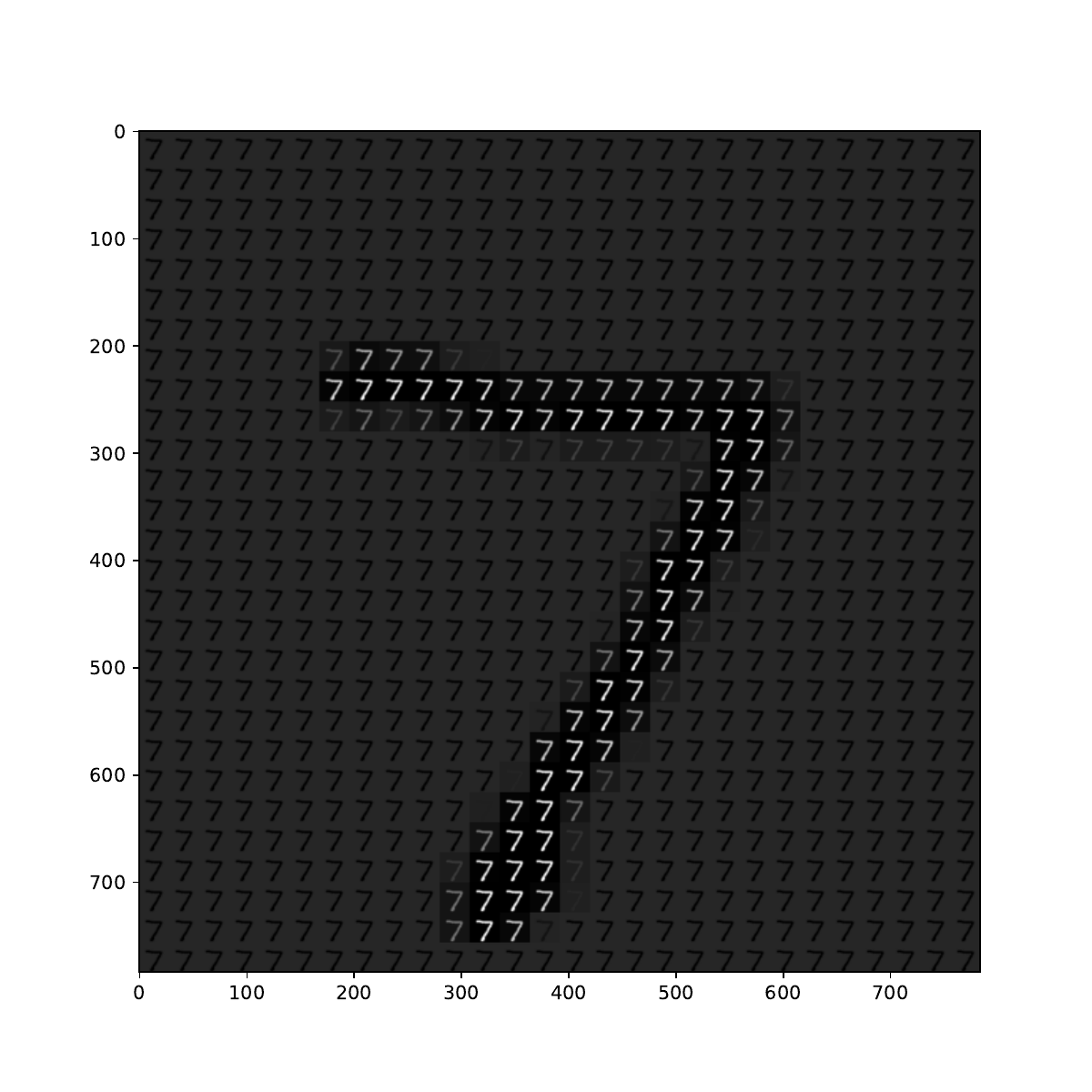}
        \caption{Expanded Data}
    \end{subfigure}
    \begin{subfigure}{0.48\textwidth}
        \centering
        \includegraphics[width=\textwidth]{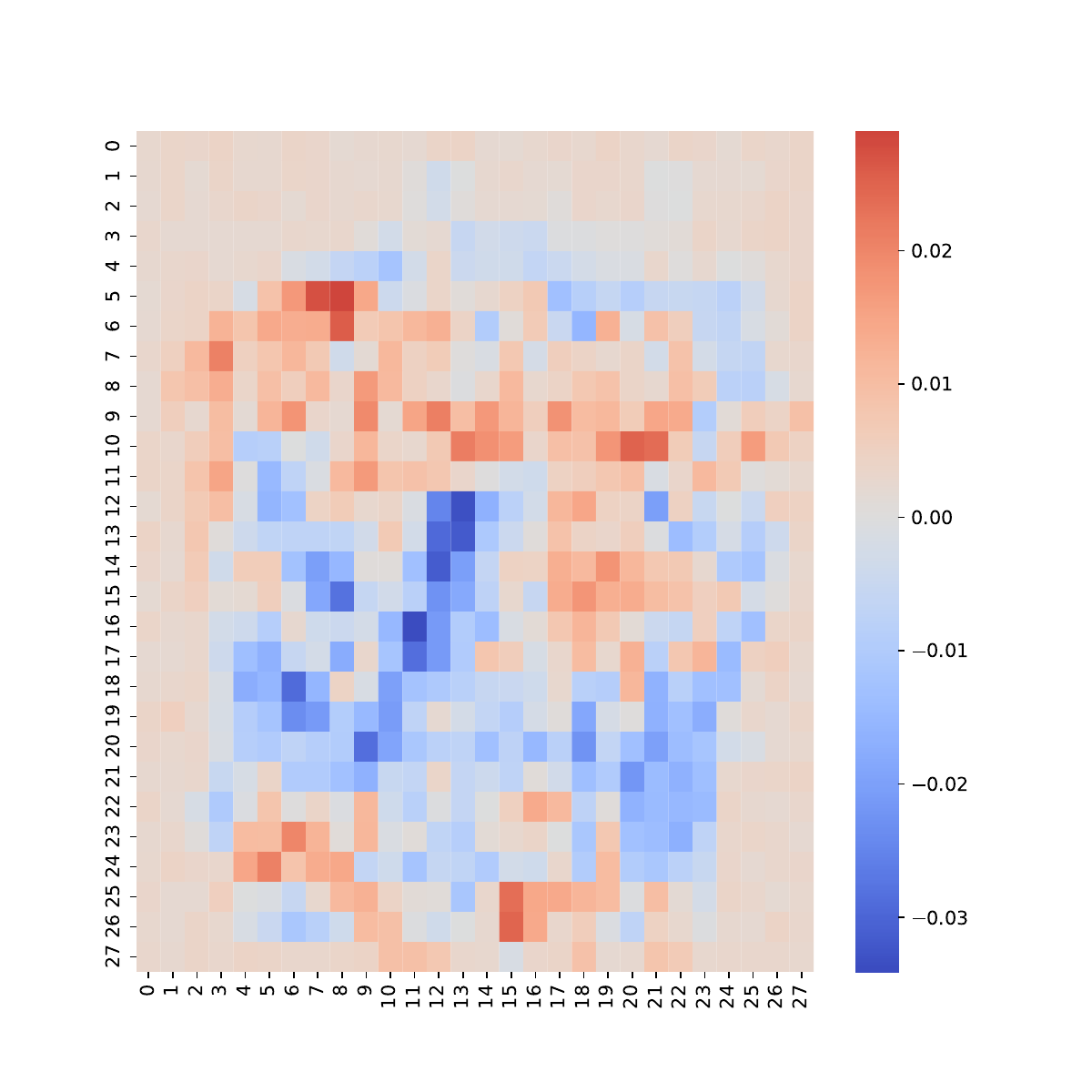}
        \caption{Parameters for Raw Data}
    \end{subfigure}
    \hfill
    \begin{subfigure}{0.48\textwidth}
        \centering
        \includegraphics[width=\textwidth]{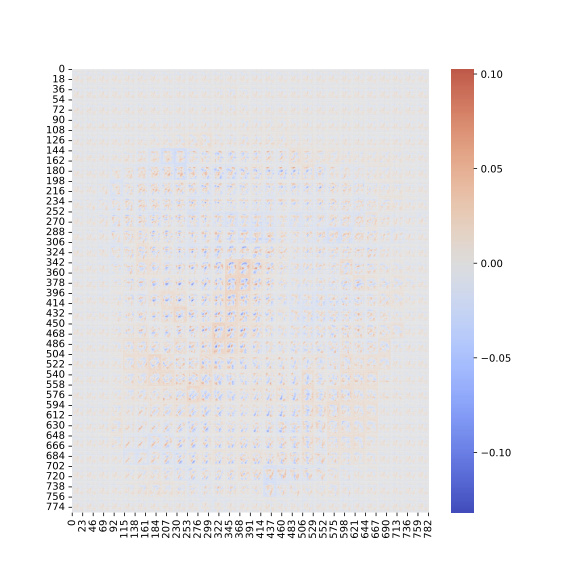}
        \caption{Parameters for Expanded Data}
    \end{subfigure}
    \caption{An illustration of an image with label $7$ randomly selected from the MNIST dataset. Plat (a): raw image data; Plot (b): expanded data; Plot (c): parameter corresponding to output neuron of label $7$ for raw image data; and Plot (d): parameter corresponding to output neuron of label $7$ for expanded image data.}
    \label{fig:mnist_visualization_appendix_7}
\end{figure}
%------------------------------------------------------------------------------------------

%------------------------------------------------------------------------------------------
\newpage
\begin{figure}[h]
    \centering
    \begin{subfigure}{0.48\textwidth}
        \centering
        \includegraphics[width=\textwidth]{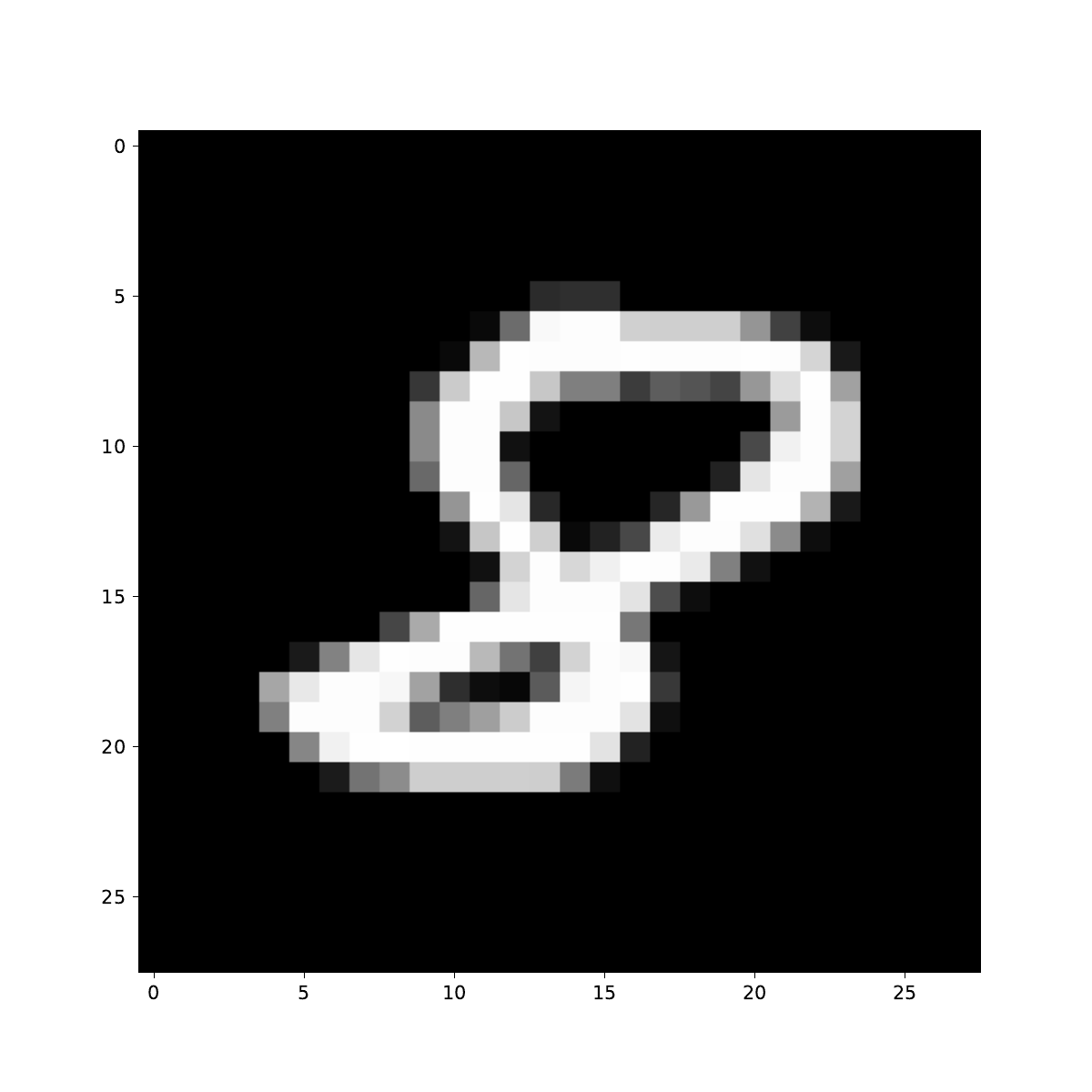}
        \caption{Input Raw Data}
    \end{subfigure}
    \hfill
    \begin{subfigure}{0.48\textwidth}
        \centering
        \includegraphics[width=\textwidth]{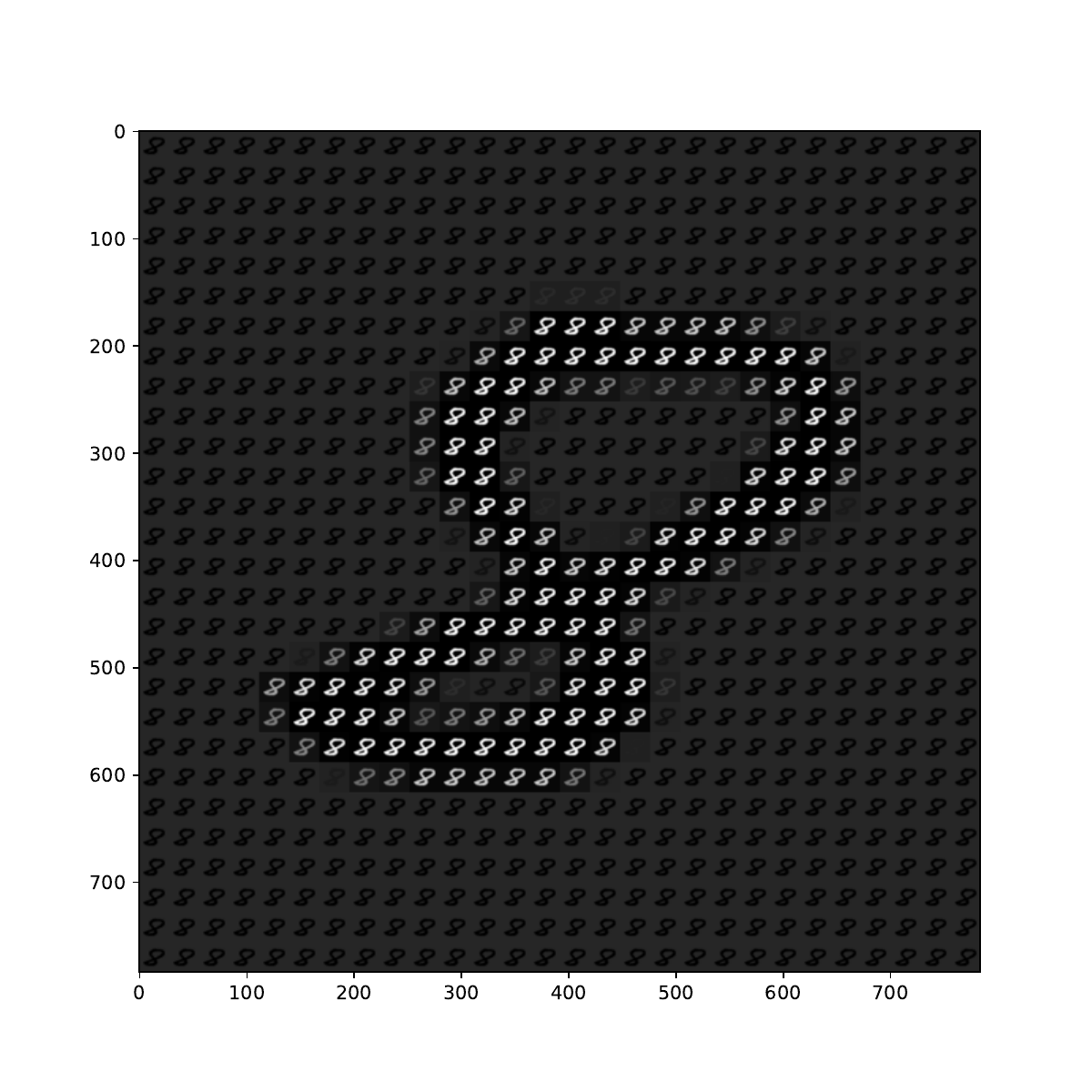}
        \caption{Expanded Data}
    \end{subfigure}
    \begin{subfigure}{0.48\textwidth}
        \centering
        \includegraphics[width=\textwidth]{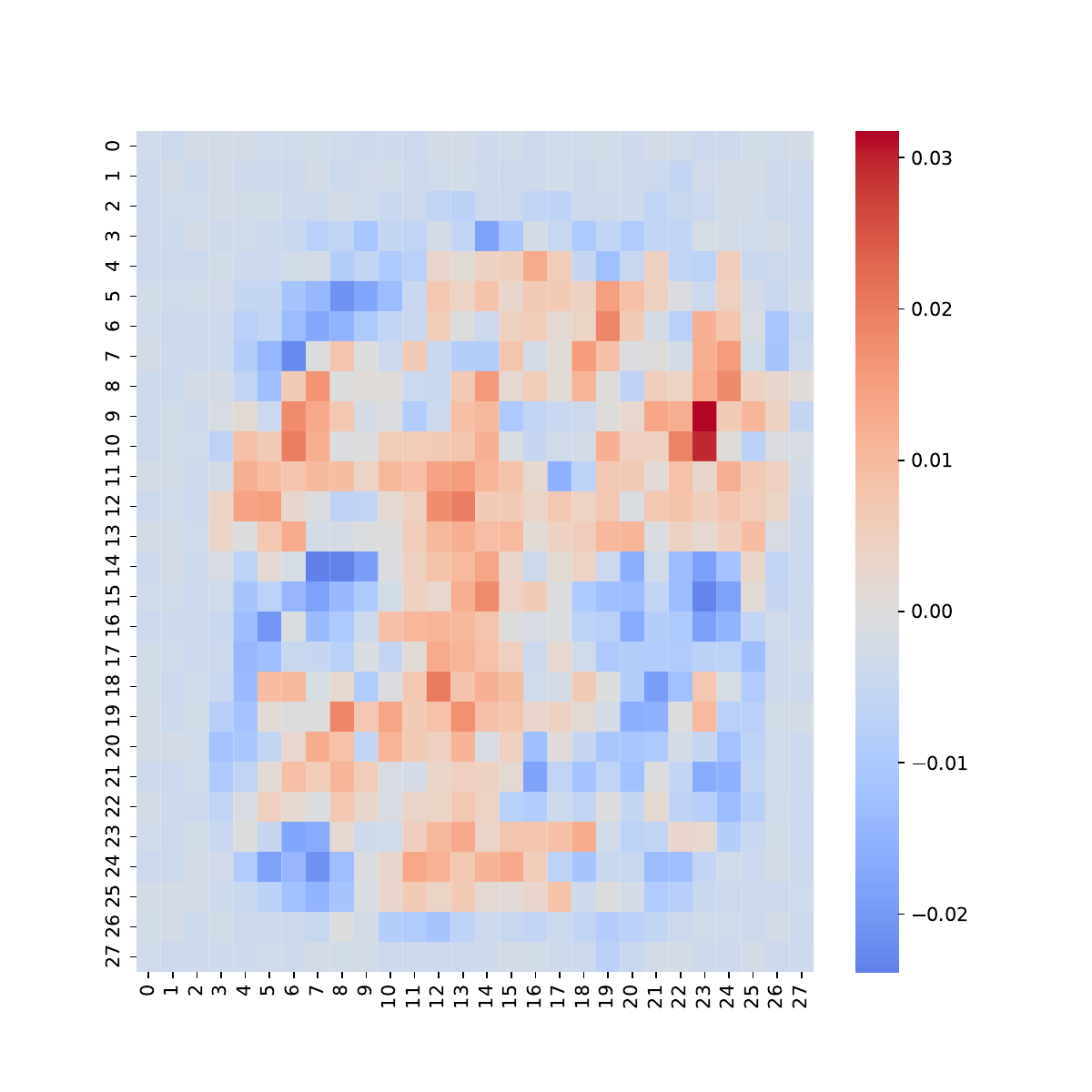}
        \caption{Parameters for Raw Data}
    \end{subfigure}
    \hfill
    \begin{subfigure}{0.48\textwidth}
        \centering
        \includegraphics[width=\textwidth]{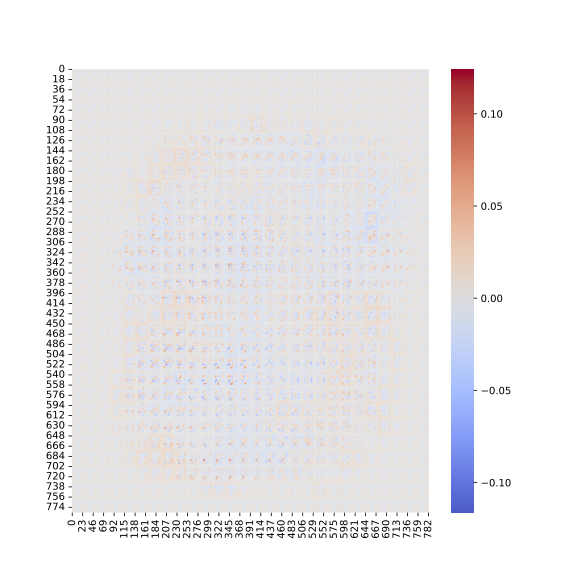}
        \caption{Parameters for Expanded Data}
    \end{subfigure}
    \caption{An illustration of an image with label $8$ randomly selected from the MNIST dataset. Plat (a): raw image data; Plot (b): expanded data; Plot (c): parameter corresponding to output neuron of label $8$ for raw image data; and Plot (d): parameter corresponding to output neuron of label $8$ for expanded image data.}
    \label{fig:mnist_visualization_appendix_8}
\end{figure}
%------------------------------------------------------------------------------------------

%------------------------------------------------------------------------------------------
\newpage
\begin{figure}[h]
    \centering
    \begin{subfigure}{0.48\textwidth}
        \centering
        \includegraphics[width=\textwidth]{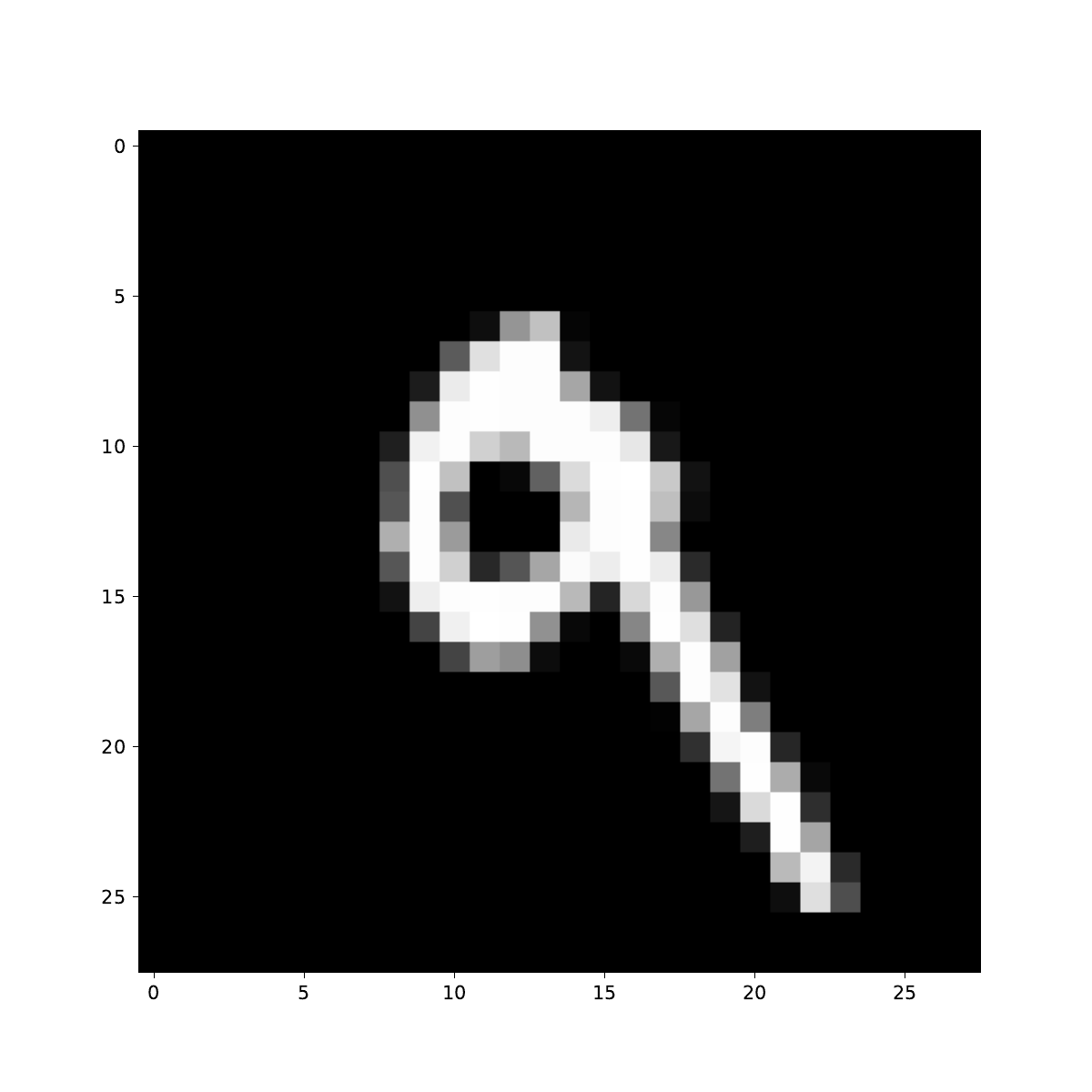}
        \caption{Input Raw Data}
    \end{subfigure}
    \hfill
    \begin{subfigure}{0.48\textwidth}
        \centering
        \includegraphics[width=\textwidth]{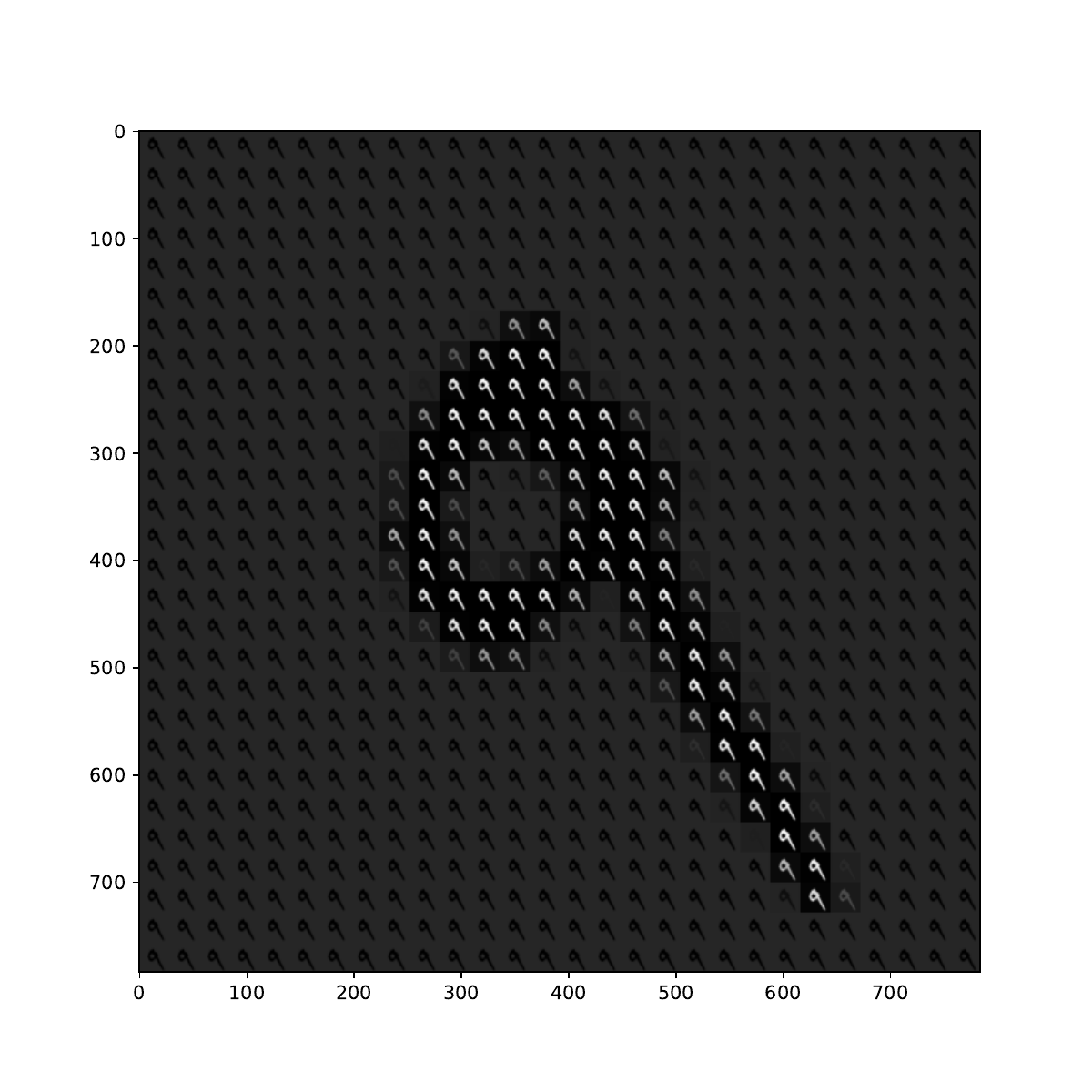}
        \caption{Expanded Data}
    \end{subfigure}
    \begin{subfigure}{0.48\textwidth}
        \centering
        \includegraphics[width=\textwidth]{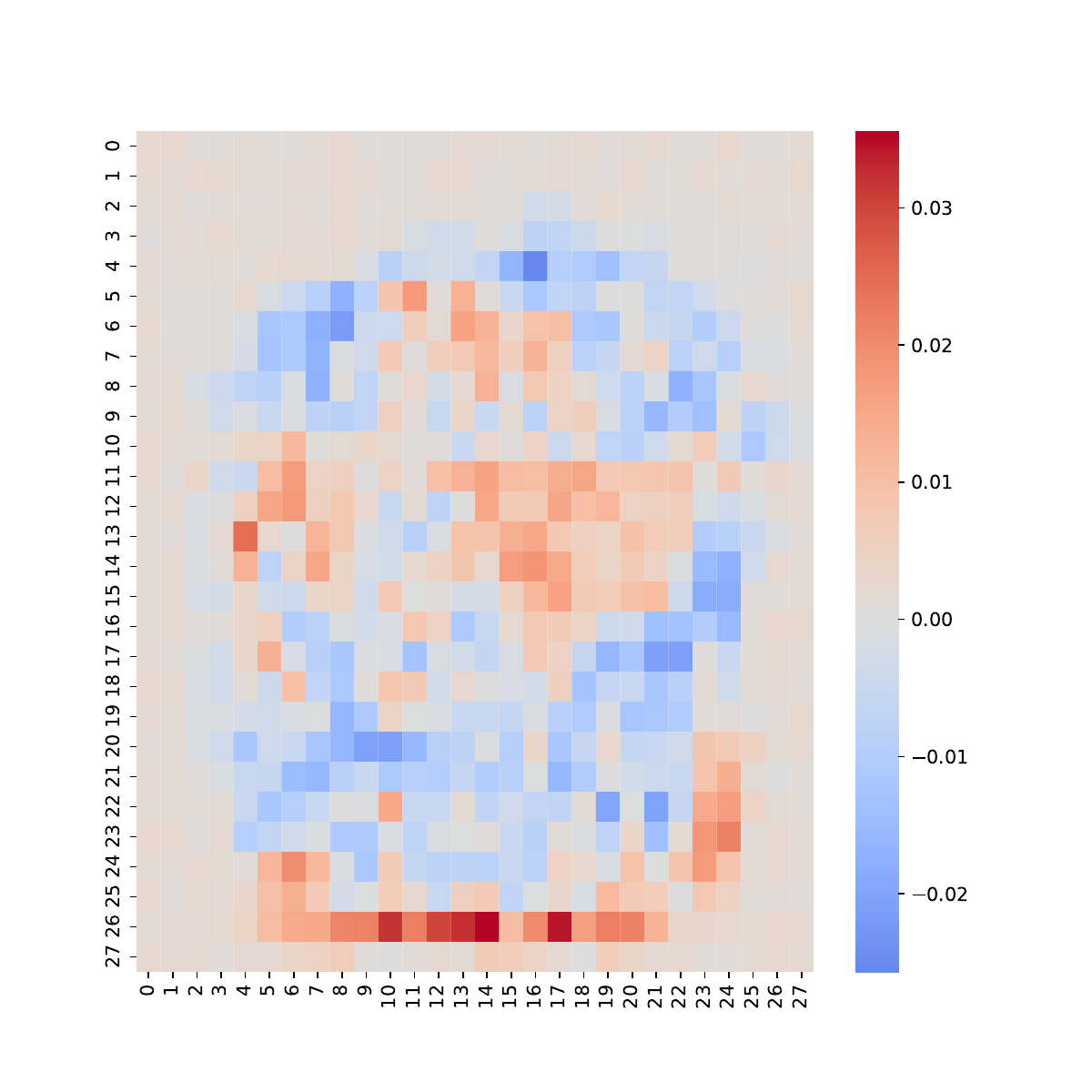}
        \caption{Parameters for Raw Data}
    \end{subfigure}
    \hfill
    \begin{subfigure}{0.48\textwidth}
        \centering
        \includegraphics[width=\textwidth]{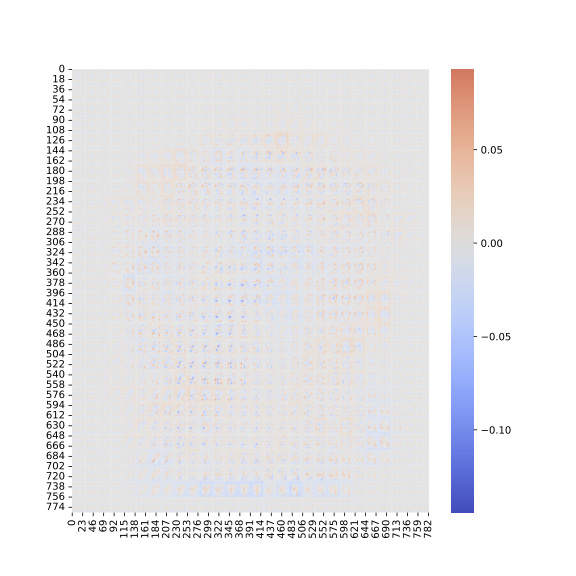}
        \caption{Parameters for Expanded Data}
    \end{subfigure}
    \caption{An illustration of an image with label $9$ randomly selected from the MNIST dataset. Plat (a): raw image data; Plot (b): expanded data; Plot (c): parameter corresponding to output neuron of label $9$ for raw image data; and Plot (d): parameter corresponding to output neuron of label $9$ for expanded image data.}
    \label{fig:mnist_visualization_appendix_9}
\end{figure}
%------------------------------------------------------------------------------------------

%-----------------------------------------

%==============================================================
\newpage
\subsection{Licensing Rights of Using BioRender Created Contents in This Paper}

\noindent \textbf{Descriptions}: For the bio-medical image contents used in the previous Figures~\ref{fig:neurons}-\ref{fig:action_potential} in Section~\ref{sec:interpretation}, we have obtained the permissions to use them in publications. The licensing rights granting confirmation letters from BioRender for using these generated contents are attached in the following pages.

\newpage


\section*{Supplementary material}

# Confirmation of Publication and Licensing Rights

**May 15th, 2024**
**Science Suite Inc.**

*Subscription:* Institution
*Agreement number:* AN26TMJDKE
*Journal name:* arxiv

To whom this may concern,

This document is to confirm that Jiawei Zhang has been granted a license to use the BioRender content, including icons, templates and other original artwork, appearing in the attached completed graphic pursuant to BioRender's Academic License Terms. This license permits BioRender content to be sublicensed for use in journal publications.

All rights and ownership of BioRender content are reserved by BioRender. All completed graphics must be accompanied by the following citation: "Created with BioRender.com".

BioRender content included in the completed graphic is not licensed for any commercial uses beyond publication in a journal. For any commercial use of this figure, users may, if allowed, recreate it in BioRender under an Industry BioRender Plan.

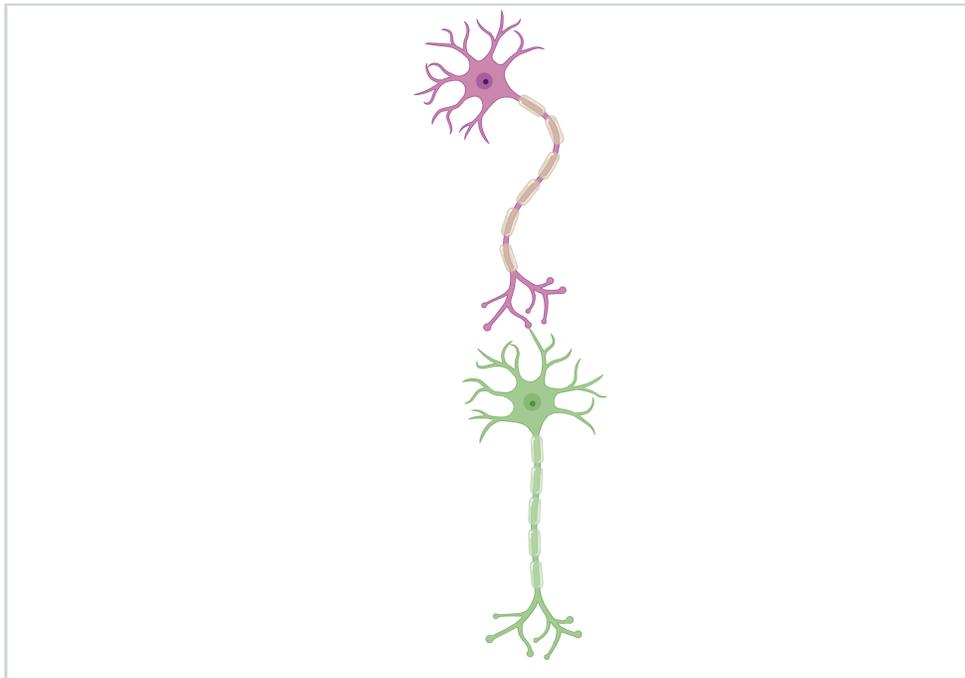

*For any questions regarding this document, or other questions about publishing with BioRender refer to our BioRender Publication Guide, or contact BioRender Support at support@biorender.com.*

![BioRender logo]

# Confirmation of Publication and Licensing Rights

**May 15th, 2024**
**Science Suite Inc.**

***Subscription:*** *Institution*
***Agreement number:*** *FK26TMK5JY*
***Journal name:*** *arxiv*

To whom this may concern,

This document is to confirm that Jiawei Zhang has been granted a license to use the BioRender content, including icons, templates and other original artwork, appearing in the attached completed graphic pursuant to BioRender's [Academic License Terms](). This license permits BioRender content to be sublicensed for use in journal publications.

All rights and ownership of BioRender content are reserved by BioRender. All completed graphics must be accompanied by the following citation: "Created with BioRender.com".

BioRender content included in the completed graphic is not licensed for any commercial uses beyond publication in a journal. For any commercial use of this figure, users may, if allowed, recreate it in BioRender under an Industry BioRender Plan.

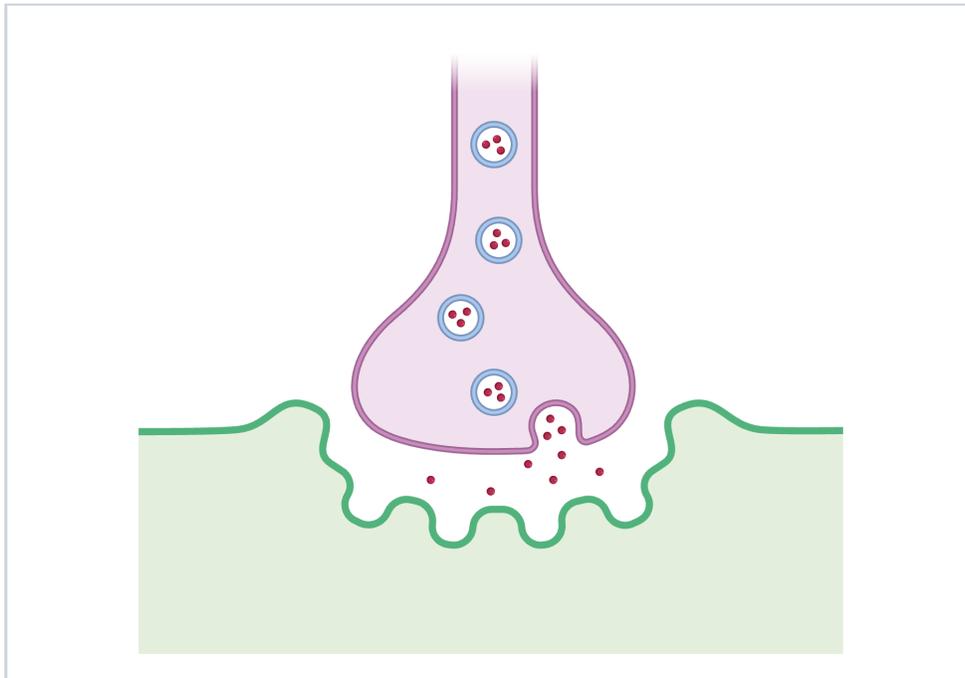

*For any questions regarding this document, or other questions about publishing with BioRender refer to our [BioRender Publication Guide](), or contact BioRender Support at [support@biorender.com]().*

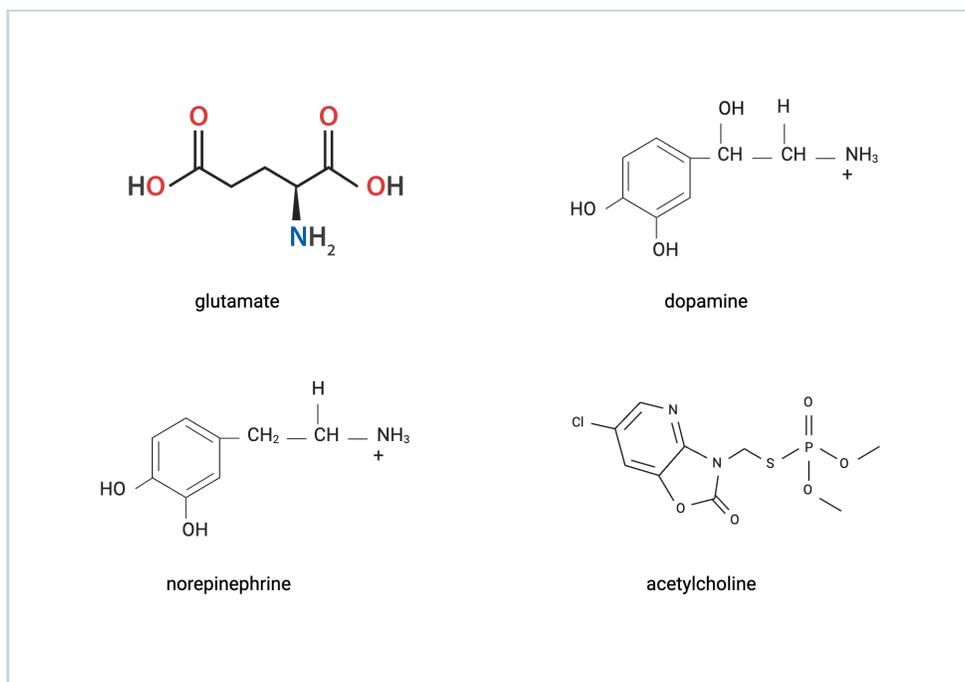

# Confirmation of Publication and Licensing Rights

49 Spadina Ave. Suite 200
Toronto ON M5V 2J1 Canada
www.biorender.com

May 15th, 2024
Science Suite Inc.

**Subscription:** Institution
**Agreement number:** BD26TMLQQI
**Journal name:** arxiv

To whom this may concern,

This document is to confirm that Jiawei Zhang has been granted a license to use the BioRender content, including icons, templates and other original artwork, appearing in the attached completed graphic pursuant to BioRender's Academic License Terms. This license permits BioRender content to be sublicensed for use in journal publications.

All rights and ownership of BioRender content are reserved by BioRender. All completed graphics must be accompanied by the following citation: "Created with BioRender.com".

BioRender content included in the completed graphic is not licensed for any commercial uses beyond publication in a journal. For any commercial use of this figure, users may, if allowed, recreate it in BioRender under an Industry BioRender Plan.

For any questions regarding this document, or other questions about publishing with BioRender refer to our BioRender Publication Guide, or contact BioRender Support at support@biorender.com.

# Confirmation of Publication and Licensing Rights

**May 15th, 2024**
**Science Suite Inc.**

***Subscription:*** Institution
***Agreement number:*** WA26TNN08G
***Journal name:*** arxiv

To whom this may concern,

This document is to confirm that Jiawei Zhang has been granted a license to use the BioRender content, including icons, templates and other original artwork, appearing in the attached completed graphic pursuant to BioRender's [Academic License Terms](#). This license permits BioRender content to be sublicensed for use in journal publications.

All rights and ownership of BioRender content are reserved by BioRender. All completed graphics must be accompanied by the following citation: "Created with BioRender.com".

BioRender content included in the completed graphic is not licensed for any commercial uses beyond publication in a journal. For any commercial use of this figure, users may, if allowed, recreate it in BioRender under an Industry BioRender Plan.

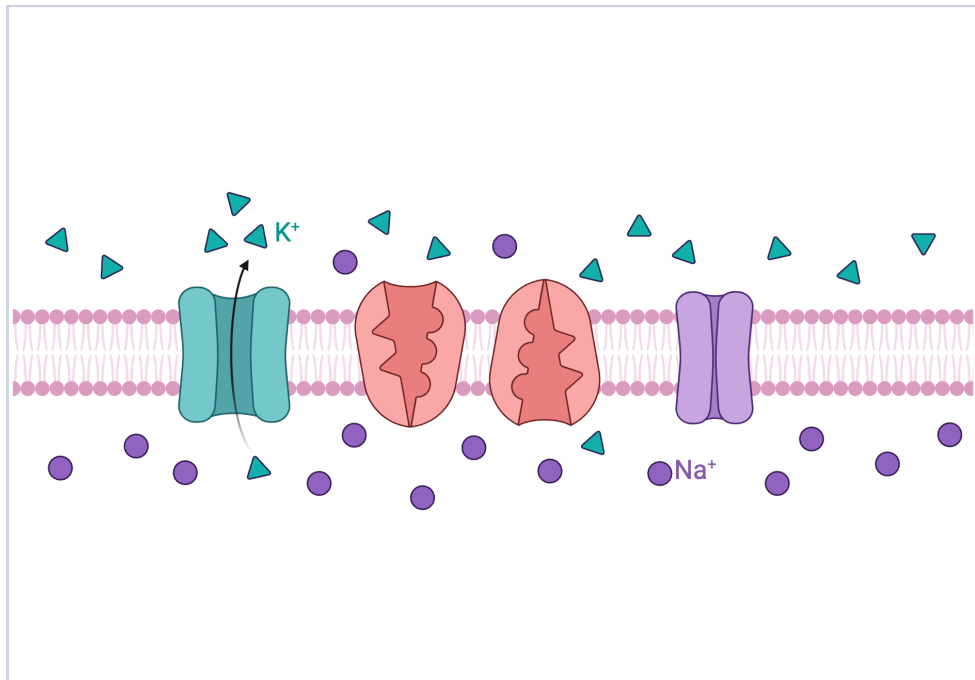

*For any questions regarding this document, or other questions about publishing with BioRender refer to our [BioRender Publication Guide](#), or contact BioRender Support at [support@biorender.com](mailto:support@biorender.com).*

# Confirmation of Publication and Licensing Rights

May 15th, 2024
Science Suite Inc.

**Subscription:** Institution
**Agreement number:** LJ26TOMY8Y
**Journal name:** arxiv

To whom this may concern,

This document is to confirm that Jiawei Zhang has been granted a license to use the BioRender content, including icons, templates and other original artwork, appearing in the attached completed graphic pursuant to BioRender's Academic License Terms. This license permits BioRender content to be sublicensed for use in journal publications.

All rights and ownership of BioRender content are reserved by BioRender. All completed graphics must be accompanied by the following citation: "Created with BioRender.com".

BioRender content included in the completed graphic is not licensed for any commercial uses beyond publication in a journal. For any commercial use of this figure, users may, if allowed, recreate it in BioRender under an Industry BioRender Plan.

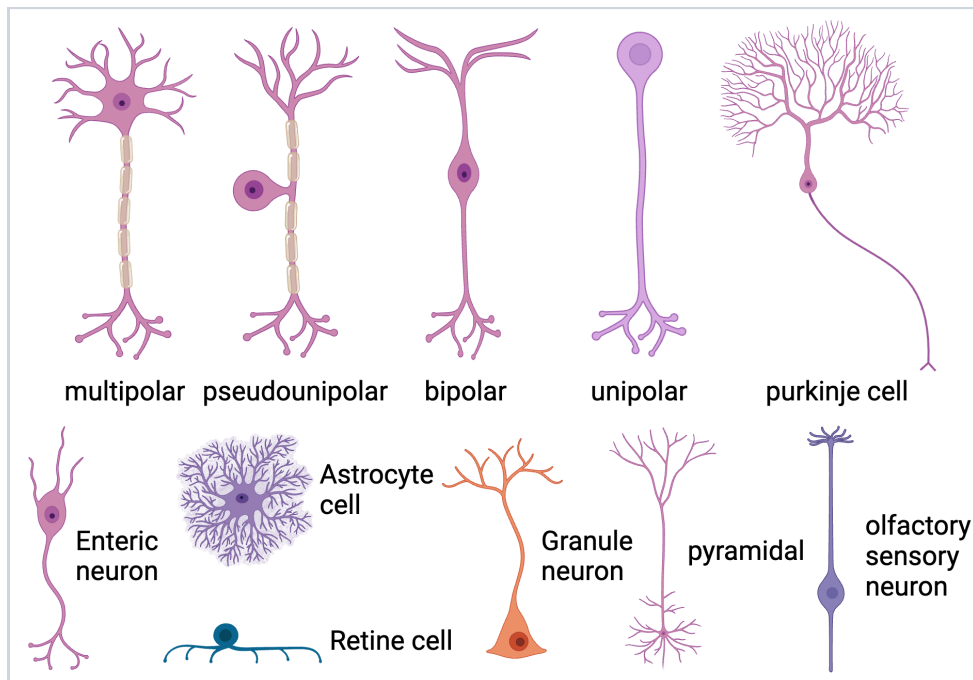

*For any questions regarding this document, or other questions about publishing with BioRender refer to our BioRender Publication Guide, or contact BioRender Support at support@biorender.com.*


}
%--------------------------------------------------------------------------

%---- optional ----
%\input{sec_appendix.tex}
%--------------------------------------------------------------------------

\end{document}